\documentclass[11pt,a4paper]{book} 
\PassOptionsToPackage{backref=page,breaklinks=true}{hyperref}
\usepackage{hyperref}

\PassOptionsToPackage{compress}{natbib}
\usepackage{natbib}
\usepackage{setspace} %

\usepackage{placeins} %

\usepackage{Configs/paper_config}
\usepackage{Configs/package}

\usepackage[english]{babel}

\usepackage{graphicx}
\usepackage{float}
\usepackage{mathrsfs}

\usepackage[T1]{fontenc}
\usepackage{domitian,euler}

\usepackage{bbm}
\usepackage{enumitem}
\usepackage{lipsum}

\usepackage{tikz}

\usepackage[					
	paper=a4paper,			
	includefoot,			
	includemp,				
	bindingoffset=0.5cm,	
	top=2.25cm,				
	left=2.5cm,			
	right=0.8cm,			
	bottom=1.5cm,			
	marginparwidth=1.75cm,	
	marginparsep=10pt,		
	footskip=2cm,			
]{geometry}

\newcommand*{\geomdoc}{\newgeometry{includefoot,includemp,bindingoffset=-0.2cm,top=3cm,left=3.55cm,right=0.55cm,bottom=1.5cm,marginparwidth=1.75cm,marginparsep=10pt,footskip=2cm
}}

\usepackage[skip=10pt plus1pt]{parskip}

\usepackage{silence}
\usepackage{epigraph}
\usepackage{fancyhdr}
\usepackage{emptypage}

\DeclareFixedFont{\titlefont}{T1}{ppl}{bx}{n}{0.35in}

\makeatletter                       
\def\printauthor{%
    {\large \@author}}              
\makeatother
\author{ \fontsize{16pt}{16pt}\selectfont \bfseries Zhijing Jin}

\usepackage{framed}
\usepackage{titletoc}

\patchcmd{\tableofcontents}{\contentsname}{\bfseries\contentsname}{}{}

\renewenvironment{leftbar}
  {%
    \MakeFramed{\parshape 1 0cm \dimexpr\textwidth-6em\relax\FrameRestore}\vskip2pt%
  }
 {\endMakeFramed}

\titlecontents{chapter}
  [0.3em]{\Large\bfseries\vspace*{2\baselineskip}}
  {\hspace{-1.2cm}\parbox{6.5em}{%
    \hfill\Huge\bfseries\color{black}\thecontentslabel}%
   \vspace*{-1.9\baselineskip}\leftbar\bfseries}
  {}{\endleftbar}
\titlecontents{section}
  [11.4em]
  {\contentslabel{3em}}{}{} 
  {\phantom{    }\dotfill \nobreak\itshape\color{black}\contentspage}
\titlecontents{subsection}
  [13em]
  {\contentslabel{3em}}{}{}  
  {\phantom{    }\dotfill \nobreak\itshape\color{black}\contentspage}

\usepackage{titlesec}

\titleformat{\chapter}
{\vspace{-1.2cm}\normalfont\fontsize{30}{20}\bfseries}{}{0pt}{}

\makeatletter
\newcommand{\mainchapter}[1]{%
  \chapter{#1}%
  \vspace{-0.7cm}%
}
\makeatother

\newcommand{\prechapter}[1]{
    \newpage
    \thispagestyle{empty}
    \begin{Huge}
        \bfseries \selectfont #1 \\
        \rule[0.5ex]{\linewidth}{0.55pt}
    \end{Huge}
}

\titleformat{\section}
{\normalfont\Large\bfseries}{~\thesection}{1em}{}

\usepackage{fancyhdr}
\usepackage{tocloft} %

\newcommand{\mytitle}{Causality for Natural Language Processing\xspace}
\newcommand{\myname}{Zhijing Jin\xspace}

\renewcommand{\contentsname}{\vspace{0cm} Contents \vspace{-2cm}}

\newcommand{\titleblock}{
\vspace*{1em}
\begin{spacing}{1.6}  %
	{\huge
		\textbf{\mytitle}
	}
\end{spacing}
}

\newif\iffinal
\finalfalse

\begin{document}

\pagenumbering{roman}

\begin{titlepage}

\begin{center}

	\vspace{2cm}

\titleblock
	
	\vspace{.5cm}

		{\Large
	\myname
	
 \vspace{.5cm}
 
	2024
}
\end{center}
\end{titlepage}

\newpage

\thispagestyle{empty}
\vspace*{\fill}
\begin{center}
    \textit{To my parents, Xuefang Jin and Jian Wu, \\
    for their unconditional love and endless inspirations to me.}
\end{center}
\vspace*{\fill}

\newpage

\prechapter{Abstract}

Causal reasoning is a cornerstone of human intelligence and a critical capability for artificial systems aiming to achieve advanced understanding and decision-making. This thesis delves into various dimensions of causal reasoning and understanding in large language models (LLMs). It encompasses a series of studies that explore the causal inference skills of LLMs, the mechanisms behind their performance, and the implications of causal and anticausal learning for natural language processing (NLP) tasks. Additionally, it investigates the application of causal reasoning in text-based computational social science, specifically focusing on political decision-making and the evaluation of scientific impact through citations. Through novel datasets, benchmark tasks, and methodological frameworks, this work identifies key challenges and opportunities to improve the causal capabilities of LLMs, providing a comprehensive foundation for future research in this evolving field.

\prechapter{Zusammenfassung}
\textit{The ``Zusammenfassung'' is a machine-translated version of the abstract via \url{https://deepl.com} and corrected with the help of Jan Schneider.}

Kausales Denken ist ein Grundpfeiler menschlicher Intelligenz und eine entscheidende Fähigkeit für künstliche Systeme, die ein fortgeschrittenes Verständnis und fundierte Ent\-scheidungsfindung anstreben. Diese Arbeit untersucht verschiedene Aspekte des kausalen Denkens und Verstehens in großen Sprachmodellen, nämlich LLMs. Sie umfasst eine Reihe von Studien, die die Fähigkeiten von LLMs zur kausalen Inferenz, die Mechanismen hinter ihrer Leistungsfähigkeit und die Auswirkungen von kausalem und antikausalem Lernen auf Aufgaben der natürlichen Sprachverarbeitung, oder NLP beleuchten. Darüber hinaus wird die Anwendung kausalen Denkens in der textbasierten computergestützten Sozialwissenschaft untersucht, wobei der Fokus auf politischer Entscheidungsfindung und der Bewertung des wissenschaftlichen Einflusses durch Zitationen liegt. Durch neuartige Datensätze, Benchmark-Aufgaben und methodische Rahmenbedingungen identifiz\-iert diese Arbeit zentrale Herausforderungen und Chancen zur Verbesserung der kausalen Fähigkeiten von LLMs und bietet eine umfassende Grundlage für zukünftige Forschung in diesem sich entwickelnden Bereich.

\prechapter{Acknowledgements}

Throughout my PhD, I am deeply grateful to Professor Bernhard Schölkopf for being a constant source of inspiration and a remarkable role model as a scientist. 
His foundational work on causal inference for machine learning inspired me to start my own journey of causality for NLP. 
His broad knowledge across AI,
statistics, physics, and philosophy often bring interesting insights in the way we do science.
Moreover, as an incredible supervisor, Bernhard creates one of the best nurturing environments to do curiosity-driven research, gives me extensive support, and keeps inspiring with his sharp insights.

I really appreciate my co-supervisor, Professor Mrinmaya Sachan, for his detailed mentorship, unwavering support, and guidance throughout my PhD. His close supervision during the design of my research agenda and teaching of how to write well have been crucial to my success. Thanks to him, I have enjoyed a resourceful and fulfilling PhD experience across Germany and Switzerland.

I owe endless thanks to Professor Rada Mihalcea, whose loving and inspiring presence has been invaluable to me. She is one of the kindest people I have ever met, and being her mentee is one of the best experiences a person could ever imagine. I appreciate our shared vision and passion, which drive us to establish various lines of research on NLP for Social Good, and the ACL Year-Round Mentorship program for our community.

I have also been fortunate to collaborate with and receive mentorship from several people. I really admire Mona Diab for her caring nature and strong sense of social responsibility, which made me really enjoy our responsible AI research together. Collaborating with Kun Zhang has shown me the brilliance of technical causal inference research and the importance of responsible community service. Ryan Cotterell has taught me how to connect formal methods with NLP, and how to write precise, beautiful academic papers. Working with Asli Celikyilmaz at Meta has taught me novel idea generation and efficient project management to bring those ideas to realization.

My PhD journey has been supported by Max Tegmark's recognition of my research agenda and his belief in my potential to contribute greatly to AI safety. The fellowship and funding support from his Future of Life Institute have significantly expedited my research. Collaborating with and learning from Yejin Choi has shown me what it means to be a great NLP researcher with a sharp mind and how to clearly position papers within the community. A final big moment at the end of my PhD was my encounter with Geoffrey Hinton, who identified my potential and encouraged me to apply to the University of Toronto, where I am fortunate to become an assistant professor, in the hometown of deep learning.

Rome wasn't built in a day, and I owe a lot of thanks to the people who have guided me into research. During my undergrad years, I was fortunate to receive smart and patient side-by-side guidance from Di Jin, who was a PhD student in Peter Szolovits's lab at MIT CSAIL at the time. His ability to identify research problems and work through technical methods led to my first burst of papers and, more importantly, helped me understand what research truly is.
My second burst of papers was inspired by Qipeng Guo during our time working together at Zheng Zhang's Amazon AI lab. Qipeng, a senior PhD student back then, was excellent in teaching, philosophical thinking, and research skills. It was during this period that I began to develop my research tastes.
As I began my PhD in causal inference, I am deeply grateful to Julius von Kügelgen and Luigi Gresele, who have strong technical backgrounds and kindly spent tremendous one-on-one time exploring the multifaceted beauty of statistical causal inference with me. Another turning point in the middle of my PhD was my collaboration with Sydney Levine and Max Kleiman-Weiner, who sparked my strong curiosity about moral reasoning. This curiosity evolved into one of the three pillars of my knowledge enterprise, alongside logic and causality.
To deepen my exploration of these three pillars, the many inspiring discussions with Felix Leeb and Ari Holtzman, fueled by our shared interest in the philosophy of science and understanding how things around us work, have been invaluable.
At the end of my PhD in 2024, I had great fortune to deliver several talk tours which gave the opportunity for me to talk to many profound thinkers, and I was intensively inspired by the chats with Peter Spirtes, Judea Pearl, Geoffrey Hinton, Yoshua Bengio, Caroline Uhler, Dominik Janzing, Elias Bareinboim, and Eric Lander, for many aspects of causality, from its philosophical foundation, mathematical formalization, to interdisciplinary applications. As an important source of knowledge, I feel really grateful to many life-changing books I encountered during my PhD, especially Bertrand Russell's \textit{The History of Western Philosophy}, and Walter Isaacson's biography of \textit{Leonardo da Vinci}, which opens my awareness of the fundamental methodology that people use for thinking.

I want to acknowledge all my talented mentees that I am very fortunate to have advised and worked with: Abhinav Lalwani, Ahmad Khan, Amélie Reymond, Andreas Opedal, András Strausz, Ayush Kaushal, David Guzman, David Jenny, Dmitrii Kharlapenko, Fernando Gonzalez, Flavio Schneider, Francesco Ortu, Giorgio Piatti, Harshvardhan Srivastava, Hongyuan Liu, Ishan Kumar Agrawal, Jad Beydoun, Jiarui Liu, Jiayi Zhang, Jingwei Ni, João Afonso Miguel, Justus Mattern, Kevin Blin, Lea Künstler, Luise Woehlke, Navreet Kaur, Neemesh Yadav, Nils Heil, Ojasv Kamal, Roberto Ceraolo, Rongwu Xu, Samuel Simko, Sawal Acharya, Steven Wang, Sumon Kanti Dey, Tejas Vaidhya, Vetha Raghuram, Xiaoyu Xing, Yahang Qi, Yihuai Hong, Yiwen Ding, Yongyi Yang, Yuchun Dai, Yuen Chen, Yuxin Ren, and Zhiheng Lyu.

Science could not be made possible without the people behind. At Max Planck Institute for Intelligent Systems and ETH, I received lots of help from directors/professors, colleagues, our incredible administrative assistants, Sabrina, Ann-Sophie, Lidia, Linda, Patrizia, Paulina, and our research engineer Vincent, along with many others. I am also grateful to Professors Georg Martius and Seong Joon Oh for kindly being on my thesis committee, and administrative support at the University of Tuebingen. 

Finally, my PhD years would not have been as inspiring without the 24/7 support of my dad. His intelligence and love provided not only emotional support but also practical help in solving unforeseen out-of-distribution problems beyond both our life experiences. As my PhD coincided with the COVID-19 period challenging for physical and mental health, the optimistic and resilient character you see in me today owes to my parents' unconditional love and the cherished friendship with my 90+ year-old German neighbor, Reinhard Beyer.

\newpage
\pdfbookmark[1]{Publications}{Publications}

\begingroup
\let\clearpage\relax
\let\cleardoublepage\relax
\let\cleardoublepage\relax

\prechapter{Publications}

\definecolor{airforceblue}{rgb}{0.36, 0.54, 0.66}
\definecolor{pubcolor}{HTML}{0F587D} %
\definecolor{menteecolor}{RGB}{80,80,80}

\newcommand{\cvpub}[5]{{\textbf{#1}}\\#2.~
\textcolor{airforceblue}{\textbf{{#3}} \textbf{{#4}}}}

\renewcommand{\ul}[1]{\textcolor{menteecolor}{\textbf{\textit{#1}}}}

\newcommand{\mystar}{\hspace{-2pt}\textsuperscript{$\star$}\hspace{3pt}}
\newcommand{\cosupervise}{\hspace{0pt}\textsuperscript{$\dag$}\hspace{0pt}\xspace}
\newcommand{\highlight}[1]{\textcolor[rgb]{0.76, 0.13, 0.28}{\textbf{#1}}}

\newenvironment{thinglist}
{
  \begin{itemize}[noitemsep,label={},leftmargin=2em,itemindent=-2em]
}{
  \end{itemize}
}

The following peer-reviewed publications are at the core of my PhD research and covered in this dissertation:

\small

\begin{center}
  Equal Contribution (*);
  Co-Supervision (\cosupervise); Mentees I Advised (\textit{\ul{Name}}); Oral (\oralpres).
\end{center}

\subsection*{Part I. Causal Reasoning in LLMs}

\begin{itemize}
  \item 
\cvpub{\href{http://arxiv.org/abs/2306.05836}{Can Large Language Models Infer Causation from Correlation?}} {\textbf{Zhijing Jin},\mystar \ul{Jiarui Liu},\mystar \ul{Zhiheng Lyu}, Spencer Poff, Mrinmaya Sachan, Rada Mihalcea, Mona Diab\cosupervise, Bernhard Schölkopf\cosupervise}{ICLR}{2024}{I proposed the idea based on discussions with Bernhard, ran the experiments together with Jiarui and Zhiheng, and wrote the entire paper. All co-authors helped the conceptualization and gave writing suggestions.}
  \item 
  \cvpub{\href{http://arxiv.org/abs/2312.04350}{CLadder: Assessing Causal Reasoning in Language Models}} {\textbf{Zhijing Jin},\mystar \ul{Yuen Chen},\mystar Felix Leeb,\mystar Luigi Gresele,\mystar \ul{Ojasv Kamal}, \ul{Zhiheng Lyu}, \ul{Kevin Blin}, \ul{Fernando Gonzalez}, Max Kleiman-Weiner, Mrinmaya Sachan, Bernhard Schölkopf}{NeurIPS}{2023}{The conceptualization and design of this project was led by Zhijing, Felix, and Luigi, and supervised
by Bernhard on the causality part, and Mrinmaya on the NLP part. The co-authors contributed to implementation (including substantial work from Yuen and Felix), idea discussions, and writing.}

\end{itemize}

\subsection*{Part II. Causal Understanding of How LLMs Work}

\begin{itemize}

  \item 
  \cvpub{\href{http://arxiv.org/abs/2402.11655}{Competition of Mechanisms: Tracing How Language Models Handle Facts and Counterfactuals}} {\ul{Francesco Ortu},\mystar \textbf{Zhijing Jin},\mystar Diego Doimo, Mrinmaya Sachan, Alberto Cazzaniga\cosupervise, Bernhard Schölkopf\cosupervise}{ACL}{2024}{Me and Francesco developed the idea and orgranized the experiments. Bernhard and Alberto co-supervised this work, and gave insightful research guidance. Diego and Mrinmaya provided crucial research suggestions. All contributed significantly to the writing of this paper.
  }
  \item 
  \cvpub{\href{https://arxiv.org/abs/2210.12023}{A Causal Framework to Quantify the Robustness of Mathematical Reasoning in Language Models}} {Alessandro Stolfo,\mystar \textbf{Zhijing Jin},\mystar Kumar Shridhar, Bernhard Schölkopf, Mrinmaya Sachan}{ACL}{2023}{Me and Alessandro designed the project. I proposed the fundamental idea of framing robustness as the discrepancy across two causal mechanisms, settling down the technical ideas. Alessandro conducted all the experiments and analyses, and we wrote the paper together, with valuable research guidance from the co-authors.}

\end{itemize}

\subsection*{Part III. Causality among the Learning Variables}

\begin{itemize}

    \item 
  \cvpub{\href{https://arxiv.org/abs/2110.03618}{Causal Direction in Data Collection Matters: Implications of Causal and Anticausal Learning in NLP}} {\textbf{Zhijing Jin},\mystar Julius von Kuegelgen,\mystar \ul{Jingwei Ni}, \ul{Tejas Vaidhya}, \ul{Ayush Kaushal}, Mrinmaya Sachan, Bernhard Schölkopf}{EMNLP}{2021. \highlight{~\oralpres~Oral Presentation}}{I proposed the idea together with Julius and Bernhard. Me, Jingwei, Tejas, and Ayush conducted different parts of the experiments. I wrote the main content, Julius wrote the technical details, and Mrinmaya and Bernhard improved the writing.}
  \item 
  \cvpub{\href{http://arxiv.org/abs/2404.11055}{On the Causal Nature of Sentiment Analysis}} {\ul{Zhiheng Lyu},\mystar \textbf{Zhijing Jin},\mystar \ul{Fernando Gonzalez}, Rada Mihalcea, Bernhard Schölkopf, Mrinmaya Sachan}{Findings of EMNLP}{2024}{I designed the main idea of the paper, inspired by discussions with Bernhard following up our causal and anti-causal learning work above. Zhiheng and I together worked on the experimental design, he and Fernando conducted all the experiments, and Rada, Bernhard and Mrinmaya closely guided the project.}

\end{itemize}

\subsection*{Part IV. Causality for Text-Based Computational Social Science}

\begin{itemize}

  \item 
  \cvpub{\href{https://aclanthology.org/2021.findings-emnlp.27/}{Mining the Cause of Political Decision-Making from Social Media: A Case Study of COVID-19 Policies across the US States}} {\textbf{Zhijing Jin}, Zeyu Peng, \ul{Tejas Vaidhya}, Bernhard Schölkopf, Rada Mihalcea}{Findings of EMNLP}{2021}{I initiated the idea together with Bernhard and Rada on using NLP for COVID research. Zeyu and I designed the research question and method, contextualizing it in political science literature. Me and Tejas conducted the experiments. I wrote the paper, with guidance from Zeyu, Bernhard, and Rada.}
  \item 
  \cvpub{\href{http://arxiv.org/abs/2311.02790}{CausalCite: A Causal Formulation of Paper Citations}} {\ul{Ishan Kumar},\mystar \textbf{Zhijing Jin},\mystar Ehsan Mokhtarian, Siyuan Guo, \ul{Yuen Chen}, Mrinmaya Sachan, Bernhard Schölkopf}{Findings of ACL}{2024}{Bernhard and I designed the main idea of NLP for causal analysis in papers. I closely guided Ishan for the experiment design and code implementation. Siyuan and Ehsan further improved the technical idealization, and Mrinmaya and Yuen helped the evaluation design. Me, Ehsan, and Ishan wrote the paper, together with all the coauthors improving the writing.}

\end{itemize}

\bigskip

My entire list of publications is available at

\vspace{-0.5em}
\quad \quad \textbf{Google Scholar:} \href{https://scholar.google.com/citations?user=Mdr6wjUAAAAJ}{scholar.google.com/citations?user=Mdr6wjUAAAAJ}

\vspace{-.7em}
\quad \quad \textbf{Citations:}	3,266 \quad \quad \quad \quad \quad \quad 
    \textbf{h-index:}	23 \quad \quad \quad (by December 2024)

\normalsize

\endgroup     

\vfill

\let\cleardoublepage=\clearpage

\pagenumbering{gobble}
\selectlanguage{english} %
\setcounter{tocdepth}{1}
\tableofcontents
\thispagestyle{empty}
\newpage
\geomdoc %

\pagenumbering{arabic}
\mainchapter{Introduction}

\section{Overview}

Causality is a fundamental aspect of human cognition and intelligence, underlying our understanding of the world and our ability to make decisions. In the field of natural language processing (NLP), the capability to infer and reason about causality is increasingly recognized as a critical component of intelligent systems.

Despite the recent advancement of large language models (LLMs)~\citep[][\textit{inter alia}]{radford2019language,devlin-etal-2019-bert,brown2020gpt3,zhang2022opt,openai2023gpt4,ignat-etal-2024-has}, a key question still remains: Can these models understand and reason about causality? 
This is a critical skill before we can trust AI agents to be integrated into decision-making systems.
Moreover, even if LLMs succeed at some extent of reasoning, they still lack transparency of how their decisions are made, forming a strong need for interpretability \citep{luo2024understanding, rauker2023toward,zou2023representation}.

To bridge the gap, this thesis explores various facets of causal reasoning in LLMs.
We present a series of studies that collectively advance the knowledge of how well these models perform causal reasoning (\cref{part:1}), how their decisions are made (\cref{part:2}), how causality among learning variables influences NLP tasks (\cref{part:3}), and how causality and NLP can together analyze social problems (\cref{part:4}).
Below we introduce an overview of the four parts and their corresponding chapters.

\subsection{Causal Reasoning in LLMs (\cref{part:1})}

In \cref{part:1}, we investigate two formal causal reasoning skills in LLMs: causal discovery (\cref{ch:corr2cause}) and causal effect reasoning (\cref{ch:cladder}), which existing models struggle to address. We propose these pure reasoning tasks independent of empirical knowledge, and contribute symbolically grounded datasets that are carefully constructed.
Based on our proposed data and test pipeline, we first look into the initial performance of LLMs off the shelf, explore how fine-tuning improves their performance, and propose chain-of-thought reasoning to ground the inference skills of LLMs in formal steps.

\paragraph{Causal Discovery in LLMs (\cref{ch:corr2cause})}
The ability to distinguish between correlation and causation is crucial for intelligent AI systems. In \cref{ch:corr2cause}, we propose a benchmark dataset, \textsc{Corr2Cause}, specifically designed to test the causal discovery skills of LLMs. This novel task requires models to determine causal relationships from a set of correlational statements. Our large-scale dataset consists of over 200K samples, on which we evaluate seventeen existing LLMs. The findings reveal a significant shortcoming: these models perform close to random on this task, indicating a fundamental gap in their causal inference abilities. Although fine-tuning improves performance somewhat, the models still struggle with generalization, performing well only on in-distribution settings but failing on out-of-distribution queries. This study underscores the challenges in guiding future research to enhance the pure reasoning skills and generalizability of LLMs.

\paragraph{Causal Effect Reasoning in LLMs (\cref{ch:cladder})}

To investigate the other skill, causal effect reasoning, we propose a new task in \cref{ch:cladder} inspired by the ``causal inference engine'' concept postulated by \citet{pearl2018book}. Our dataset contains 10K samples derived from causal graphs and queries, which are translated into natural language. We evaluate LLMs using this dataset and introduce a chain-of-thought prompting strategy, \textsc{CausalCoT}. The results highlight the challenges LLMs face in causal reasoning, providing deep insights into their limitations and suggesting directions for future improvements.

\subsection{Causal Understanding of How LLMs Work (\cref{part:2})}

Beyond understanding the causal reasoning skills in LLMs, a natural following question is to interpret how models make their decisions, which leads to our exploration in \cref{part:2}. We look into two types of interpretation, one to inspect and intervene their internal states, called intrinsic interpretability (\cref{ch:compmech}), and the other to perturb the input-output space to capture behavioral tendencies, called
behavioral interpretability (\cref{ch:mathrobust}). We will introduce each in the following.

\paragraph{Intrinsic Interpretability (\cref{ch:compmech})}
We present a novel formulation to understand the inner mechanisms of LLMs.
Most existing research focuses on analyzing a single mechanism, such as how models copy or recall factual knowledge. In this work, we propose a formulation of \textit{competition of mechanisms}, which focuses on the interplay of multiple mechanisms instead of individual mechanisms and traces how one of them becomes dominant in the final prediction.
We uncover how and where mechanisms compete within LLMs using two interpretability methods: logit inspection and attention modification. Our findings show traces of the mechanisms and their competition across various model components and reveal attention positions that effectively control the strength of certain mechanisms.%

\paragraph{Behavioral Interpretability (\cref{ch:mathrobust})}
Different from previous work that uses heuristics to perform behavioral tests, we systematically propose a pipeline to probe the difference between the desired causal mechanisms, and model-learned causal mechanisms.
To implement our framework, we focus on mathematical reasoning problems. By grounding our analysis in a causal graph, we assess the causal effect of different input factors such as problem text, operands, and operators on the model's output. 
 This framework enables a detailed analysis of the models' robustness and highlights the critical role of causal factors in shaping their responses.  
 The study reveals that robustness does not continuously improve with model size; however, the GPT-3 Davinci model (175B) shows significant advancements in both robustness and sensitivity compared to other variants. Despite the relative improvement, our results indicate that there is still substantial room for enhancing the robustness and generalization capabilities of LLMs through a better understanding of their causal mechanisms.

\subsection{Causality among the Learning Variables (\cref{part:3})}

Apart from causal inference to improve performance and to interpret the models, we also explore the causal relationships between the input and output variables in NLP tasks in \cref{part:3}, which provides valuable insights into the design and evaluation of models. This part includes studies that examine the implications of causal and anticausal learning and their impact on NLP tasks (\cref{ch:icm}), and also provide an example discovering the causal relationship between the input-output learning variables (\cref{ch:psychcausal}).

\paragraph{Implications of Causal and Anticausal Learning in NLP (\cref{ch:icm})}

We introduce the principle of independent causal mechanisms (ICM) to the NLP community. By inspecting the causal directions between the input and output variables, we categorize common NLP tasks into clearly causal (where the input variable serves as a cause of the output variable, or is collected before the annotation of the output), anticausal (where the output variable serves as a cause of the input variable, or is collected before the annotation of the input), or mixed.
After an extensive meta-analysis of over 130 published studies, the paper demonstrates that the causal direction of data collection significantly affects learning outcomes. This work is the first to apply the ICM principle to NLP and offers constructive suggestions for future modeling choices based on causal insights.

\paragraph{Discovering Causal and Anticausal Sentiment Analysis (\cref{ch:psychcausal})}

We further extend this exploration to cases where the variable causal relations are not evident \textit{a priori}, but discovered using interdisciplinary insights. In \cref{ch:psychcausal}, we reformulate sentiment analysis (SA) a combination of causal discovery and prediction tasks. For the first task, we discover the causal relation in SA tasks, namely between a review and its sentiment, by using the peak-end rule from psychology. In this way, we classify reviews based on whether the review primes the sentiment or vice versa. Knowing the causal direction, we inspect how it affects the prediction performance of LLMs when the prediction task aligns with the same causal direction or is opposite.
To improve model performance, we construct causal prompts that reflect the underlying causal graph, which lead to substantial improvements, demonstrating the importance of considering causal mechanisms in SA tasks.

\subsection{Causality for Text-Based Computational Social Science (\cref{part:4})}

Finally, we explore in \cref{part:4} the application of causal inference to text-based computational social science to address real-world problems. We conduct two studies that examine the causes behind political decision-making (\cref{ch:covidtwitter}) and paper citations (\cref{ch:causalcite}).

\paragraph{Causal Policy Analysis (\cref{ch:covidtwitter})}
As the first application study, we investigate how public opinion on social media influences policy decisions during the COVID-19 pandemic. By analyzing textual Twitter data and controlling for confounders such as case increases and unemployment rates, the study conducts causal inference to identify trends in political decision-making across different states. This research highlights the dynamic interaction between public sentiment and political actions in a rapidly evolving context.

\paragraph{Causal Analysis of Paper Citations (\cref{ch:causalcite})}

For the second case study of paper citations, we propose a new method to evaluate the significance of scientific papers through causal impact on subsequent research. Using \textsc{TextMatch}, a novel causal inference method adapted for high-dimensional text embeddings, we assess the causal influence of papers on their follow-ups. The effectiveness of our framework is demonstrated through various criteria, providing a more accurate measure of a paper's true impact and suggesting ways for future researchers to utilize this metric.

\section{Structure of the Thesis}

This thesis is organized into four main parts, corresponding to the thematic areas outlined above. Each part includes a detailed examination of the relevant studies, methodologies, and findings, offering a comprehensive view of causal methods for NLP:
\begin{enumerate}
    \item \textbf{Causal Reasoning in LLMs} (\cref{part:1}):
    Assessing the ability of LLMs on causal inference by introducing formal benchmarks, and improving the model performance by chain-of-thought reasoning and finetuning.
    \item \textbf{Causal Understanding of How LLMs Work} (\cref{part:2}):
    Investigating the internal mechanisms and robustness of LLMs by causal interventions and causal mediation analysis.
    \item \textbf{Causality among Learning Variables} (\cref{part:3}):
    Exploring the causal or anticausal relation among the learning variables, and drawing insights to understand and improve model performance.
    \item \textbf{Causality for Text-Based Computational Social Science} (\cref{part:4}):
    Applying causal inference on text data to uncover social and political insights.
\end{enumerate}

\section{Contributions and Impact}

This body of work contributes to the growing field of causal inference in machine learning, providing critical insights into the capabilities and limitations of LLMs in reasoning about causality. By using causal methods to intervene on LLMs, we pave the way for developing more robust and interpretable AI systems. Moreover, the applications in computational social science highlight the broader societal implications of causal reasoning with text data, offering new tools and perspectives for analyzing complex social phenomena.

\part{Causal Reasoning in LLMs} \label{part:1}

\mainchapter{Can LLMs Infer Causation from Correlation?}\label{ch:corr2cause}

\newcommand{\ourdata}{{\textsc{Corr2Cause}}\xspace}

Causal inference is the study that discovers cause-effect relationships from interventional and/or observational data \citep{pearl2009causality,spirtes2000causation,zhang2009causality}. Causality started as a philosophical subject \citep{beebee2009oxford,russell2004history,kant1781critique}, and in the recent centuries integrated into statistics with established concepts and tools \citep{fisher1927,rubin1980,spirtes1993causation,pearl2009causality}. 
In the era of LLMs, we formulate such formally-grounded pure causal inference as a target reasoning capability for language models.

In this work, we propose the first benchmark dataset to test the 
pure causal inference skills of large language models (LLMs).
Specifically, we formulate a novel task \ourdata, which takes a set of correlational statements
and determines the causal relationship between the variables.
We curate a large-scale dataset of more than 200K samples, on which we evaluate seventeen existing LLMs. Through our experiments, we identify a key shortcoming of LLMs in terms of their causal inference skills, and show that these models achieve almost close to random performance on the task. This shortcoming is somewhat mitigated when we try to re-purpose LLMs for this skill via finetuning, but we find that these models still fail to generalize -- they can only perform causal inference in in-distribution settings when variable names and textual expressions used in the queries are similar to those in the training set, but fail in out-of-distribution settings generated by perturbing these queries.
\ourdata is a challenging task for LLMs, and can be helpful in guiding future research on improving LLMs' pure reasoning skills and generalizability. Our data is available at  { \url{https://huggingface.co/datasets/causalnlp/corr2cause}}, and our code is at { \url{https://github.com/causalNLP/corr2cause}}.

\section{Introduction}

Causal inference, i.e., the ability to establish the correct causal relationships between variables or events, is fundamental to human intelligence. %
There are two distinct ways this causal inference capability can be acquired: one through empirical knowledge, e.g., we know from common sense that touching a hot stove will get us burned;
the other through \textit{pure causal reasoning}, 
as causality can be formally argued and reasoned about using known procedures and rules from causal inference \citep{spirtes2001causation,pearl2009causality,peters2017elements}.
One example is that we have the a priori knowledge that the correlation between A and B does not necessarily imply causality. This is a formal rule that holds true regardless of the realizations of the variables A and B.

With the rise of large language models (LLMs) \citep[][\textit{inter alia}]{radford2019language,devlin-etal-2019-bert,ouyang2022instructGPT,
zhang2022opt,openai2023gpt4}, a crucial research question is whether they can do causal reasoning well. 
Recent studies have pointed out that LLMs are ``causal parrots,'' which recite the causal knowledge in the training data \citep{zevcevic2023causal}. Moreover, the vast majority of studies frame causal reasoning as a skill to navigate around empirical knowledge \citep{gordon-etal-2012-semeval,sap2019atomic,Sap2019SocialIQA,qin-etal-2019-counterfactual,bhagavatula2020abductive}, and also
treat LLMs as a knowledge base when evaluating its causal skills \citep{kiciman2023causal,tu2023causal,xie2023echo}.
However, all the above lines of research frame causality as empirical knowledge, thus relying heavily on the quality and the coverage of the training data,  overlooking the great potential of the formal causal reasoning skills 
to process correlational information to causal conclusions.

\begin{figure}[t]
\centering
    \includegraphics[width=\textwidth]{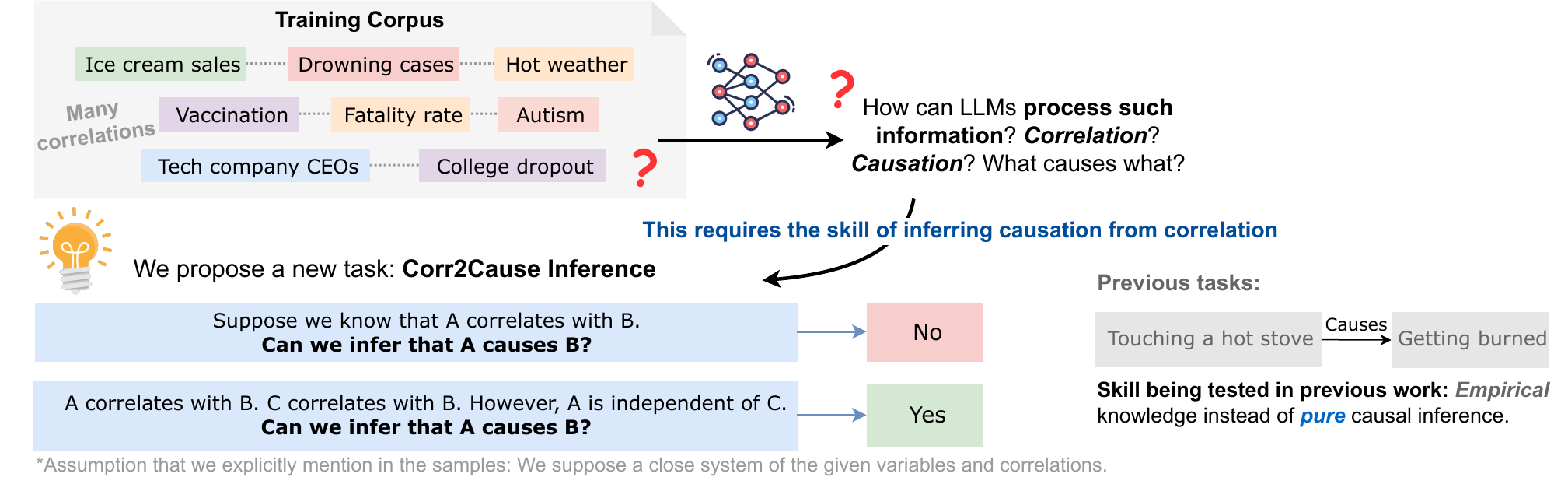}
    \caption{Illustration of the motivation behind our task and dataset.}
    \label{corr2cause:fig:ex}
\end{figure}

Drawing inspirations from technical studies on causal discovery \citep{spirtes2001causation,spirtes2016causal,clark2019review},
we formulate a novel task for NLP, \textit{correlation-to-causation inference }(\ourdata), which is an important skill for LLMs. Imagine the scenario in \cref{corr2cause:fig:ex}, where the training corpus does not tediously cover every causal relation, but more pervasively talk about correlations, such as which events tend to co-occur. Learning a good \ourdata skill can enable LLMs to draw causal relations behind the mere correlational information on the surface. For example, several decades ago, there might be an observation that female university students tend to perform better, but behind the correlational statistics is the causal graph that female students have to achieve extra good performance to get into universities as the first place.

To this end, we collect the \ourdata dataset, the first dataset %
to test the pure causal reasoning abilities of LLMs.
All the questions in this dataset are centered around testing
when it is valid or invalid to infer causation from correlation. To 
systematically compose this dataset, we ground our generalization process in the formal framework of causal discovery \citep{spirtes2001causation,glymour2016causal,spirtes2016causal}, which provides rules about how to deduce 
causal relations
among variables 
given their statistical correlation in the observational data.
We generate more than 200K data points, 
and label a correlation-causation statement pair as valid if and only if there is a bijective mapping between the statistical correlation and the underlying causality.

Based on our \ourdata dataset with 200K samples, we investigate two main  research questions: 
(1) How well do existing LLMs perform on this task? 
(2) Can existing LLMs be re-trained or re-purposed on this task and obtain robust causal inference skills?
Through extensive experiments, we show empirically that none of the 17 existing LLMs we investigate perform well on this pure causal inference task. We also show that although LLMs can demonstrate better performance after being finetuned on the data, the %
causal inference skills attained by them are not robust. 
In summary, our contributions are as follows:
\begin{enumerate}
[topsep=0px,itemsep=0.5px]
    \item We propose the novel task of \ourdata, 
    to probe an aspect of LLM's reasoning
    ability, \textit{pure causal inference};
    \item We compose a dataset of over 200K samples, using insights from causal discovery;
    \item We evaluate the performance of 17 LLMs on our dataset, finding that all of them perform  poorly, close to the random baseline;
    \item We further explored whether LLMs can learn the skill through finetuning, and find that 
     LLMs fail to robustly acquire this skill in out-of-distribution settings. Finally, we suggest future work to explore more ways to enhance the pure causal inference skill in LLMs.
\end{enumerate}

\section{Preliminaries: Causal Inference} \label{corr2cause:sec:causality}

\subsection{Directed Graphical Causal Models (DGCMs)}\label{corr2cause:sec:graph_notations}
A directed graphical causal model (DGCM) is a commonly used representation to express the causal relations among a set of variables. 
Given a set of $N$ variables $\bm{X} = \{X_1, \dots, X_N \}$, we can encode the causal relations among them using a directed graph $\mathcal{G} := (\bm{X}, \bm{E})$, where $\bm{E}$ is the set of directed edges. Each edge $e_{i,j} \in \bm{E}$ represents a causal link $X_i \rightarrow X_j$, meaning that $X_i$ is a direct cause of $X_j$. 
In the context of this work, we take the common assumption of directed acyclic graphs (DAGs), which most causal discovery methods use \citep{clark2019review}, as graphs with cycles can make the causal discovery process arbitrarily hard.

Following the graph-theoretic terminology, we use an analogy of the ancestry tree to denote the relations between two variables.
For example, we call $X_i$ as a \textit{parent} of $X_j$ if there is a directed edge $X_i \rightarrow X_j$ in the graph, and, thus, $X_j$ is a \textit{child} of $X_i$. Similarly, we denote $X_i$ as an \textit{ancestor} of $X_j$ if there exists a directed path from $X_i$ to $X_j$, and, thus, $X_j$ is a \textit{descendent} of $X_i$. Note that a parent is a special case of an ancestor where the directed path has a length of 1. 

For convenience, we also introduce the notions for some special three-variable relations.
Given two variables $X_i$ and $X_j$, we call a third variable $X_k$ a \textit{confounder} (i.e., \textit{common cause}) if $X_k$ is a parent of both $X_i$ and $X_j$; a \textit{collider} (i.e., \textit{common effect}) if $X_k$ is a child of both $X_i$ and $X_j$; and a \textit{mediator} if $X_k$ is both a child of $X_i$, and a parent of $X_j$. 

\subsection{D-Separation and Markov Property}

\paragraph{D-Separation}
D-separation \citep{pearl1988probabilistic} is a fundamental concept in graphical models used to determine whether two sets of nodes $\bm{X}$ and $\bm{Y}$ in a DAG $\mathcal{G}$ are conditionally independent given a third set of nodes
$\bm{Z}$, where the three sets are disjoint.
We say that $\bm{X}$ and $\bm{Y}$ are d-separated by $\bm{Z}$ if all paths between any node in $\bm{X}$ and any node in $\bm{Y}$ are \textit{blocked} by the conditioning set $\bm{Z}$. A path between $\bm{X}$ and $\bm{Y}$ is blocked by $\bm{Z}$ if there exists a node $A\in \bm{Z}$ which satisfies one of the following conditions: $A$ is the 
parent node in a fork structure on the path (i.e., $\cdot \leftarrow A \rightarrow \cdot$); $A$ is the 
mediator node in a chain structure on the path (i.e., $\cdot \rightarrow A \rightarrow \cdot$); or in any collider structure on the path (i.e., $\cdot \rightarrow A \leftarrow \cdot$), $\bm{Z}$ does not contain $A$ or its descendants.

\paragraph{Markov Property}
The Markov property in a DAG $\mathcal{G}$ states that each node $X_i$ is conditionally independent of its non-descendants given its parents, namely
$X_i \perp\!\!\!\perp \NonDe(X_i) | \PA(X_i)$,
where $\NonDe(X_i)$ denotes the non-descendants of $X_i$ excluding itself, and $\PA(X_i)$ denotes the parents of $X_i$.
Using the Markov property, we can factorize the joint distribution of all the nodes in the graph into
$P(X_1, \dots, X_N) = \prod_{i=1}^N P(X_i | \bm{\mathrm{PA}}(X_i) )$. To infer the causal graph from probability distributions, a common assumption is faithfulness, namely the validity to infer all the d-separation sets in the graph from the independence relations in the probability distribution. In our work, we also take this broadly taken assumption which holds for most real-world scenarios.

\paragraph{Markov Equivalence of Graphs}
We denote two DAGs as Markov equivalent if they induce the same joint distribution $P(\bm{X})$. The set of DAGs that are Markov equivalent to each other is called a Markov equivalence class (MEC).
Causal graphs in the same MEC can be easily identified since they have the same skeleton (i.e., undirected edges) and V-structures (i.e., structures in the form of $A\rightarrow B \leftarrow C$ where $A$ and $C$ are not connected).

Obviously, there is a one-to-many mapping (i.e., surjection) between the causal graph and statistical distribution. 
Namely, each causal graph sufficiently determines a statistical distribution, but from a statistical distribution, we cannot necessarily induce a unique causal graph. This is why we say ``correlation does not necessarily mean causation''.

\subsection{Causal Discovery} 

Causal discovery aims to learn the causal relations by analyzing statistical
properties in the observational data \citep{spirtes2001causation,glymour2016causal,spirtes2016causal,clark2019review}. It can be achieved through constraint-based methods \citep{spirtes2001causation}, score-based methods 
\citep{chickering2002optimal}, or other methods taking advantage of the functional causal models \citep{shimizu2006linear,hoyer2008nonlinear,zhang2009causality}.

To fit for the spirit of this paper to infer from correlation (expressed in natural language) to causation, we base our dataset design on the widely-used Peter-Clark (PC) algorithm \citep{spirtes2001causation}. 
The PC algorithm is based on the principles of conditional independence and the causal Markov assumption, which allows it to efficiently identify causal relationships among variables in a given dataset. 
The algorithm first starts with a fully connected undirected graph among all the variables. Then it removes the edge between two variables if there is an unconditional or conditional independence relationship between them. Afterwards, it orients the directed edges whenever there is a V-structure. And finally, it iteratively checks the direction of the other edges until the entire causal graph is consistent with all the statistical correlations.

\section{Dataset Construction}
\begin{figure}[t]
    \centering
    
    \includegraphics[width=0.85\columnwidth]{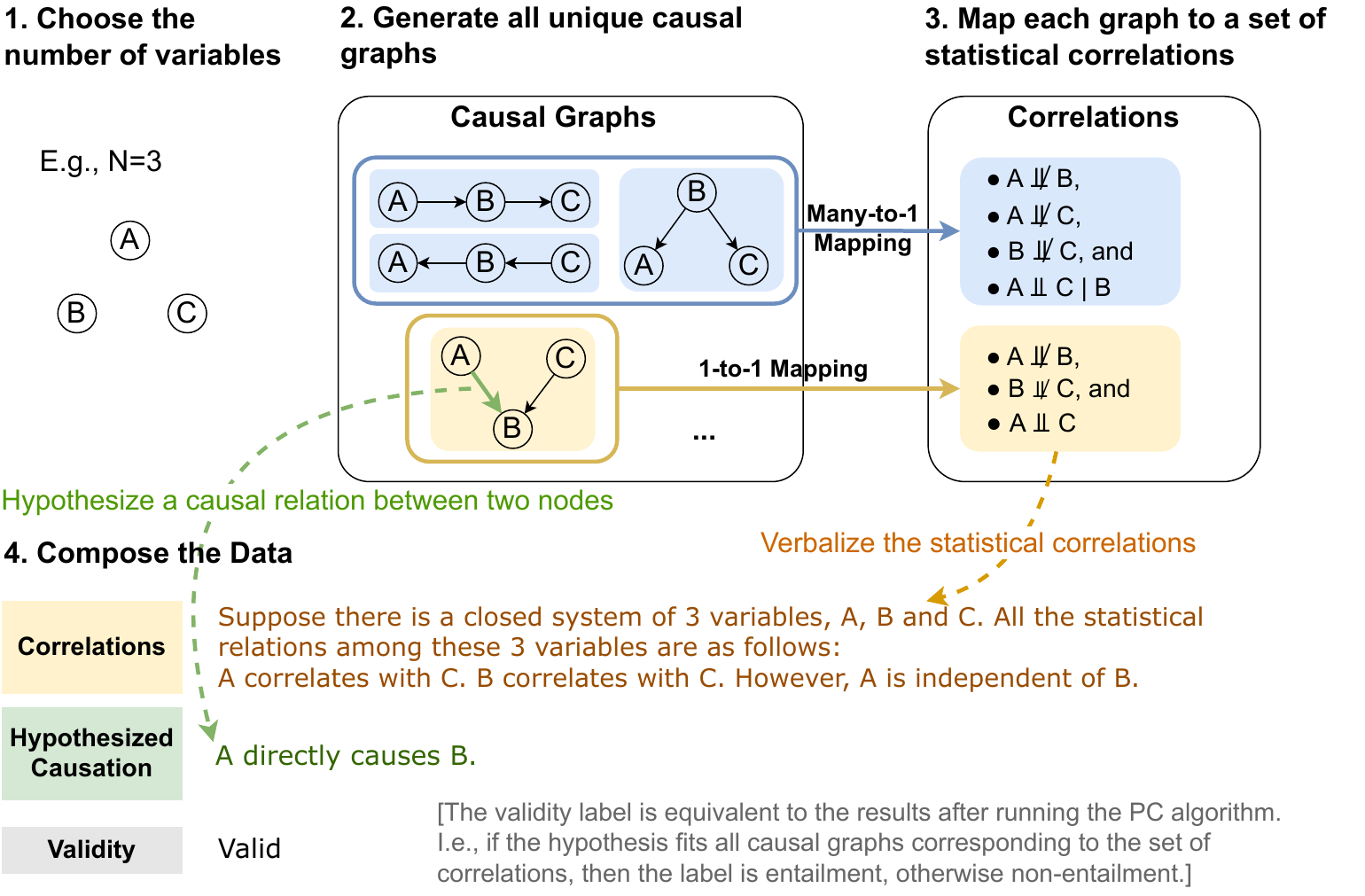}
    \caption{Pipeline of the data construction process.}
    \label{corr2cause:fig:construction}
\end{figure}
We introduce the construction of our dataset in this section. We start with our task formulation for \ourdata, and then briefly give an overview of the data generation process, followed by detailed descriptions of each step. We conclude the section with the overall statistics of the dataset.

\subsection{Task Formulation}
Given a set of $N$ variables $\bm{X}=\{X_1, \dots, X_N\}$, we have a statement $\bm{s}$ about all the correlations among the variables, and a hypothesis $\bm{h}$ describing the causal relation $r$ between the pair of variables $X_i$ and $X_j$. The task is to learn a function $f: (\bm{s}, \bm{h}) \mapsto v$ which maps the correlation statement $\bm{s}$ and the causal relation hypothesis $\bm{h}$ to their validity $v \in \{0, 1\}$, which takes the value 0 if this inference is invalid, and the value 1 if this inference is valid.

\subsection{Overview of the Data Generation Process}
We base the construction our dataset on several concepts of causal inference, including the DGCM, d-separation, and MECs, as introduced in \cref{corr2cause:sec:causality}. 

As in the overview of our data generation process in \cref{corr2cause:fig:construction}, we first choose the number $N$ of variables (Step 1) and generate all the unique DGCMs with $N$ nodes (Step 2), which we will introduce in the \cref{corr2cause:sec:graph_generation}. Then we collect all the d-separation sets from these graphs to identify MECs (Step 3) in \cref{corr2cause:sec:dsep_generation}.
Then, in Step 4, we create the formal form of data in \cref{corr2cause:sec:form_generation}. For each correspondence of the MEC to causal graphs, we compose the correlation statement based on the statistical relations in the MEC, and  hypothesize a causal relation between two variables, and produce the validity $v=1$ if the hypothesis is a shared property of all causal graphs in the MEC, and $v=0$ if the hypothesis is not necessarily true for all the MEC graphs. Finally, we introduce the verbalization process in \cref{corr2cause:sec:verbalization}.

\subsection{Constructing the Graphs with Isomorphism Checks}\label{corr2cause:sec:graph_generation}

The first step of the data generation is to compose the causal graphs, as in Step 1 and 2 of \cref{corr2cause:fig:construction}.
For a set of $N$ variables $\bm{X} = \{X_1, \dots, X_N\}$, there are $N(N-1)$ possible directed edges, since each node can link to any node other than itself. 
To remove cycles in the graph, we make the nodes in topological order, which only allows edges $X_i \rightarrow X_j$, where $i<j$. We achieve this by limiting the adjacency matrix of the graph to only having non-zero values above the diagonal, resulting in $N(N-1)/2$ possible directed edges for the DAGs.

At the first glance, for $N$ nodes, there should be $2^{N(N-1)/2}$ possible DAGs (i.e., the power set of all edges).  However, there could be isomorphic graphs in this set. To avoid this,  we perform a graph isomorphism check \citep{mckay2014practical}, and reduce the set so that only unique DAGs are retained, and we show their statistics in \cref{corr2cause:tab:step1_causal_graph}. Although we can handle large graphs, we mostly focus on smaller graphs that can still lead to a reasonably sized dataset, so we empirically set $N=6$, but future work can use our open-sourced codes to extend to more nodes.

\begin{table}[t]
    \centering \small
    \begin{tabular}{clccccc}
    \toprule
    \# Nodes & \# Unique DAGs & \# Edges/DAG & \# MECs & \# DAGs/MEC \\ \midrule
    2 & 2 out of $2$ & 0.50 & 2 & 1.0 \\
    3 & 6 out of $2^{3}$ & 1.67 & 5 & 1.2 \\
    4 & 31 out of $2^{6}$ & 3.48 & 20 & 1.55 \\
    5 & 302 out of $2^{10}$ & 5.89 & 142 & 2.13 \\
    6 & 5,984 out of $2^{15}$ & 8.77 & 2,207  & 2.71 \\
    Total & 6,325 & 8.60 & 2,376  & 2.66 \\
    \bottomrule
    \end{tabular}
    \caption{Statistics about the source causal graphs in our dataset. Given the number of nodes, we report the number of unique DAGs, average number of edges per DAG, number of MECs, and average number of DAGs per MEC.}
    \label{corr2cause:tab:step1_causal_graph}
\end{table}

\subsection{Programmatically Generating the D-Separation Sets}\label{corr2cause:sec:dsep_generation}
Based on the set of unique DAGs, we then programmatically generate the d-separation sets by graph theoretical conditions, as in Step 3 of \cref{corr2cause:fig:construction}. 
To realize this step, we code an efficient graph-theoretic algorithm to check for all the chain, fork, and collider structures to automatically identify the set of nodes that d-separate each pair of nodes.
Using the d-separation sets and the faithfulness assumption, we form the statistical correlations as follows. For each pair of nodes, they are conditionally independent given the variables in the d-separation set. If the d-separation set is empty, then the two nodes are unconditionally independent. If no d-separation set can be found for the two nodes, then they are directly correlated.

Moreover, using the d-separation sets, we are able to cluster causal graphs to MECs. We achieve it by tracing the mapping between the causal graphs and the set of statistical correlations, and backtracking the graphs with the same d-separation sets to group them in the same MEC. We show in \cref{corr2cause:tab:step1_causal_graph} that each MEC contains on average 2.66 DAGs.

\subsection{Composing the Hypotheses and Label}\label{corr2cause:sec:form_generation}

After generating the set of correlations based on the d-separation sets, we now generate the causal hypotheses. For the causal relation $r$, we focus on six common causal relations between two nodes introduced in \cref{corr2cause:sec:graph_notations}: Is-Parent, Is-Child,
Is-Ancestor (excluding the parents), Is-Descendant (excluding the children), 
Has-Confounder (i.e., there exists a confounder, or common cause, of the two nodes), and
Has-Collider (i.e., there exists a collider, or common effect, of the two nodes). 
In this way, the set of hypotheses contains 
all six meaningful causal relations between every pair of variables, resulting in a total size of
$6 \cdot N (N-1)/2 = 3N(N-1)$ hypotheses for a graph with $N$ variables.

To generate the ground-truth validity label, we start from the correlation sets in Step 3, then look up all the causal graphs in the same MEC corresponding to the given set of correlations, and check the necessity of the hypothesized causal relation. If the causal relationship proposed in the hypothesis is valid for all causal graphs within the MEC, then we generate the validity $v=1$; otherwise, we generate $v=0$. A special case of valid samples is that when the size of the MEC is 1, then there is a bijective mapping between the causal graph and the d-separation sets, so any hypothesis stating the causal properties of that unique causal graph is valid.

\subsection{Verbalizing into Language}\label{corr2cause:sec:verbalization}

Finally, as in the last step of \cref{corr2cause:fig:construction}, we convert all the information above to text data for our \ourdata task.
For the correlation statement, we verbalize
the set of correlations in Step 3 into a natural language statement $\bm{s}$.
When two variables cannot be d-separated, i.e., $A \not\perp\!\!\!\perp B$, then we describe them as ``$A$ correlates with $B$'' since they are directly correlated and cannot be independent by any condition. And if two variables have a valid d-separation set $\bm{C}$, then we describe them as ``$A$ is independent of $B$ given $\bm{C}$.'' In the special case when the d-separation set is empty, we directly say ``$A$ is independent of $B$.''
In addition, we disambiguate the setting by starting the correlation statement with the setup of a closed system of the given variables, and no hidden variables: ``Suppose there is a closed system of $N$ variables, A, B, \dots ~All the statistical relations among these $N$ variables are as follows:''.
Finally, to verbalize the hypothesis, we feed the causal relation triplet ($X_i$, $r$, $X_j$) into their hypothesis templates in \cref{corr2cause:tab:template}. For example, we turn the triplet ($A, \text{Is-Parent}, B$) into ``$A$ directly causes $B$'', as in the example of \cref{corr2cause:fig:construction}.
\begin{table}[t]
    \centering \small
    \begin{tabular}{ll}
    \toprule
Causal Relation & Hypothesis Template \\\midrule
Is-Parent & \texttt{\{Var i\}} directly causes \texttt{\{Var j\}}. \\
Is-Ancestor & \texttt{\{Var i\}} causes something else which causes \texttt{\{Var j\}}. \\
Is-Child & \texttt{\{Var j\}} directly causes \texttt{\{Var i\}}. \\
Is-Descendant & \texttt{\{Var j\}} is a cause for \texttt{\{Var i\}}, but not a direct one. \\
Has-Collider & There exists at least one collider (i.e., common effect) of \texttt{\{Var i\}} and \texttt{\{Var j\}}. \\
Has-Confounder & There exists at least one confounder (i.e., common cause) of \texttt{\{Var i\}} and \texttt{\{Var j\}}. \\
\bottomrule
    \end{tabular}
    \caption{Templates for each causal relation in the hypothesis. We use \texttt{\{Var i\}} and \texttt{\{Var j\}} as placeholders for the two variables.}
    \label{corr2cause:tab:template}
\end{table}

\subsection{Statistics of the Resulting Data}\label{corr2cause:sec:stats} 

We show the statistics of our \ourdata{} dataset in \cref{corr2cause:tab:stats}. 
Overall, our dataset contains 207,972 samples, where 18.57\% of the samples have the positive label (i.e., with validity=1). 
The average length of the premise is 424.11 tokens, and hypothesis 10.83 tokens. We split the data into 205,734 training samples, 1,076 development and 1,162 test samples.\footnote{\textit{Note for our dataset v2.0:} We notice that our original data (v1.0) has duplication due to symmetric relations and verbalizations of the hypothesis. E.g., Is-Parent(A, B) has the exact hypothesis verbalization as Is-Child(B, A). Hence, for our data v2.0, we perform a careful de-duplication, and update the data statistics in \cref{corr2cause:tab:stats}. See more version comparison details in \cref{corr2cause:appd:version}. Note that, due to the symmetry, the current version is a random sample half of the size of the original version, so the modeling results in the experiment section roughly hold.} Since the main purpose of the dataset is
to benchmark the performance of LLMs, we prioritize 
the test and development sets to have a comprehensive coverage over all sizes of graphs. Specifically, we iterate through the subset of our data for each $N$, and split it entirely for only the
test and development sets if the data is less than 1K, which is the case for $N=2$ and $3$. For the other subsets that are larger, we randomly sample up to 1K or 10\% of the data, whichever is smaller, to the test and development sets. We set the cap to be 1K in order to form a reasonable computation budget, since many LLMs are expensive to query in the inference mode. Aside from the test and valid sets, all the rest of the data goes into the training set.

\begin{table}[ht]
    \centering\small
    \begin{tabular}{lcccccccccccc}
    \toprule
& \multirow{2}{*}{Overall} & \multicolumn{5}{c}{Statistics by the Number of Nodes $N$} \\
\cline{3-7}
&  & $N=2$ & $N=3$ & $N=4$ & $N=5$ & $N=6$ \\ \midrule
\# Samples & 207,972 & 12 & 90 & 720 & 8,520 & 198,630  \\
\quad \# Test  & 1,162 & 6 & 48 & 72 & 514 & 522  \\
\quad \# Dev & 1,076 & 6 & 42 & 72 & 482 & 474  \\
\quad \# Train & 205,734 & 0 & 0 & 576 & 7,524 & 197,634  \\
\# Tokens/Premise & 424.11 & 31.5 & 52.0 & 104.0 & 212.61 & 434.54  \\
\# Tokens/Hypothesis & 10.83 & 10.83 & 10.83 & 10.83 & 10.83 & 10.83  \\
\% Positive Labels & 18.57 & 0.00 & 3.33 & 7.50 & 13.01 & 18.85  \\
Vocab Size   & 65 & 49 & 53 & 55 & 57 & 61 \\
\bottomrule
    \end{tabular}
    \caption{Statistics of our \ourdata dataset, and by subsets. We report the total number of samples (\# Samples); splits of the test (\# Test), developement (\# Dev) and training sets (\# Train); number of tokens per premise (\# Tokens/Premise) and hypothesis (\# Tokens/Hypothesis); percentage of the positive labels (\% Positive Labels), and vocabulary size by the number of unique tokens (Vocab Size). Note that the number of unique graphs and MECs are in \cref{corr2cause:tab:step1_causal_graph}.}
    \label{corr2cause:tab:stats}
\end{table}

\section{Experiments}

\subsection{Experimental Setup}

\newif\ifaccuracy
\accuracytrue

\ifaccuracy
\begin{table}[t]
    \centering \small
    \begin{tabular}{lccccc}
    \toprule
    & F1 & Precision & Recall
    & Accuracy\\ \midrule
    \multicolumn{5}{l}{\textit{\textbf{Random Baselines}}} \\
\quad    Always Majority & 0.0 & 0.0 & 0.0 & 84.77\\ 
\quad    Random (Proportional) & 13.5 & 12.53 & 14.62 & 71.46\\
\quad    Random (Uniform) & \ul{20.38} & 15.11 & 31.29 & 62.78\\
    \midrule
    \textit{\textbf{BERT-Based Models}} \\
\quad    BERT MNLI & 2.82 & 7.23 & 1.75 & 81.61\\
\quad    RoBERTa MNLI & 22.79 & 34.73 & 16.96 & 82.50\\
\quad    DeBERTa MNLI & 14.52 & 14.71 & 14.33 & 74.31\\ 
\quad    DistilBERT MNLI & 20.70 & 24.12 & 18.13 & 78.85\\ 
\quad    DistilBART MNLI & 26.74 & 15.92 & 83.63 & 30.23\\ 
\quad    BART %
    MNLI & \textbf{\ul{33.38}} & 31.59 & 35.38 & 78.50\\  \midrule
    \textit{\textbf{LLaMa-Based Models}} \\
\quad LLaMa-7B & 26.81 & 15.50 & 99.42 & 17.36\\ 
\quad Alpaca-7B & \ul{27.37} & 15.93 & 97.37 & 21.33\\
\midrule
    \textit{\textbf{GPT-Based Models}} \\
\quad GPT-3 Ada & 0.00 & 0.00 & 0.00 & 84.77\\ 
\quad GPT-3 Babbage & 27.45 & 15.96 & 97.95 & 21.15\\ 
\quad GPT-3 Curie & 26.43 & 15.23 & 100.00 & 15.23\\ 
\quad GPT-3 Davinci & 27.82 & 16.57 & 86.55 & 31.61\\ 
\quad GPT-3 Instruct (text-davinci-001) & 17.99 & 11.84 & 37.43 & 48.04\\ 
\quad GPT-3 Instruct (text-davinci-002) & 21.87 & 13.46 & 58.19 & 36.69\\ 
\quad GPT-3 Instruct (text-davinci-003) & 15.72 & 13.4 & 19.01 & 68.97\\
\quad     GPT-3.5 & 21.69 & 17.79 & 27.78 & 69.46\\ 
\quad GPT-4 & \ul{29.08} & 20.92 & 47.66 & 64.60\\ 
    \bottomrule
    \end{tabular}
    \caption{Overall performance. We report F1 (main metric), precision, recall and accuracy. For the main metric, F1 score, we use the \textbf{bold} font to highlight the overall best performance, and \ul{underline} to highlight the best performance within each category of models. 
    }
    \label{corr2cause:tab:res}
\end{table}

\else

\begin{table}[t]
    \centering \small
    \begin{tabular}{lcccc}
    \toprule
    & F1 & Precision & Recall
    \\ \midrule
    \multicolumn{4}{l}{\textit{\textbf{Random Baselines}}} \\
\quad    Always Majority & 0.0 & 0.0 & 0.0 \\ 
\quad    Random (Proportional) & 13.5 & 12.53 & 14.62 \\
\quad    Random (Uniform) & \ul{20.38} & 15.11 & 31.29 \\
    \midrule
    \textit{\textbf{BERT-Based Models}} \\
\quad    BERT MNLI & 2.82 & 7.23 & 1.75 \\
\quad    RoBERTa MNLI & 22.79 & 34.73 & 16.96 \\
\quad    DeBERTa MNLI & 14.52 & 14.71 & 14.33 \\ 
\quad    DistilBERT MNLI & 20.70 & 24.12 & 18.13 \\ 
\quad    DistilBART MNLI & 26.74 & 15.92 & 83.63 \\ 
\quad    BART %
    MNLI & \textbf{\ul{33.38}} & 31.59 & 35.38 \\  \midrule
    \textit{\textbf{LLaMa-Based Models}} \\
\quad LLaMa-7B & 26.81 & 15.50 & 99.42 \\ 
\quad Alpaca-7B & \ul{27.37} & 15.93 & 97.37 \\
\midrule
    \textit{\textbf{GPT-Based Models}} \\
\quad GPT-3 Ada & 0.00 & 0.00 & 0.00 \\ 
\quad GPT-3 Babbage & 27.45 & 15.96 & 97.95 \\ 
\quad GPT-3 Curie & 26.43 & 15.23 & 100.00\\ 
\quad GPT-3 Davinci & 27.82 & 16.57 & 86.55 \\ 
\quad GPT-3 Instruct (text-davinci-001) & 17.99 & 11.84 & 37.43 \\ 
\quad GPT-3 Instruct (text-davinci-002) & 21.87 & 13.46 & 58.19 \\ 
\quad GPT-3 Instruct (text-davinci-003) & 15.72 & 13.4 & 19.01\\
\quad     GPT-3.5 & 21.69 & 17.79 & 27.78 \\ 
\quad GPT-4 & \ul{29.08} & 20.92 & 47.66\\ 
    \bottomrule
    \end{tabular}
    \caption{Overall performance. We report F1 (main metric), precision, recall and accuracy. For the main metric, F1 score, we use the \textbf{bold} font to highlight the overall best performance, and \ul{underline} to highlight the best performance within each category of models. 
    }
    \label{corr2cause:tab:res}
\end{table}

\fi

We set up a diverse list of LLMs for the experiments on our \ourdata dataset. 
To \textit{test existing LLMs}, we first include six commonly used BERT-based NLI models in the transformers library \citep{wolf-etal-2020-transformers}: 
BERT \citep{devlin-etal-2019-bert}, RoBERTa~\citep{liu2019roberta}, BART \citep{lewis-etal-2020-bart}, DeBERTa~\citep{he2021deberta}, DistilBERT~\citep{sanh2019distilbert}, and DistilBART \citep{shleifer2020pretrained}. Apart from these BERT-based NLI models, we also evaluate the general-purpose autoregressive LLMs based on GPT \citep{radford2019language}: GPT-3 Ada, Babbage, Curie, Davinci \citep{brown2020gpt3}; its instruction-tuned versions \citep{ouyang2022instructGPT}, text-davinci-001, text-davinci-002, and text-davinci-003; and GPT-3.5 (i.e., ChatGPT), and the latest GPT-4 \citep{openai2023gpt4} by April 2023, using the OpenAI API
({\small \url{https://openai.com/api/}})
with temperature 0. We also evaluate the recent, more efficient models, LLaMa \citep{touvron2023llama} and Alpaca \citep{taori2023alpaca}.

When inspecting the behavior of \textit{finetuned models}, 
we adopt a large set of models, including GPT-based models (GPT-3 Ada, Babbage, Curie, and Davinci) using the OpenAI finetuning API for classification at {\small \url{https://platform.openai.com/docs/guides/fine-tuning}},
open-sourced decoder-only models (GPT2, GPT2-Large, GPT2-XL, LLaMA-7B, and LLaMA2-7B), BERT-based models from scratch (BERT-Base, BERT-Large, RoBERTa-Base, and RoBERTa-Large), and BERT-Based NLI models (BERT-Base MNLI, BERT-Large MNLI, RoBERTa-Base MNLI, and RoBERTa-Large MNLI) using the transformers library \citep{wolf-etal-2020-transformers}. See training details
in \cref{corr2cause:appd:implementation}.

For the \textit{random baselines}, we provide ``always majority'' to predict the majority class 100\% of the time, ``random (uniform)'' to uniformly sample a label (i.e., 50\% for each), and ``random (proportional)'' to sample a label from a Bernouli distribution proportional to the development set label distribution.

\subsection{The \ourdata Skill in Existing LLMs}\label{corr2cause:sec:0shot}

We show the performance of seventeen LLMs in \cref{corr2cause:tab:res}. We can see that pure causal inference is a very challenging task across all existing LLMs.
Among all the LLMs, the best performance is 33.38\% F1 by BART MNLI, which is even higher than the latest GPT-based model, GPT-4.
Notably, many models are worse than random guess, which means that they totally fail at this pure causal inference task.
The observation still holds for few-shot chain-of-thought prompts tested in \cref{corr2cause:appd:optimization}.

\subsection{Finetuned Performance}
Next, we address the question: \textit{Can we re-purpose LLMs to learn this task?}
The experimental results in \cref{corr2cause:tab:finetune} of 17 models finetuned on our \ourdata seem very strong at first sight. Most models see a substantial increase, among which the finetuned BERT-based NLI models demonstrate the strongest performance.
The best-performing one, RoBERTa-Large MNLI, achieves 94.74\% F1 score on this task, as well as very high precision, recall and accuracy scores.

\begin{table}[t]
    \centering \small
    \begin{subtable}[t]{0.66\textwidth}
        \centering
    \begin{tabular}{lccccc}
    \toprule
    & F1 & Precison & Recall
    & Accuracy  \\ \midrule
    \multicolumn{5}{l}{\textit{\textbf{Finetuned GPT-Based Models Using OpenAI API}}} \\

GPT-3 Ada & 79.85 & 70.47 & 92.11 & 92.92 \\ 
GPT-3 Babbage & 78.19 & 69.98 & 88.60 & 92.48 \\ 
GPT-3 Curie & 81.23 & 75.00 & 88.60 & 93.77 \\ 
GPT-3 Davinci & \ul{85.52} & 80.26 & 91.52 & 95.28 \\ \midrule
    \multicolumn{5}{l}{\textit{\textbf{Finetuned Open-Sourced Decoder-Only Models}}} \\
GPT2 & 89.18 & 88.03 & 90.35 & 96.66 \\
GPT2-Large & 94.29 & 92.18 & 96.49 & 98.22 \\
GPT2-XL & \ul{94.30} & 91.94 & 96.78 & 98.22 \\
LLaMA-7B &  91.98 & 88.62 & 95.61 & 97.46 \\
LLaMA2-7B & 92.92 & 90.11 & 95.91 & 97.77 \\ \midrule
    \multicolumn{5}{l}{\textit{\textbf{Finetuned BERT-Based Models}}} \\
    BERT-Base & 69.29 & 54.42 & 95.32 & 87.13 \\ 
BERT-Large & 85.26 & 77.51 & 94.74 & 95.01 \\ 
RoBERTa-Base & 87.60 & 78.47 & 99.12 & 95.73 \\ 
RoBERTa-Large & \ul{89.10} & 82.54 & 96.78 & 96.39 \\
    
\midrule
    \multicolumn{5}{l}{\textit{\textbf{Finetuned BERT-Based NLI Models}}} \\
    BERT-Base MNLI & 89.88 & 85.49 & 94.74 & 86.51 \\ 
    BERT-Large MNLI & 90.19 & 84.44 & 96.78 & 96.79 \\
    RoBERTa-Base MNLI & 94.27 & 90.35 & 98.54 & 98.17 \\
    RoBERTa-Large MNLI & \textbf{\ul{94.74}} & 92.24 & 97.37 & 98.35 \\

    \bottomrule
    \end{tabular}
    \caption{Performance of finetuned models on the original test set.
    }
    \label{corr2cause:tab:finetune}
\end{subtable}%
    \hfill
    \begin{subtable}[t]{0.3\textwidth}
        \centering
    \begin{tabular}{ccc}
    \toprule
    F1 (Paraph.) & F1 (Var. Ref.)  \\ \midrule
 \\
61.73 & 41.57 \\ 
62.34 & 43.28 \\ 
64.93 & 45.32 \\ 
65.01 & 46.96 \\ \midrule
\\
56.76 &   31.70 \\
55.95 &     31.99 \\
60.32 &    43.95 \\
56.41 &  53.92 \\
52.24 &     49.47 \\ \midrule
\\
61.13 & 35.20 \\
63.64 & 38.54 \\
65.58 & 53.12 \\
65.05 & 60.20 \\
    
\midrule
\\
65.56 & 31.50 \\
67.24 & 52.04 \\
57.42 & 62.83 \\
55.45 & 67.87 \\
    \bottomrule
    \end{tabular}
    \caption{F1 scores of finetuned models on the perturbed test sets by paraphrasing (Paraph.) and variable refactorization (Var. Ref.).
    }
    \label{corr2cause:tab:goodhart}
\end{subtable}%
\caption{Performance of finetuned models on the original test set and perturbed test sets.}
\end{table}

\begin{table}[t]
    \centering \small
    
    \setlength{\tabcolsep}{5pt}
    \begin{subtable}[t]{0.6\textwidth}
    \begin{tabular}{lccccc}
\toprule
Relation Type & F1 & Precision & Recall & Accuracy \\ \midrule
Is-Parent &  96.18 &  95.45 &  96.92 & 98.67 \\
Is-Ancestor &  93.94 &  93.94 &  93.94 & 98.93 \\
Is-Child &  95.73 &  94.92 &  96.56 & 98.67 \\
Is-Descendant &  96.55 &  93.33 &  100 & 99.47 \\
Has-Collider &  92.19 &  87.41 &  97.52 &  94.64 \\
Has-Confounder &  98.67 &  97.37 &  100 & 99.73 \\

\bottomrule
    \end{tabular}
    \caption{Fine-grained performance of RoBERTa-Large by causal relation type on the original test set.
    }
    \label{corr2cause:tab:finetune_by_class}
\end{subtable}%
    \hfill
    \begin{subtable}[t]{0.38\textwidth}
        \centering
        
    \setlength{\tabcolsep}{5pt}
    \begin{tabular}{cccccc}
    \toprule

F1 & Precision & Recall & Accuracy \\ \midrule

74.80 & 79.31 & 70.77 & 91.73 \\
45.45 & 90.91 & 30.30 & 93.60 \\
73.39 & 78.43 & 68.97 & 92.27 \\
29.41 & 83.33 & 17.86 & 93.60 \\
70.70 & 75.00 & 66.90 & 82.04 \\
70.42 & 73.53 & 67.57 & 94.37 \\

    \bottomrule
    \end{tabular}
    \caption{Its fine-grained performance by relation type after variable refactorization.
    }
    \label{corr2cause:tab:perturb_by_class}
\end{subtable}%
\caption{Fine-grained analysis of the best-performing model, RoBERTa-Large MNLI.}
\end{table}

\subsection{Fine-Grained Performance by Causal Relation}
In addition to the overall results mentioned above, we conduct a fine-grained analysis to check the performance of the strongest finetuned model, RoBERTa-Large MNLI, by our six causal relation types. As in \cref{corr2cause:tab:finetune_by_class}, the model is very good at judging relations such as Is-Parent, Is-Descendant and Has-Confounder, all with more than 96\% F1 scores, whereas it is several points weaker on the Has-Collider relations. This could be due to that the collider relation is the most special type, requiring identification of the V-structure based on both the unconditional independence based on the two variables only and correlations whenever conditioned on a common descendant. We also conduct error analysis for non-finetuned models in \cref{corr2cause:appd:error}.

\subsection{Robustness Analysis
}

Looking at the very high performance of the finetuned models, we raise the next question: \textit{Did the models really robustly learn the causal inference skills?} 

\paragraph{Two Robustness Tests}
We design two simple robustness tests: (1) paraphrasing, and (2) variable refactorization. For (1) paraphrasing, we simply paraphrase the hypothesis by changing the text template for each causal relation to some semantically-equivalent alternatives in \cref{corr2cause:appd:templates}. For (2) variable refactorization, we reverse the alphabet of the variable names, namely flipping A, B, C, to Z, Y, X and so on.
The inspiration behind the two robustness tests comes from the spurious correlation analysis described in \cref{corr2cause:appd:spurious}.

Specifically, we adopt the common setup of text adversarial attack \citep{morris-etal-2020-textattack,jin2020bert} to preserve the training set and keep the same saved models, but run the inference on the perturbed test set.
In this way, we separate the possibilities of the models only overfitting on the training data vs. mastering the reasoning skills.

\paragraph{Results after Perturbation}
We can see from \cref{corr2cause:tab:goodhart} that all the models drop drastically, by up to 39.29 on the paraphrased test set, and up to 62.30 after variable refactorization. 
The best-performing model, RoBERTa-Large MNLI, is especially sensitive towards paraphrasing, demonstrating the most drop among all models; however, it is the most robust against the variable refactorization, maintaining a high F1 score of 67.87. We conduct fine-grained  analysis for RoBERTa-Large MNLI under perturbation in \cref{corr2cause:tab:perturb_by_class}. We can see the the main source of the performance drop of the model comes from the two classes, Is-Ancestor (decreasing to 45.45\%) and Is-Descendant (decreasing to 29.41\%), while the other classes stay relatively robust, keeping their F1 scores over 70\%.

From this analysis, we make the following suggestions to future studies testing this \ourdata skill of LLMs. First, it is safe to use it as a test set to benchmark existing LLMs' performance, since the data we generate is out-of-distribution from the training data of the current LLMs. Then, when testing finetuned models, it is very important to accompany adversarial attack together with the i.i.d. test set. We open-source our perturbed test sets for future work to test the generalizability skill.

\ifperfect
\subsection{Effect of In-Context Learning}
\subsection{Adding Signals to the context}
\subsection{Example Reasoning}

\fi
\subsection{Extension to Natural Stories}

We envision our \ourdata dataset to be a foundation for future extensions to various settings, such as instantiating the variables with actual phenomena and situating the story in a more natural setting.
For example, the \textit{correlation does not imply causation} rule can be instantiated with the ice cream sales and swimming pool attendance as the two variables, and argue that
ice cream sales does not necessarily affect swimming pool attendance, because their correlation could be due to a third variable, such as hot weather. We provide a case study for how to instantiate the symbolic expressions in our dataset to more natural stories, and find that LLMs such as GPT-4 can generate realistic, daily life stories that has foreseeably broad applications. See more details
in \cref{corr2cause:appd:story}.

\section{Related Work}
\paragraph{Existing Causal Reasoning Tasks}
A large body of existing research of causal reasoning in NLP focuses on leveraging empirical knowledge to do tasks such as inferring the cause and effect of why an agent perform certain tasks \citep{sap2019atomic}, the motivation and emotional reaction in a social context \citep{Sap2019SocialIQA}, how people achieve a given goal with a set of concrete steps \citep{zhang-etal-2020-reasoning}, the development of a story given a different beginning \citep{qin-etal-2019-counterfactual}, and how in general LLMs serve as a knowledge base of cause and effect \citep{willig2023probing,kiciman2023causal}.
In contrast, our \ourdata task focuses on the pure causal inference skill of models, which is a knowledge-dependent reasoning skill based on formally correct rules from causal inference.

\paragraph{Existing Logical and Inference Tasks}
Another related area of literature is logical and inference tasks, of which a well-established one is natural language inference (NLI), to identify the semantic relationship between a pair of
sentences \citep{maccartney-manning-2008-modeling,bowman-etal-2015-large}. NLI datasets mainly focus on the set and paraphrase relations. For example, ``a group of boys are playing football'' can entail ``some guys are playing football,'' where ``boys'' are a sub-concept of ``guys,'' and ``a group of'' and ``some'' are paraphrases. 
Recently, there have been increasing efforts to extend the inference task to various logical inference skills such as deductive logic and propaganda techniques \citep{jin-etal-2022-logical,alhindi-etal-2022-multitask}. Our \ourdata dataset is the first dataset testing the correlation-to-causation inference skill, which is unique of its type.

\section{Conclusion}
In this work, we introduced a novel task, \ourdata, to infer causation from correlation, and collected a large-scale dataset of over 200K samples.
We evaluated an extensive list of LLMs on this new task, and showed that off-the-shelf LLMs perform poorly on this task. We also show that it is possible to re-purpose LLMs on this task by finetuning, but future work needs to be aware of the out-of-distribution generalization problem. To avoid the Goodhart's law, we recommend using this dataset to benchmark the pure causal inference skills for LLMs that have not seen this dataset. 
Given the limited reasoning abilities of current LLMs, and the difficulty of separating actual reasoning from training-corpus-derived knowledge, it is imperative that our community focus on work aiming to accurately disentangle and measure both abilities. We believe the present work is a first such step.

\section*{Limitations and Future Work}
We identify several limitations of this work and open future directions:
First, in the context of this work, we limit the causal graphs to two to six nodes, but future work can feel free to explore larger graphs.
Another aspect is that we do not assume hidden confounders in this inference problem, so we welcome future work to generate an even more challenging dataset to infer the existence of hidden confounders, analogous to the causal discovery algorithm of fast causal inference (FCI)  \citep{spirtes2001causation}. And also in general, explorations of other causal discovery algorithms are welcomed too.
Finally, a lot of motivation behind proposing this task is inspired by the problem of invalid reasoning patterns in our daily reasoning \citep{jin-etal-2022-logical}, which could fertilize the ground for more pervasive spread of fake news. We believe false causal inference is a prevalent type of fallacious beliefs, and welcome future work to connect the idea of this benchmark to more real-world false beliefs based on confusing correlation with causation.

\mainchapter{CLadder: Assessing Causal Reasoning in LLMs}\label{ch:cladder}
\renewcommand{\ourdata}{{\modelfont {CLadder}}\xspace}
\newcommand{\ourmodel}{{\textsc {CausalCoT}}\xspace}

\newcommand{\stepone}{\ding{172}\xspace}
\newcommand{\steptwo}{\ding{173}\xspace}
\newcommand{\stepthree}{\ding{174}\xspace}
\newcommand{\stepfour}{\ding{175}\xspace}
\newcommand{\stepfive}{\ding{176}\xspace}
\newcommand{\stepsix}{\ding{177}\xspace}

Apart from causal discovery covered in the previous chapter, we further explore causal effect reasoning in natural language, inspired by the {\em ``causal inference engine''} postulated by \citet{pearl2018book}. To this end, we compose a large dataset, \ourdata, with 10K samples: based on a collection of causal graphs and queries (associational, interventional, and counterfactual), we obtain symbolic questions and ground-truth answers, through an oracle causal inference engine. These are then translated into natural language.
We evaluate multiple LLMs on our dataset, and we introduce and evaluate a bespoke chain-of-thought prompting strategy, \ourmodel.
We show that our task is highly challenging for LLMs, and we conduct an in-depth analysis to gain deeper insights into the causal reasoning abilities of LLMs.
Our data is open-sourced at \url{https://huggingface.co/datasets/causalNLP/cladder}, and our code can be found at \url{https://github.com/causalNLP/cladder}.

\section{Introduction}
\label{cladder:sec:intro}
\begin{quote}
    \textit{Once we really understand the logic behind causal thinking, we could emulate it on modern computers and create an ``artificial scientist''.}
    
    \hspace*{\fill}--- \textit{\citeauthor{pearl2018book} \citeyearpar{pearl2018book}}
\end{quote}

Causal reasoning is believed to be one of the hallmarks of human intelligence~\citep{penn2007causal, harari2014sapiens}.
The ability to draw causal inferences from available information is crucial for scientific understanding and rational decision-making: 
for example, knowing whether smoking causes cancer might enable consumers to make a more informed decision~\citep{doll1950smoking,doll1954mortality}; assessing the causal effect of a vaccine is essential for effective policy-making during a pandemic%
~\citep{plotkin2005vaccines,
de2013largest,
whitney2014benefits, kekic2023evaluating}; and  understanding the  interplay behind family background, education and income helps devise effective education policies~\citep{card1999causal,chetty2011does,heckman2006earnings,psacharopoulos2004returns}.

\begin{figure}[t]
    \centering
    \includegraphics[width=\textwidth]{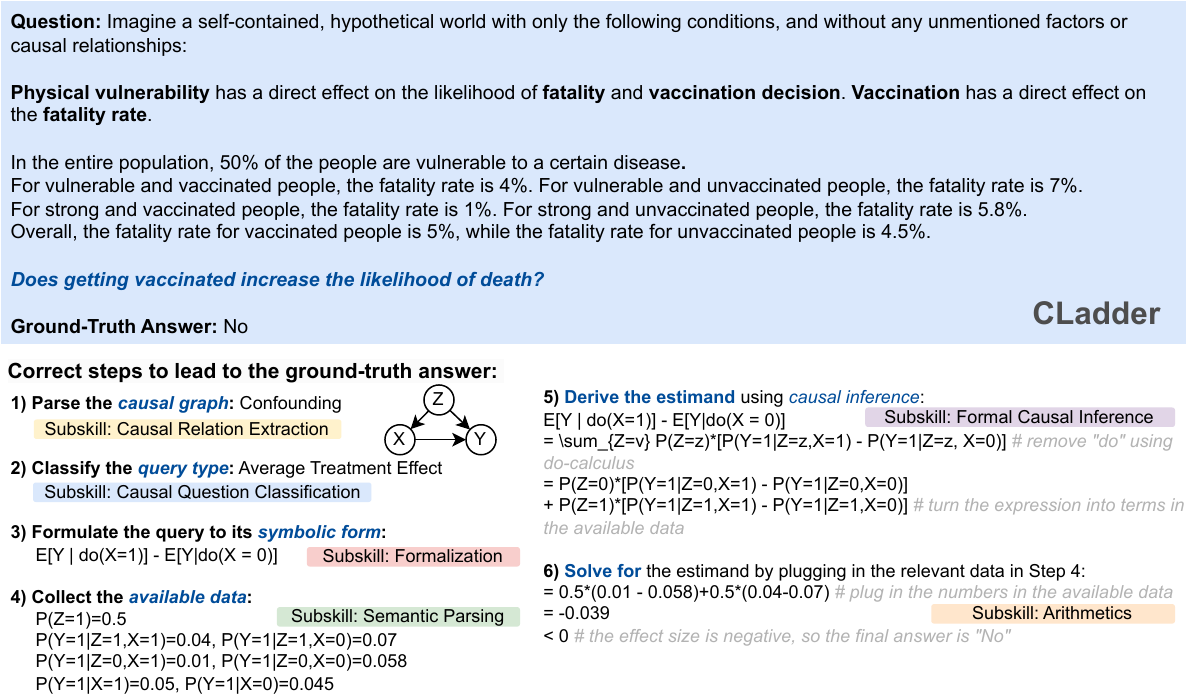}
    \caption{Example question in our \ourdata dataset featuring an instance of {\em Simpson's paradox}~\citep{pearl2022comment}. We generate the following (symbolic) triple: (i) the causal query; (ii) the ground-truth answer, derived through a {\em causal inference engine}~\citep{pearl2018book}; and (iii) a step-by-step explanation. 
    We then \textit{verbalize} these questions by turning them into stories, inspired by examples from the causality literature, which can be expressed in natural language.
    }
    \label{cladder:fig:question_example}
\end{figure}

Our opening quote therefore mirrors the aspirations of many scientists in artificial intelligence and causal inference: to construct a machine capable of performing sound causal reasoning, and able to answer causal questions at scale and with ease.
Recent advances in large language models (LLMs) have brought about a paradigm shift in natural language processing (NLP) and artificial intelligence~\citep[][\textit{inter alia}]{radford2019language,devlin-etal-2019-bert,brown2020gpt3,zhang2022opt,openai2023gpt4,ignat-etal-2024-has}. 
These transformative developments raise the question of whether these machines are already capable of causal reasoning: 
    \textit{Do LLMs understand causality?} 

Many previous works addressed the above question by focusing on {\em commonsense} causality~\citep{zevcevic2023causal,zhang2023understanding,ho2022wikiwhy}, inspired by the literature that explores LLMs as {\em knowledge bases}~\citep{petroni-etal-2019-language,shin-etal-2020-autoprompt,jiang-etal-2020-know} (we refer to this line of work as \textit{causality as knowledge}). 
This involves assessing the alignment between commonsense knowledge about causal relationships in humans and LLMs. This line of work generally does not focus on evaluating how well models are capable of {\em causal reasoning}. For example, it may be difficult to rule out the possibility that LLMs perform potentially unreliable {\em amortized causal inference}, answering causal questions by a simple repetition of verbal patterns present in the texts composing their training data:\footnote{which may itself contain instances of fallacious causal reasoning.}%
\footnote{The extent to which this would imply an inaptitude of LLMs for causal reasoning has been questioned~\citep{Huszár_2023}.}
in other words, LLMs may just be {\em ``causal parrots''}~\citep{zevcevic2023causal}.%

In this work, we %
introduce a way to test the {\em formal causal reasoning in LLMs}. %
To this end, we introduce the \ourdata dataset. The specificity of \ourdata %
is that 
causal questions posed in natural language are {\em grounded in symbolic questions and ground truth answers}: the latter are derived through an oracle \textit{causal inference engine (CI engine)}~\citep{pearl2018book}, which abides by the rules of the causal inference approach described by~\citet{pearl2009causality}, based on graphical models and structural causal models (SCMs)~\citep{pearl1995causal,spirtes2001causation,pearl2009causality,glymour2016causal,peters2017elements}. 
We compose more than 10,000 causal questions that cover a variety of causal queries across the three rungs of the {\em Ladder of Causation}~\citep{pearl2018book, bareinboim2022pearl}---i.e., {\em associational (Rung 1)}, {\em interventional (Rung 2)}, and {\em counterfactual (Rung 3)}. We consider several causal graphs, giving rise to scenarios which require different causal inference abilities. 
Additionally, we generate ground-truth explanations with step-by-step reasoning for more in-depth analysis of LLM behavior.
Our symbolic questions and answers are then {\em verbalized}, by turning them into stories which can be expressed in natural language. To probe whether LLMs employ amortized causal inference%
, we construct stories with commonsensical, as well as anti-commonsensical and with nonsensical causal relations: in these latter cases, amortized causal inference is expected to fail, whereas formal causal reasoning would still yield the correct answer.
An example question
from \ourdata is shown in~\cref{cladder:fig:question_example}%
.

Exploiting \ourdata, w%
e also introduce a method to elicit sound causal reasoning in LLMs and help them solve challenging causality questions. %
Specifically, we develop \textbf{\ourmodel}, a chain-of-thought prompting strategy~\citep{wei2022chain} inspired by the CI engine, which prompts the LLM to extract the causal graph, causal query, and available ``data'' (e.g., conditional or interventional {\em do}-probabilities~\citep{goldszmidt1992rank}) from the question, formalize them precisely, and perform correct causal inferences.
Our experiments indicate that \ourmodel %
achieves an accuracy of 70.40\%, which substantially improves the performance of vanilla GPT-4 by 8.37 points
on \ourdata.

We summarize the {\em main contributions} of our work:
\begin{enumerate}
[itemsep=0.5em,topsep=0em
]
    \item \looseness=-1 In contrast to most other works on causality in LLMs, 
    focusing on {\em commonsense causal knowledge},
    our goal is to assess the LLMs' ability to perform {\em formal causal reasoning} (briefly reviewed in~\cref{cladder:sec:preliminaries}). %
    \item We introduce \ourdata~(\cref{cladder:sec:cladder}), a dataset containing more than 10K causal questions, spanning all three rungs of the ladder of causation, several causal graphs, and various stories for verbalization.
    \item We develop \ourmodel~(\cref{cladder:sec:causal_cot}), a chain-of-thought prompting strategy to elicit formal causal reasoning in LLMs, inspired by the {\em causal inference engine}.
    \item We perform extensive experiments on eight LLMs~(\cref{cladder:sec:experiments}), analyze fine-grained errors to showcase the limitations of LLMs in formal causal reasoning, and suggest directions for future research.
\end{enumerate}

\section{Preliminaries on Causal Inference}\label{cladder:sec:preliminaries}

\looseness=-1 Our dataset design takes inspiration from the {\em Causal Inference Engine} as postulated by~\citet{pearl2018book}, see also~\citep{pearl1995causal}. %
We begin with a brief overview of the causality framework by~\citet{pearl2009causality}.\footnote{We refer to~\citep{pearl2016causal, bareinboim2022pearl} for a comprehensive introduction. See also~\cref{cladder:app:more_preliminaries} for further details.}
This framework was largely developed within the field of artificial intelligence, and therefore puts particular emphasis on {\em algorithmic} aspects of causal reasoning (e.g.,~\citep{pearl2011algorithmization})---which makes it particularly suited for our work, where we want to algorithmically generate ground truth answers to causal queries, without having to appeal to common sense to assess the correctness of an answer. 

\subsection{The Ladder of Causation}
The {\em Ladder of Causation}, introduced by \citet{pearl2018book}, is a proposed taxonomy, and hierarchy, of causal inference tasks~\citep{bareinboim2022pearl}. It consists of three distinct rungs. %

\paragraph{Rung 1 ({\em ``seeing''}).} \looseness=-1 This describes statistical associations %
({\em ``How often do I take an aspirin when I have a headache?''}). Rung 1 %
deals with
statistical dependences among random variables, and involves
probabilistic reasoning 
about
joint and conditional distributions, ${P(X=x, Y=y)}$ and $P{(Y=y|X=x)}$, which can be formalised through {\em Bayesian Networks}~\citep{pearl1988probabilistic, cowell2007probabilistic} representing a set of variables and their conditional dependencies via a directed acyclic graph (DAG).
\paragraph{Rung 2 ({\em ``doing''}).} This enables us to formalize the concept of actively intervening in the world, and modifying it toward some end ({\em ``If I take an aspirin now, will my headache subside?''}).
Interventions can be formalized using the {\em do-operator}~\citep{goldszmidt1992rank} and {\em Causal Bayesian Networks}~\citep{pearl2009causality} to represent, for example, the distribution over $Y$ when intervening on $X$ to set its value to $x$ as $P(Y=y|\mathrm{do}(X=x))$.

\paragraph{Rung 3 ({\em ``imagining''}).}%
\looseness=-1 %
This rung deals with counterfactual reasoning, i.e., reasoning about alternative scenarios in which the world could have been different, possibly even contradicting the factual state
({\em ``Would my headache have subsided, if I had taken an aspirin?''}).
Counterfactual probabilities can be written as $P(Y_{x}=y)$, 
representing the probability that ``$Y$ would be $y$, had $X$ been $x$''.
Reasoning about Rung 3 quantities requires the introduction of {\em Structural Causal Models (SCMs)}~\citep{pearl2009causality}.
SCMs are especially powerful as they enable any quantity in Rungs 1, 2, and 3 to be formulated precisely~\citep{bareinboim2022pearl}.

\subsection{Causal Inference}
\paragraph{Identification.}
Causal inference is especially difficult since we typically only have measurements from {\em lower} rungs, but want to reason about {\em higher} ones.
A crucial question is then %
under what conditions are such inferences possible, i.e., what assumptions and measurements are required to unambiguously answer a causal query of interest: this is the question of {\em identification}.
As argued in~\citep{bareinboim2022pearl}, {\em ``it is generically impossible to draw higher-layer inferences using only lower-layer information''}. One may
be able to draw inferences at a higher layer given a combination of partial knowledge of the %
underlying SCM, in the form of a causal graph, and data at lower layers. The graphical structure therefore plays a crucial role in bridging the rungs of the Ladder of Causation, and many prior works have been dedicated to exploiting properties of the graph to transform higher-rung queries into expressions which can be estimated based on lower-rung quantities~\citep{huang2006pearl, shpitser2006identification, pearl2022external}.

\paragraph{Causal Inference Engine.} \looseness=-1 An overarching objective of this research is the construction of a {\em Causal Inference Engine (CI Engine)}~\citep{pearl1995causal, pearl2018book, hunermund2019causal},
which takes as input a query, a graph, and some available data (typically from lower rungs than the query); and outputs whether a solution exists, and, if so, %
an equivalent expression of the query which is estimable from the available data. 
While some previous works refer to the CI engine in the context of Rung 2 queries, where it corresponds to the {\em do}-calculus~\citep{shpitser2006identification, huang2006pearl}, here we refer to it in a more general sense, encompassing all three rungs.

\section{Composing the \ourdata Dataset}
\label{cladder:sec:cladder}
\begin{figure}[t]
    \centering
    \includegraphics[width=0.98\textwidth]{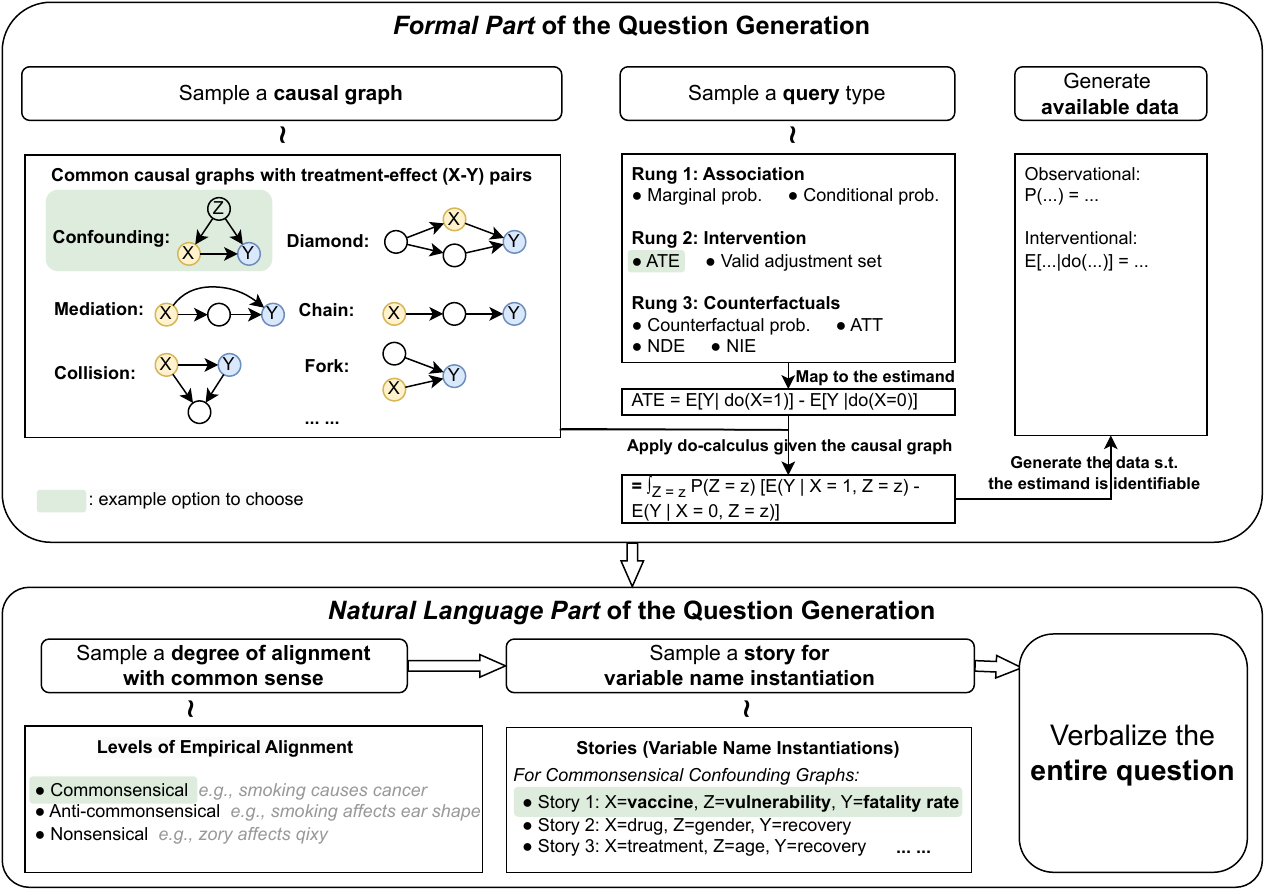}
    \caption{The data-generating process of the \ourdata dataset. The upper part of the figure describes the {\em formal part} of the question  generation, which samples inputs for the CI Engine and derives a ground truth answer. The bottom part describes the {\em natural language part} of the question generation---i.e., its verbalization, based on multiple stories and different degrees of alignment with commonsense knowledge.}
    \label{cladder:fig:generator}
\end{figure}

\paragraph{Task Formulation.} 
\looseness=-1
Like in the example of~\cref{cladder:fig:question_example}, our dataset $\mathcal{D}:= \{(\bm{q}_i, \bm{a}_i, \bm{e}_i)\}_{i=1}^N$ consists of $N$ triples, each containing a question $\bm{q}_i$, binary answer $a_i \in \{\text{Yes}, \text{No}\}$, and an explanation~$\bm{e}_i$.
Our main task is to test the accuracy of the prediction function $f: \bm{q} \mapsto a$, i.e., a LLM which maps a natural language causal question to an answer.
Apart from directly evaluating the answer, we also compose the ground-truth explanations $\bm{e}$ to evaluate the reasoning steps of LLMs. 

\paragraph{Design Principles.}
\looseness=-1
In the composition of our dataset, we adhere to the following
 design principles. 
First, we ensure broad coverage of all rungs of the ladder of causation. %
Second, 
we avoid settings that involve continuous variables and use binary variables instead: this is partly due to the large availability of identifiability results for binary and categorical variables, and partly because queries involving binary variables lend themselves to more natural-sounding verbalization.
Moreover, since LLMs struggle with calculation-heavy tasks~\citep{hendrycks2021measuring,stolfo-etal-2023-causal},
and we are chiefly interested in causal reasoning abilities, %
we focus on graphs with few (three to four) variables, in various common configurations, to produce %
questions which are identifiable from the outset.
Lastly, we carefully design a rich set of templates to translate the abstract formulas into grammatically correct and
 natural-sounding, fluent prompts.

\paragraph{Overall Pipeline.}

\looseness=-1
The generation pipeline for \ourdata, depicted in~\cref{cladder:fig:generator}, consists of two parts: 
\begin{enumerate}[itemsep=0em,topsep=0em,leftmargin=1.25em]
    \item In the \textit{Formal Part} (which we illustrate in~\cref{cladder:sec:sampling_graph}), we specify all the required inputs (query, model, data) and the ground truth answer generated by the
    CI Engine.%
\item In the \textit{Natural Language Part} (in~\cref{cladder:sec:nat_lang}),
we verbalize the formal queries and %
specification of the causal model and data %
by associating them to a story or narrative,
using a rich set of templates.
\end{enumerate}

\subsection{
Formal Part of the Question Formulation
}\label{cladder:sec:sampling_graph}

The first step of our data generating process is to construct a set of inputs to the CI Engine such that 
{\em by design}
there exists a well-defined ground truth answer: i.e., we
construct triples of causal queries, graphs, and data such that the query can be unambiguously answered based on the available data (ensuring {\em identifiability} by construction).\footnote{We use the term ``data'' to denote numerical values of conditional or {\em do}-probabilities, and not as collections of data samples. This is in line with how the term is used in other descriptions of the CI Engine~\citep{pearl2018book, hunermund2019causal}.}
The ground truth causal models, which specify all quantities which are considered measurable in our questions, %
are causal Bayesian networks (CBNs), where each causal mechanism (i.e., conditional probability of a variable given its parents in the factorization according to the causal graph $G$) corresponds to a Bernoulli distribution.
We compile a selection of graphs $G$ based on examples drawn from multiple sources from the literature~\citep{spirtes2001causation,peters2017elements,pearl2009causality,pearl2018book}, where suitable graph structures are used to illustrate toy problems in causal inference.
The complete list of structures we consider can be found in~\cref{cladder:appd:causal_graphs}; the complete list of sources in~\cref{cladder:appd:books}.

\paragraph{Selecting Query Types.}
We again draw from the causal inference literature to collect common query types in each rung. As illustrated in the {\em ``Sample a query type''} box in \cref{cladder:fig:generator}, for Rung 1, we can ask about probability distributions such as marginal probabilities and conditional probabilities. For Rung 2 questions, we can enquire \textit{average treatment effects (ATE)} ({\em ``how will $Y$ change if $X$ changes from $x$ to $x'$?''}), or what constitutes a valid adjustment set that can block all backdoor spurious correlations between $X$ and $Y$%
. Lastly, for Rung 3, we include \textit{counterfactuals} ({\em ``what would happen to $Y$ had $X$ been $x'$ instead of $x$?''}), \textit{average treatment effect on the treated (ATT)} ({\em ``for the subpopulation whose $X$ changed from $x$ to $x'$, how does their $Y$ change on average?''}), \textit{natural direct effect (NDE)} ({\em ``what is the direct effect of $X$ in $Y$, but not through the mediators?''}), and \textit{natural indirect effect (NIE)} ({\em ``what is the effect from $X$ to $Y$ through the mediators?''}).

\paragraph{Applying the Causal Inference Engine for the Ground-truth answer.}

By construction, the causal processes we define encapsulates all necessary information to make the causal quantities of the query types identifiable. 
This allows us to apply the rules of causal inference to obtain an estimand for each causal graph and query type, and evaluate the estimand to get a ground truth answer. 
The Rung 2 queries simplify to Rung 1 terms using the rules of {\em do}-calculus \citep{pearl1995causal}, and, for the Rung 3 queries, we apply methods of counterfactual causal inference~\citep{pearl2009causality} (with details in \cref{cladder:appd:ci_engine}).
The estimand also specifies exactly which terms are necessary to include in the prompt as {\em ``available data''} in order to ensure that enough information is provided to answer the question correctly (i.e., for identifiability), provided the correct causal reasoning is applied.
Our entire code base of the data generation process can be found at our GitHub repository,~\href{https://github.com/causalNLP/cladder}{\texttt{https://github.com/causalNLP/cladder}}%
.

\subsection{Natural Language Part of the Question Formulation}
\label{cladder:sec:nat_lang}
\looseness=-1
While~\cref{cladder:sec:sampling_graph} describes a way to generate the ground-truth causal model, query and answers, computed through a causal inference engine, real-world causal reasoning problems are expressed in natural language rather than symbolic expressions. The next part of the data generation pipeline therefore focuses on the verbalization of all these components with a plausible narrative in natural language.

\paragraph{Generating the Stories.}
\looseness=-1
For each causal graph, we collect a set of two to five \textit{stories} which consist of a list of variable names for each node in the graph. The stories are primarily selected from examples in commonly cited causal inference books and papers (see \cref{cladder:appd:books}), which ensures that the stories and corresponding causal graph structures adhere to empirical common sense 
(e.g., the drug-gender-recovery example of \citet{pearl2018book}).
However, it is very likely that at least some of the stories appear in the training data of many LLMs. Therefore, we also generate various {\em anti-common sense} and {\em nonsensical} variants of the stories, meant to isolate the effects of memorization.
For the anti-commonsensical stories, we randomly do one of the actions: (1) replace the effect variable $Y$ with an unusual attribute,  that would not be an effect variable in any of the stories (e.g., ``ear shape''); 
or (2) create an irrelevant treatment variable $X$ that does not play a causal role in any of our commonsensical stories, such as ``playing card games'' (see \cref{cladder:appd:anti}). 
For the nonsensical variants, we invent artificial words as variable names such as ``zory'' and ``qixy'' (see~\cref{cladder:appd:stories}).
.

\paragraph{Verbalizing the Prompts.} 

The verbalization procedure applies the mapping of symbolic variables to semantic concepts to form a plausible narrative for the underlying causal process and then translates the symbolic expressions from the underlying causal process to natural language using carefully designed templates.

Specifically, we use several different grammatical forms for each semantic concept $\bm{t}$ in the story to make the resulting prompt sound natural and grammatically correct. We first have the overall variable name $v_{\mathrm{overall}}(\bm{t})$ (e.g., the recovery status), and, then, for each binary value $i \in \{0, 1\}$, we compose its noun $v_{\mathrm{noun}}(\bm{t}=i)$ (e.g., recovery), verb (e.g., to recover), sentence $v_{\mathrm{sent}}(\bm{t}=i)$ (e.g., the patients recover), noun with attributive clause $v_{\mathrm{attr}}(\bm{t}=i)$ (e.g., patients who recover), and third conditional $v_{\mathrm{cond}}(\bm{t}=i)$ (e.g., if the patient had recovered).

Using these elements, we first verbalize the causal graph by iterating through each node and its outgoing edges, using the template ``$\bm{t}$ has a direct effect on $\CH(\bm{t}$).'', where $\CH(\cdot)$ denotes the set of direct effects (children) of a variable. 
Then, for the available data $\bm{d}$, we verbalize each conditional probability by ``For $v_{\mathrm{attr}}(\bm{t}_m=i)$, the probability of $v_{\mathrm{noun}}(\bm{t}_n=1)$ is $p$.'', and each marginal probability by ``The overall probability of $v_{\mathrm{attr}}(\bm{t}=1)$ is $p$.'' Note that our distributions are Bernoulli, so it is adequate to just introduce the parameter $p$, which is the likelihood of $\bm{t}=1$.
For example, we generate sentences such as ``The overall probability of recovery is 60\%.'' and ``For patients who have small kidney stones, the probability of recovery is 70\%.''
Finally, for the query $\bm{q}$, we instantiate each query type in our dataset following our question templates in \cref{cladder:appd:query} such that the questions can always be answered with ``yes'' or ``no''.

\paragraph{Generating the Explanations.}

Apart from the question-answer pairs, we also generate the step-by-step explanations.
Our goal is to provide all intermediate reasoning steps a student of causal inference would use to answer the questions, so that each necessary subskill necessary for causal inference can be evaluated individually. We identify the following six subskills: \stepone{} causal graph extraction; \steptwo{} correct query type interpretation; \stepthree{} symbolic formalization of the query; \stepfour{} semantic parsing to compile the available data; \stepfive{} estimand derivation; and \stepsix{} arithmetic calculation to solve the estimand, as in the colored boxes in~\cref{cladder:fig:question_example}.
Our explanation $\bm{e}$ verbalizes all the elements \stepone{}-\stepsix{} as sequential steps using our template in~\cref{cladder:appd:expl}.

\subsection{Dataset Statistics}
\begin{figure}[ht]
  \centering
  \begin{minipage}[b]{0.6\textwidth}
    \centering \small
    \begin{tabular}{lc|ccccc}
\toprule
& Total & Rung 1 & Rung 2 & Rung 3 \\ \hline
Size &&&&&\\
\quad \# Samples & 

10,112 & 3,160 & 3,160 & 3,792
\\
Question & \\ 
\quad \# Sentences/Sample & 6.01 & 5.88 & 5.37 & 6.65 \\
\quad \# Words/Sample & 80.9 & 73.43 & 76.95 & 90.42 \\
\quad \# Nodes/Graph & 3.52 & 3.5 & 3.5 & 3.54 \\
\quad \# Edges/Graph & 3.38 & 3.3 & 3.3 & 3.5\\
Answer & \\
\quad Positive Class (\%) & 50 & 50 & 50 & 50 \\
Explanations & \\
\quad \# Sentences/Sample & 9.11 & 9.1 & 8.1 & 9.96 \\
\quad \# Words/Sample & 47.95 & 49.87 & 32.8 & 58.97 \\
\bottomrule
    \end{tabular}
    \captionof{table}{Statistics of our \ourdata dataset v1.5.}
    \label{cladder:tab:stats}
  \end{minipage}
  \hfill
  \begin{minipage}[b]{0.25\textwidth}
    \centering
    \includegraphics[width=\textwidth]{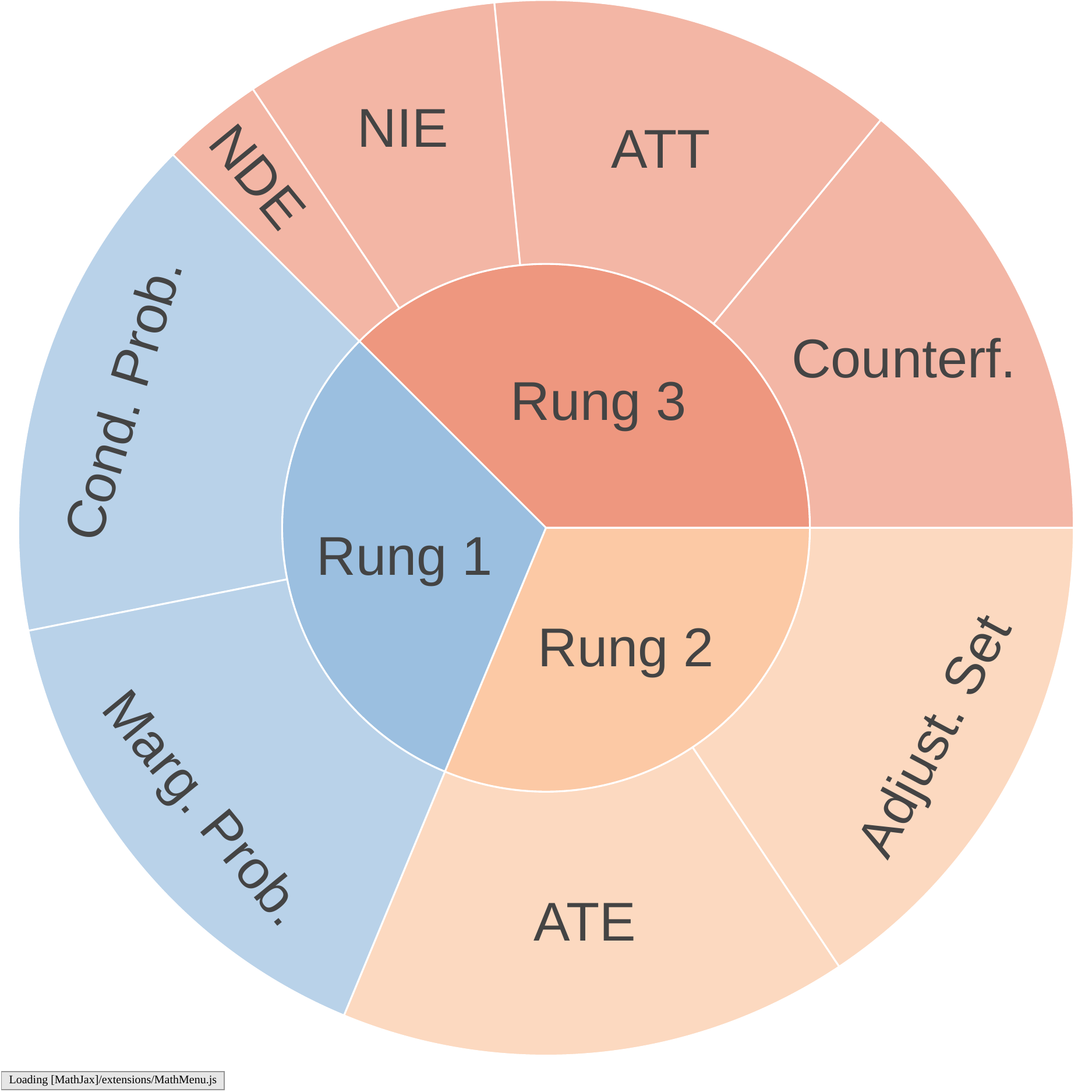}
    \caption{Distributions of query types in our 10K data. 
    }\label{cladder:fig:query}
  \end{minipage}
\end{figure}

Our data-generating procedure has the potential to algorithmically generate a vast large number of questions. 
In practice, we pick a dataset size that is large enough to be representative, and at the same time not too large to be problematic given the expensive inference costs of LLMs. We therefore set our dataset size to be 10K, and report the statistics %
in \cref{cladder:tab:stats}.

The dataset roughly balance across the query types, graph structures, stories, and ground truth answers (as seen in \cref{cladder:fig:query}). 
Note that some causal queries are only compatible with a subset of the graphs, thereby resulting in a slightly lower representation of those queries (such as the NDE and NIE).
More details on our design choices can be found in~\cref{cladder:appd:querycoverage}.

\subsection{Data Quality Check}

Our dataset is generated through an algorithmic procedure, which has the following potential benefits: formal correctness; zero human annotation cost; and, most importantly, controllability---e.g., for the question distribution, as well as for making it more unlikely that the data was previously seen by the model.
However, since the dataset is different from common NLP datasets collected from human natural language writing, we also need to perform additional data quality checks.
We therefore checked for a list of non-formal, natural language properties: grammaticality; human readability; naturalness/perplexity; and how well humans perform on this task.

For grammaticality, we ran a grammatical error check on our dataset using the LanguageTool package~\citep{naber2003rule}, and got on average 1.26 grammatical errors per 100 words (i.e., 98.74\% correctness), which shows that most of the language in our dataset follows English grammar.
For human readability, we checked how comprehensible the questions are to students who have taken causality courses. We selected a random subset of 50 questions from the dataset, and let a graduate student annotator go through the questions to judge whether they could understand them or not: 96\% of the questions were deemed readable. Next, for the naturalness/perplexity score, we used the open-sourced GPT-2 model and obtained a perplexity score of 21.17 on our dataset, which is substantially lower (i.e., closer to the distribution of natural human-written text) than the one of MATH~\citep{hendrycks2021measuring}, a commonly used dataset of maths questions. Lastly, we conducted a sanity check where one expert evaluator tried to solve a random sample of 50 questions from the dataset, and we recorded an %
accuracy of 82\% on this task.

\section{Our \ourmodel Model}
\label{cladder:sec:causal_cot}
\begin{figure}[ht]
    \centering
    \includegraphics[width=\textwidth]{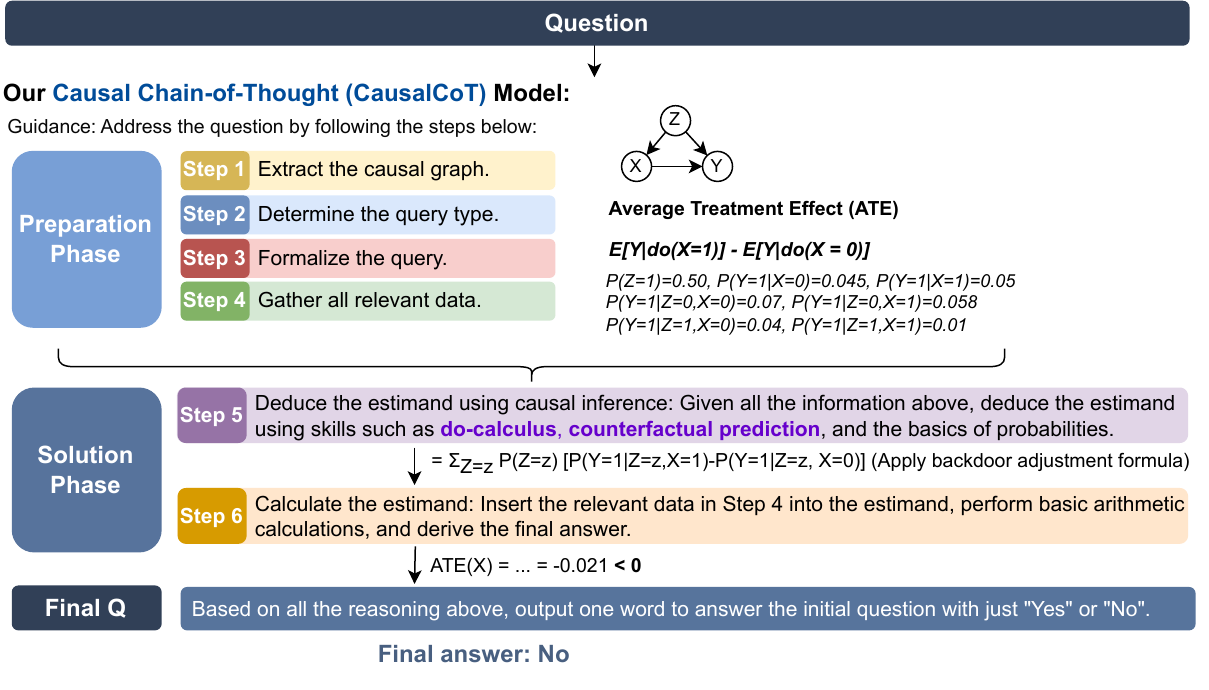}
    \caption{Illustration of our \ourmodel prompting strategy, which designs a chain of subquestions inspired by the idea of a CI engine \citep{pearl2018book}.}
    \label{cladder:fig:model}
    
\end{figure}

In order to guide LLMs in correctly answering the questions in \ourdata, 
we draw inspiration from the ideal functioning of the CI engine \citep{pearl2018book},
which breaks down a causal reasoning problem into multiple symbolically-grounded, simpler steps.
We develop \ourmodel, a multi-step causal chain-of-thought prompt in \cref{cladder:fig:model}, which combines formal causal reasoning skills with the idea of chain-of-thought prompting \citep{wei2022chain} and the use of scratch pads for solving more complicated problems requiring a long list of steps \citep{nye2021show} for LLMs. 

We base our prompt design on the multi-step reasoning process of causal inference as shown in \cref{cladder:fig:model}, first starting with four preparation steps: \stepone{} identifying the causal graph structure; \steptwo{} determining the causal query type;%
\footnote{This step amounts to a multi-class classification problem, where each class is a different causal query.} %
\stepthree{} formulating the query symbolically precisely; and
\stepfour{} extracting relevant data from the prompt.
Then, given all the information collected in the preparation stage, we introduce the formal solution:
\stepfive
correctly deducing the estimand using causal inference techniques;
and finally \stepsix{} evaluating the estimand to answer the question.
This set of steps require both \textit{natural language understanding} to parse the question (as in most steps in the preparation phase), as well as \textit{formal causal reasoning} to derive the correct estimand (as in the solution phase).

We build our \ourmodel prompting strategy using GPT-4 \citep{openai2023gpt4}, a recent autoregressive LLM that achieves state-of-the-art performance on many tasks. This latest model builds upon the previous series of general pretrained models (GPT) \citep{radford2019language,brown2020gpt3} and adds reinforcement learning with human feedback, or instruction-tuning \citep{ziegler2019finetuning,ouyang2022instructGPT,bai2022training}, to align the model responses to free-form questions with human preferences. It has achieved human-competitive performance over a list of tasks \citep{openai2023gpt4,bubeck2023sparks,nori2023capabilities,katz2023gpt,ziems2023large}, among which the more formal tasks unseen in the training data still remain elusive \citep{razeghi-etal-2022-impact,stolfo-etal-2023-causal,jin2024large}.

Given a causal question $\bm{q}$, we provide the LLM a list of instructions $\bm{\ell}:=(\bm{s}_1, \dots, \bm{s}_6)$ consisting of the detailed descriptions of the six steps $\bm{s}_1, \dots, \bm{s}_6$ in \cref{cladder:fig:model}.
As the model $f_{\mathrm{LLM}} : \bm{s}_i \mapsto \bm{r}_i$ autoregressively produces responses $\bm{r}_1, \cdots, \bm{r}_6$ sequentially corresponding to the six steps, we concatenate all the above before asking the final question 
``Based on all the reasoning above, output one word to answer the initial question with just `Yes' or `No'.''
 See the complete prompt in \cref{cladder:appd:prompt}.
In the end, we obtain the binary answer $a\in \{\text{Yes}, \text{No}\}$ as the final result.

Compared with the standard strategy of directly prompting the LLMs a question, we impose  an \textit{inductive bias} upon LLMs by using the causal inference framework, thus incorporating some of the powerful, principled insights of the causal inference community for NLP tasks. In this way, we enhance the strong natural language ability of LLMs with formal causal reasoning skills.

\section{Testing LLMs with \ourdata}
\label{cladder:sec:experiments}
\subsection{Experimental Setup}
Our empirical investigation focuses on some of the most recent language models. We include the latest GPT-4 \citep{openai2023gpt4} with 1T parameters by the time we conduct the experiments (i.e., gpt-4-1106-preview), the previous ChatGPT (i.e., GPT-3.5) with 175B parameters, and then a series of earlier models with instruction-tuning on the 175B GPT-3 (text-davinci-001, -002, and -003) \citep{ouyang2022instructGPT}. 
As baselines, we also include the non-instruction-tuned GPT-3 (davinci). We use the OpenAI API with temperature 0 when querying these models.
We also include open-source, more efficient models like LLaMa \citep{touvron2023llama} and its instruction-tuned version Alpaca \citep{taori2023alpaca}, both with the same number of parameters, 6.7B. 

\subsection{Main Results}
\begin{table}[ht]
    \centering \small
    \setlength\tabcolsep{5pt}
    \begin{tabular}{lc|ccc|ccccccccccc}
\toprule
& \multirow{2}{*}{Overall Acc.} & \multicolumn{3}{c|}{Acc. by Rung} & \multicolumn{3}{c}{Acc. by Commonsense Alignment} \\
&& 1 & 2 & 3 & Comm. & Nonsens. & Anti-C. \\
\hline
Random & 49.27 & 50.28 & 48.40 & 49.12 & 49.01 & 49.69 & 49.12\\
LLaMa & 44.03 & 48.23 & 29.46 & 52.66 & 45.14 & 44.22 & 42.67 \\
Alpaca & 44.66 & 52.03 & 29.53 & 51.13 & 44.86 & 44.40 & 44.77\\
GPT-3 Non-Instr. (davinci) & 
49.92 
& 50.00 & 49.75 & 50.00 & 49.06 & 49.97 & 50.72 \\
GPT-3 Instr. (text-davinci-001) & 
51.40
& 51.30 & 52.63 & 50.47 & 54.31 & 50.13 & 50.05 \\
GPT-3 Instr. (text-davinci-002) & 
53.15
& 50.85 & 56.96 & 51.90 & 55.33 & 52.47 & 51.81 \\
GPT-3 Instr. (text-davinci-003) & 
56.26
& 51.11 & 62.97 & 54.96 & 56.83 & 54.79 & 57.49 \\
GPT-3.5 & 
52.18
& 51.80 & 54.78 & 50.32 & 54.09 & 50.68 & 52.09 \\
GPT-4 & 

62.03 & 63.01 & 62.82 & 60.55 & 62.27 & 63.09 & 60.47
\\
+ \ourmodel & 
\textbf{70.40} & \textbf{83.35} & \textbf{67.47} & \textbf{62.05} & \textbf{69.25} & \textbf{71.58} & \textbf{70.12} \\
\bottomrule
    \end{tabular}
    \caption{Performance of all models on our \ourdata dataset {v1.5}. We report the overall accuracy (Acc.), and also fine-grained accuracy by rung, and by degree of commonsense alignment, from commonsensical (Comm.), nonsensical (Nonsens.), to  anti-commonsensical (Anti-C.).}
    \label{cladder:tab:main_res}
\end{table}
We compare the performance of all models in \cref{cladder:tab:main_res}. %
    First, 
we can see that the causal reasoning task in \ourdata is in general very challenging for all models. Models such as the earlier, non-instruction-tuned GPT-3, and both LLaMa and Alpaca are around random performance.
With instruction-tuning, models start to show some improvement. And amongst all, our \ourmodel achieves the highest performance of 70.40\%, which is substantially better than the vanilla GPT-4 by 8.37 points. Moreover, \ourmodel also achieve the best performance across all three rungs of causal questions, with a monotonically decreasing performance as the rungs get higher, i.e., the questions get more difficult.
See \cref{cladder:appd:v1} for experiments on our earlier dataset v1.0.

\subsection{Isolating the Effect of Data Contamination}
A well-known problem with evaluating LLMs on question-answering tasks is the data contamination problem, i.e., that LLMs perform well on a test set because the test set is (unintentionally) contained partially or even entirely in the training data \citep{openai2023gpt4,brown2020gpt3}. We address this problem by creating not only the commonsensical subset of our dataset, but also anti-commonsensical and nonsensical,
both of which, by construction, are very likely not in the training data of LLMs.
From the accuracy by commonsense alignment degree in \cref{cladder:tab:main_res}, we can see the original GPT-4 model performs the worst on the anti-commonsensical subset (1.8 points lower than that on the commonsensical subset). However, our \ourmodel enhances the reasoning ability across all levels, with  substantial improvement on anti-commonsensical data by 9.65 points, highlighting the strength of \ourmodel on unseen data.

\subsection{Error Analysis by Subquestions} \label{cladder:sec:exp_steps}

\begin{table}[ht]
    \centering \small
    
    \setlength\tabcolsep{2.5pt}
    \begin{tabular}{ccccccccccccccccc}
\toprule
\multicolumn{3}{c}{Step \stepone{}} & & \multicolumn{4}{c}{Step \steptwo{}} && \multirow{1}{*}{Step \stepthree{} \& \stepfive} && \multirow{1}{*}{Step \stepfour{}} && {Step \stepsix} \\
\cline{1-3} \cline{5-8} \cline{10-10} \cline{12-12}
\cline{14-14}
Node & Edge & Dist. ($\downarrow$) & & Overall F1 & Rung 1 & Rung 2 & Rung 3 && Estimand && F1 && Arithmetic \\ \midrule
99.34 & 97.01
& 1.69 & & 50.65 & 69.99 & 59.14 & 42.12 && 53 && 47.53 && 99
\\
\bottomrule
    \end{tabular}
    \caption{Performance for each step in \ourmodel. For Step \stepone{}, we report the F1 score of node prediction, edge prediction, and also the graph edit distance (Dist.) with the true graph. See more details in \cref{cladder:appd:caption_error}.}
    \label{cladder:tab:subquestions}
\end{table}

We conduct a fine-grained error analysis by looking into the performance of different steps of \ourmodel in \cref{cladder:tab:subquestions}.%
\footnote{We experienced some rate-limiting in the fine-grained analysis of LLMs that are only accessible through a web API. As a result, we occasionally had to evaluate on a subset of 2K random samples.}
We can see that the model is good at Step \stepone{} to extract causal graph $\mathcal{G}$, achieving high F1 scores for predicting both the nodes and the edges correctly, although not perfect, still leaving a graph edit distance of 1.69 between the ground truth causal graph and the model-identified graph.
The other steps are more challenging for the model. Among those, Steps \steptwo{}, \stepthree and \stepfive{} require careful and correct application of causal inference, where the model struggles. This reveals a notable weakness of current LLMs to perform formal causal reasoning, which is an important direction for future work on improving and enhancing LLMs.
To better understand the reasoning abilities of LLMs, we also perform an extensive analysis taking the entire reasoning chain of our \ourmodel and the ground-truth explanations, to produce 20 fine-grained scores about the multi-step reasoning quality using the ROSCOE framework \citep{golovneva2022roscoe}, and show detailed results in \cref{cladder:par:roscoe}.

\subsection{Effect of In-Context Learning}
\begin{wrapfigure}{r}{0.3\textwidth}
  \begin{center}
\includegraphics[width=0.3\textwidth]{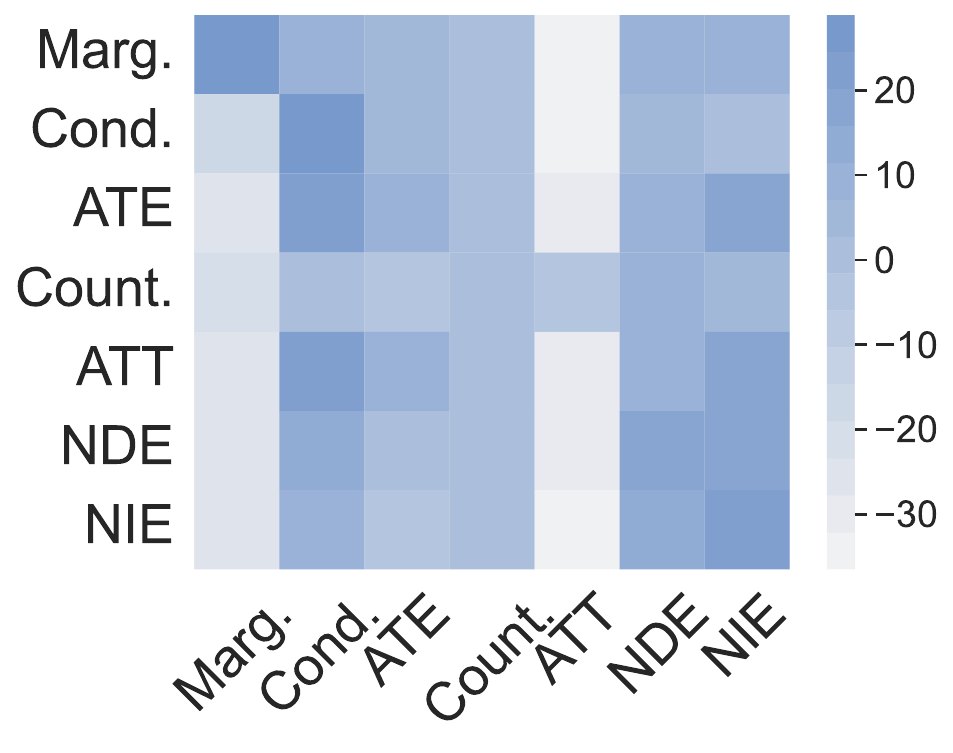}
   \end{center}
    \caption{Heatmap showing the how helpful each query type is to solving subsequent query types.}
    \label{cladder:fig:heatmap}
\end{wrapfigure}
As an additional analysis, we look into the effect of in-context learning (ICL) by providing an example solution before asking the question.
The interesting question to us is whether models can generalize across different query types. Namely, we keep our \ourmodel framework, and prepend a reasoning example of query type $i$, and then calculate how much improvement it can bring when models answer new questions of query type $j$. In \cref{cladder:fig:heatmap}, we can see that conditional probability and NIE are the questions that benefit the most from ICL, and showing examples of marginal probability and ATT are among the most helpful to all questions in general.

\section{Related Work}

\paragraph{Skill evaluation for LLMs.}
Our work may be seen as part of the literature aimed at evaluating
the performance of current LLMs~\citep[][\textit{inter alia}]{radford2019language,devlin-etal-2019-bert,brown2020gpt3,
zhang2022opt,openai2023gpt4}, focusing on understanding their strengths and weaknesses.
Various studies into the capabilities of LLMs \citep{bubeck2023sparks,qin2023chatgpt,openai2023gpt4,ignat-etal-2024-has} change people's perception of domains such as education \citep{baidoo2023education,rudolph2023chatgpt}, medicine \citep{singhal2022large,nori2023capabilities}, law \citep{katz2023gpt}, and computational social science \citep{ziems2023large}.
However, most work evaluates new models on existing datasets from previously-curated large-scale benchmarks \citep{wang2019superglue,wang2022supernaturalinstructions,srivastava2022beyond}, or human exams \citep{katz2023gpt,openai2023gpt4,jin-etal-2022-logical} which is becoming increasingly unreliable due to training set contamination.

\paragraph{Causality-related skills for NLP.}
With the increasing attention on LLMs and causality \citep{zevcevic2023causal,zhang2023understanding}, we review several formulations of causality-related skills for NLP, which we summarize into (1) causality as knowledge, (2) causality as language comprehension, and (3) causality as reasoning.
In the \textit{causality-as-knowledge} line of work, many existing studies investigate how well NLP models understand commonsense causality, such as the cause and effect of an agent's action \citep{sap2019atomic}, motivation and emotional reaction in a social context \citep{Sap2019SocialIQA}, correspondence of a set of steps with a high-level goal \citep{zhang-etal-2020-reasoning}, development of a story given a different beginning \citep{qin-etal-2019-counterfactual}, and how in general LLMs serve as a knowledge base of causality \citep{zevcevic2023causal}. 
Concurrent work~\citep{kiciman2023causal} focuses on evaluating LLMs on various causality related tasks %
by leveraging the conceptual knowledge accrued from the training data, rather than formal causal inference, except for their causal sufficiency analysis which is close to our counterfactual questions.
Importantly, most work in this line does not define explicit causal graphs, making it difficult to quantitatively define the ground-truth causal relationships in a principled way.
The {\em causality-as-language-comprehension} line of work stems from traditional linguistic studies on causal connectives and causal language usage \citep{stede-2008-connective,cao-etal-2022-cognitive,yu-etal-2019-detecting}, to the recent causal relation extraction \citep{bethard-etal-2008-building,hidey-mckeown-2016-identifying,xu-etal-2020-review} to identify cause-effect pairs as a subtask of information extraction from text. 

Finally, for \textit{causality as formal reasoning}, our \ourdata work formulates the task of causal inference for NLP, and our other work, \textsc{Corr2Cause} \citep{jin2024large}, addresses the causal discovery problem to infer causation from correlation. Together, they cover the two major branches of causal reasoning investigated in existing technical literature on causality.
See a comprehensive comparison of literature in \cref{cladder:appd:related}.

\section{Discussion of Limitations and Future Work}

\paragraph{A Natural Language {\em ``Mini Turing Test''} for Causality.} \citet{pearl2018book} describe an ideal {\em ``mini-Turing test''} to assess understanding of causal inference, and argue that if a machine can answer all possible questions correctly, then it ``understands'' causality. According to the authors, 
this is because there are no possible shortcuts when you consider all possible combinations of queries, graphs and data in this ideal test: due to their combinatorial explosion, the machine can only answer all questions right if it correctly applies causal reasoning. From this point of view, our work constitutes a {\em first step towards a mini-Turing test formulated in natural language}. However, we cover only some of the commonly studied causal queries spanning all three rungs. Future work may extend this to further queries, such as, e.g., path-specific effects other than NDE and NIE~\citep[]{nabi2018fair}, thereby increasing the number of potential questions and moving closer to the ideal test. 
\paragraph{LLMs and Causal Reasoning.}
\looseness=-1
It has been claimed that LLMs understand causality well (e.g.,~\citep{kiciman2023causal} report high performance, such as 97\% and 92\%). In contrast, our work suggests that LLMs may still be far from reasoning reliably about causality (reaching only 60+\% on~\ourdata). 
As argued in~\cref{cladder:sec:intro}, we believe that investigating this aspect may be of particular importance, since causal inference is crucial in many policy-relevant scenarios, where reliable AI systems could assist decision-making: from epidemiology~\citep{glass2013causal, rothman2005causation} to economics~\citep{card1999causal, hunermund2019causal} to fairness~\citep{loftus2018causal, plecko2022causal}.
Testing the abilities 
of these systems
in semi-realistic scenarios is therefore crucial, motivating some of the design choices in our dataset: e.g., the example in~\cref{cladder:fig:question_example} was 
inspired by similar questions which arose in the context of the COVID-19 pandemic, where incorrect causal reasoning resulted in a fallacy where vaccinations were considered to be harmful instead of beneficial~\citep{morris2021israeli, Ellenberg2021}. 
Further work may be dedicated to making the questions and verbalizations even closer to realistic instances of causal inference problems.

\paragraph{A CI Engine Plug-in for LLMs.}
An interesting direction for future research could be to provide the LLM access to an actual implementation of the CI engine. For example,~\citet{davis2023testing} tested the improvement of math abilities in LLMs augmented with plug-ins (i.e., 
external modules that extend the model's capabilities by adding specific functionality or customizing its behaviour for particular tasks, like a calculator), suggesting that they
significantly enhance the model's ability to solve these problems. However, %
even with plug-ins, there are still often {\em ``interface''} failures: that is, {\em ``[the LLM] often has trouble formulating problems in a way that elicits useful answers from the plug-ins''}.
We hypothesise that something similar would happen for causal inference: even once suitable plug-ins are built, the language-to-tool interface may still be a non-trivial research question.

\section{Conclusion}

We proposed formal causal reasoning as a new task to evaluate LLMs, and created the \ourdata benchmark, covering several aspects of causal inference across all rungs of the ladder of causation %
and verbalizations involving semi-realistic scenarios. 
To address the task, we proposed a prompting strategy, \ourmodel, inspired by the principles of formal causal inference, which introduces multistep chain-of-thought reasoning for causal questions. Extensive experiments indicate that this dataset is highly challenging, thus offering a principled tool to gain a better understanding of the reasoning abilities of LLMs and to develop better models for causal reasoning in natural language.

\part{Causal Understanding of How LLMs Work}\label{part:2}

\mainchapter{Competition of Mechanisms: Tracing How LLMs Handle Facts and Counterfactuals}\label{ch:compmech}

\newcommand{\tfact}{t_{\mathrm{fact}}\xspace}
\newcommand{\talt}{t_{\mathrm{cofa}}\xspace}
\newcommand{\dalt}{\Delta_{\mathrm{cofa}}\xspace}

Interpretability research aims to bridge the gap between empirical success and our scientific understanding of the inner workings of large language models (LLMs).
However, most existing research focuses on analyzing a single mechanism, such as how models copy or recall factual knowledge. In this work, we propose a formulation of \textit{competition of mechanisms}, which focuses on the interplay of multiple mechanisms instead of individual mechanisms and traces how one of them becomes dominant in the final prediction.
We uncover how and where mechanisms compete within LLMs using two interpretability methods: logit inspection and attention modification. Our findings show traces of the mechanisms and their competition across various model components and reveal attention positions that effectively control the strength of certain mechanisms.%

Our code is available at {\href{https://github.com/francescortu/comp-mech}{https://github.com/francescortu/comp-mech}, and our data at \href{https://huggingface.co/datasets/francescortu/comp-mech}{https://huggingface.co/datasets/francescortu/comp-mech}}.

\section{Introduction}
Recent advancements in large language models (LLMs) have brought unprecedented performance improvements to NLP \cite[\textit{inter alia}]{brown2020gpt3,touvron2023llama,openai2023gpt4,anil2023gemini}. However, the black-box nature of these models obfuscates our scientific understanding of \textit{how these models achieve certain capabilities}, and \textit{how can we trace the problem when they fail at other tasks}.
This has brought an increasing focus on interpretability research to help us understand the inner workings of LLMs.
\begin{figure}[ht]
    \centering
        \centering
        \includegraphics[width=.5\linewidth]{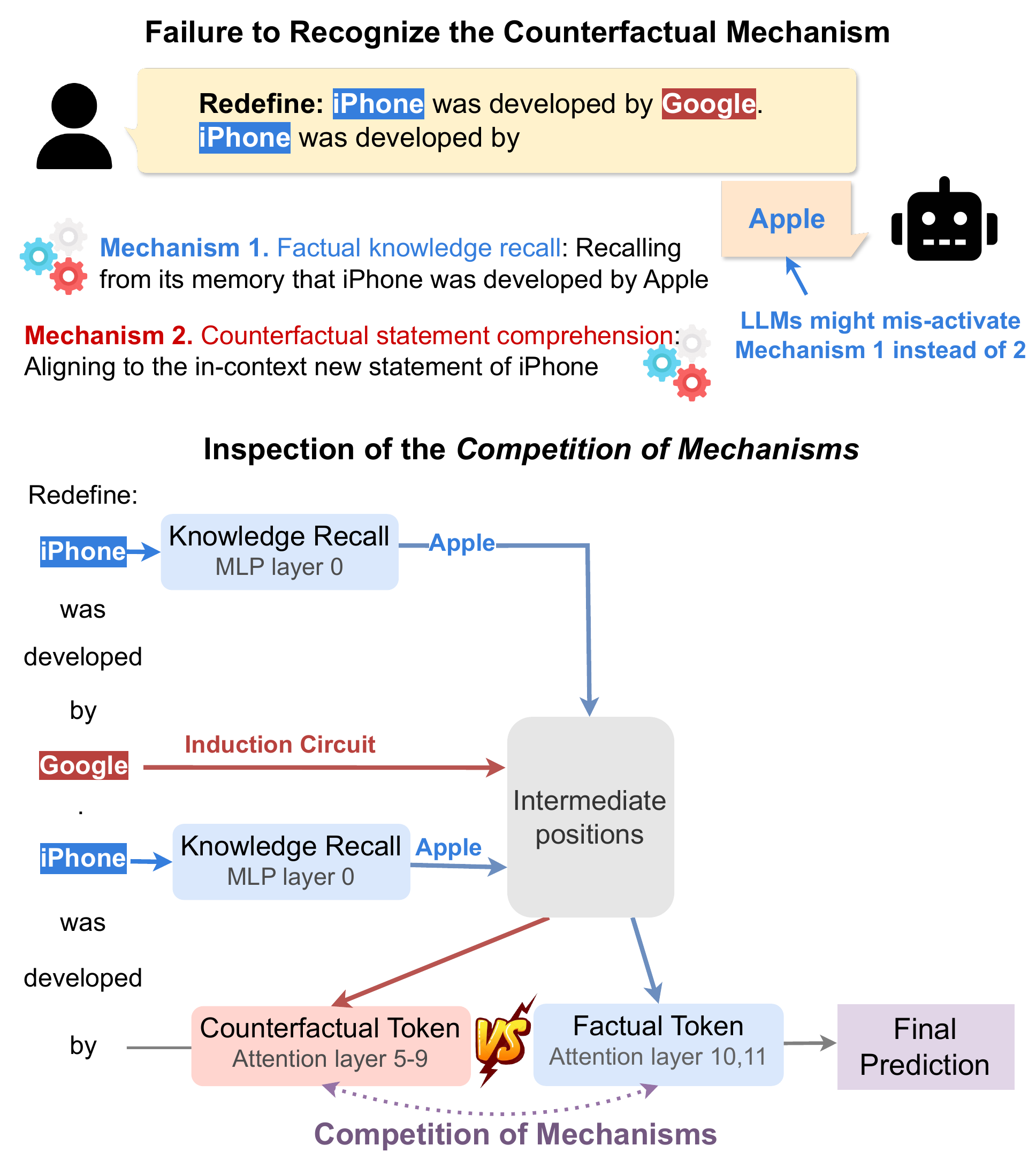}
    \caption{Top: An example showing that LLMs can fail to recognize the correct mechanism when multiple possible mechanisms exist.
    Bottom: Our mechanistic inspection of where and how the competition of mechanisms takes place within the LLMs.
    }\label{fig:examples}
\end{figure}

Existing interpretability research has been largely focused on discovering the \textit{existence} of single mechanisms, such as the copy mechanism in induction heads of LLMs \cite{elhage2021mathematical,olsson2022incontext}, and factual knowledge recall in the MLP layers \citep{geva-etal-2021-transformer,meng2022locating,geva2023dissecting}. However, different from discovering \textit{what mechanisms exist in LLMs}, we propose a more fundamental question: \textit{how do different mechanisms interact in the decision-making of LLMs?}
We show a motivating example in \cref{fig:examples},
where the model fails to recognize the correct mechanism when it needs to judge between two possible mechanisms: whether to recall the factual knowledge on who developed the iPhone (i.e., Mechanism 1) or to follow its counterfactual redefinition in the new given context (i.e., Mechanism 2).

We propose a novel formulation of \textit{competition of mechanisms}, which focuses on tracing each mechanism in the model and understanding how one of them becomes dominant in the final prediction by winning the ``competition''. Specifically, we build our work on two single mechanisms that are well-studied separately in literature: (1) the factual knowledge recall mechanism, which can be located in the MLP layers \citep{geva-etal-2021-transformer,meng2022locating,geva2023dissecting}; and (2) the in-context adaptation to a counterfactual statement, which is enabled by the copy mechanism conducted by induction heads of attention layers \cite{elhage2021mathematical,olsson2022incontext}.
Based on the latest tools to inspect each of these two mechanisms \cite{logitlens2020lesswrong,wang2023IOI,geva2023dissecting}, we then unfold \textit{how and where} the competition of the two mechanisms happen, and how it leads to the overall success or failure of LLMs.

Technically, we deploy two main methods: logit inspection \cite{logitlens2020lesswrong,geva-etal-2022-transformer} by projecting the outputs of each model component by an unembedding matrix, and attention modification 
\cite{geva2023dissecting,wang2023IOI}.
Using these methods, we assess the contributions of various model components, both from a macroscopic view (e.g., each layer) and a microscopic view (e.g., attention heads), and identify
critical positions and attention heads involved in the competition of the two mechanisms.
Moreover, we locate a few localized positions of some attention head matrices that can significantly control the strength of the factual mechanism. 
We summarize our main findings as follows:

\begin{enumerate}
    \item In early layers, the factual attribute is encoded in the subject position, while the counterfactual is in the attribute position (\cref{sec:residual_stream});
    \item The attention blocks write most of the factual and counterfactual information to the last position (\cref{sec:attn_mlp_contrbution});
    \item All the highly activated heads attend to the attribute position regardless of the specific type of information they promote. The factual information flows by penalizing the counterfactual attribute rather than promoting the factual one (\cref{sec:inspection_heads});
    \item 
    We find that we can up-weight the value of a few very localized values of the attention head matrix to strengthen factual mechanisms substantially
    (\cref{sec:improving_factual_recall}).
\end{enumerate}

\section{Related  Work on  Interpretability
}
As deep learning approaches show increasingly impressive performance in NLP, their black-box nature has hindered the scientific understanding of these models and their effective future improvements. To this end, interpretability research has been a rising research direction to understand the internal workings of these models.

\paragraph{Interpreting the Representations.}
One major type of work in interpretability has focused on understanding what has been encoded in the representations of deep learning models.
This is usually achieved by  a \textit{probe}, namely by training a supervised classifier to predict features from its representations
\citep[\textit{inter alia}]{alain2016probe,conneau-etal-2018-cram,hupkes2018visualisation, hewitt-liang-2019-designing,tenney2019WhatDY, jiang-etal-2020-know, elazar2021pararel}, or by geometric methods \cite{doimo2020hierarchical,valeriani2024geometry,park2024geometry,cheng2024emergence}. 
Example features of interest include part of speech \cite{belinkov-etal-2017-neural}, 
verb tense \cite{conneau-etal-2018-cram},
syntax \cite{hewitt-manning-2019-structural}, 
and factual knowledge \cite{petroni-etal-2019-language}.

\paragraph{Interpreting the Mechanisms/Functions.}
Beyond interpreting the representations in the hidden states of the black-box models, another research direction is to interpret the mechanisms or functions that the models have learned, giving rise to the field of mechanistic interpretability
\citep[\textit{inter alia}]{olah2020zoom, elhage2021mathematical, olsson2022incontext,nanda2023grokking}. 
Some example mechanisms decoded in recent work include mathematical operations such as modular addition
\cite{nanda2023grokking} and the greater-than operation \citep{Hanna2023greater-then}; natural language-related operations such as the copy mechanism achieved by induction heads in LLMs \citet{olsson2022incontext} and factual knowledge recall achieved by MLP layers \citep{geva-etal-2021-transformer,meng2022locating,geva2023dissecting}, which we describe below.

\textit{The Single Mechanism of Copy:}
One of the basic actions in LLMs is the copy mechanism, which is found to be operationalized by attending to the copied token in the attention heads and passing it on to the next token prediction \cite{elhage2021mathematical, olsson2022incontext}. This foundational mechanism enables further research to decode more complex mechanisms, such as 
indirect object identification \cite{wang2023IOI}.

\textit{The Single Mechanism of Factual Knowledge Recall:} 
Another major direction is understanding how LLMs mechanistically recall factual information \citep{geva-etal-2021-transformer,meng2022locating,geva2023dissecting}. 
For example, \citet{meng2022locating} develop the \emph{causal tracing} method to show that the factual information is found in the mid-layer MLP units in GPT-2. 
A followup work \cite{geva2023dissecting} shows that MLPs of early layers enrich the subject embeddings with related attributes, and late attention blocks select and write the correct factual information to the sentence's last position.

\textit{Interplay of Multiple Mechanisms:}
In the final stage of our project in December 2023, we noticed a related study by \citet{yu2023characterizing}, which also investigates the role of two different mechanisms in LLMs.
Specifically, they inspect a type of prompt whose subjects are the capital cities and whose attributes are the countries, 
examine the dynamics of the factual recall mechanism and the effect of the in-context counterfactual statement,
and find that the subject and attribute frequency in the pre-training set can affect the ability of factual recall.
Differently, the methods in our work are applied to a broader set of prompts; 
moreover, we also establish novel analyses of the underlying mechanistic details of the competition, and precisely localize the path where the information flows at the level of single attention map activations, based on which we discover new findings that are unique to our study.

\section{Problem Setup}\label{sec:setup}
Following the setup of many existing interpretability studies \citep[\textit{inter alia}]{olah2020zoom, elhage2021mathematical, olsson2022incontext,nanda2023grokking}, we look into the next token prediction behavior of autoregressive LLMs in their inference mode, namely 
\begin{align}
    P(t_k | t_{<k})
    ,
\end{align}
which predicts the $k$-th token $t_k$ given all the previous tokens in the context.

Next, we design the task to incorporate the competition of mechanisms as in \cref{fig:examples}.
Specifically, for each factual statement $\bm{f}:=(t_1^f, \dots, t_k^f)$ consisting of $k$ tokens (e.g., ``iPhone was developed by Apple.''), we compose a corresponding counterfactual statement $\bm{c}:=(t_1^c, \dots, t_{k'}^c)$ (e.g., ``iPhone was developed by Google.''). Then, we compose a prompt connecting the two statements as ``Redefine: $\bm{c}$. $\bm{f}_{1:k-1}$.'', such as \textit{``Redefine: iPhone was developed by Google. iPhone was developed by \_\_\_''}.

The two mechanisms can be traced by inspecting the rise and fall of the factual token $t_k^f$ and the counterfactual token $t_{k'}^c$. For the simplicity of notation, we take the tokens out of the context of their exact position and denote them as $\tfact$ and $\talt$, respectively, in the rest of the paper.

\section{Method and Background}
\label{sec:methods}
\paragraph{Method 1: Logit Inspection.} To inspect the inner workings of the two mechanisms, we trace the \textit{residual stream} \cite{elhage2021mathematical}, or logits of each component in the LLM. 
Given a text sequence of $k$ tokens, LLMs map it
into the residual stream, namely a matrix $\bm{x} \in \mathbb{R}^{d \times k}$, where $d$ is the dimension of the internal states of the model.
We use the term $\bm{x}_{i}^{l}$ to specify the residual stream at position $i$ and layer $l$. 

An LLM produces the initial residual stream $\bm{x}_{i}^{0}$ by applying an embedding matrix $W_E \in \mathbb{R}^{|V| \times d}$ to each token $t_i$, where $|V|$ is the size of the vocabulary. 
Then, it modifies the residual stream by a sequence of $L$ layers, each consisting of an attention block $\bm{a}^l$ and MLP $\bm{m}^l$. Finally, after the last layer, it projects the internal state of the residual stream back to the vocabulary space with an unembedding matrix $W_U \in \mathbb{R}^{d \times |V|}$.
Formally, the update of the residual stream at the $l^{th}$ layer is:
\begin{equation}
\bm{x}^{l} = \bm{x}^{l-1} + \bm{a}^{l} + \bm{m}^{l}
~,
\label{eq:residual_stream}
\end{equation}
where both the attention and the MLP block take as input the $\bm{x}$ after layer normalization $\mathrm{norm}$:
\begin{align}
\bm{a}^{l} &= \bm{a}^{l}(\mathrm{norm}(\bm{x}^{l-1}))
~,\label{eq:attention_out} \\
\bm{m}^{l} &= \bm{m}^{l}(\mathrm{norm}(\bm{x}^{l-1}  + \bm{a}^{l}))
~.\label{eq:mlp_out}
\end{align}

To understand which token the residual stream $\bm{x}^l$ favors, we follow the common practice in previous work \citep{halawi2023overthinking_the_truth, geva2023dissecting, dar2023embeddingspace, geva-etal-2022-transformer} 
to project it to the vocabulary space using the aforementioned unembedding matrix $W_{U}$ which maps the latent embeddings to actual tokens in the vocabulary, enabling us to obtain the logits of the factual $\tfact$ and counterfactual token $\talt$. 

Known as the \emph{Logit Lens} \citep{logitlens2020lesswrong}, this method is broadly adopted due to its consistent success in yielding interpretable results, demonstrating its effectiveness through broad empirical usage. However, it is important to note that it can occasionally fail to reflect the actual importance of vocabulary items, especially in the early layers of the network \citep{belrose2023tunedlens}.

\paragraph{Method 2: Attention Modification.}
Modifying or ablating the activation of a specific model component is also a strategy used to improve the understanding of the information flow within LLMs, 
including techniques such as causal tracing \cite{meng2022locating} and attention knockout 
\citep{wang2023IOI, geva2023dissecting}.

In our work, we focus on modifying a small number of entries in the attention matrix. Namely, in the attention matrix ${A}^{hl}$ of the $h$-th head of the $l$-th attention layer $\bm{a}^{l}$, we focus on a certain entry, e.g., at the $(i, j)$ position, where $j<i$, which is the attention value of the token $\bm{x}_i^{l}$ attending to one of its earlier tokens $\bm{x}_j^{l}$. Following recent work \cite{yu2023characterizing} 
, the modification is after the softmax layer, so the other attention values of the matrix stay unchanged.
For the target entry ${A}_{ij}^{hl}$, we scale it up by a multiplier of $\alpha$:
\begin{equation}
{A}_{ij}^{hl} \leftarrow \alpha \cdot {A}_{ij}^{hl} , \quad \text{ where }j<i~.
\label{eq:alpha}
\end{equation}

\section{Experimental Setup}\label{sec:experimental}

\paragraph{Data Creation}
To compose the factual and counterfactual statements as introduced in \cref{sec:setup}, we adopt 
\textsc{CounterFact}\footnote{\href{https://rome.baulab.info/data/}{https://rome.baulab.info/data/}} \citep{meng2022locating}, commonly used dataset to interpret models' ability of factual knowledge recall. 
We select 10K data points by considering only examples where the attributes are represented by a single token and where the model completes the sentence in a factually accurate manner.

Each instance of \textsc{CounterFact}  expresses a relation $r$ between a subject $s$ and an attribute $a$:  $(s, r, a)$. For example, in the sentence \textit{``iPhone was developed by Apple''}, $s=$ \textit{``iPhone''}, $r=$ \textit{``was developed by''}, $a=$\textit{``Apple''}. Moreover, each $(s, r)$ instance is provided two values of the attribute $a$, namely a factual token $\tfact$, and a counterfactual token $\talt$, representing a false fact.

Using this source data, we compose each instance of our test set in the format of $(\text{\textit{``Redefine:''}}, s,r,{\talt},s,r,\_)$, such as \textit{``Redefine: iPhone was developed by Google. iPhone was developed by \_\_\_''}. We preprocess the original dataset by keeping only the data points whose attribute is a single token (for the simplicity of our implementation), and where the model correctly predicts the factual token $\tfact$ when completing the sentence $(s,r,\_)$. We randomly select 10K test samples into our test set from 219,180 such samples.
We open-source our dataset at {\href{https://huggingface.co/datasets/francescortu/comp-mech}{https://huggingface.co/datasets/francescortu/comp-mech}}.

\begin{figure*}[ht]
    \centering
    \begin{subfigure}{.47\textwidth}
        \centering
        \includegraphics[width=.82\linewidth]{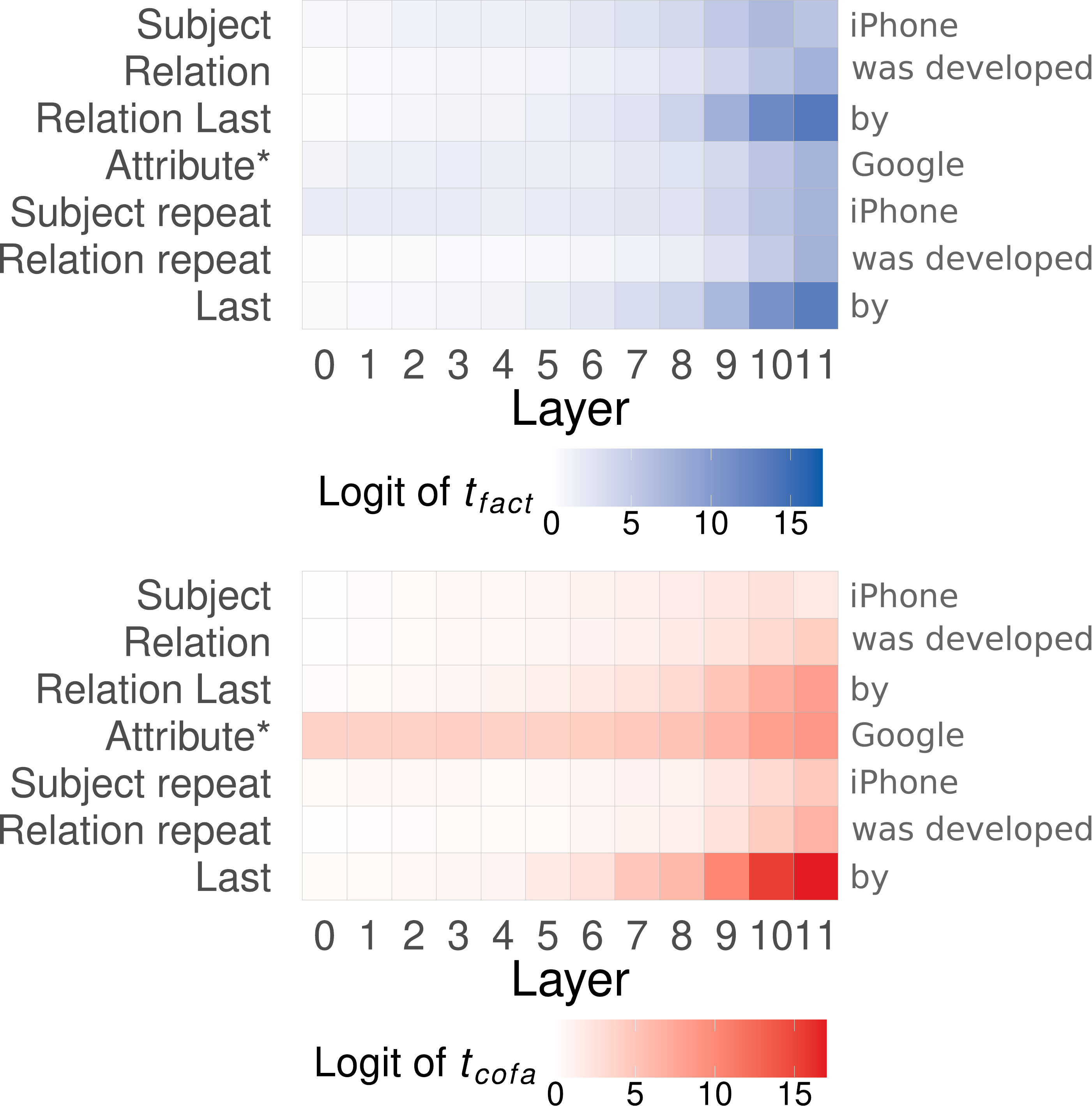}
        \caption{The logit values for the factual token $\tfact$ (blue), and counterfactual token $\talt$ (red).}\label{fig:token_logit}
    \end{subfigure}
    \hfill
    \begin{subfigure}{.47\textwidth}
        \centering
        \includegraphics[width=1\linewidth]{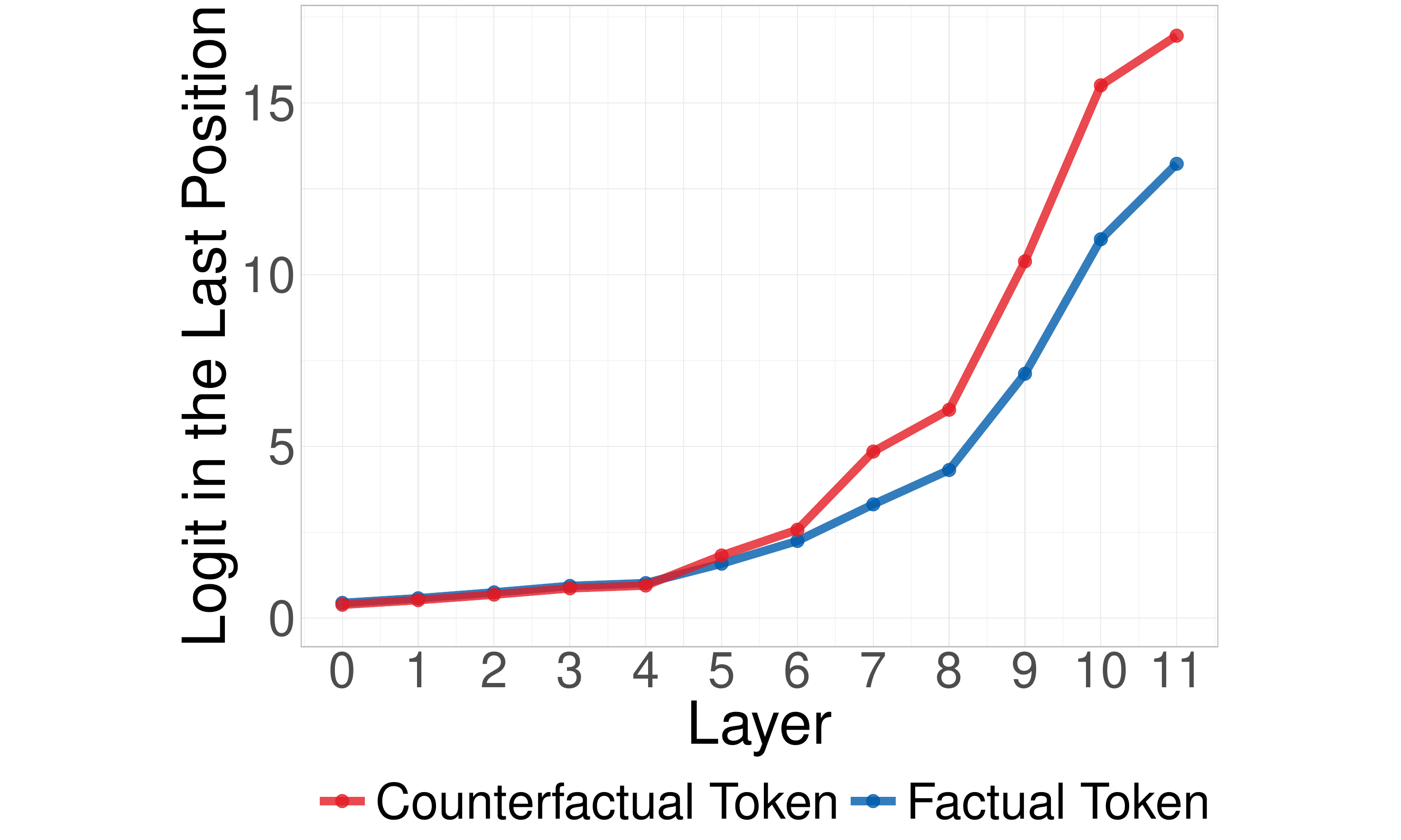}
        \caption{The average logits of $\tfact$ (blue) and $\talt$ (red) in the last token position.}\label{fig:layer_residual}
    \end{subfigure}
    \caption{
    Logits of the factual token $\tfact$ and counterfactual token $\talt$ across different positions and layers in GPT-2. 
    The logit of $\tfact$ is higher in the subject position in the initial layers and in the last position of the premise and second sentence in the final layers. 
    The logit of $\talt$ is higher in the attribute position in the first layers and in the last position of the second sentence at the end of the network. 
    }
    \label{fig:residual_stream}
\end{figure*}
\paragraph{Models}
For this work, we first choose the GPT-2 small \citep{radford2019language} model as it is the most commonly used one in previous interpretability studies  \citep[e.g.,][]{meng2022locating,wang2023IOI,conmy2023towards_automated_circuit_discovery, Hanna2023greater-then}. Aligning the same model with those studies can communicate the findings of this work better in the context of existing literature. Then, in addition to GPT-2, we check the generalizability of our work by provide supplemental results of Pythia-6.9B \citep{pythia2023} in \cref{appendix:Pythia}, to show the robustness of our findings across the two LLMs of different architectures and scales. In this way, having similar results across the two very diverse models makes the finding stronger than existing studies, most of which are only on GPT-2.

\paragraph{Implementation Details
}
As for experimental details, GPT-2 small has 117M parameters, consisting of 12 layers with 12 self-attention heads each and a residual stream of 768 dimensions.
Pythia-6.9B has 32 layers with 32 self-attention heads each and a model dimension of 4,096, with a 30x increase in the number of parameters.
For all our experiments, we deploy the pre-trained models from the Huggingface Hub \cite{wolf-etal-2020-transformers}, and inspect the residual streams by the LogitLens tool in the {TransformerLens} library \citep{nanda2022transformerlens}.

\section{Results and Findings}
\label{sec:results}
In this section, we trace the competition of the mechanisms within the LLM via the two methods introduced in \cref{sec:methods}, i.e., inspecting the residual stream and intervening on the attention.
We provide mechanistic analyses on five research questions in the following subsections:
\begin{enumerate}[topsep=0pt]
    \item Macroscopic view: Which layers and token positions contribute to the two mechanisms? (\cref{sec:residual_stream})
\item
Intermediate view: How do we attribute the prediction to attention and MLP blocks? (\cref{sec:attn_mlp_contrbution})
\item
Microscopic view:
How do individual attention heads contribute to the prediction? (\cref{sec:inspection_heads})

\item
Intrinsic intervention:
Can we edit the model activations to modify the strength of a certain mechanism?  (\cref{sec:improving_factual_recall})

\item
Behavioral analysis: What word choice varies the strength of the counterfactual mechanism in the given context? (\cref{sec:similarity_and_competition})

\end{enumerate}

\subsection{Macroscopic Inspection across Layers and Token Positions}
\label{sec:residual_stream}

In the main model that we inspect, GPT-2, we find that it can usually identify the counterfactual mechanism in 96\% of the 10K test examples.
This means that, in the last sequence position, at the output of the network, the counterfactual token, $\talt$, gets most of the times a higher probability than $\tfact$. In the following, we will inspect how the ``winning'' of the counterfactual mechanism happens across the layers of the LLM.

\paragraph{Method.}
We study how $\tfact$ and $\talt$ are encoded in the residual stream using the logit inspection method described in \cref{sec:methods}.
Specifically, for a given token position $i$ and a layer $l$, we project the embedding $\bm{x}_i^{l}$, i.e., the residual stream in \cref{eq:residual_stream}, to the vocabulary space by
$\bm{\tilde{x}}_i^{l} = W_U \cdot \mathrm{norm}(\bm{x}_i^{l}$), where $W_U$ is the unembedding matrix and $\mathrm{norm}$ is the normalization of the last layer of the model. 
By varying $l$, we measure the values of the logits of $\tfact$ and $\talt$ as they evolve in the residual stream after the first attention block.

\begin{figure*}
    \centering
    \begin{subfigure}{.47\textwidth}
    \centering
        \includegraphics[width=0.83\linewidth]{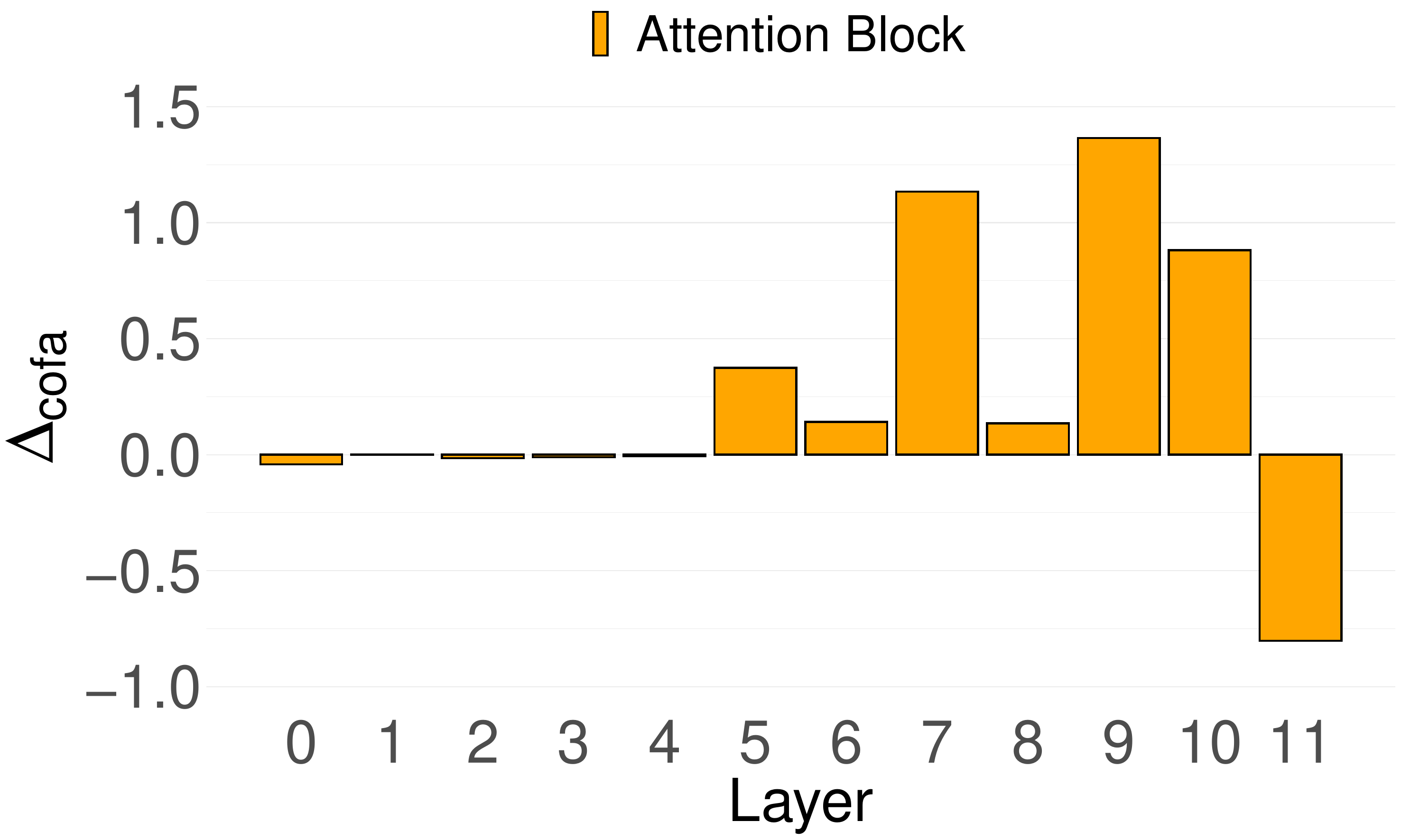}
        \caption{Logit difference $\dalt$ of the last token position after the attention block in each layer of GPT-2.}
\label{fig:attn}

    \end{subfigure}%
    \hfill
    \begin{subfigure}{.47\textwidth}
        \centering
        \includegraphics[width=0.83\linewidth]{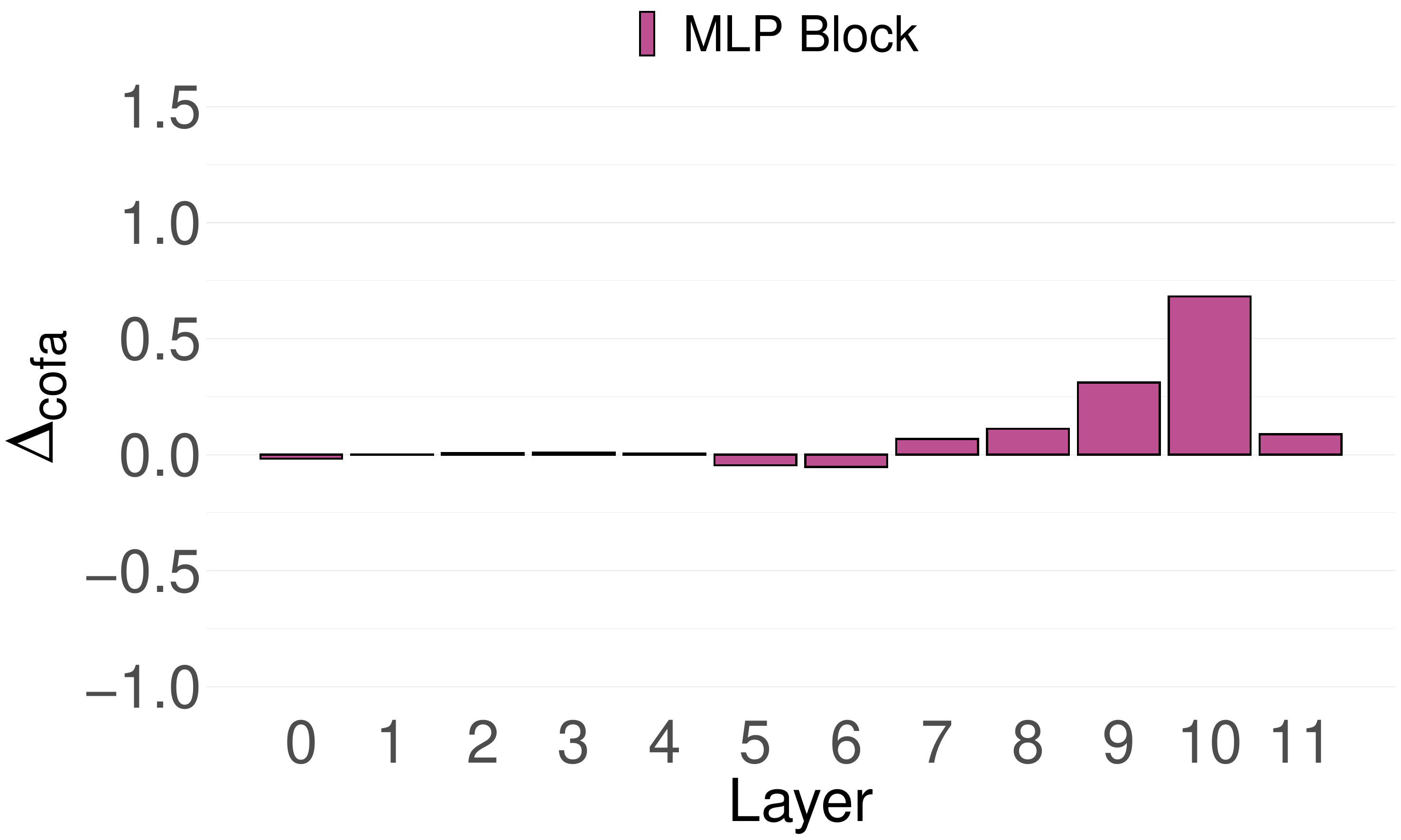}
        \caption{Logit difference $\dalt$ of the last token position after the MLP block in each layer of GPT-2.}
        \label{fig:mlp}
    \end{subfigure}%
    \caption{{Contributions of the attention and MLP blocks to the competition of the mechanisms.} 
    The attention blocks (left) contribute more to the marginal win of the counterfactual mechanism than the MLP blocks (right). 
    }
    \label{fig:attn_mlp_contribution}
\end{figure*}

\paragraph{Results.}
Our results reveal the prevalence of each mechanism by varying the layer $l$ and position $i$.

\textit{Finding 1: Information flows from different tokens for different mechanisms.}
We analyze the role of previous context at various token positions with respect to different depths of the layers. In \cref{fig:token_logit}, the blue heatmap above shows the logits of the factual token $\tfact$, and the red heatmap below shows those of the counterfactual token $\talt$. 

Looking at the blue heatmap, we see that the \textit{subject} position is the main contributor to the logits of $\tfact$ in early layers, which is consistent with a previous finding \cite{geva2023dissecting}. Specifically, we also locate the factual attribute in the subject positions by the first MLP layer, and find they increase on average the value of $\tfact$ from $0.38$ to $0.74$ in the premise and from $0.9$ to $1.93$ in the second sentence.
Then, in the later layers, the strongest contributor is the last tokens before the attribute, as the last token position is used to predict the attribute.
From the red heatmap, we see the evolution of $\talt$'s logits. The observations of later layers are similar across two mechanisms, in that the last token contributes the most.
However, in early layers, the counterfactual mechanism's $\talt$ token is best encoded in the \textit{attribute} position instead of the subject position for the factual mechanism.

Such information flow between different token positions suggests a major role played by the attention mechanism in moving such information to the last position, resonating with observations in \citet{geva2023dissecting}.

\textit{Finding 2: Both the individual mechanisms and competition take place in late, but not early layers.}
We trace the competition of the two mechanisms across the layers by plotting in \cref{fig:layer_residual} the scale of the logits corresponding to the two mechanisms in the last token position.
The first observation is that the strength of each individual mechanism increases monotonically across the layers, from a relatively small logit below 1
in early layers to large values of around 15 in the final layer.

Another observation is that, although both mechanisms increase in strength, stronger signals of the competition (where the counterfactual mechanism prevails the factual one) start after the fifth layer, and this prevalence gradually grows in later layers. The logits of the counterfactual mechanism are,
in most of the examples, the highest in the 50K-dimensional vocabulary of GPT-2, making $\tfact$ dominant in 96\% of the examples.

\subsection{Intermediate Inspection of Attention and MLP Blocks}
\label{sec:attn_mlp_contrbution}
Behind the overall win of the counterfactual mechanism, we want to trace the contributions from the attention and MLP blocks in each layer.

\paragraph{Method.}
For each attention or MLP block, it processes the input embedding and outputs the logits of $\tfact$ and $\talt$ to be added to the residual stream. We can consider the contribution of each block as its added logit values to the residual stream. Intuitively, if the added logit value for $\talt$ is higher than that of $\tfact$, then this block pushes the overall prediction to lean towards the counterfactual mechanism; otherwise, this block suppresses the counterfactual mechanism.

Hence, we inspect the margin of the added logit of $\talt$ over that of $\tfact$ in each block, represented by $\dalt := \text{BlockLogit}(\talt)-\text{BlockLogit}(\tfact)$.
To this end, we apply the logit inspection method to analyze the logit distribution at $W_U \bm{a_N}^{l}$ and $W_U \bm{m_N}^{l}$, where $N$ denotes the last token position in the sequence.
The logit contribution of the attention block is the sum over that of all the attention heads.
As for the result, a positive value of $\dalt$ for a block means that it supports the counterfactual mechanism in the competition, and a negative value indicates suppression.

\begin{figure*}
\centering
     \begin{subfigure}{.47\textwidth}
        \centering
        \includegraphics[width=0.7\linewidth]{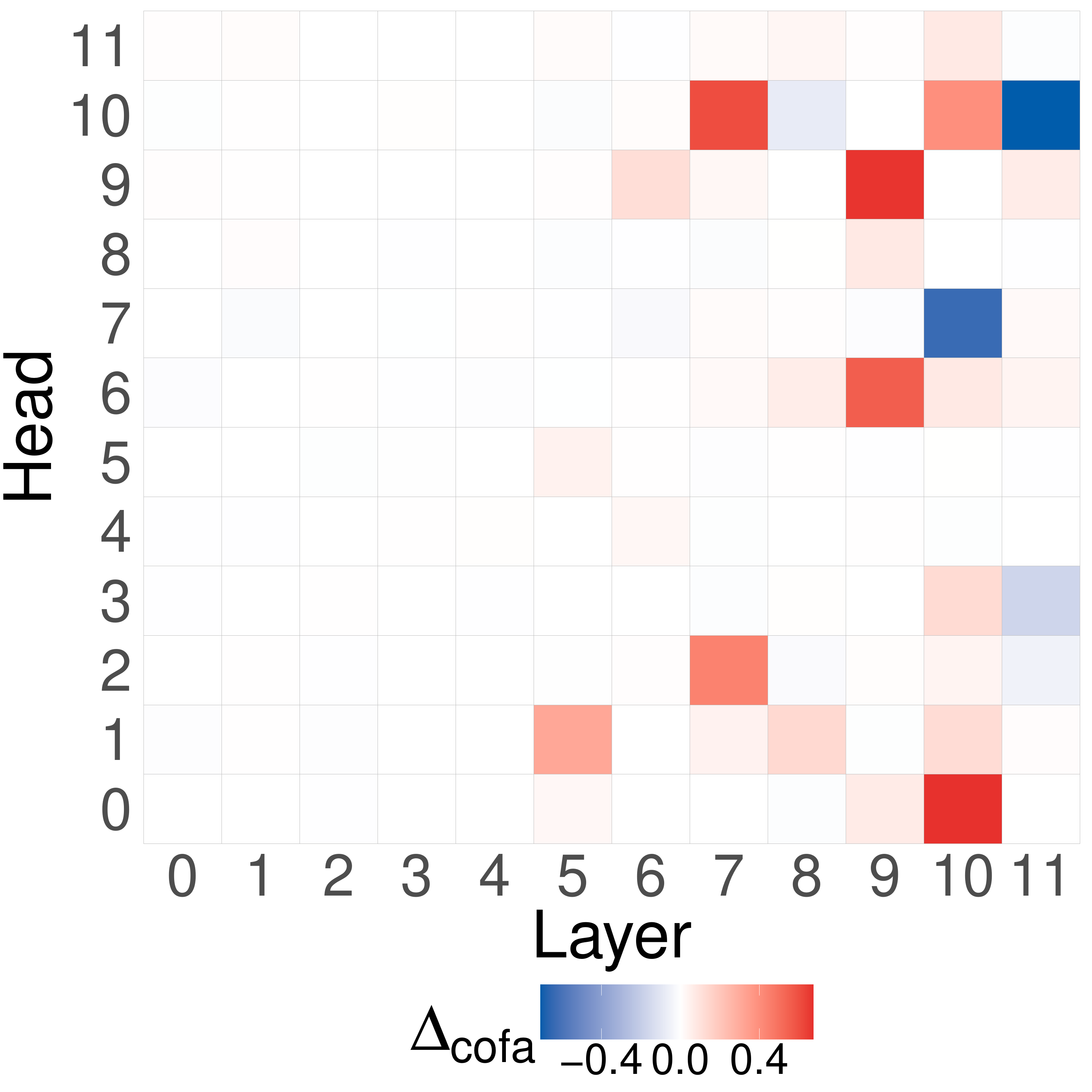}
        \caption{Direct contribution to
    $\dalt$ of all heads in GPT-2. Heads favoring $\tfact$ are colored in blue, and those favoring $\talt$ in red.}
    \label{fig:all_heads}
    \end{subfigure}%
    \hfill
    \begin{subfigure}{.47\textwidth}
    \centering
        \includegraphics[width=0.8\linewidth]{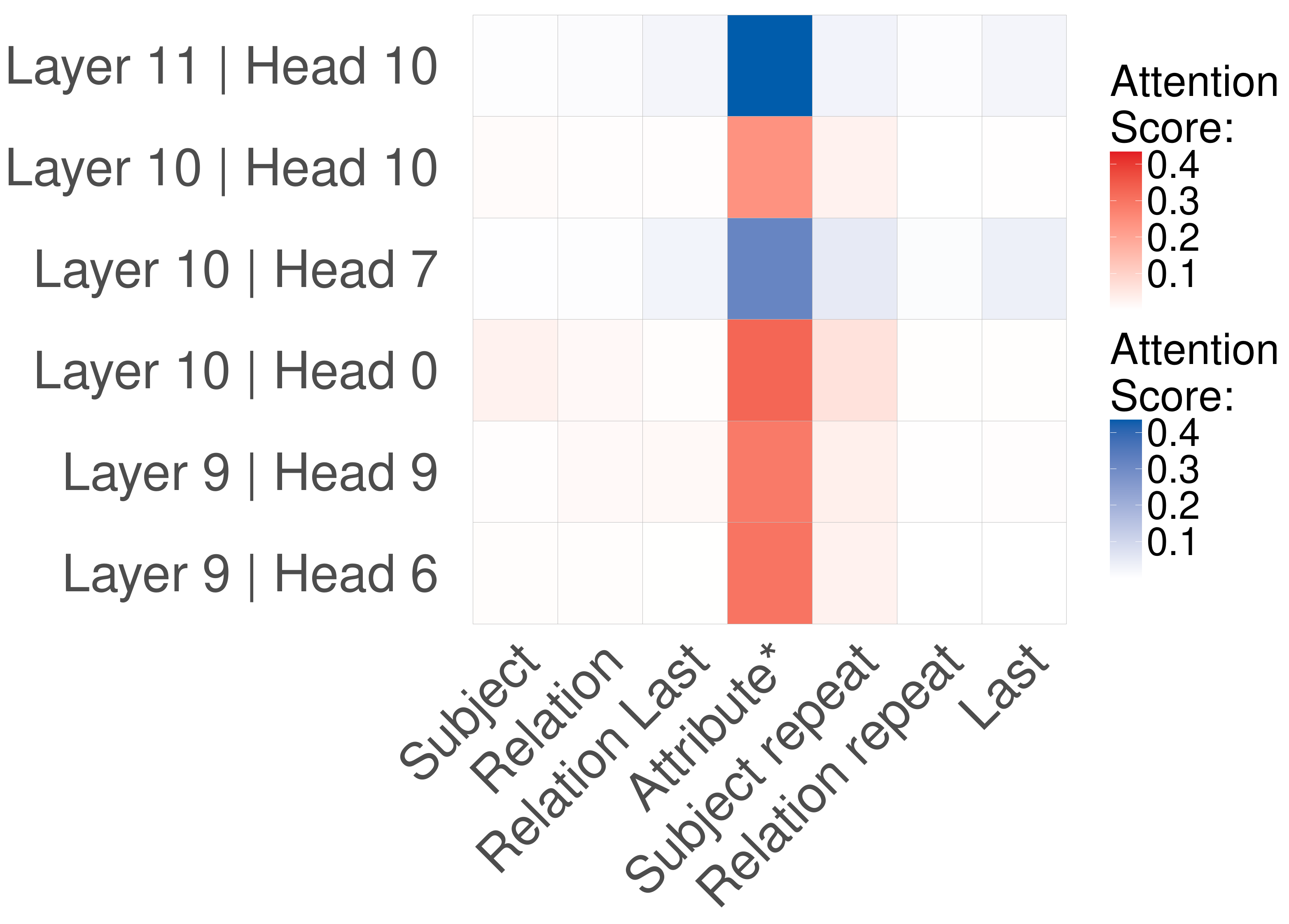}
        \caption{Attention scores of relevant heads for the last position. A large attention score in the attribute position is found in all highly activated heads.}
        \label{fig:main_heads}
    \end{subfigure}%
    \caption{{Attention pattern for relevant attention heads.}}
    \label{fig:attention_heads_analysis}
\end{figure*}
\paragraph{Results.}
We quantify the contribution of each block in each layer by plotting the $\dalt$ values in \cref{fig:attn_mlp_contribution}.

\textit{Finding 1: The attention blocks play a larger role in the competition of mechanisms than the MLP blocks.}
Contrasting the $\dalt$ margin of the added logits of the attention blocks in \cref{fig:attn} and MLP blocks in \cref{fig:mlp}, we see that the size of $\dalt$ is almost always larger in the attention blocks than in MLP blocks. This is consistent with the work of \citet{geva2023dissecting} showing that the attention blocks adds most of the information about $\talt$ to the residual stream.

\textit{Finding 2: Only late but not early layers contribute to the competition of the mechanisms.}
We find that {the early layers have almost no contribution} to the competition of the mechanisms, reflected by the close-to-zero margin $\dalt$ in Layer 0-4 for both types of blocks. However, later layers contribute to substantially to the increase of the margin $\dalt$, by a relatively smaller rate for the MLP blocks, and a larger overall rate for the attention blocks, together with a large variance.

Note that we observe a negative $\dalt$ around -0.8 in the last attention block, somewhat favoring $\tfact$, which might be since the factual information is moved to the last position in the last layers, as already noted by \citealt{geva2023dissecting}.

\subsection{Microscopic Inspection of Individual Attention Heads}
\label{sec:inspection_heads}
Beyond the overall contributions of the attention block, we further study the contribution of each individual attention head in this section.

\paragraph{Method.}
We analyze the effect of each individual attention head with the logit inspection method by projecting the outputs of each attention head to the last sequence position $N$ in the vocabulary space. 
Formally, we consider $\dalt = \text{HeadLogit}(\talt)-\text{HeadLogit}(\tfact)$ with the logits from the projection $W_U \bm{a_N}^{h, l}$ of each head $h$. Here $\bm{a_N}$ is the output of the attention head $h$ after it has been processed by the output matrix of the Attention Block but before its sum to the residual stream.

\paragraph{Results.}
We plot the contributions of individual attention heads to $\dalt$ in 
\cref{fig:attention_heads_analysis}, and introduce the main findings as follows.

\textit{Finding 1:}
\textit{A few specialized attention heads contribute the most to the competition}. As we can see from the overall contributions of all attention heads across all the layers in \cref{fig:all_heads}, several attention heads (e.g., L9H6, L9H9, L10H0, and L10H10) strongly promote the counterfactual mechanism, i.e., with a positive value of $\dalt$ colored in dark red, and two attention heads (L10H7 and L11H10) strongly support the factual mechanism instead, reflected by the large negative $\dalt$ in dark blue.

For example, the sum of L7H2 and L7H10 equals 75\% of the large positive $\dalt$ contribution of Layer 7. The sum of L9H6 and L9H9 explains 65\% of the $\dalt$ at Layer 9. 

On the other hand, the two attention heads, L10H7 and L11H10, explain almost the 70\% of the total negative contribution to $\dalt$ in the entire network (33\% and 37\% respectively). This also explains the reason behind the negative $\dalt$ in \cref{fig:attn} of the previous section.
Our study is consistent with \citet{mcdougall2023copysupression} showing that these two heads are responsible for suppressing the copy mechanisms in GPT-2 small. In our setting, the joint ablation of these two heads decreases the factual recall of GPT-2 small from 4.13\% to 0.65\%.

\textit{Finding 2: All the highly activated heads attend to the same position -- the attribute token.} 
Focusing on the heads with large absolute values of $\dalt$, we show the attention scores of the last position $N$ to different tokens in \cref{fig:main_heads}. 
Expectedly, the major heads supporting the counterfactual mechanism (those in red) attend to the attribute position because they need to copy this token for the prediction, which echoes the findings in \cref{sec:residual_stream}.

However, it is surprising to see the other heads supporting the factual mechanism (those in blue) also mainly attend to the counterfactual attribute token.
We find that those heads
read from the attribute position
to give a lower value to the logit of $\talt$, which might be an easier operation for it to learn than increasing the logit of the factual token. 
The evidence is that, in these two heads, the logit of $\tfact$ 
is smaller than the mean of the two layers, but the logit of $\talt$ (which is -1.13 for L10H7 and -1.05 for L11H10) are the lowest of all the heads in the network.

We include supplementary analyses showing the consistency of Finding 2 on Pythia in \cref{appendix:Pythia}, and provide
the full attention maps with attention scores between every pair of tokens for these heads in Appendix \cref{app-subsec:full_attn_pattern_gpt2}.

\subsection{Intrinsic Intervention by Attention Modification}
\label{sec:improving_factual_recall}
After \textit{understanding} where the two mechanisms take place, we use the insights to \textit{intervene} on the model internal states. Specifically, we perform model editing to alter the factual mechanism, which concentrates on a few strongly activated attention heads (L10H7 and L11H10 in GPT-2, and mostly L17H28, L20H18, and L21H8 in Pythia, see \cref{app-subsec:attention_heads_pythia}),
and has most of the information flowing from the attribute position (see \cref{fig:attention_heads_analysis}-right and \cref{sec:attn_mlp_contrbution}).
In the following, we show that enlarging the value of a few well-localized attention values can largely improve the factual recall of the model.

\paragraph{Method.}
We utilize the attention modification method in \cref{eq:alpha} to apply a multiplier of
$\alpha$ to the attention weights of the last token to the attribute position in L10H7 and L11H10 for GPT-2, and L17H28, L20H18, and L21H8  in Pythia. 
To choose the $\alpha$ value, we perform a grid search over $[2, 5, 10, 100]$ to maximize the factual recall rate of the model. We find that $\alpha = 5$ is the best value for both GPT-2 for Pythia.

\begin{figure}[t]
    \centering    \includegraphics[width=.5\linewidth]{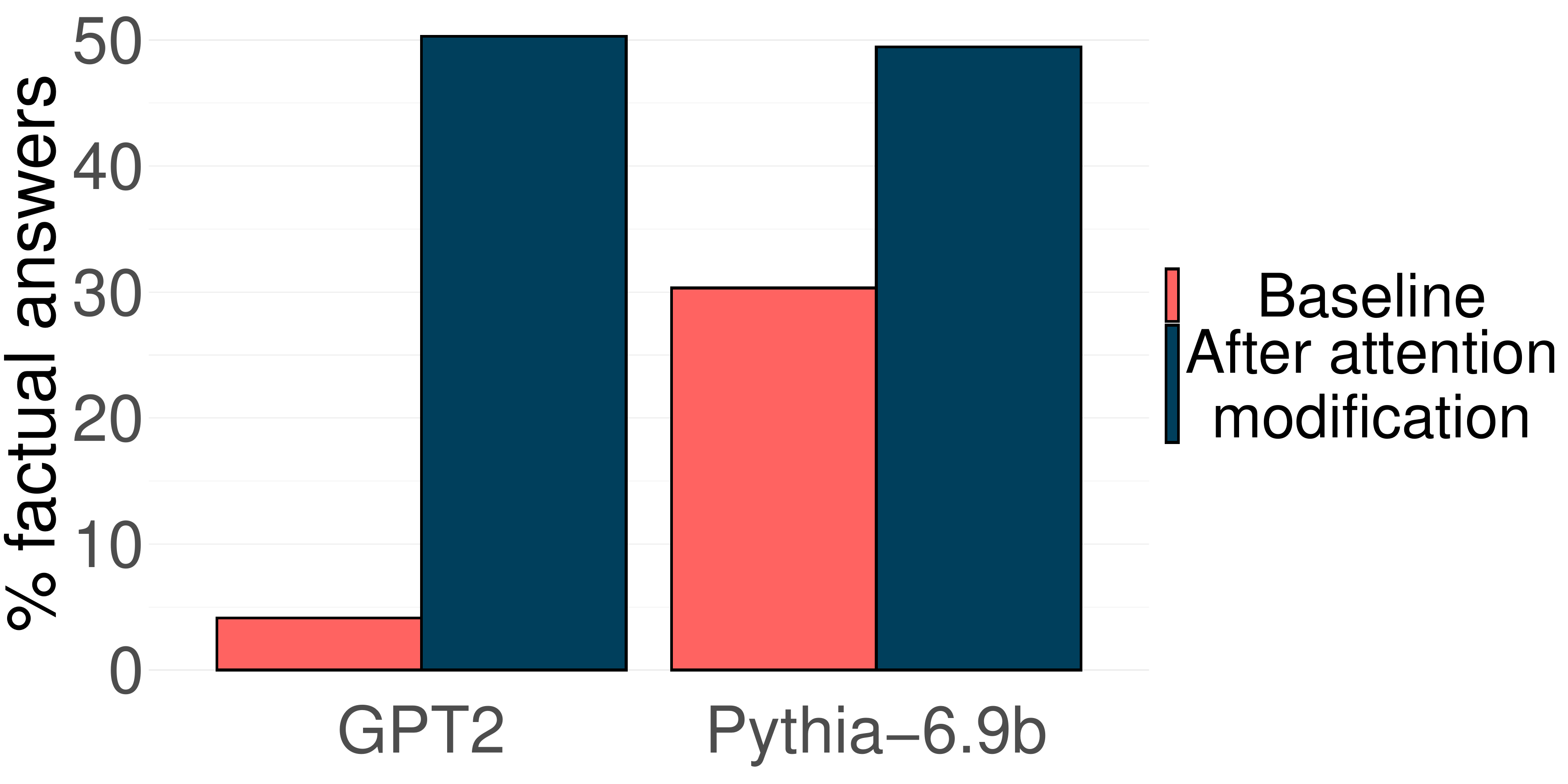}
\caption{{The factual recall mechanism increases substantially across GPT-2 and Pythia after attention modification.} 
}
\label{fig:improved_factual_recall}
\end{figure}

\paragraph{Results.}
We highlight the effect of our model editing method on the strength of the factual recall mechanism in
\cref{fig:improved_factual_recall}.
Originally, GPT-2 has only 4\% of the cases where the factual mechanism prevails the counterfactual one, and Pythia only 30\%. However, after modifying the attention weights of the entries mentioned above, the strength of the factual mechanism increases drastically that it wins over the other mechanism in 50\% of the cases for both models.
This result is remarkable since we modify only two entries in the attention map out of the 33,264 attention values of GPT-2 (117M parameters) and three entries out of the 270,848 attention values of Pythia (6.9B parameters). 
This highlights the importance of the interpretability analysis in \cref{sec:inspection_heads,sec:attn_mlp_contrbution}, which enables us to find the detailed role played by the individual units of each transformer layer.

\subsection{What Word Choices Intensify the Competition?}
\label{sec:similarity_and_competition}
After the intrinsic intervention to edit the internal states of the model, we explore how the similarity between $\tfact$ and $\talt$ in our dataset affects the mechanism described in the previous sections.

\paragraph{Method.}
We divided the dataset into 10 equal bins based on the similarity between the vectors for 
$\tfact$ and $\talt$, with each bin containing $1000$ items. Starting from the lowest, each group represents a $10\%$ segment of the dataset, arranged by increasing similarity scores. %
For our word similarity metric, we calculate the
cosine similarity of the 300-dimensional word embeddings from the pre-trained Word2Vec model \citep{mikolov2013word2vec} implemented in the Gensim Python package \citep{gensim}.
\begin{figure}[t]
    \centering
    \includegraphics[width=0.5\linewidth]{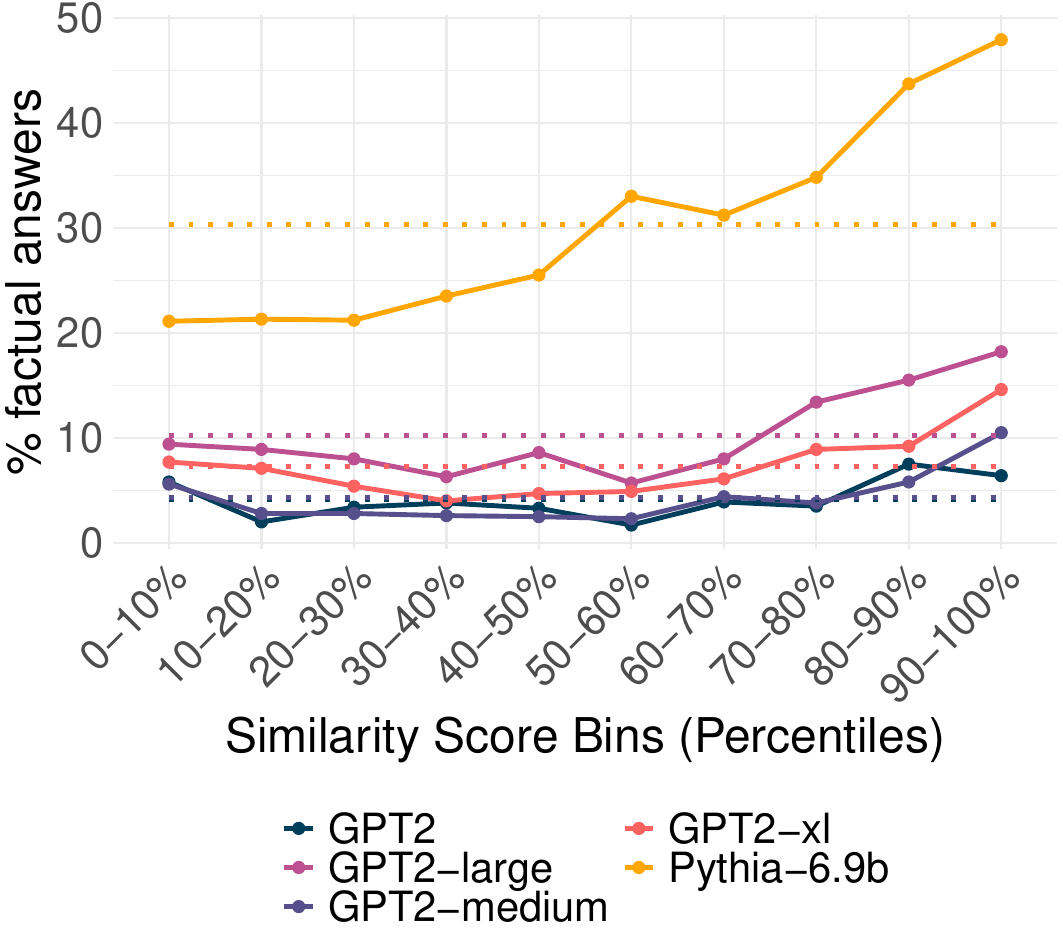}
\caption{{Prediction frequency of factual token by similarity level}. We show the percentage of $\tfact$ predictions within each bin compared to the entire dataset (represented by a dotted line) across various model sizes. 
We can notice that more similar $\tfact$ and $\talt$ are, and the factual mechanism is stronger.}
\label{fig:similarity_semantics}
\end{figure}

\paragraph{Results.}
As a result of the varying similarity of the two tokens, we see a drastic change in the dominance of the factual mechanism in 
\cref{fig:similarity_semantics}.

\textit{Finding 1: Similar tokens confuse the model more easily.}
Consistently across all the models, the more similar the two tokens are, the more likely the model is to be confused and mistakenly let the factual mechanism dominate, to predict $\tfact$ as its output.

\textit{Finding 2:}
\textit{Larger models suffer more from such confusion.} For example, the largest one, Pythia-6.9B, demonstrates a very strong attachment to the factual information, letting the factual mechanism win almost 45\% of the cases when the token similarity reaches 90\%.
Even when the similarity is low, larger models are still more likely to confuse and lean towards the factual mechanism. This finding resonates with the observations from the inverse scaling prize \cite{mckenzie2023inverse} that larger models have a greater capacity to store and retrieve factual information, thus more influenced by the factual mechanism.

\section{Discussion and Future Work}

\paragraph{Situating Our Findings in Related Work.}
Our findings about the late attention blocks are consistent with \citet{geva2023dissecting}, showing that late attention blocks write most of the information to the last layer when adding a counterfactual premise. 
Surprisingly, however, we find that the largest contribution to the factual prediction of the network mostly comes from the suppression of the counterfactual token read from the attribute position rather than the promotion of the factual token from the subject position. 

Consistently with \citet{mcdougall2023copysupression}, we find that few highly specialized heads suppress the counterfactual information. Moreover, we make a unique contribution up-weighting only two or three attention entries of these heads to increase substantially the number of factual responses of the model.

With an approach similar to ours, \citet{yu2023characterizing} find that more heads can promote the factual mechanism, also in early layers, but found it challenging to improve the factual responses by scaling up the weights of the attention maps. 
This discrepancy can be due to the broader set of topics we include in our prompts, which allowed us to select fewer, more specialized heads, to the different ways the prompts are framed, or also to our more focused modification of the attention maps.

\paragraph{Future Work}
For future research directions, we aim to analyze more in depth how our findings depend on the prompt structure and whether the promotion of factual responses by suppressing the counterfactuals generalizes to larger models and a more comprehensive variety of datasets.

\section{Conclusion}
In this work, we have proposed the formulation of the \textit{competition of mechanisms} as a powerful interpretation when LLMs need to handle multiple mechanisms, only one of which leads to the correct answer. We deployed two mechanistic interpretability tools, logit inspection and attention modification, and identified critical positions and model components involved in competing for the mechanisms. Finally, we discovered a few localized positions in the attention map, which largely control the strength of the factual mechanism. Our study sheds light on future work on interpretability research for LLMs.

\section*{Limitations}
\label{sec:limitations}
\textit{Limited models:}
Our study aligns with most existing work in mechanistic interpretability to use GPT-2 small. However, we understand that this is a small model with far fewer parameters than current state-of-the-art LLMs. Future work is welcome to extend to larger-sized models, which might generalize our conclusion to a certain extent, and also reveal interesting behavior once the models get beyond a specific size, maybe also seeing a U-shaped curve \cite{wei2023ushapescaling} for the dominance of the counterfactual mechanism.

\textit{Interpretability method:}
Furthermore, our experiments and insights are heavily grounded in the interpretability within the embedding space of the model's inner components. This approach is reliable and extensively employed in mechanistic interpretability research \citep{dar2023embeddingspace, geva-etal-2022-transformer, halawi2023overthinking_the_truth}.
The logit inspection method, although commonly employed in previous work,
can occasionally fail to reflect the actual importance of some vocabulary items, especially in the early layers of the network \citep{belrose2023tunedlens}.

\textit{Simplicity of the prompts:}
Our prompts have a relatively simple structure for the controllability of the counterfactual information tracing, as it is very challenging to follow the information flow in a more diversified set of prompts. We welcome future work to explore methodological advances to enable analyses over more diverse prompts.

\section*{Ethical Considerations}
The aim of our study is to enhance comprehension of the interplay among mechanisms within language models that may yield unforeseen and undesirable outcomes. Additionally, our research serves as a conceptual demonstration of methods to guide model behavior under such conditions. We believe that recognizing and dissecting the mechanisms by which LLMs produce unpredictable responses is crucial for mitigating biases and unwanted results. Moreover, understanding the competitive dynamics under investigation is critical for improving the safety of LLMs. Specifically, inputting a prompt with an inaccurate redefinition may lead the model to inadvertently reveal sensitive factual information.

\mainchapter{A Causal Framework to Quantify the Robustness of Mathematical Reasoning in LLMs}\label{ch:mathrobust}

We have recently witnessed a number of impressive results on hard mathematical reasoning problems with language models. At the same time, the robustness of these models has also been called into question; recent works have shown that models can rely on shallow patterns in the problem description when generating a solution.
Building on the idea of behavioral testing, we propose a novel framework, which pins down the causal effect of various factors in the input, e.g., the surface form of the problem text, the operands, and math operators on the output solution.
By grounding the behavioral analysis in a causal graph describing an intuitive reasoning process, we study the behavior of language models in terms of robustness and sensitivity to direct interventions in the input space. We apply our framework on a test bed of math word problems.
Our analysis shows that robustness does not appear to continuously improve as a function of size, but
the GPT-3 Davinci models (175B) achieve a dramatic improvement in both robustness and sensitivity compared to all other GPT variants.
Our code and data
are available at \url{https://github.com/alestolfo/causal-math}.

\section{Introduction}
\begin{figure}[t]
    \centering
    \includegraphics[width=.6\columnwidth]{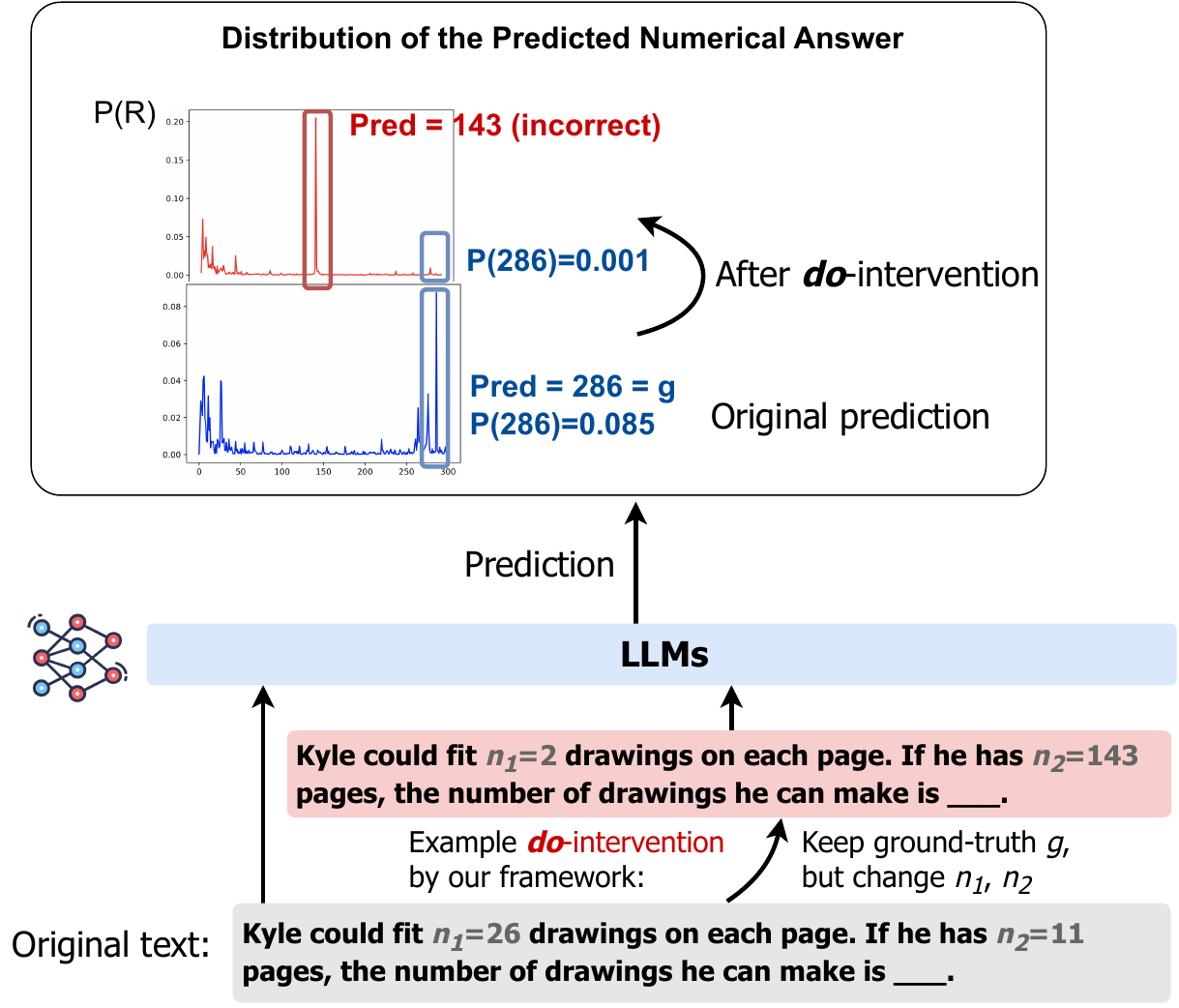}
    \caption{Through our framework, we conduct $\mathrm{do}$-interventions on the input and evaluate the change in the distribution $\mathbb{P}(R)$ of the prediction $R$ by LLMs, in this figure, GPT-J. This allows us to measure the causal effect of each factor in the input on the model's response.}
    \label{mathrobust:fig:intro}
\end{figure}
Many natural language understanding situations, such as understanding the financial news, require reasoning with text that includes numbers.
However, such mathematical reasoning is challenging for NLP models \citep{cobbe2021training, mishra-etal-2022-numglue}. %
Mathematical reasoning for text has been an active area of research for a while \citep[\textit{inter alia}]{seo-etal-2015-solving,sachan2017learning,sachan2017textbooks,sachan2018learning}, and has also emerged as a key task to track the capabilities of large language models (LLMs) in recent years \citep[\textit{inter alia}]{brown2020gpt3,ouyang2022instructGPT,wei2022emergent}.

\begin{figure*}[t]
    \centering
    \includegraphics[width=0.95\textwidth]{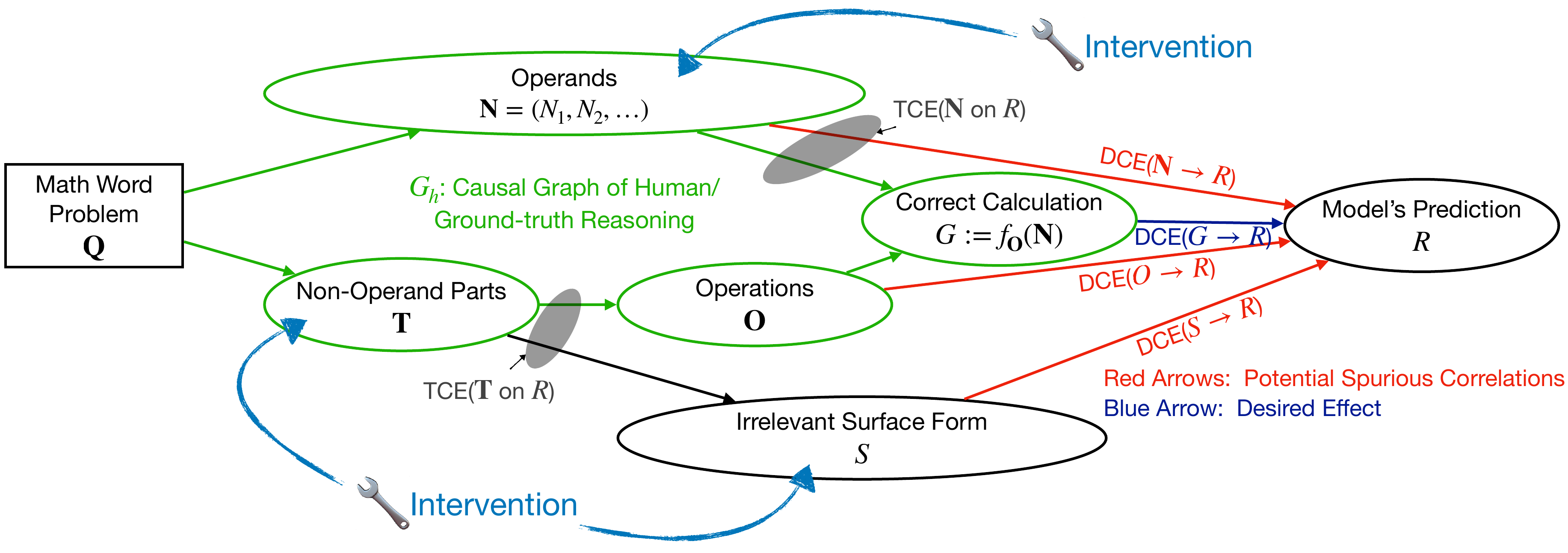}
    \caption{Causal graph of model predictions on math questions. We highlight the difference between a cognitively-inspired correct reasoning path (\green{$\mathcal{G}_h$}) and the undesired effects that some factors might have on the model's prediction (red arrows). By performing controlled interventions of the numerical values ($\bm{N}$) and on the textual framing of the problem ($\bm{T}$, $S$), we are able to quantify the causal effects of each factor.
    }
    \label{mathrobust:fig:causal_graph}
\end{figure*}

However, despite the impressive performance of LLMs on various math reasoning benchmarks \citep[e.g.,][]{ouyang2022instructGPT,chowdhery2022palm}, it remains unclear whether these models have %
learned mere artifacts in the data
or have truly mastered the mathematical concepts needed to consistently solve all variations of the same problem \citep{patel-etal-2021-nlp, razeghi-etal-2022-impact,welleck2022symbolic}. In sharp contrast with a large number of papers on improving the performance of LLMs on various types of math-based problems, there has been little effort on behavioral analysis of LLMs for these tasks. Existing methods for understanding the robustness of these models \citep{patel-etal-2021-nlp} rely on manually constructing variations of math problems, and we do not yet have a principled, comprehensive framework for quantifying such robustness.%

Thus, in this work, we propose a formal framework based on causal inference, to quantify the robustness of NLP models' math reasoning abilities. Specifically, we describe a causal graph formulation of math reasoning, where the graph allows us to measure the difference in the structural causal models of human reasoning and model judgment. We consider various causal factors such as the textual framing of the question, numerical operands, and operation types. Then, we identify a set of interventions  in the context of math word problems (an example of which is illustrated in \cref{mathrobust:fig:intro}), and provide a causal inference framework to obtain causal effects of each factor via direct $\mathrm{do}$-interventions \citep{pearl1995causal} and causal mediation analysis \citep{pearl2001direct}. While our approach is reminiscent of recent studies using causal analysis for LLMs \citep{finlayson-etal-2021-causal,vig2020investigating,meng2022locating}, in this work, we provide a new theoretical analysis framework specifically suitable for math reasoning.
Using our framework, we disentangle factors affecting the model's predictions and measure their influences. This way, we are able to provide insights into the model’s reasoning in terms of \emph{robustness} and \emph{sensitivity} with respect to changes in these factors.

We apply our framework to study a set of thirteen GPT models with various sizes and training procedures (i.e., instruction-tuned and non-instruction-tuned).
We observe that, among non-instruction-tuned language models, the larger ones tend to be more sensitive to changes in the ground-truth result of a math word problem, but not necessarily more robust.
However, we observe a different behavior in the instruction-tuned GPT-3 models \citep{ouyang2022instructGPT}, which show a remarkable improvement in both sensitivity and robustness, although the robustness reduces when problems get more complicated.
We additionally investigate the role of size and instruction tuning on the model's performance with three models of the LLaMA family \citep{touvron2023llama} and Stanford Alpaca \citep{taori2023alpaca}.

\section{Problem Setup}
\label{mathrobust:sec:problem_setup}

We consider a dataset $\mathcal{D}$ of math word problems (MWPs), where each MWP is denoted as a question $\bm{Q}$. $\bm{Q}$ is a list $(\bm{T}, \bm{N})$ consisting of a question template $\bm{T}$ and an ordered list of operands $\bm{N} = (N_1, N_2, \dots, N_m)$. Each question template $\bm{T} := (\bm{O}, S)$ further contains two types of information: a set of arithmetic operations $\bm{O}$ implicitly expressed in the question, and the text surface form $S$ irrelevant to the arithmetic operations. $\bm{O}$ incorporates the information relative to the operations as a collection of tuples $\{(O_1, i_1, j_1), (O_2, i_2, j_2), \dots\}$, where $O_k \in \{+,-,\times,\div \}$ ($k \in \mathbb{N}$) and $i_k, j_k \in \mathbb{N}$ represent the indices of the operands to which operator $O_k$ should be applied.\footnote{The intermediate result of operation $O_l$ is indicated by $i_k = m + l$.}
The ground-truth result $G = f_{\bm{O}}(\bm{N})$ is calculated by computing the function $f_{\bm{O}}$, which represents the application of all the operators in $\bm{O}$ to the respective operands.
We illustrate the factors in $\bm{Q}$ and their inter-dependency in the causal graph in \cref{mathrobust:fig:causal_graph}.
A two-operand instance $\bm{q}$ of $\bm{Q}$ in this form from \citet{patel-etal-2021-nlp} is:
\begin{quote}
    \textbf{Template $\bm{t}$}: Mark has $n_1$ trees in his backyard. If he plants $n_2$ more,
    how many trees will he have? \\
    \textbf{Operands $\bm{n}$}: $(n_1 = 12, n_2 = 13)$ \\
    \textbf{Operations} $\bm{o}$: \{(``$+$'', 1, 2)\} \\
    \textbf{Result}: $g= f_{\bm{o}}(\bm{n}) = n_1 + n_2 = 25$ \\
\end{quote}%

Our goal is to quantify the robustness of a model $\mathcal{M}$ on the set of problems $\bm{q} \in \mathcal{D}$. Ideally, $\mathcal{D}$ should be a dataset not seen by the model during training.
We assume that a model takes $\bm{q}$ as input and predicts a probability distribution of the result $R$: $\mathbb{P}(R \ |\ \bm{t}, \bm{n})$.
Our formulation below will be easier to understand using this finite discrete set and can be generalized to any kind of data pairing a natural language template with a function that maps a set of operands to a result (e.g., a Python program; \citealt{mishra2022lila}).

\section{A Causal Framework}
In this section, we describe our framework in three steps. First, we define the idea of model robustness on MWPs.
Then, we identify possible $\mathrm{do}$-interventions \citep{pearl1995causal} that we can perform. Finally, we describe the causal effects that we measure to quantify the robustness of various models.

\subsection{Step 1. Question Reformulation}
We address the research question ``\textit{Is a model reasoning robustly on MWPs?}'' by comparing the causal mechanisms of the model's decisions to a hypothesized human reasoning mechanism.
Note that we do not claim to know how humans reason about these problems. We simply propose a reasonable and intuitive way to judge model robustness given a reasonable and intuitive human reasoning mechanism inspired by findings regarding the independence of language and mathematical reasoning in humans \citep{doi:10.1073/pnas.0500328102, monti2012thought}.

\paragraph{Human Reasoning Mechanisms.} The causal mechanisms of how humans might solve $\bm{q}$ include
\begin{align}
    \bm{o} & = f_{\mathrm{abstract}}(\bm{q})
    ~,
    \\
    g & = f_{\bm{o}}(\bm{n})
    ~,
\end{align}
where they first abstract the arithmetic operations $\bm{o}$ from the problem $\bm{q}$ by some cognitive process $f_{\mathrm{abstract}}$, and then apply the operation to the operands to obtain the result $g$. We show these mechanisms in the green subgraph \green{$\mathcal{G}_h$} of \cref{mathrobust:fig:causal_graph}.

\paragraph{Model Reasoning Mechanisms.} In contrast, the causal mechanisms of how a model might solve $\bm{q}$ are as follows:
\begin{align}
    r = f_{\mathrm{blackBox}} (\bm{t}, \bm{n})
    ~,
\end{align}
where we are unsure about (1) \textit{what} part(s) of $\bm{t}$ the model takes into account, and (2) \textit{how} it operates over the relevant variables.

Thus, we draw all possible causal mechanisms that might take place in the black-box model $f_{\mathrm{blackBox}}$ in the complete causal graph in \cref{mathrobust:fig:causal_graph}. Some possible fine-grained causal mechanisms are
\begin{enumerate}[nolistsep]
    \item The model might attend over the question template $\bm{t}$ in two ways: paying attention to the text surface form $s$ via the causal path $\bm{T} \rightarrow S \rightarrow R$, or text relevant to the math operations $\bm{o}$ via the causal path $\bm{T} \rightarrow \bm{O} \rightarrow R$.
    \item The model might also attend to the operands $\bm{n} := (n_1, n_2, \dots)$ via a causal path $\bm{N} \rightarrow R$.
    \item If the model learns the correct causal mechanisms as in the human cognitive process, it should capture how the operator and the operands matter to the ground-truth result $g$ (via $\bm{O} \rightarrow G$ and $\bm{N} \rightarrow G$) and then the model prediction should be sensitive to any changes in the ground truth, namely $G \rightarrow R$. No spurious correlations can directly affect $R$ without going through the mediator $G$.
\end{enumerate}

Hence, to answer the question ``How robust is the mathematical reasoning of a model on MWPs?'' we can answer the following subquestions:
\begin{enumerate}
    \item How does $R$ change in response to $G$? By quantifying this, we assess the \emph{sensitivity} (correct responsiveness) of the model to changes in the problem. In other words, does the model correctly adjust its prediction in response to a change in the correct solution of the problem?
    \item What is the (unwanted) direct causal effect size of $S \rightarrow R$,  and $\bm{N} \rightarrow R$? We see the quantities as a measure of the \emph{brittleness} (i.e., wrong responsiveness) of the model to result-preserving changes in the input. The lower the direct causal effect of $S$ and $\bm{N}$, the more \emph{robust} the model is.
\end{enumerate}

\subsection{Step 2. Causal Intervention List}\label{mathrobust:sec:quantities_of_interest}
After formulating the cognitively-inspired subgraph \green{$\mathcal{G}_h$} and defining the undesired causal paths in \cref{mathrobust:fig:causal_graph}, we list all feasible limited actions that allow us to perform our causal analysis. In the context of MWPs, we use the following interventions:
\begin{enumerate}
    \item Direct intervention on all possible $n_1, n_2, \dots$;
    \item Partially controllable interventions on $\bm{T}$. We can replace the template $\bm{T}$ in two ways:
    \begin{enumerate}
        \item both $S$ and $\bm{O}$ are affected, or
        \item $S$ is affected but $\bm{O}$ is not affected.
    \end{enumerate}
\end{enumerate}

\subsection{Step 3. Turning Limited Actions into Causal Effect Sizes}
Next, we explain how we can obtain the causal effect sizes we want (listed in Step 1) from the limited set of interventions we can do (listed in Step 2).
Specifically, we first start from all the feasible interventions, and for variables that we cannot directly intervene on, we
apply deductions from $\mathrm{do}$-calculus~\citep{pearl1995causal} to obtain or approximate the direct causal effect sizes.
In the following, we describe a list of causal effect sizes that we need.

\paragraph{General Formulation.}

Let us consider an intervention $\mathrm{do}(X: x \rightarrow x')$, where $X \in \{\bm{T}, S, \bm{N}\}$ and a problem $\bm{Q} = \{\bm{T}, \bm{N}\}$.
The support of the numerical values $N_i$'s and $R$ is $\mathcal{I} \subseteq \mathbb{N}$, and we consider $\bm{N}$ to be distributed uniformly over the set $\{ \bm{n} \in \mathcal{I}^2 \ | \ f_{\bm{O}}(\bm{n}) \in \mathcal{I}\}$.
We denote the distribution before the intervention $\mathbb{P}(R \ | \ \bm{T}, \bm{N})$ as $P$ and the distribution after the intervention
as $P'$.

Following the distributional definition of causal effect by \citet{pearl1995causal},
we quantify the effect of factor $X$ in our causal graph using a distance metric $\delta$ between the distributions $P$ and $P'$. That is,
\begin{align}
    \mathrm{CE} = \delta(P, P'),
\end{align}

where $\mathrm{CE}$ can refer to the \textbf{total causal effect} (TCE, i.e., the joint effect through all the directed causal paths from a variable to another), or the \textbf{direct causal effect} (DCE, i.e., the effect from the directed causal path from a variable to another that does not go through any intermediate variables) \citep{pearl2001direct}.
We describe our choices for $\delta$ in \cref{mathrobust:sec:distr_diff_metrics}.

\paragraph{Causal Effects of the Operands.}
When intervening on the operands $\bm{N}:=(N_1, N_2, \dots)$, we can obtain the size of the total causal effect of $\bm{N}$ on $R$, namely
\begin{align}
    \label{mathrobust:eq:tce_n_r}
     & \mathrm{TCE}(\bm{N} \text{ on } R)
    := \mathbb{E}_{\bm{n}'\sim \mathbb{P}(\bm{N})} [\delta(P, P')],\\
    & \text{where } P' = \mathbb{P}(R|\bm{T}, \mathrm{do}(\bm{N} = \bm{n}'))
    ~.
\end{align}
Note that this TCE is not the exact desired quantity, because we want to separate two different paths of how $\bm{N}$ affects $R$: (1) the path $\bm{N} \rightarrow G \rightarrow R$, which is the correct decision path that we want the model to pick up (where the model reacts to the change in the ground-truth answer), and (2) the path $\bm{N} \rightarrow R$, which is the spurious correlation that the model might have learned (where the model relies on some spurious correlations with certain numerical values, which could be traced to perhaps their frequencies in the training corpus).

We can quantify the \textbf{direct causal effect} (DCE, i.e., the effect from the directed causal path from a variable to another that does not go through any intermediate variables) \citep{pearl2001direct} of $\bm{N}$ on $R$, namely the strength of the direct causal path $\bm{N} \rightarrow R$, by controlling for $G$ to be fixed every time we intervene on $\bm{N}$:
\begin{align}
    \label{mathrobust:eq:dce_n_r}
     & \mathrm{DCE}(\bm{N} \rightarrow R)
    := \mathbb{E}_{\bm{n}' \sim \mathbb{P}(\bm{N} | G)} [\delta(P, P')],\\
    & \text{where } P' = \mathbb{P}(R|\bm{T}, \mathrm{do}(\bm{N} = \bm{n}'))
    ~.
\end{align}
For example, if we observe a model doing $100+100=200$ correctly, we want to separate the math ability here into (1) the model's sensitivity towards the ground-truth answer, and (2) the model's decisions based on its familiarity with just the operand $100$.
Here, the overall effect is the calculable $\mathrm{TCE}(\bm{N} \text{ on } R) $ by \cref{mathrobust:eq:tce_n_r}, and one of the subeffects is the calculable $\mathrm{DCE}(\bm{N} \rightarrow R)$ by \cref{mathrobust:eq:dce_n_r}.

\paragraph{Causal Effects of the Text Surface Form.}

As for the operands, we can compute both the direct and indirect effects of the surface form representing the math problem. In particular, intervening on $\bm{T}$ without controlling for $\bm{O}$ (intervention 2a in \cref{mathrobust:sec:quantities_of_interest}), we can compute the total effect, i.e.,
\begin{align}
     & \mathrm{TCE}(\bm{T} \text{ on } R)
    := \mathbb{E}_{\bm{t}'\sim \mathbb{P}(\bm{T})} [\delta(P, P')],\\
    & \text{where } P' = \mathbb{P}(R|\bm{N}, \mathrm{do}(\bm{T} = \bm{t}'))
    ~.
\end{align}

Controlling for the operations $\bm{O}$ (intervention 2b in \cref{mathrobust:sec:quantities_of_interest}) will instead allow us to obtain the direct causal effect of the surface text:
\begin{align}
     & \mathrm{DCE}(S \rightarrow R)
    := \mathbb{E}_{\bm{t}'\sim \mathbb{P}(\bm{T} | O)} [\delta(P, P')],\\
    & \text{where } P' = \mathbb{P}(R|\bm{N}, \mathrm{do}(\bm{T} = \bm{t}'))
    ~.
\end{align}
Note that since there is no mediator between $S$ and $R$, the $\mathrm{DCE}(S \rightarrow R)$ is also TCE of $S$ on $R$.
The only adaptation that we need to make with regard to the MWPs is that it is not feasible to enumerate all possible perturbations of $S$. Therefore, the practical results that researchers can achieve are over a certain subset of $S$. In practice, we obtain this by intervening on $\bm{T}$ without affecting $\bm{O}$.

\paragraph{Causal Effects of the Operators.}

The ideal way to obtain the TCE of $\bm{O}$ on $R$ is through some careful human annotation that minimally changes the templates as \citet{kaushik2020learning} do for sentiment classification.
The challenge for MWPs in our case is that with all our possible interventions, we cannot \textit{only} intervene on $\bm{O}$ without introducing changes to the irrelevant surface form.
However, we might get some information about
$\mathrm{TCE}(\bm{O} \text{ on } R)$ because, on the causal graph, the total causal influence of $\bm{T}$ on $R$ actually flows into two directed paths, one through $S$ to $R$ (which is the $\mathrm{DCE}(S \rightarrow R)$), and the other from $\bm{O}$ to $R$, which is our interested quantity $\mathrm{TCE}(\bm{O} \text{ on } R)$. Therefore, we compare the two quantities we know, $\mathrm{TCE}(\bm{T} \rightarrow R)$ and $\mathrm{DCE}(S \rightarrow R)$, to get a sense of the causal influence of $\bm{O}$ on $R$ that we cannot obtain in any other way.

\subsection{Step 4. Quantifying the Causal Influence}\label{mathrobust:sec:distr_diff_metrics}

Consider a realization of problem $\bm{Q}$ with operands $\bm{n}$ and ground-truth result $g = f_{\bm{o}}(\bm{n})$, and denote by $g'$ the result after the intervention $\mathrm{do}(X: x \rightarrow x')$.
We quantify the causal effect of factor $X$ on the model's prediction $R$ in two ways: by assessing the change in the predicted result, and by measuring the change in the probability assigned by the model to the correct result $g$ (or $g'$).

\paragraph{Change in the Prediction}
To account for the inability of LMs to capture the continuous property of numbers \citep{jin2021numgpt}, we measure the change in the model's prediction using an indicator of the ``change result'' event:
\begin{equation}
    \delta_{\mathrm{cp}} ({P}, {P'})
    := 
    \bm{1}_{r \neq r'}
    ~,
\end{equation}
where $r = \argmax_{x \in \mathcal{I}}P(x)$, and $r' = \argmax_{x \in \mathcal{I}}P'(x)$.

\paragraph{Relative Change in Confidence}
Inspired by \citet{finlayson-etal-2021-causal}, we also highlight the change in terms of the relative difference in the probability assigned to $g$ and $g'$. We formulate two types of relative change, one quantifying the relative change in the confidence of $g$, and the other quantifying the relative change in the confidence of $g'$:
\begin{align}
    \Delta_{\mathrm{rel}} &= \frac{ {P}(g)- {P'}(g) }{{P'}(g)} \\
    \Delta_{\mathrm{rel}}'& = \frac{ {P'}(g')- {P}(g') }{{P}(g')} ~.
\end{align}

We quantify the overall relative change in confidence (RCC) as the average of the two relative changes above:
\begin{align}
     \delta_{\mathrm{rcc}} ({P}, {P'}) = \frac{1}{2} \bigg(\Delta_{\mathrm{rel}}  +  \Delta_{\mathrm{rel}}' \bigg)~.
\end{align}

\paragraph{A Unified Form}
We are interested in the average causal effect of the intervention across all problems in $\mathcal{D}$. Thus, we measure the average of the effects over all instances $\bm{q} \in \mathcal{D}$.
We denote by the subscripts $\mathrm{TCE}_{\mathrm{cp}}$/$\mathrm{DCE}_{\mathrm{cp}}$ and $\mathrm{TCE}_{\mathrm{rcc}}$/$\mathrm{DCE}_{\mathrm{rcc}}$ the causal effects computed using the change in prediction metric and the relative change in confidence, respectively.
We describe how we construct the dataset $\mathcal{D}$ in \cref{mathrobust:sec:intervention_data}.

\section{Experimental Setup}
In this section, we describe the data used to perform the interventions and to measure the causal effects.

\subsection{Datasets}
For our analyses, we use instances of math word problems from three popular datasets: ASDiv-A \citep{miao-etal-2020-diverse}, MAWPS \citep{koncel-kedziorski-etal-2016-mawps}, and SVAMP \citep{patel-etal-2021-nlp}.
The examples contained in these collections are pairs $(\bm{t},\bm{o})$ consisting of a question template $\bm{t}$ with its annotated operations $\bm{o}$.
Each of these pairs can be instantiated multiple times into problems $\bm{q} = (\bm{t}, \bm{n})$ by filling the template with numerical values $(n_1, n_2, \dots)$ and computing the ground-truth result $g = f_{\bm{o}}(\bm{n})$ (most problems involve two to three operands, i.e., $|\bm{n}| \in \{2, 3\}$).
We select a set of 437 two-operand and 307 three-operand template-expression pairs that we use to generate pairs of prompts representing an intervention. More details about the prompt generation procedure are in \cref{mathrobust:appd:prompt_creation}.
We use $(\bm{t}, \bm{n})$ to refer to an instantiated template that we use as a prompt.

\subsection{Intervention Data}
\label{mathrobust:sec:intervention_data}

Given an MWP $\bm{q} = (\bm{t}, \bm{n})$ and its solution $g$, we generate a second problem-solution instance $(\bm{q}', g')$
depending on the type of causal effect $\mathrm{CE}$ we want to measure and on the considered variable.
When intervening on the operands of the problem, the text of the problem is kept unaltered and a set of new operands $\bm{n}$ is sampled in such a way that the result $g$ is affected or not depending on the effect that is being measured.
When changing the textual description of the problem, we change $\bm{t}$ such that either $\bm{o}' = \bm{o}$, or $\bm{o}' \neq \bm{o}$. In the former case, we sample a different template $\bm{t}' = (s', \bm{o})$ from the set of templates describing the same operations $\bm{o}$, in the latter case we sample a new $\bm{t}'$ describing a different operation. In \cref{mathrobust:appd:examples} we report some examples of $(\bm{q}, \bm{q}')$ pairs representing the different types of interventions.

Given a model, we use the question pair $(\bm{q}, \bm{q}')$ to obtain a pair of answer distributions $\mathbb{P}(R| \bm{t}, \bm{n})$ and $\mathbb{P}(R| \bm{t}', \bm{n}')$, which we use to measure the causal effect of the intervention.
We consider the space for the numerical values to be $\mathcal{I} = \{1, 2, \dots, C\}$ consisting of integer values, following the setup of several existing MWP datasets \citep{miao-etal-2020-diverse,koncel-kedziorski-etal-2016-mawps,patel-etal-2021-nlp}.
To control our experimental costs and make sure the models keep the number as one token, we set $C=300$.
From all the tokens in a model's vocabulary, we focus on the probability assigned to the numbers in our numerical space $\mathcal{I}$, and thus we use $\mathbb{P}(R=r)$ to denote the normalized probability $\mathbb{P}_{\mathrm{raw}}(R=r)/Z$, where $Z=\sum_{r=1}^{C} \mathbb{P}_{\mathrm{raw}}(R=r)$, and $\mathbb{P}_{\mathrm{raw}}(x)$ is the raw probability score assigned to the vocabulary token $x$.
For each intervention type, we generate a dataset $\mathcal{D}$ consisting of $ (\bm{q}, \bm{q}')$ pairs.
Unless otherwise specified, for our experiments we generate 500 intervention pairs for each template, and results are averaged over three seeds.

\subsection{Models to Evaluate}
We use our framework to assess the robustness of reasoning in thirteen pre-trained language models. We consider five sizes of the GPT-2 model \citep{radford2019language}: distilled \citep{sanh2019distilbert}, small, medium, large, and XL. We evaluate four models from EleutherAI that were pre-trained on the Pile \citep{gao2020pile}: GPT-Neo 1.3B and 2.7B \citep{gpt-neo}, GPT-J-6B \citep{gpt-j}, and GPT-NeoX-20B \citep{black2022gpt}. We use HuggingFace Transformers \citep{wolf-etal-2020-transformers}
to access the models.
Additionally, we experiment with a set of instruction-tuned versions of GPT-3 \citep{brown2020gpt3}: Instruct \citep{ouyang2022instructGPT}, Curie, Davinci-002, and Davinci-003.\footnote{The OpenAI ids for these models are, respectively,  \texttt{davinci-instruct-beta}, \texttt{text-curie-001}, \texttt{text-davinci-002}, and \texttt{text-davinci-003}.}
Experiments with GPT-3 are carried out under the constraints set by the OpenAI APIs\footnote{\url{https://openai.com/api/}}, which prevent us from computing the causal effect using the same procedure as for the other models. We report the details about how the metrics were computed for GPT-3 in \cref{mathrobust:appd:gpt3_approx}.
In the reported results, we indicate with an asterisk ($^*$) the metrics that were influenced by this limitation.

\section{Results}
Our analyses focus primarily on two-operand problems (\cref{mathrobust:sec:n_on_r_2ops,mathrobust:sec:t_on_r_2ops}) and later extend to more complex problems that involve three operands (\cref{mathrobust:sec:3_ops}) for the models that perform best on the two-operand test bed.
We compare the direct causal effect $\mathrm{DCE}$ and the total causal effect $\mathrm{TCE}$ of $\bm{N}$ and $\bm{T}$ on $R$.  $\mathrm{DCE}$ represents the undesired effect for a model to being mistakenly responsive to a change in $\bm{N}$ or $\bm{T}$ not leading to a change in the result $g$ (low robustness), whereas higher values of  $\mathrm{TCE}$ indicate a higher ability of the model to correctly adjust the probability weight assigned to the new solution $g'$ after the intervention (high sensitivity).

\subsection{Effect of \textbf{\textit{N}} on \textit{R}}
\label{mathrobust:sec:n_on_r_2ops}
\begin{figure}\footnotesize
\centering
\begin{tikzpicture}
\begin{semilogyaxis}
[
  xtick=data,
  ymax=800000,
  ymin=0.1,
  log origin y=infty,
  ylabel=$\delta_{\mathrm{rcc}}$,
    ytick={1,10,100,1000,10000,100000,1000000},
  height=0.35\columnwidth,
  width=.7\columnwidth,
  symbolic x coords={Distilled, Small, Medium, Large, Neo-1.3B, XL, Neo-2.7B, J-6B, NeoX, \ , 3-Curie*, 3-Instruct*, 3-Davinci-002*, 3-Davinci-003*, A},
  enlarge y limits=0.0,
  enlarge x limits=0.06,
  legend style={at={(0.4,1.15)},
  anchor=north,legend columns=-1},
  ybar,
  bar width=3pt,
  grid=major,
    xmajorgrids=false,
    x tick label style={rotate=45,anchor=east}
]
\addplot+ [
   black, fill=cyan,
   error bars/.cd,
   y dir=both,
   y explicit,
   error mark options={
      rotate=90,
      mark size=0pt,
    }
]coordinates {
(J-6B, 0.834235346744347) +- (0, 0.04)
(Neo-1.3B, 0.522155872100273) +- (0, 0.027)
(Neo-2.7B, 0.546945810291668) +- (0, 0.024)
(Distilled, 0.108832419634291) +- (0, 0.001)
(Small, 0.157099819813293) +- (0, 0.00132896355741316)
(Large, 0.671930790279737) +- (0, 0.01)
(Medium, 0.569471490478381) +- (0, 0.01)
(XL, 0.56) +- (0, 0.00976850871032544)
(3-Curie*, 38.37348145494138) +- (0, 30.483747647342565)
(NeoX, 1.17916686887043) +- (0, 0.0278626018160753)
(3-Davinci-002*,3.8603336668546295) +- (0, 0.47435185634652777)
(3-Instruct*, 1.7914718574180748) +- (0, 0.5891337864978382)
(3-Davinci-003*, 11.025906942004847) +- (0, 0.8210212394041703)
};

\addplot+ [
   black, fill=orange,
   error bars/.cd,
   y dir=both,
   y explicit,
    error mark options={
      rotate=90,
      mark size=0pt,
    }
]coordinates {
(J-6B, 25.780298441897347) +- (0, 6.362351613767966)
(Neo-1.3B, 2.1197260504015434) +- (0, 0.04940484879745536)
(Neo-2.7B, 2.8916255381461853) +- (0, 0.03780760816893716)
(Distilled, 0.2434568953880113) +- (0, 0.00991243715811916)
(Small, 0.3892899513625348) +- (0, 0.011852482242094703)
(Large, 1.6354087671691948) +- (0, 0.06290616497617149)
(Medium, 1.6226962831730949) +- (0, 0.05577335022177161)
(XL, 1.5734177353962508) +- (0, 0.07869885155377497)
(3-Curie*, 485.15496502908155) +- (0, 344.38332170275)
(NeoX, 2140.875) +- (0, 885.048638522765)
(3-Davinci-002*, 11815.602815407865) +- (0, 1910.5459089278447)
(3-Davinci-003*, 421210.69134962215) +- (0, 103460.56199762033)
(3-Instruct*, 40.00695149530433) +- (0, 7.948463509136467)
};
\draw[very thick] (axis cs:\ ,0.001) -- (axis cs:\ ,10000000);,

\legend{DCE$_{\mathrm{rcc}}$ of $\bm{N}$, TCE$_{\mathrm{rcc}}$ of $\bm{N}$}
\end{semilogyaxis}
\end{tikzpicture}

\begin{tikzpicture}
\begin{axis}
[
  xtick=data,
  ymax=1,
  ymin=0.1,
  log origin y=infty,
  ylabel=$\delta_{\mathrm{cp}}$,
  height=0.7\columnwidth,
  width=.35\columnwidth,
  symbolic x coords={Distilled, Small, Medium, Large, Neo-1.3B, XL, Neo-2.7B, J-6B, NeoX, \ , 3-Curie, 3-Instruct, 3-Davinci-002, 3-Davinci-003, A},
  enlarge y limits=0.0,
  enlarge x limits=0.06,
  legend style={at={(0.4,1.15)},
  anchor=north,legend columns=-1},
  ybar,
  bar width=3pt,
  grid=major,
    xmajorgrids=false,
    x tick label style={rotate=45,anchor=east}
]
\addplot+ [
   black, fill=cyan,
   error bars/.cd,
   y dir=both,
   y explicit,
   error mark options={
      rotate=90,
      mark size=0pt,
    }
]coordinates {
(J-6B, 0.824442410373761) +- (0, 0.000732354431409923)
(Neo-1.3B, 0.7679359267734553) +- (0, 0.00001)
(Neo-2.7B, 0.802183066361557) +- (0, 0.0005354691075515)
(Distilled, 0.23335469107551488) +- (0, 0.0043971439847412495)
(Small, 0.26252631578947366) +- (0, 0.00001)
(Large, 0.777116704805492) +- (0, 0.00001)
(Medium, 0.5065812356979406) +- (0, 0.00001)
(XL, 0.6470160183066361) +- (0, 0.00001)
(3-Davinci-002, 0.28878718535469106) +- (0, 0.01774992567234188)
(3-Curie,0.8083142639206713) +- (0, 0.017886417848300213)
(NeoX, 0.8322) +- (0, 0.000721110255092798)
(3-Instruct, 0.7944317315026698) +- (0, 0.030136816154239656)
(3-Davinci-003, 0.15385202135774215) +- (0, 0.0033735079157125912)
};
\draw[very thick] (axis cs:\ ,0.001) -- (axis cs:\ ,1.1);,

\addplot+ [
   black, fill=orange,
   error bars/.cd,
   y dir=both,
   y explicit,
  error mark options={
      rotate=90,
      mark size=0pt,
    }
]coordinates {
(J-6B, 0.863734553775744) +- (0, 0.000662631111777246)
(Neo-1.3B, 0.7994279176201373) +- (0, 0.0001)
(Neo-2.7B, 0.83173767209665) +- (0, 0.000480962268167874)
(Distilled, 0.26010983981693364) +- (0, 0.0001)
(Small, 0.28039816933638445) +- (0, 0.0001)
(Large, 0.8118215102974828) +- (0, 0.0001)
(Medium, 0.5408695652173913) +- (0, 0.0001)
(XL, 0.673720823798627) +- (0, 0.0001)
(3-Curie,0.8472921434019831) +- (0, 0.0157771244041857)
(NeoX, 0.8839) +- (0, 0.000360555127546399)
(3-Davinci-002, 0.980091533180778) +- (0, 0.0021547592512319727)
(3-Instruct, 0.9118993135011441) +- (0, 0.015484738829031626)
(3-Davinci-003, 0.9933638443935927) +- (0, 0.00225374320407236)
};

\legend{DCE$_{\mathrm{cp}}$ of $\bm{N}$, TCE$_{\mathrm{cp}}$ of $\bm{N}$}
\end{axis}
\end{tikzpicture}
\caption{Comparison of $\mathrm{DCE}(\bm{N} \rightarrow R)$ and $\mathrm{TCE}(\bm{N} \text{ on } R)$. $^*$approx values, see \cref{mathrobust:appd:gpt3_approx}.}
\label{mathrobust:fig:effect_of_n}
\end{figure}

\begin{figure*}[t]
    \centering
    \includegraphics[width=0.66\columnwidth]{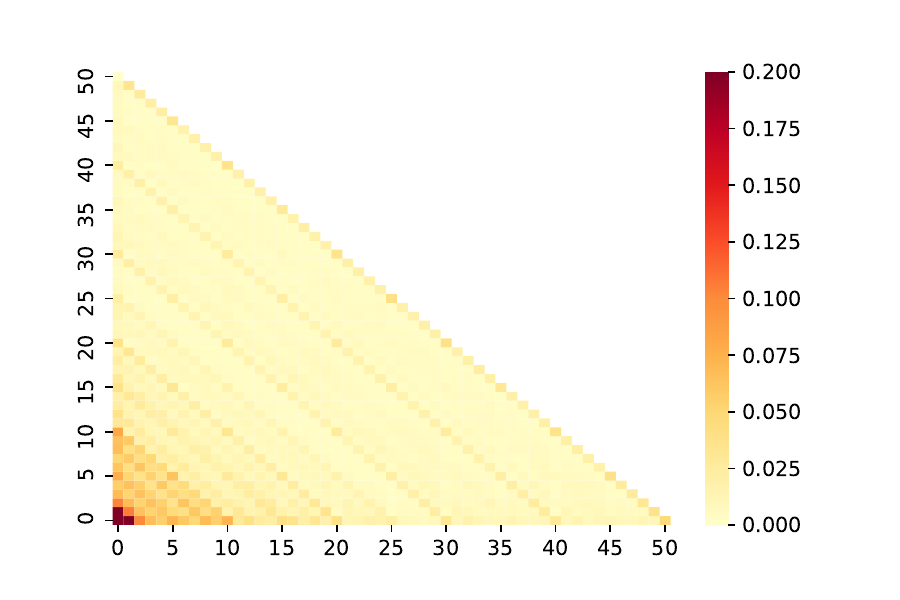}
    \includegraphics[width=0.66\columnwidth]{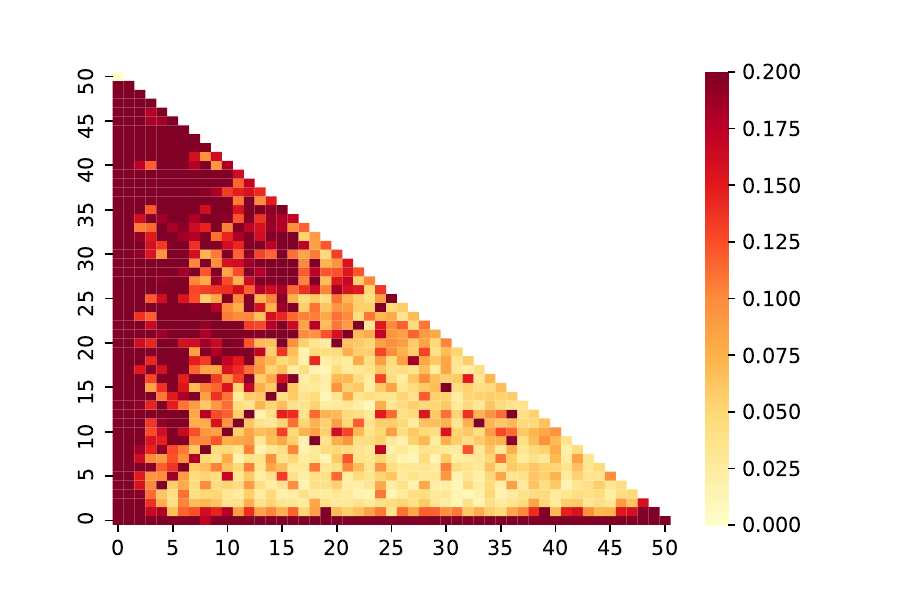}
    \includegraphics[width=0.66\columnwidth]{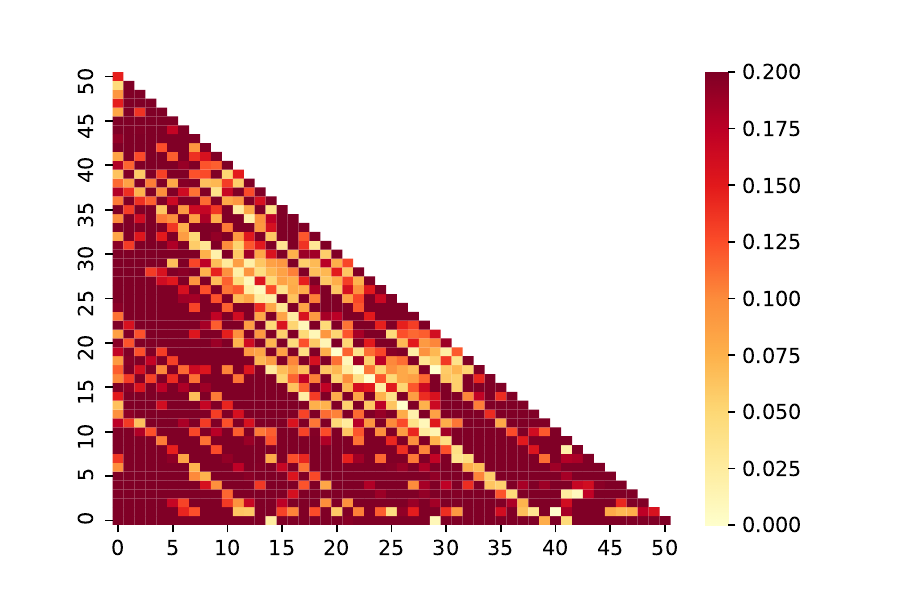}
    \caption{
    Heatmaps displaying $P(g)$ for Distil-GPT-2 (left), GPT-J-6B (center), and GPT-3 Davinci-002 (right). $g$ is the ground-truth result $g = n_1+n_2$ ($n_1$ and $n_2$ are represented by the x and y axes, respectively. The probability values for each combination of $((n_1, n_2), g)$ are averaged over 20 different templates.  Probability values over 0.2 are displayed with the darkest color.
    }
    \label{mathrobust:fig:heatmaps}
\end{figure*}

From the results in \cref{mathrobust:fig:effect_of_n}, we notice that larger models exhibit a larger TCE$_{\mathrm{rcc}}$/DCE$_{\mathrm{rcc}}$ ratio.
In particular, in GPT-J-6B and NeoX, the TCE is, respectively, 30x and 1000x larger than the DCE.
However, this improvement in sensitivity is not manifested in terms of change of prediction ($\delta_{\mathrm{cp}}$), for which the models show to be affected by result-preserving changes almost as equally as by result-altering interventions. This behavior changes significantly in instruction-tuned models. In particular, for the 175B-parameter GPT-3, performance varies depending on the type of supervision, with the PPO-trained Davinci-003 exhibiting an 84\% difference between direct and total effect.

\begin{figure}[t]\footnotesize
\begin{tikzpicture}
\begin{semilogyaxis}[
  xtick=data,
  ymax=800000,
  ymin=0.1,
    ytick={1,10,100,1000,10000,100000,1000000},
  log origin y=infty,
  ylabel=$\delta_{\mathrm{rcc}}$,
  height=0.5\columnwidth,
  width=\columnwidth,
    symbolic x coords={Distilled, Small, Medium, Large, Neo-1.3B, XL, Neo-2.7B, J-6B, NeoX, \ , 3-Curie*, 3-Instruct*, 3-Davinci-002*, 3-Davinci-003*, A},
  enlarge y limits=0.0,
  enlarge x limits=0.06,
  legend style={at={(0.4,1.15)},
  anchor=north,legend columns=-1},
  ybar,
  bar width=3pt,
  grid=major,
    xmajorgrids=false,
    x tick label style={rotate=45,anchor=east}
]
\addplot+ [
   black, fill=teal,
   error bars/.cd,
   y dir=both,
   y explicit,
     error mark options={
      rotate=90,
      mark size=0pt,
    }
]coordinates {
(J-6B, 1.66141132299187) +- (0, 0.225)
(Neo-1.3B, 0.398066705758993) +- (0, 0.022)
(Neo-2.7B, 0.551559779439723) +- (0, 0.015)
(Distilled, 0.195896660986663) +- (0, 0.007)
(Small, 0.20761336577218) +- (0, 0.00971811540502479)
(Large, 0.47599176632657) +- (0, 0.0407676565497067)
(Medium, 0.466289443075747) +- (0, 0.0234487922448366)
(XL, 0.452318654151913) +- (0, 0.017891774509494)
(3-Curie*, 207.09022634500295) +- (0, 147.48511894560136)
(NeoX, 11.4182657266601) +- (0, 5)
(3-Davinci-002*,72.48066042572056) +- (0, 14.844075549654468)
(3-Instruct*, 13.150551053840516) +- (0, 1.6771769860754733)
(3-Davinci-003*, 454.4408496416752) +- (0, 44.605656629482915)

};

\addplot+ [
   black, fill=magenta,
   error bars/.cd,
   y dir=both,
   y explicit,
     error mark options={
      rotate=90,
      mark size=0pt,
    }
]coordinates {
(J-6B, 8.786903772873094) +- (0, 0.5434826780437797)
(Neo-1.3B, 0.9964369224026975) +- (0, 0.1060857609733839)
(Neo-2.7B, 1.4947411285112124) +- (0, 0.10159453344958379)
(Distilled, 0.32129470931913434) +- (0, 0.007003333999838406)
(Small, 0.2877384149650301) +- (0, 0.013008330346946696)
(Large, 1.1563198973806559) +- (0, 0.04193074654864333)
(Medium, 0.5445143435025176) +- (0, 0.041300853778617044)
(XL, 0.9287654038449459) +- (0, 0.07269771764576204)
(3-Curie*, 2223.413615634005) +- (0, 438.820275890894)
(NeoX, 55.8513333333333) +- (0, 8.94)
(3-Davinci-002*, 40710.92153893157) +- (0, 2746.9696297021583)
(3-Instruct*, 77.02657273094938) +- (0, 2.9102600975385986)
(3-Davinci-003*, 445878.50103203097) +- (0, 126506.5178705026)
};
\draw[very thick] (axis cs:\ ,0.0001) -- (axis cs:\ ,10000000);,

\legend{DCE$_{\mathrm{rcc}}$ of $S$, TCE$_{\mathrm{rcc}}$ of $\bm{T}$}
\end{semilogyaxis}
\end{tikzpicture}

\begin{tikzpicture}
\begin{axis}
[
  xtick=data,
  ymax=1,
  ymin=0.1,
  log origin y=infty,
  ylabel=$\delta_{\mathrm{cp}}$,
  height=0.5\columnwidth,
  width=\columnwidth,
  symbolic x coords={Distilled, Small, Medium, Large, Neo-1.3B, XL, Neo-2.7B, J-6B, NeoX, \ , 3-Curie, 3-Instruct, 3-Davinci-002, 3-Davinci-003, A},
  enlarge y limits=0.0,
  enlarge x limits=0.06,
  legend style={at={(0.4,1.18)},
  anchor=north,legend columns=-1},
  ybar,
  bar width=3pt,
  grid=major,
    xmajorgrids=false,
    x tick label style={rotate=45,anchor=east}
]
\addplot+ [
   black, fill=teal,
   error bars/.cd,
   y dir=both,
   y explicit,
     error mark options={
      rotate=90,
      mark size=0pt,
    }
]coordinates {
(J-6B, 0.6338293838862559) +- (0, 0.00001)
(Neo-1.3B, 0.556957345971564) +- (0, 0.00001)
(Neo-2.7B, 0.5996113744075829) +- (0, 0.00001)
(Distilled, 0.37133649289099524) +- (0, 0.0043971439847412495)
(Small, 0.318) +- (0, 0.00001)
(Large, 0.6056872037914692) +- (0, 0.00001)
(Medium, 0.46161137440758293) +- (0, 0.00001)
(XL, 0.5986445497630332) +- (0, 0.00001)
(3-Davinci-002, 0.3726698262243286) +- (0, 0.002576541300205386)
(3-Curie,0.8175355450236967) +- (0, 0.0045622524733629)
(NeoX,0.6674) +- (0,0.00763740793725201)
(3-Instruct, 0.7205371248025276) +- (0, 0.010558549643922253)
(3-Davinci-003, 0.2218009478672985) +- (0, 0.0037617317218927885)
};

\addplot+ [
   black, fill=magenta,
   error bars/.cd,
   y dir=both,
   y explicit,
  error mark options={
      rotate=90,
      mark size=0pt,
    }
]coordinates {
(J-6B, 0.5989649178255373) +- (0, 0.00001)
(Neo-1.3B, 0.5423119469026548) +- (0, 0.00001)
(Neo-2.7B, 0.6137701485461441) +- (0, 0.00001)
(Distilled, 0.3417845290771176) +- (0, 0.00001)
(Small, 0.38670393489254107) +- (0, 0.00001)
(Large, 0.5893844816687737) +- (0, 0.00001)
(Medium, 0.5624802465233881) +- (0, 0.00001)
(XL, 0.6448324905183312) +- (0, 0.00001)
(3-Davinci-002,.9837329633475319) +- (0, 0.0038282605239010337)
(3-Curie,0.9092859658113035) +- (0, 0.007838254348368172)
(NeoX,0.657566666666667) +- (0,0.00360185137579736)
(3-Instruct, 0.8496108305391257) +- (0, 0.015687273409255294)
(3-Davinci-003, 0.9888261976296984) +- (0, 0.001492852803797162)
};
\draw[very thick] (axis cs:\ ,0) -- (axis cs:\ ,1.1);,

\legend{DCE$_{\mathrm{cp}}$ of $S$, TCE$_{\mathrm{cp}}$ of $\bm{T}$}
\end{axis}
\end{tikzpicture}
\caption{Comparison of $\mathrm{DCE}(S \rightarrow R)$ and $\mathrm{TCE}(\bm{T} \text{ on } R)$. We use $^*$ to denote approximated values, explained in \cref{mathrobust:appd:gpt3_approx}.}
\label{mathrobust:fig:effect_of_t}

\end{figure}

In \cref{mathrobust:fig:heatmaps}, we present a different visualization of the direct causal effect of $\bm{N}$ on the model's prediction. We report the heatmaps showing the probability assigned by the model to the result $g$ of a problem $(\bm{t}, (n_1, n_2), g) \ | \ g = n_1 + n_2, \ \forall g \in \{0,1,\dots,50 \}, \ \forall (n_1, n_2) \in \{0,1,\dots,50 \}^2$. For Distil-GPT-2 we observe low overall probability assigned to $g$ and diagonal patterns indicating consistency in assigning higher probability to specific results (e.g., 10, 20, 30, 40, 50). For the two larger models we notice a higher probability mass assigned to the problem's result, but less consistency on the prediction of the same result with different sets of operands (this is true for GPT-J in particular). This result is consistent with the observed higher DCE and TCE in larger models: $P(g)$ might vary more considerably when intervening on $\bm{N}$ without affecting $g$, but overall the model assigns higher probability weight to the correct result, which correlates with higher sensitivity.

\subsection{Effect of \textbf{\textit{T}} on \textit{R}}
\label{mathrobust:sec:t_on_r_2ops}
In \cref{mathrobust:fig:effect_of_t}, we report the total causal effect of the textual framing $\bm{T}$ and the direct causal effect of the irrelevant text elements $S$ on the model's prediction.
For the instruction-tuned models, the improvement in terms of prediction change ($\delta_{\mathrm{cp}}$) follows a similar trend as for $\bm{N}$, with GPT-3 Davinci-003 showing a 76\% difference between direct and total effect.
An interesting observation is that the irrelevant textual information $S$ appears to have a lower direct effect than $\bm{N}$ for all non-instruction-tuned models. However, in the GPT-3 Davinci-00x models, we observe the opposite (i.e., DCE$(\bm{N} \rightarrow R)$ $\leq$ DCE$(S \rightarrow R)$). This suggests that large instruction-based models tend to be more susceptible to variation in the textual framing of a problem, while smaller models are more responsive to changes in the numerical values (though not necessarily correctly).

\subsection{Overall Insights}
In comparison to other models, GPT-3 Davinci shows the highest DCE$_{\mathrm{rcc}}$, but low DCE$_{\mathrm{cp}}$. This discrepancy is related to the quantities that the two metrics consider. $\delta_{\mathrm{rcc}}$ takes into account the probability assigned to $g$, while $\delta_{\mathrm{cp}}$ does not consider the ground truth solution. One interpretation of this result is that GPT-3 Davinci consistently predicts the same answer $r = r'$ when $g = g'$, but the probabilities $P(g)$ and $P'(g)$ might vary significantly.

The results observed for the two kinds of intervention $\mathrm{do}(\bm{T}:\bm{t} \rightarrow \bm{t}')$ and $\mathrm{do}(\bm{N}:(n_1,n_2) \rightarrow (n_1', n_2'))$ show similar trends. Small models (Distilled and Small GPT-2) exhibit low sensitivity to interventions. Larger models (from GPT-2 Medium to GPT-Neo) appear to be more influenced by changes in both $\bm{N}$ and $\bm{T}$. However, they display similar sensitivity to both result-altering and result-preserving interventions. An improvement in sensitivity is noticeable in GPT-J and NeoX, though not accompanied by an improvement in robustness. Remarkably different behavior is instead shown by the GPT-3 Davinci models, which demonstrate substantially higher sensitivity to result-altering interventions (high TCE), and higher robustness (in terms of prediction change).
In \cref{mathrobust:appd:accuracy}, we report the accuracy of the models on the generated instances of MWPs, which exhibits a similar trend as the robustness/sensitivity changes we observed.

Possible explanations for the improved robustness and sensitivity demonstrated by the large GPT-3 models might be the dramatic size increase and extension/enhancement of the training procedure involving instructions.
The former idea is aligned with the \emph{emergent abilities} hypothesis \citep{wei2022emergent}, which postulates the existence of skills that are displayed by large-scale models but are not present in smaller-scale models.
However, our observations show different performances in versions of GPT-3 Davinci that differ in the training procedure.\footnote{A high-level description of the training procedures for the models is provided at \url{https://beta.openai.com/docs/model-index-for-researchers}.}
This raises the question of whether the capability of LLMs to reason about math problems benefits from instruction-based tuning.
We address this question in the following section.

\subsection{Extending to LLaMA-Based Models}

To further investigate the roles played by size and training method in the model's performance, we carry out our experimental procedure on three versions with different sizes (7B, 13B, and 30B) of the LLaMA model \citep{touvron2023llama}, and on Stanford Alpaca (which applies instruction tuning on LLaMA 7B) \citep{taori2023alpaca}.
We present these results separately, as the LLaMA tokenization makes the prediction setup different from the one used from the other models, and prevents us from computing the relative change in confidence ($\delta_{\mathrm{rcc}}$).\footnote{The LLaMA tokenizer considers each digit as an independent token in the vocabulary. This makes it problematic to compare the probability value assigned by the model to multi-digit numbers.}

\begin{figure}\footnotesize
\centering
\begin{tikzpicture}
\begin{axis}
[
  xtick=data,
  ymax=0.7,
  ymin=0.2,
    ytick={0.1,0.2,...,1},
  log origin y=infty,
  ylabel=$\delta_{\mathrm{cp}}$,
  height=0.5\columnwidth,
  width=0.75\columnwidth,
  symbolic x coords={\ LLaMA 7B\ \ , \ LLaMA 13B\ \ , \ LLaMA 30B\ \ , \ Alpaca \ },
  enlarge y limits=0.0,
  enlarge x limits=0.3,
  legend style={at={(0.5,1.15)},
  anchor=north,legend columns=-1},
  ybar,
  bar width=6pt,
  grid=major,
    xmajorgrids=false,
    x tick label style={rotate=45,anchor=east}
]
\addplot+ [
   black, fill=cyan,
   error bars/.cd,
   y dir=both,
   y explicit,
     error mark options={
      rotate=90,
      mark size=0pt,
    }
]coordinates {
(\ Alpaca \ , 0.2856) +- (0, 0.005338)
(\ LLaMA 7B\ \ , 0.3775) +- (0, 0.005338)
(\ LLaMA 13B\ \ , 0.3959) +- (0, 0.005231)
(\ LLaMA 30B\ \ , 0.3366) +- (0, 0.007149)
};
\addplot+ [
   black, fill=orange,
   error bars/.cd,
   y dir=both,
   y explicit,
     error mark options={
      rotate=90,
      mark size=0pt,
    }
]coordinates {
(\ Alpaca \ , 0.3801) +- (0, 0.005338)
(\ LLaMA 7B\ \ ,  0.4682) +- (0, 0.005185)
(\ LLaMA 13B\ \ , 0.6027) +- (0, 0.005234)
(\ LLaMA 30B\ \ , 0.5989) +- (0, 0.007415)
};
\legend{DCE of $\bm{N}$, TCE of $\bm{N}$}
\end{axis}
\end{tikzpicture}
\caption{Comparison of direct and total effects of $N$ on $R$ for LLaMA and Alpaca.}
\label{mathrobust:fig:llama}
\end{figure}
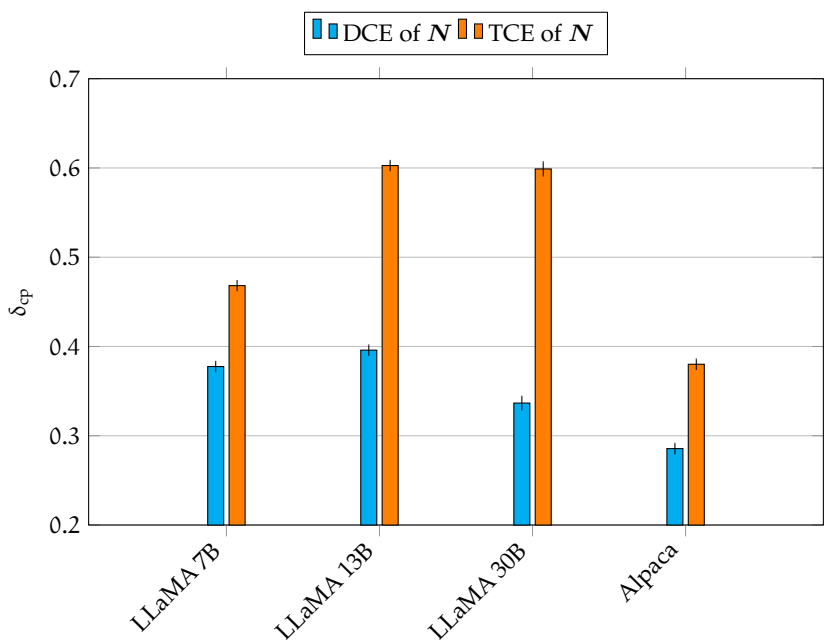

From the results (\cref{mathrobust:fig:llama}), two notable observations emerge. Firstly, the increased difference between TCE and DCE observed with the increasing size of the LLaMA models suggests that a larger number of parameters can be a significant driver behind robustness/sensitivity improvement.
However, this is not necessarily the case across different models: GPT-NeoX-20B shows a smaller TCE$_{\mathrm{cp}}$-DCE$_{\mathrm{cp}}$ gap compared to LLaMA 7B (5.2\% vs 9.0\%).
Secondly, the instruction tuning procedure of Alpaca does not seem to help significantly with mathematical computation: the decrease in both TCE and DCE shows that robustness improves at the expense of sensitivity. Nonetheless, overall, when comparing Alpaca compared to its base model, LLaMA 7B, we observe an increase in the gap between TCE and DCE, although this difference is minimal (9.5\% vs 9.0\%).

The limited improvement of Alpaca might be attributed to its instruction tuning procedure consisting of ``a list of user-oriented instructions including email writing, social media, and productivity tools'' \citep{taori2023alpaca}, which differs from reasoning-intensive tasks.
We suggest future work to examine different types of instruction tuning (e.g., focused on reasoning procedures or reinforcement learning from human feedback), which might help the model answer more complex types of questions in a step-by-step manner and more accurately.
We hypothesize that the different performances in versions of GPT-3 Davinci might be produced by the specific type of instructions used for training, by the reinforcement learning component \citep{ouyang2022instructGPT}, or simply by an extension of the language modeling pre-training. It is challenging to pinpoint the exact factor in the training procedure that contributes to this improvement, as specific methodological details are not available.

\subsection{Moving to Three-Operand Problems}
\label{mathrobust:sec:3_ops}

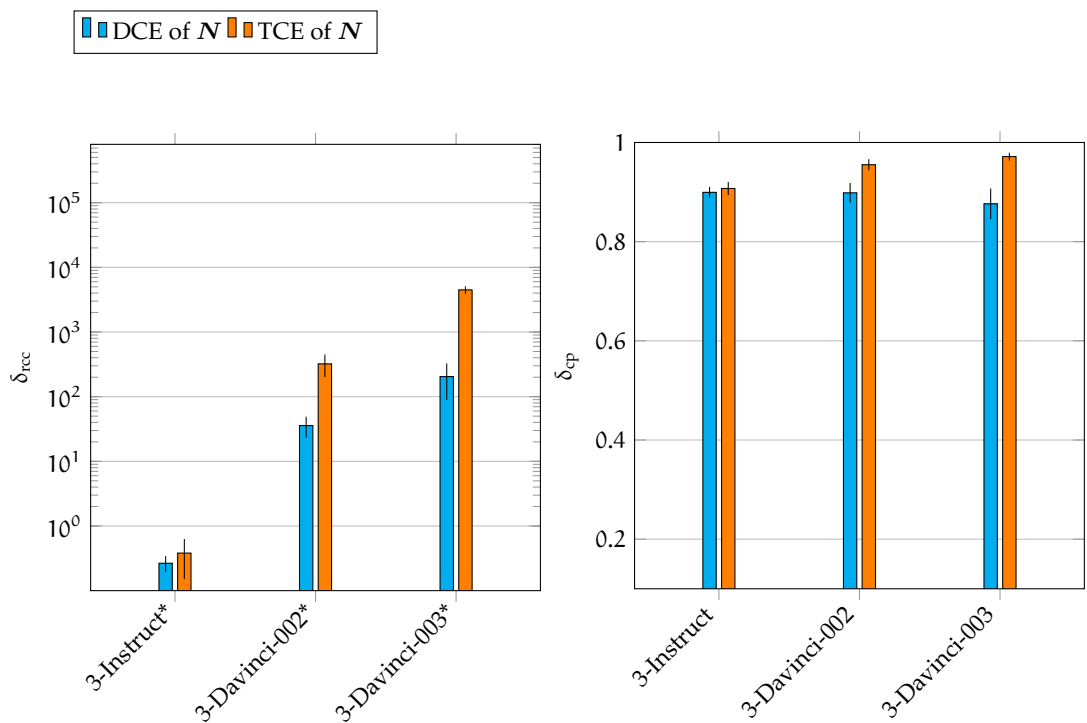
\begin{figure}\footnotesize
\begin{tikzpicture}
\begin{semilogyaxis}
[
  xtick=data,
  ymax=800000,
  ymin=0.1,
  log origin y=infty,
  ylabel=$\delta_{\mathrm{rcc}}$,
    ytick={1,10,100,1000,10000, 100000, 1000000},
  height=0.5\columnwidth,
  width=0.5\columnwidth,
  symbolic x coords={3-Instruct*, 3-Davinci-002*, 3-Davinci-003*},
  enlarge y limits=0.0,
  enlarge x limits=0.3,
  legend style={at={(0.3,1.3)},
  anchor=north,legend columns=-1},
  ybar,
  bar width=5pt,
  grid=major,
    xmajorgrids=false,
    x tick label style={rotate=45,anchor=east}
]
\addplot+ [
   black, fill=cyan,
   error bars/.cd,
   y dir=both,
   y explicit,
     error mark options={
      rotate=90,
      mark size=0pt,
    }
]coordinates {
(3-Davinci-002*,35.73510970674531) +- (0, 11.576964545214814)
(3-Instruct*, 0.2654215096374149) +- (0, 0.06448478904336268)
(3-Davinci-003*, 204.42033846659515) +- (0, 112.14213366968319)
};

\addplot+ [
   black, fill=orange,
   error bars/.cd,
   y dir=both,
   y explicit,
     error mark options={
      rotate=90,
      mark size=0pt,
    }
]coordinates {
(3-Davinci-002*, 321.9426983017017) +- (0, 113.4233710415126)
(3-Davinci-003*, 4486.165919771869) +- (0, 440.25242681937243)
(3-Instruct*, 0.3811035764990454) +- (0, 0.22529671387767283)
};
\legend{DCE of $\bm{N}$, TCE of $\bm{N}$}
\end{semilogyaxis}
\end{tikzpicture}
\begin{tikzpicture}
\begin{axis}
[
  xtick=data,
  ymax=1,
  ymin=0.1,
  log origin y=infty,
  ylabel=$\delta_{\mathrm{cp}}$,
  height=0.5\columnwidth,
  width=0.5\columnwidth,
  symbolic x coords={\ 3-Instruct\ \ , \ 3-Davinci-002\ \ , \ 3-Davinci-003\ \ ,},
  enlarge y limits=0.0,
  enlarge x limits=0.3,
  legend style={at={(0.4,1.15)},
  anchor=north,legend columns=-1},
  ybar,
  bar width=5pt,
  grid=major,
    xmajorgrids=false,
    x tick label style={rotate=45,anchor=east}
]
\addplot+ [
   black, fill=cyan,
   error bars/.cd,
   y dir=both,
   y explicit,
     error mark options={
      rotate=90,
      mark size=0pt,
    }
]coordinates {
(\ 3-Instruct\ \ , 0.899457111834962) +- (0, 0.008925847519897764)
(\ 3-Davinci-003\ \ , 0.8764386536373507) +- (0, 0.029015632994062374)
(\ 3-Davinci-002\ \ , 0.8984799131378937) +- (0, 0.017821952658976598)
};
\addplot+ [
   black, fill=orange,
   error bars/.cd,
   y dir=both,
   y explicit,
     error mark options={
      rotate=90,
      mark size=0pt,
    }
]coordinates {
(\ 3-Davinci-002\ \ , 0.955157437567861) +- (0, 0.009145047422039358)
(\ 3-Instruct\ \ , 0.9072747014115091) +- (0, 0.011171366876779134)
(\ 3-Davinci-003\ \ , 0.9715526601520087) +- (0, 0.005996395774803076)
};
\end{axis}
\end{tikzpicture}
\caption{Comparison of direct and total effects of $\bm{N}$ on $R$ for three-operand problems.}
\label{mathrobust:fig:effect_of_n_3ops}
\end{figure}

We extend our evaluation to consider the three-operand problems in the dataset. In these experiments, we consider only the GPT-3 175B-parameter models, as they are the only models performing well on the simpler bivariate problems. The results regarding the effects of $\bm{N}$ are reported in \cref{mathrobust:fig:effect_of_n_3ops}.
We notice that the large difference between the desired (TCE) and undesired (DCE) effects observed on simpler problems shrinks significantly for both metrics. In particular, for Davinci-003, the direct effect of $\bm{N}$ (measured as $\delta_{\mathrm{cp}}$) grows from 0.17 to 0.87. That is, GPT-3 Davinci-003 predicts a different result 87\% of the time after an intervention that does not affect the ground-truth solution.
The increase in direct effect indicates a performance degradation in terms of brittleness: even the models that show good performance on two-operand problems, now display an unstable behavior after result-preserving interventions.

\section{Related Work}

\paragraph{Causal NLP}
Causal inference aims to study the cause and effect from observational and interventional data \citep{pearl2009causality,peters2017elements}. Traditionally, researchers usually apply causal techniques to phenomena in nature and human society.
With the rise of powerful models in NLP, recent research has started to explore the intersection of causal inference and NLP, forming the study of Causal NLP \citep{jin-etal-2022-causalnlp,feder2021causal}.

There are several formulations for Causal NLP: the \textit{causality for NLP} thread involves using the causal framework for data collection and task formulation \citep{jin-etal-2021-causal}, inspecting the (path-specific) causal effect of certain neurons on predictions \citep{vig2020investigating,meng2022locating}, understanding the causal effect of data and learning paradigm for model performance \citep{ni-etal-2022-original}, and as a way to frame prompts \citep{lyu2024causal}; and \textit{NLP for causality} involves testing the pure causal inference skills of LLMs \citep{jin2023cladder,jin2024large}, and use text as a variable for causal effect estimation \citep{roberts2020adjusting,veitch2020adapting,jin-etal-2021-mining-cause,jin2023ai}.

The most similar line of research to our work is the application of causal effect estimation on interpreting models' behavior, such as how models understand syntactic agreement \citep{finlayson-etal-2021-causal}, and how interventions in the representations and weights affect the model prediction \citep{feder-etal-2021-causalm}.
To the best of our knowledge, our work is the first to formulate a causal framework for robustness behavioral tests, and also we are the first to introduce the idea to quantify the differences in the causal mechanisms of human reasoning and model decisions.

\paragraph{Math Reasoning in NLP}
A growing body of work
tries to improve the math reasoning capability in NLP models \citep{zhang-etal-2020-language-embeddings, geva-etal-2020-injecting, spokoyny2021masked}, and prompting techniques for LLMs \citep{cobbe2021training,shen2021generate,kojima2022large,wei2022chain,chowdhery2022palm}. For analysis, significant attention has been given to models' ability to understand numerical quantities \citep{wallace-etal-2019-nlp, thawani-etal-2021-representing} and numerical operations \citep{pal-baral-2021-investigating-numeracy, berg-kirkpatrick-spokoyny-2020-empirical, piekos-etal-2021-measuring, razeghi-etal-2022-impact}.

\section{Conclusion}
We developed a framework to disentangle and separately measure the effect of different factors influencing the predictions of LLMs for math reasoning.
Our results indicate that a drastic increase in both robustness and sensitivity emerges in the GPT-3 Davinci models. Additionally, we study the contribution of  size and instruction tuning in the models of the LLaMA family, observing that the Alpaca instruction tuning, while increasing the model's robustness, does not significantly improve the overall performance.
Our framework provides a formalized theory of behavioral testing for math reasoning models and opens new future directions to design behavioral tests of models in a principled way.

\section*{Ethical Considerations}
As for the ethical practice in this work,
the data involved are from existing MWP datasets with no private user information, and available under the MIT license.
As for the ethical impact of the use of this work, the study is about providing a metric and analyzing existing models' robustness, so there is less concern over harmful usage. Rather, it is more about putting checks on existing AI models and helping humans understand them better before use. Potential stakeholders that could benefit from this research include NLP researchers working on math models, practitioners working on various applications involving mathematical reasoning with text, and e-learning design.

\section*{Limitations}
A key limitation in our work is that LLMs might have seen these math problems. Our work theoretically assumes this is not the case.
Another limitation is that for the sake of simplicity, our work makes some assumptions. For example, we assume all numbers in the range of integers 0 to $C=300$. This would not cover every MWP out there. And future work is needed to generalize our framework to other forms of MWPs.
In this work, we are also constrained by the limitations of the OpenAI policy on the GPT-3 API. This limits the number of perturbations we consider in this work as well as the accuracy with which we can estimate our causal distributions. %
Finally, our work is restricted to English, and extending it to other languages will require us to create an MWP dataset in that language.

\part{Causality among the Learning Variables}\label{part:3}

\mainchapter{Causal Direction in Data Matters: Implications of Causal and Anticausal Learning in NLP}\label{ch:icm}

The principle of independent causal mechanisms (ICM) states that generative processes of real world data consist of independent modules which do not influence or inform each other. While this idea has led to fruitful developments in the field of causal inference, it is not widely-known in the NLP community. In this work, we argue that the causal direction of the data collection process bears nontrivial implications that can explain a number of published NLP findings, such as differences in semi-supervised learning (SSL) and domain adaptation (DA) performance across different settings. We categorize common NLP tasks according to their causal direction and empirically assay the validity of the ICM principle for text data using minimum description length. We conduct an extensive meta-analysis of over 100 published SSL and 30 DA studies, and find that the results are consistent with our expectations based on causal insights. This work presents the first attempt to analyze the ICM principle in NLP, and provides constructive suggestions for future modeling choices. Our code is available at \url{https://github.com/zhijing-jin/icm4nlp}.

\section{Introduction}\label{icm:sec:intro}
NLP practitioners typically do not pay great attention to the causal direction of the data collection process.
As a motivating example, consider the case of collecting a dataset to train a machine translation (MT) model to translate from
English (En) to Spanish (Es): 
it is common practice to
mix all available En-Es sentence pairs together and train 
the model on the entire pooled data set~\citep{bahdanau2014neural,cho2014learning}. 
However, such mixed corpora actually consist of two distinct types of data: (i)
sentences that originated in English and have been translated (by human translators) into Spanish (En$\rightarrow$Es); and (ii) 
sentences that originated in Spanish and have subsequently been translated 
into English (Es$\rightarrow$En).\footnote{There is, in principle, a third option: both could be translations from a third language, but this occurs less frequently.}

Intuitively, these two subsets are qualitatively different, and 
an increasing number of observations by the NLP community indeed suggests that 
they exhibit
different properties~\citep{freitag-etal-2019-ape,edunov-etal-2020-evaluation,riley-etal-2020-translationese,shen-etal-2021-source}.
In the case of MT, for example, researchers find that training models 
on each of these two types of data separately leads to
different test performance, as well as different performance improvement by semi-supervised learning (SSL)~\citep{bogoychev2019domain,graham-etal-2020-statistical,edunov-etal-2020-evaluation}.
Motivated by this observation that the data collection process seems to matter for model performance, in this work, we provide an explanation of this
phenomenon
from the perspective of causality~\citep{pearl2009causality,peters2017elements}.

\begin{figure}[ht]
    \centering 
    \includegraphics[width=.6\columnwidth]{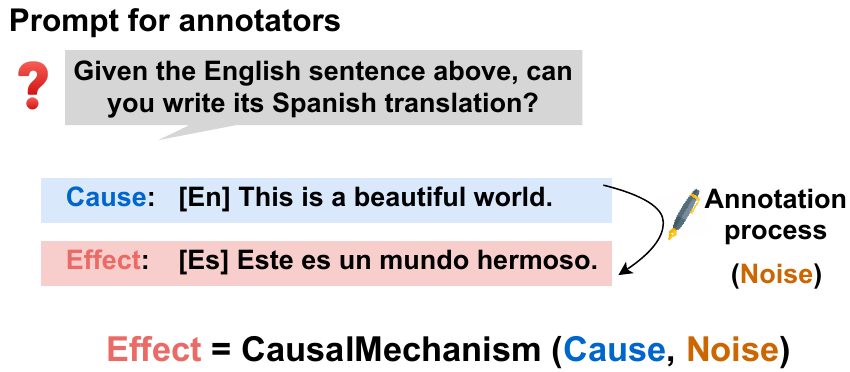}
    \caption{Annotation process for NLP data: the random variable that exists first is typically the cause (e.g., a given prompt), and the one generated afterwards is typically the effect (e.g., the annotated answer).}
    \label{icm:fig:prompt}
\end{figure}
First, we introduce the notion of the \textit{causal direction} for a given NLP task, see~\cref{icm:fig:prompt} for an example.
Throughout, we denote the
input of a learning task by~$X$ and the output which is to be predicted by~$Y$.
If, during the data collection process,  $X$ is generated first, and then $Y$ is collected based on
$X$ (e.g., through annotation), we say that $X$ causes $Y$, and denote this by $X\rightarrow Y$.
If, on the other hand,
 $Y$ is generated first, and then $X$ is collected based on $Y$, we say that $Y$ causes $X$ ($Y\rightarrow X$).\footnote{This corresponds to an \textit{interventional} notion of causation: if one were to manipulate the cause,  the annotation process would lead to a potentially different effect. A manipulation of the effect, in contrast, would not change the cause.}

\begin{figure}[ht]
    \centering
    \includegraphics[width=.6\columnwidth]{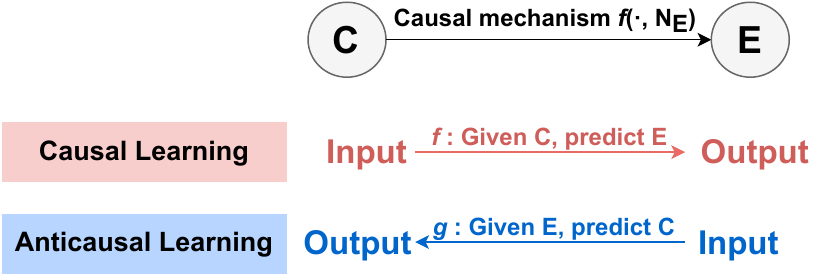}
    \caption{\textit{(Top)} A causal graph $C \rightarrow E$, where $C$ is the cause and $E$ is the effect. The function $f(\cdot, N_E)$ denotes the causal process, or mechanism, $P_{E|C}$ by which the effect $E$ is generated from $C$ and  unobserved noise $N_E$. 
    \textit{(Bottom)} Based on whether the direction of prediction aligns with the direction of causation or not, we distinguish two types of tasks: (i)
    {causal learning}, i.e., predicting the effect from the cause;
    and (ii)
    {anticausal learning}, i.e., predicting the cause from the effect.
    }
    \label{icm:fig:causal_graph}
\end{figure}

Based on whether the direction of prediction
aligns with the causal direction of the data collection process or not, 
 \citet{schoelkopf2012causal} categorize these types of tasks as \textit{causal learning} ($X\rightarrow Y$), or \textit{anticausal learning} ($Y\rightarrow X$), respectively; see~\cref{icm:fig:causal_graph} for an illustration.
 In the context of our motivating MT example
 this means that,
 if the  goal is to translate from English ($X=\text{En}$) 
 into Spanish ($Y=\text{Es}$),  training \textit{only} on subset (i) of the data
 consisting of En$\rightarrow$Es pairs 
 corresponds to \textit{causal learning} ($X\rightarrow Y$), whereas training \textit{only} on subset (ii) consisting of Es$\rightarrow$En pairs is categorised as \textit{anticausal learning} ($Y\rightarrow X$).

Based on the principle of independent causal mechanisms (ICM)~\citep{janzing2010causal,peters2017elements}, it has been hypothesized that the causal direction of data collection (i.e., whether a given NLP learning task can be classified as causal or anticausal) has implications for the effectiveness of commonly used techniques such as SSL and domain adaptation (DA)~\citep{schoelkopf2012causal}. We will argue that this can explain performance differences reported by the NLP community across different data collection processes and tasks.
In particular, we make the following contributions:
\begin{enumerate}[itemsep=-4pt,topsep=1pt]
    \item We categorize a number of common NLP tasks according to the causal direction of the underlying data collection process~(\cref{icm:sec:causal_categorization}).
    \item We review the ICM principle and its implications for common techniques of using unlabelled data such as SSL and DA
    in the context of causal and anticausal NLP tasks~(\cref{icm:sec:implications_of_causal_anticausal_learning_for_NLP}).
    \item We empirically assay the validity of ICM for NLP data using minimum description length in a machine translation setting~(\cref{icm:sec:mdl}).
    \item We verify experimentally and through a meta-study of over respectively 100 (SSL) and 30 (DA) published findings that the difference in SSL~(\cref{icm:sec:ssl}) and domain adaptation (DA)~(\cref{icm:sec:da}) performance on causal vs anticausal datasets reported in the literature
    is consistent with what is predicted by the ICM principle.
    \item We make suggestions on how to use findings in this paper for future work in NLP~(\cref{icm:sec:how_to_use}).
\end{enumerate}

\section{Categorization of Common NLP Tasks}
\label{icm:sec:causal_categorization}

We start by categorizing common NLP tasks which use an input variable $X$ to predict a target or output variable $Y$ into causal learning ($X\rightarrow Y$), anticausal learning ($Y\rightarrow X$), and other tasks that do not have a clear underlying causal direction, or which typically rely on mixed (causal and anticausal)
types of data, as summarised in~\cref{icm:tab:nlp_task_class}.

\newcommand{\ra}[1]{\renewcommand{\arraystretch}{#1}}
\begin{table}[t]
    \centering \small
    \ra{1.2}
    \small
    \begin{tabular}{m{4.5cm}m{10cm}l}
    \toprule
    {Category} & {Example NLP Tasks} \\ \midrule
    {Causal learning} & Summarization, parsing, tagging, data-to-text generation, information extraction \\ \hline
    {Anticausal learning} & Author attribute classification, review sentiment classification \\ \hline
    {Other/mixed (depending on data collection)} & Machine translation, question answering, question generation, text style transfer, intent classification \\
    \bottomrule
    \end{tabular}
    \caption{Classification of typical NLP tasks into causal (where the model takes the cause as input and predicts the effect), and anticausal (where the model takes the effect as input and predicts the cause) learning problems, as well as other tasks which do not have a clear causal interpretation of the data collection process, or where a mixture of both types of data is typically used.}
    \label{icm:tab:nlp_task_class}
\end{table}

Key to this categorization is determining whether the input $X$ corresponds to the cause or the effect in the data collection process.
As illustrated in~\cref{icm:fig:prompt},
if the input $X$ and output $Y$ are generated at two different time steps, then the variable that is generated first is typically the cause, and the other that is subsequently generated
 is typically the effect, provided it is generated based on the previous one (rather than, say, on a common confounder that causes both variables).
If $X$ and $Y$ are generated jointly, then we need to distinguish based on the underlying generative process whether one of the two variables is causing the other variable.

\paragraph{Learning Effect from Cause (Causal Learning)}
Causal ($X\rightarrow Y$) NLP tasks typically aim to predict a post-hoc generated human annotation (i.e., the target $Y$ is the effect) from a given input $X$ (the cause).
Examples include: summarization (\textit{article$\rightarrow$summary}) where the goal is to produce a summary $Y$ of a given input text $X$;
parsing and tagging (\textit{text$\rightarrow$linguists' annotated structure}) where the goal is to predict an annotated syntactic structure $Y$ of a given input sentence $X$; data-to-text generation (\textit{data$\rightarrow$description}) where the goal is to produce a textual description $Y$ of a set of structured input data $X$; 
and information extraction (\textit{text$\rightarrow$entities/relations/etc}) where the goal is to extract structured information
from a given text.

\paragraph{Learning Cause from Effect (Anticausal Learning)}
Anticausal ($Y\rightarrow X$) NLP tasks typically aim to predict or infer some latent target property $Y$ such as an unobserved prompt from an observed input $X$ which takes the form of one of its effects.
Typical anticausal NLP learning problems include, for example, author attribute identification (\textit{author attribute$\rightarrow$text}) where the goal is to predict some unobserved attribute $Y$ of the writer of a given text snippet $X$;
and review sentiment classification (\textit{sentiment$\rightarrow$review text})  where the goal is to predict the latent sentiment $Y$ that caused an author to write a particular review $X$.

\paragraph{Other/Mixed}
Some tasks can be categorized as either causal or anticausal, depending on how exactly the data is collected.
In~\cref{icm:sec:intro}, we discussed the example of MT where different types of (causal and anticausal) data are typically mixed.
Another example is the task of intent classification: if 
the \textit{same} author reveals their intent before the writing (i.e., \textit{intent$\rightarrow$text}),  it can be viewed as an anticausal learning task; if, on the other hand, the data is annotated by \textit{other} people who are not the original author (i.e., \textit{text$\rightarrow$annotated intent}),  it can be viewed as a causal learning task.
A similar reasoning applies to question answering and generation tasks which respectively aim to provide an answer to a given question, or vice versa: if first a piece of informative text is selected and annotators are then asked to come up with a corresponding question (\textit{answer$\rightarrow$question}) as, e.g., in the SQuAD dataset~\citep{rajpurkar-etal-2016-squad}, then question answering is an anticausal and question generation a causal learning task; if, on the other hand, a question such as a search query is selected first and subsequently an answer is provided (\textit{question$\rightarrow$answer}) as, e.g., in the Natural Questions dataset~\citep{kwiatkowski2019natural}, then question answering is a causal and question generation an anticausal learning task. 
Often, multiple such datasets are combined without regard for their causal direction.

\section{Implications of ICM for Causal and Anticausal Learning}
\label{icm:sec:implications_of_causal_anticausal_learning_for_NLP}

\begin{figure*}[t]
    \centering
    \includegraphics[width=\textwidth]{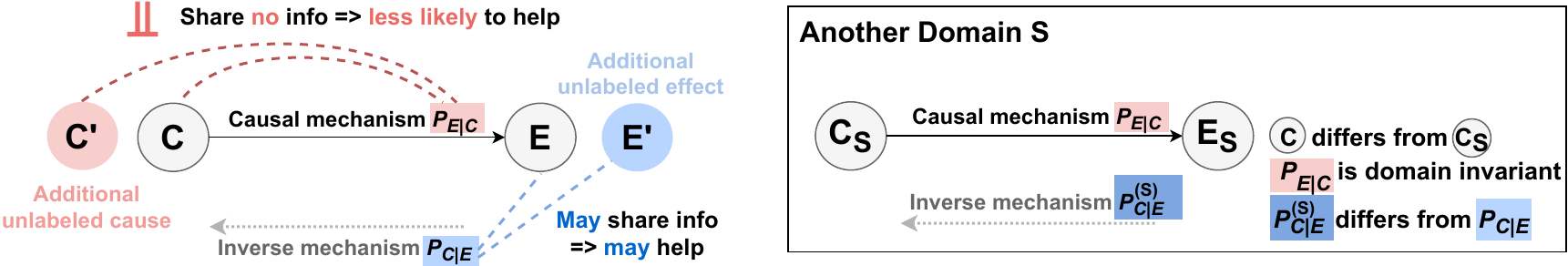}
    \caption{
    The ICM principle
    assumes that \textit{the generative process $P_C$ of the cause $C$ is independent of the causal mechanism $P_{E|C}$}: the two distributions share no information and each may be changed or manipulated without affecting the other. In the anticausal direction, on the other hand, the effect distribution \textit{$P_E$ is (in the generic case) not independent of the inverse mechanism $P_{C|E}$}: they may share information and change dependently.
    \textit{(Left)} SSL, which aims to improve an estimate of the target conditional $P_{Y|X}$ given additional unlabelled input data from $P_X$, should therefore not help for causal learning ($X\rightarrow Y$), 
    but may help in the anticausal direction ($Y\rightarrow X$).
    \textit{(Right)} DA,
    which aims to
    adapt a model of $P_{Y|X}$ from a source domain to a target domain (e.g., fine-tuning on a smaller dataset), should work better for causal learning settings where a change in $P_C$ is not expected to lead to a change in the mechanism $P_{E|C}$, whereas in the anticausal direction $P_E$ and $P_{C|E}$ may change in a \textit{dependent} manner.
    }
    \label{icm:fig:icm}
\end{figure*}

Whether we are in a causal or anticausal learning scenario has important implications for  semi-supervised learning (SSL)
and domain adaptation (DA)~\citep{schoelkopf2012causal,sgouritsa2015inference,zhang2013domain,zhang2015multi,gong2016domain,kugelgen2019semi,kugelgen2020semi}, which are techniques also commonly used in NLP.
These implications are derived from the principle of independent causal mechanisms (ICM)~\citep{schoelkopf2012causal,lemeire2006causal} which states that ``\textit{the causal generative process of a system's variables is composed of autonomous modules that do not inform or influence each other}''~\citep{peters2017elements}.

In the bivariate case, this amount to a type of independence assumption between 
the distribution~$P_C$ of the cause~$C$, and the causal process, or mechanism,%
~$P_{E|C}$ that generates the effect from the cause. 
For example, for a question answering task, the generative process~$P_C$ by which one person comes up with a question~$C$ is ``independent'' of the process $P_{E|C}$ by which another person produces an answer~$E$ for question~$C$.\footnote{The validity of this is meant in an approximate sense, and one can imagine settings where it is questionable. E.g., if the person asking the question has prior knowledge of the respondent (e.g., in a classroom setting), then she might adjust the question accordingly which would violate the assumption.}

Here, ``independent'' is not meant in the sense of \textit{statistical} independence of random variables, but rather as \textit{independence at the level of generative processes or distributions} in the sense that $P_C$ and $P_{E|C}$ \textit{do not share information} (the person asking the question and the one answering may not know each other) and \textit{can be manipulated independently of each other} (we can swap either of the two for another participant without the other one being influenced by this). 
Crucially, this type of independence is generally violated in the opposite, i.e., \textit{anticausal}, direction: $P_E$ and $P_{C|E}$ may share information and change dependently~\citep{DanJanMooZscSteZhaSch10,janzing2012informationgeometric}.
This has two important implications for common learning tasks~\citep{schoelkopf2012causal} which are illustrated in~\cref{icm:fig:icm}. 

\paragraph{Implications of ICM for SSL}
First, if $P_C$ shares no information with $P_{E|C}$, SSL---where one has additional unlabelled input data from $P_X$ and aims to improve an estimate of the target conditional $P_{Y|X}$---should not work in the causal direction ($X\rightarrow Y$), but may work in the anticausal direction ($Y\rightarrow X$), as $P_E$ and $P_{C|E}$ may share information.
Causal NLP tasks should thus be less likely  to show improvements over a supervised baseline when using SSL than anticausal tasks.

\paragraph{Implications of ICM for DA}
Second, according to the ICM principle, the causal mechanism $P_{E|C}$ should be invariant to changes in the cause distribution $P_C$, so domain---specifically, covariate shift~\citep{shimodaira2000improving,sugiyama2012machine}---adaptation, where $P_X$ changes but $P_{Y|X}$ is assumed to stay invariant, should work in the causal direction, but not necessarily in the anticausal direction.
Hence, 
DA should be easier for causal NLP tasks 
than for anticausal NLP tasks.

\section{Validity of ICM for NLP Data Using MDL}
\label{icm:sec:mdl}
Traditionally, the ICM principle is thought of in the context of \textit{physical} processes or mechanisms, rather than \textit{social} or \textit{linguistic} ones such as language.
Since ICM amounts to an independence assumption that---while well motivated in principle---may not always hold in practice,\footnote{E.g., due to confounding influences from unobserved variables, or mechanisms which have co-evolved to be dependent} we now assay its validity on NLP data.

Recall, that ICM postulates a type of independence between $P_C$ and $P_{E|C}$. One way to formalize this uses Kolmogorov complexity  $K(\cdot)$ as a measure of algorithmic information, which can be understood as the length of the shortest program that computes a particular algorithmic object such as a distribution or a function~\citep{
solomonoff1964formal,
kolmogorov1965three}.
ICM then reads~\citep{janzing2010causal}:\footnote{Here, $\overset{+}{=}$ and $\overset{+}{\leq}$ hold up a constant due to the choice of a Turing machine in the definition of algorithmic information.}
\begin{align}
\label{icm:eq:ICM_komogorov}
\begin{split}
K(P_{C,E})&\overset{+}{=} K(P_C) + K(P_{E|C}) \\
&\overset{+}{\leq} K(P_E)+K(P_{C|E})\,.
\end{split}
\end{align}
In other words, the shortest description of the joint distribution $P_{C, E}$ corresponds to describing $P_C$ and $P_{E|C}$ separately (i.e., they share no information), whereas there may be redundant (shared) information in the non-causal direction such that a  separate description of $P_E$ and $P_{C|E}$ will generally be longer than that of the joint distribution $P_{C, E}$.

\subsection{Estimation by MDL}\label{icm:sec:estimate_by_mdl}

Since Kolmogorov complexity is not computable \citep{li2008introduction}, we adopt a commonly used proxy, the minimum description length (MDL) \citep{grunwald2007minimum}, to test the applicability of ICM for NLP data.
Given an input, such as a collection of observations $\{(c_i,e_i)\}_{i=1}^n\sim P_{C,E}$, MDL returns the shortest codelength (in bits) needed to compress the input, as well as the
parameters needed to decompress it.
We use MDL to approximate~\cref{icm:eq:ICM_komogorov} as follows:
\begin{align}
    &\mathrm{MDL}(\bm{c}_{1:n},\bm{e}_{1:n})= \mathrm{MDL}(\bm{c}_{1:n}) + \mathrm{MDL}(\bm{e}_{1:n}|\bm{c}_{1:n}) 
    \nonumber
    \\
   & \quad\quad  \leq \mathrm{MDL}(\bm{e}_{1:n}) + \mathrm{MDL} (\bm{c}_{1:n} | \bm{e}_{1:n}),
\label{icm:eq:mdl_ineq}
\end{align}%
where MDL($\cdot|\cdot$) denotes a conditional compression where the second argument is treated as ``free parameters'' which do not count towards the compression length of the first argument.
\cref{icm:eq:mdl_ineq} can thus be interpreted as a comparison between two ways of compressing the same data $(\bm{c}_{1:n},\bm{e}_{1:n})$: either we first compress $\bm{c}_{1:n}$ and then compress $\bm{e}_{1:n}$ conditional on $\bm{c}_{1:n}$, or vice versa.
According to the ICM principle, 
the first way should tend to be more ``concise'' than the second.

\subsection{Calculating MDL Using Machine Translation as a Case Study}
\label{icm:sec:mdl_seq2seq}

To  empirically assess the validity
of ICM for NLP data using MDL as a proxy, we turn to MT as a case study.
We choose MT because the input and output spaces of MT are relatively symmetric, as opposed to other NLP tasks such as text classification where the input space is sequences, but the output space is a small set of labels. 

There are only very few studies which calculate MDL on NLP data, so we extend the method of~\citet{voita2020information} to calculate MDL using online codes~\citep{rissanen1984universal} for deep learning tasks \citep{blier2018description}. Since the original calculation method for MDL by~\citet{voita2020information} was developed for classification, we extend it to sequence-to-sequence (Seq2Seq) generation.
Specifically, given a translation dataset $D = \{(\bm{x}_1, \bm{y}_1), \dots, (\bm{x}_n, \bm{y}_n) \}$ of $n$ pairs of sentences $\bm{x}_i$ with translation $\bm{y}_i$, denote the size of the vocabulary of the source language 
by $V_x$, and the size of the vocabulary of the target language
by $V_y$.
In order to assess whether~\cref{icm:eq:mdl_ineq} holds, we need to calculate four different terms: two marginal terms $\mathrm{MDL}(\bm{x}_{1:n})$ and $\mathrm{MDL}(\bm{y}_{1:n})$, and two conditional terms $\mathrm{MDL}(\bm{y}_{1:n}|\bm{x}_{1:n})$ and $\mathrm{MDL}(\bm{x}_{1:n}|\bm{y}_{1:n})$.

\paragraph{Codelength of the Conditional Terms}
To calculate the codelength of the two conditional terms,
we extend the method of \citet{voita2020information} from classification to Seq2Seq generation. Following the setting of \citet{voita2020information}, we break the dataset $D$ into 10 disjoint subsets with increasing sizes and denote the end index of each subset as~$t_i$.\footnote{The sizes of the 10 subsets are 0.1, 0.2, 0.4, 0.8, 1.6, 3.2, 6.25, 12.5, 25, and 50 percent of the dataset size, respectively. E.g., $t_1=0.1\% n, t_2 = (0.1\%+0.2\%) n, \dots$.
}
We then estimate  $\mathrm{MDL}(\bm{y}_{1:n} |\bm{x}_{1:n})$ as
\begin{align}
    & \widehat{\mathrm{MDL}}(\bm{y}_{1:n} |\bm{x}_{1:n}) = {\textstyle \sum_{i=1}^{t_1}} \mathrm{length}(\bm{y}_i) \cdot
\log_2 V_y
\nonumber\\ 
& - {\textstyle\sum_{i=1}^{n-1}} \log_2 p_{\theta_i} (\bm{y}_{1+t_i:t_{i+1}} | \bm{x}_{1+t_i:t_{i+1}})
~, 
\label{icm:eq:mdl_yx}
\end{align}
where $\mathrm{length}(\bm{y}_i)$ refers to the number of tokens in the sequence $\bm{y}_i$, $\theta_i$ are the parameters of a translation model $h_i$ trained on the first $t_i$ data points, and $\bm{\mathrm{seq}}_{\mathrm{idx}_1: \mathrm{idx}_2}$ refers to the set of sequences from the $\mathrm{idx}_1$-th to the $\mathrm{idx}_2$-th sample in the dataset $D$, where $\bm{\mathrm{seq}} \in \{\bm{x}, \bm{y}\}$ and $\mathrm{idx}_i \in \{1, \dots, n\}$.
Similarly, when calculating $\mathrm{MDL}(\bm{x}_{1:n}|\bm{y}_{1:n})$, we simply swap the roles of $\bm{x}$ and $\bm{y}$.

\begin{table}[t]
\ra{1.2}
    \centering
    \small
    \begin{tabular}{llll}
    \toprule
    \textbf{Dataset} & \textbf{Size} 
    & \textbf{Note}
    \\ \midrule
    En$\rightarrow$Es & 81K & Original English, Translated Spanish \\
    Es$\rightarrow$En & 81K & Original Spanish, Translated English \\
    En$\rightarrow$Fr & 16K & Original English, Translated French \\
    Fr$\rightarrow$En & 16K & Original French, Translated English \\
    Es$\rightarrow$Fr & 15K & Original Spanish, Translated French \\
    Fr$\rightarrow$Es & 15K & Original French, Translated Spanish \\
    \bottomrule
    \end{tabular}
    \caption{Details of the CausalMT corpus.
    }
    \label{icm:tab:causalmt}
\end{table}

\begin{table*}[th!]
    \centering
    \small
    \ra{1.1}
    \resizebox{\textwidth}{!}{
    \begin{tabular}{>{\centering\arraybackslash}m{2cm}cccc>{\centering\arraybackslash}m{6.5cm}}
    \toprule
    \textbf{Data (X$\rightarrow$Y)} & \textbf{MDL(X)} & \textbf{MDL(Y)} & \textbf{MDL(Y|X)} & \textbf{MDL(X|Y)} & \textbf{MDL(X)+MDL(Y|X)} 
    \textbf{vs. MDL(Y)+MDL(X|Y)} \\ \midrule
    En$\rightarrow$Es & 46.54 & 105.99 & 2033.95 & 2320.93 & 2080.49 $<$ 2426.92 \\
    Es$\rightarrow$En & 113.42 & 55.79 & 3289.99 & 3534.09 & 3403.41 $<$ 3589.88 \\
    En$\rightarrow$Fr & 20.54 & 53.83 & 503.78 & 535.88 & 524.32 $<$ 589.71 \\
    Fr$\rightarrow$En & 53.83 & 21.6 & 705.28 & 681.12 & 759.11 $>$ 702.72 \\
    Es$\rightarrow$Fr & 58.26 & 55.66 & 701.04 & 755.5 & 759.30 $<$ 811.16 \\
    Fr$\rightarrow$Es & 56.14 & 54.34 & 665.26 & 706.53 & 721.40 $<$ 760.87 \\
    \bottomrule
    \end{tabular}
    }
    \caption{Codelength (in kbits) of $\mathrm{MDL}(X)$, $\mathrm{MDL}(Y)$, $\mathrm{MDL}(Y|X)$, and $\mathrm{MDL}(X|Y)$ on six CausalMT datasets.}
    \label{icm:tab:mdl_res}
\end{table*}

\paragraph{Codelength of the Marginal Terms}
When calculating the two marginal terms, $\mathrm{MDL}(\bm{x}_{1:n})$ and $\mathrm{MDL}(\bm{y}_{1:n})$, we make two changes from the above calculation of conditional terms: first, we replace the \textit{translation models} $h_i$ with \textit{language models}; second, we remove the conditional distribution. That is, we calculate $\mathrm{MDL}(\bm{x}_{1:n})$ as
\begin{equation}
\begin{split}
\widehat{\mathrm{MDL}}(\bm{x}_{1:n}) & = 
{\textstyle\sum_{i=1}^{t_1}} \mathrm{length}(\bm{x}_i) \cdot \log_2 V_x
\\ 
& - {\textstyle\sum_{i=1}^{n-1}} \log_2 p_{\theta_i} (\bm{x}_{1+t_i:t_{i+1}} )
~,
\end{split}
\end{equation}
where $\theta_i$ are the parameters of a language model~$h_i$ trained on the first $t_i$ data points. We apply the same method to calculate $\mathrm{MDL}(\bm{y}_{1:n})$.

For the language model, we use GPT2 \citep{radford2019language}, and for the translation model, we use the Marian neural machine translation model \citep{mariannmt} trained on the OPUS Corpus \citep{tiedemann-nygaard-2004-opus}. 
For fair comparison, all models adopt the transformer architecture \citep{vaswani2017attention}, and 
have roughly the same number of parameters.
See \cref{icm:appd:mdl} for more experimental details.

\subsection{CausalMT Corpus}
\label{icm:sec:mdl_mt}

For our MDL experiment,
we need datasets for which the causal direction of data collection is known, i.e., for which we have ground-truth annotation of which text is the original and which is a translation, instead of a mixture of both.
Since existing MT corpora do not have this property as discussed in~\cref{icm:sec:intro}, we curate our own corpus, which we call the CausalMT corpus.

Specifically, we consider the existing MT dataset WMT'19,\footnote{\href{http://www.statmt.org/wmt19/parallel-corpus-filtering.html}{Link to WMT'19}.} and identify some subsets that have a clear notion of causality. The subsets we use are the EuroParl~\citep{koehn2005europarl} and Global Voices translation corpora.\footnote{\href{http://casmacat.eu/corpus/global-voices-tar-balls/training.tgz}{Link to Global Voices}.}
For EuroParl, each text has meta information such as the speaker's language; for Global Voices, each text has meta information about whether it is translated or not. We regard  text that is in the same language as the speaker's native language in EuroParl (and non-translated text in Global Voices) as the original (i.e., the cause). We then retrieve a corresponding effect by using the cause text to match the parallel pairs in the processed dataset.
In this way, we compile six translation datasets with clear causal direction as summarized in~\cref{icm:tab:causalmt}. For each dataset, we use 1K samples each as test and validation sets, and use the rest for training.

\subsection{Results}
The results of our MDL experiment on the six CausalMT datasets are summarised in~\cref{icm:tab:mdl_res}. 
If ICM holds, we expect the sum of codelengths to be smaller for the causal direction than for the anticausal one, see~\cref{icm:eq:mdl_ineq}.
As can be seen from the last column, this is the case for five out of the six datasets.
For example, on one of the  largest datasets (En$\rightarrow$Es), the MDL difference is 346 kbits.\footnote{
As far as we know, determining statistical significance in the investigated setting remains an open problem. While, in theory, one may use information entropy to estimate it, in practice, this may be inaccurate since (i) MDL is only a proxy for algorithmic information; and (ii) ICM may not hold exactly, but only approximately. We evaluate on six different datasets, so that the overall results can show a general trend.
}

Comparing the dataset sizes in~\cref{icm:tab:causalmt} and results in~\cref{icm:tab:mdl_res}, we observe that the absolute MDL values are roughly proportional to  dataset size, but other factors such as language and task complexity also play a role.
This is inherent to the nature of MDL being the sum of codelengths of the model and of the data given the model. 
Since we use equally-sized datasets for each language pair in the CausalMT corpus (i.e., in both the $X\rightarrow Y$ and $Y\rightarrow X$ directions, see~\cref{icm:tab:causalmt}), numbers for the same language pair in~\cref{icm:tab:mdl_res}, including the most important column ``MDL(X)+MDL(Y|X) vs.\ MDL(Y)+MDL(X|Y)'', 
form a valid comparison. 
That is, En\&Es experiments are comparable within themselves, so are the other language pairs.

For some of the smaller differences in the last column in~\cref{icm:tab:mdl_res}, and, in particular the reversed inequality in row 4, a potential explanation may be the relatively small dataset size, as well as the fact that text data may be confounded (e.g., through shared grammar and semantics). 

\section{SSL for Causal vs. Anticausal Models}\label{icm:sec:ssl}
In semi-supervised learning (SSL), we are given a typically-small set of $k$ labeled observations $D_L = \{(\bm{x}_1, \bm{y}_1), \dots, (\bm{x}_k, \bm{y}_k) \}$, and a typically-large set of $m$ unlabeled observations of the input $D_U = \{\bm{x}_1^{(u)}, \dots, \bm{x}_m^{(u)}\}$.
SSL then aims to use the additional information about the input distribution
$P_X$
from the unlabeled dataset $D_U$ to improve a model of $P_{Y|X}$ learned on the labeled dataset $D_L$.

As explained in~\cref{icm:sec:implications_of_causal_anticausal_learning_for_NLP}, SSL should only work for anticausal (or confounded) learning tasks, according to the ICM principle.
\citet{schoelkopf2012causal} have observed this trend on a number of classification and regression tasks on small-scale numerical inputs, such as predicting Boston housing prices from quantifiable neighborhood features (causal learning), or breast cancer from lab statistics (anticausal learning). 
However, there exist no studies investigating the implications of ICM for SSL on NLP data, which is of a more complex nature due to the high dimensionality of the input and output spaces, as well as potentially large confounding. In the following, we use a sequence-to-sequence decipherment experiment~(\cref{icm:sec:decipherment}) and a meta-study of existing literature~(\cref{icm:sec:meta_ssl}) to showcase that the same phenomenon also occurs in 
NLP.

\subsection{Decipherment Experiment}
\label{icm:sec:decipherment}
To have control over causal direction of the data collection process, we use a synthetic decipherment dataset to test the difference in  SSL improvement between causal and anticausal learning tasks.

\paragraph{Dataset}
We create a synthetic dataset of encrypted sequences. Specifically, we (i) adopt a monolingual English corpus (for which we use the English corpus of the En$\rightarrow$Es in the CausalMT dataset, for convenience), (ii) apply the ROT13 encryption algorithm~\citep{schneier1996applied} to obtain the encrypted corpus, and  then (iii) apply noise on the corpus that is chosen to be the effect corpus.

In the encryption step (ii), for each English sentence $\bm{x}$, its encryption $\mathrm{ROT13}(\bm{x})$ replaces each letter with the 13th letter after it in the alphabet, e.g., ``A''$\rightarrow$``N,'' ``B''$\rightarrow$``O.''
Note that we choose ROT13 due to its invertibility, since $\mathrm{ROT13}(\mathrm{ROT13}(\bm{x}))=\bm{x}$.
Therefore, without any noises, the corpus of English and the corpus of encrypted sequences by ROT13 are symmetric.

In the noising step (iii), we apply noise either to the English text or to the ciphertext, thus creating two datasets Cipher$\rightarrow$En, and En$\rightarrow$Cipher, respectively.
When applying noise to a sequence, we use the implementation of the Fairseq library.\footnote{
\href{https://github.com/pytorch/fairseq/blob/master/fairseq/data/denoising\_dataset.py}{Link to the Fairseq implementation}.
}
Namely, we mask some random words in the sequence (word masking), permute a part of the sequence (permuted noise), randomly shift the endings of the sequence to the beginning (rolling noise), and insert some random characters or masks to the sequence (insertion noise). We set the probability  of all noises to $p=5\%$. 

\begin{table}[t]
    \centering \small
    \ra{1.1}
    \small
    \begin{tabular}{lllll}
    \toprule
    \textbf{Causal Data} & \textbf{Learning Task} & \textbf{Sup. BLEU} & \textbf{$\Delta$SSL (BLEU)} \\ \midrule
    \multirow{2}{*}{En$\rightarrow$Cipher} & Causal
    & 19.20 & +1.84 \\
    & Anticausal & 7.75 & +38.02 \\ \hline
    \multirow{2}{*}{Cipher$\rightarrow$En} & Causal
    & 17.08 & +4.05 \\
    & Anticausal & 7.97 & +38.01 \\ 
    \bottomrule
    \end{tabular}
    \caption{SSL improvements ($\Delta$SSL) in BLEU score across causal vs.\ anticausal learning tasks on the synthetic decipherment datasets.
    }
    \label{icm:tab:cipher}
\end{table}

\paragraph{Results}
For each of the two datasets En$\rightarrow$Cipher and Cipher$\rightarrow$En, we perform SSL in the causal and anticausal direction by either treating the input $X$ as the cause and the target $Y$ as the effect, or vice versa.
Specifically, we use a standard Transformer architecture for the supervised model, and for SSL, we multitask the translation task with an additional denoising autoencoder~\citep{vincent2008extracting} using the Fairseq Python package.
The results are shown
in~\cref{icm:tab:cipher}.
It can be seen that in both cases, anticausal models show a substantially larger SSL improvement  than causal models.

We also note that there is a substantial gap in the supervised performance between causal and anticausal learning tasks on the same underlying data. 
This is also expected as 
causal learning is typically easier than anticausal learning since it corresponds to learning the ``natural'' forward function, or causal mechanism, while anticausal learning corresponds to learning the less natural, non-causal inverse mechanism.

\subsection{SSL Improvements in Existing Work}\label{icm:sec:meta_ssl}

After verifying the different behaviour in SSL improvement predicted by the ICM principle on the decipherment experiment, we conduct an extensive meta-study to survey whether this trend is also reflected in published NLP findings.
To this end, we consider a diverse set of tasks, and SSL methods. The tasks covered in our meta-study include machine translation, summarization, parsing, tagging, information extraction, review sentiment classification, text category classification, word sense disambiguation, and chunking. The SSL methods include self-training, co-training~\citep{blum1998combining}, tri-training~\citep{zhou2005tri}, transductive support vector
machines~\citep{joachims1999transductive}, expectation maximization~\citep{nigam2006semi}, multitasking with language modeling~\citep{dai2015semi}, multitasking with sentence reordering (as used in~\citet{zhang-zong-2016-exploiting}), and cross-view training~\citep{clark2018semi}. Further details on our meta study are explained in \cref{icm:appd:meta_ssl}.

\begin{table}[t]
    \centering
    \ra{1.1}
    \small
    \begin{tabular}{lclc}
    \toprule
    \textbf{Task Type} & \textbf{Mean} \textbf{$\Delta$SSL ($\pm$std)} & \textbf{According to ICM} \\ \midrule
    Causal & +0.04 ($\pm$4.23) & Smaller or none \\
    Anticausal & +1.70 ($\pm$2.05) & Larger \\
    \bottomrule
    \end{tabular}
    \caption{Meta-study of SSL improvement ($\Delta$SSL) across 55  causal and 50 anticausal NLP tasks. 
    }
    \label{icm:tab:meta_ssl}
\end{table}

We covered 55 instances of causal learning and 50 instances of anticausal learning. A summary of the trends of causal SSL and anticausal SSL are listed in \cref{icm:tab:meta_ssl}. Echoing with the implications of ICM stated in~\cref{icm:sec:implications_of_causal_anticausal_learning_for_NLP}, for causal learning tasks, the average improvement by SSL is only very small, 0.04\%. In contrast, the anticausal SSL improvement is larger, 1.70\% on average. We use Welch's t-test~\citep{welch1947generalization} to assess whether the difference in mean between the two distributions of SSL improvment (with unequal variance) is significant 
and obtain a p-value of 0.011.

\section{DA for Causal vs. Anticausal Models}\label{icm:sec:da}

We also consider a supervised domain adaptation (DA) setting in which the goal is to adapt a model trained on a  large labeled data set 
from a source domain, to a potentially different target domain from which we only have a
a small labeled data set.
As explained in~\cref{icm:sec:implications_of_causal_anticausal_learning_for_NLP}, DA 
should only work well for causal learning, but not necessarily for anticausal learning, according to the ICM principle.

Similar to the meta-study on SSL, we also review existing NLP literature on DA. We focus on DA improvement, i.e., the performance gain of using DA over an unadapted baseline that only learns from the source data and is tested on the target domain. 
Since the number of studies on DA that we can find is smaller than for SSL, we cover 22 instances of DA on causal tasks, and 11 instances of DA on anticausal tasks.

\begin{table}[t]
\small\ra{1.1}
    \centering
    \begin{tabular}{lclc}
    \toprule
    \textbf{Task Type} & \textbf{Mean} $\Delta$\textbf{DA} ($\pm$std) & \textbf{According to ICM} \\ \midrule
    Causal & 5.18 ($\pm$6.57) & Larger \\
    Anticausal & 1.26 ($\pm$1.79) & Smaller \\
    \bottomrule
    \end{tabular}
    \caption{Meta-study of DA improvement ($\Delta$DA) across 22 causal and 11 anticausal NLP tasks.
    }
    \label{icm:tab:da}
\end{table}

The results are summarised in~\cref{icm:tab:da}. We find that the observations again echo with our expectations (according to ICM) that DA should work better for causal, than for anticausal learning tasks.
Again, we use Welch's t-test~\citep{welch1947generalization} to verify that the DA improvements of causal learning and anticausal learning are statistically different, and obtain a p-value of 0.023.

\section{How to Use the Findings in this Study}\label{icm:sec:how_to_use}
\paragraph{{Data Collection Practice in NLP}}
{Due to the different implications of causal and anticausal learning tasks, \textit{we strongly suggest annotating the causal direction when collecting new NLP data.}
One way to do this is to only collect data from one causal direction and to mention this in the meta information.
For example, summarization data collected from the TL;DR of scientific papers \text{SciTldr}~\citep{cachola-etal-2020-tldr} should be \textit{causal}, as the TL;DR summaries on OpenReview (some from authors when submitting the paper, others derived from the beginning of peer reviews) were  likely composed after the original papers or reviews were written.
Alternatively, one may allow mixed corpora, but label the causal direction for each $(\bm{x}, \bm{y})$ pair, e.g., which is the original vs.\ translated text in a translation pair.
Since more data often leads to better model performance, it is common to mix data from both causal directions, e.g., training on both En$\rightarrow$Es and Es$\rightarrow$En data.
Annotating the causal direction for each pair allows future users of the dataset to potentially 
handle the causal and anticausal parts of the data differently.
}

\paragraph{Causality-Aware Modeling}
{
When building NLP models, the causal direction provides additional information that can
potentially be built into the model.
In the MT case, since
causal and anticausal learning can lead to different
performance~\citep{ni-etal-2022-original}, 
one way to take advantage of the known causal direction is to add a prefix such as ``[Modeling-Effect-to-Cause]'' to the original input, so that the model can learn from causally-annotated input-output pairs. For example,~\citet{riley-etal-2020-translationese} use labels of the causal direction
to elicit different behavior at inference time.
Another option is to carefully design a combination of different modeling techniques, such as limiting self-training (a method for SSL) only to the anticausal direction and allowing back-translation in both directions, as preliminarily explored by~\citet{shen-etal-2021-source}.
}

\paragraph{Causal Discovery}
{
Suppose that we are given measurements of two types of NLP data $X$ and $Y$ (e.g., text, parse tree, intent type) whose collection process is unknown, i.e., which is the cause and which the effect. 
One key finding of our study is that there is typically a causal footprint of the data collection process which manifests itself, e.g., when computing the description length in different directions~(\cref{icm:sec:mdl}) or when performing SSL~(\cref{icm:sec:ssl}) or DA~(\cref{icm:sec:da}). Based on which direction has the shorter MDL, or allows better SSL or DA, we can thus infer one causal direction over the other.
}

\paragraph{Prediction of SSL and DA Effectiveness}
Being able to predict the effectiveness of SSL or DA for a given NLP task can be very useful, e.g., to set the weights in an ensemble of different models~\citep{sogaard2013semi}.
While predicting SSL performance has previously been studied from a non-causal perspective~\citep{nigam2000analyzing, asch2016predicting}, our findings suggest that a simple qualitative description of the data collection process in terms of its causal direction (as summarised for the most common NLP tasks in~\cref{icm:tab:nlp_task_class}) can also be surprisingly effective to evaluate whether SSL or DA should be expected to work well.

\section{{Limitations and Future Work}}
{
We note that ICM---when taken strictly---is an idealized
assumption that may be violated and thus may not hold exactly for a given
real-world data set, e.g., due to 
confounding, i.e., when both variables are influenced by a third, unobserved variable.
In this case, one may observe less of a difference between causal and anticausal learning tasks.
}

{We also note that, while we have made an effort to classify different NLP tasks as \textit{typically} causal or anticausal, our categorization should not be applied blindly without regard for the specific generative process at hand: deviations are possible as explained in the Mixed/Other category.}

Another limitation is that the SSL and DA settings considered in this paper are only a subset of the various settings that exist in NLP. Our study does not cover, for example, SSL that uses additional output data (e.g.,~\citet{jean2015montreal,gulcehre2015using,sennrich-zhang-2019-revisiting}), or unsupervised DA (as reviewed by~\citet{ramponi-plank-2020-neural}).
In addition, in our meta-study of published  SSL and DA findings, the improvements of causal vs.\ anticausal learning might be amplified by the scale of research efforts on different tasks and potentially suffer from selection bias.

{Finally, we remark that, in the present work, we have focused on bivariate prediction tasks with an input $X$ and output $Y$. 
Future work may also apply ICM-based reasoning to more complex NLP settings, for example, by (i) incorporating additional (sequential/temporal) structure of the data (e.g., for MT or language modeling) or (ii) considering settings in which the input $X$ consists of both cause $X_\textsc{cau}$ and effect $X_\textsc{eff}$ features of the target $Y$~\citep{kugelgen2019semi,kugelgen2020semi}.}

\section{Related Work}

\noindent\textbf{NLP and Causality}
Existing work on NLP and causality mainly focuses on the extracting text features  for causal inference. Researchers first propose a causal graph based on domain knowledge, and then use text features to represent some elements in the causal graph, e.g., the cause~\citep{egami2018make,jin-etal-2021-mining-cause}, effect~\citep{fong2016discovery}, and confounders~\citep{roberts2020adjusting,veitch2020adapting,keith2020text}. Another line of work mines causal relations among events from textual expressions, and uses them to perform relation extraction~\citep{do-etal-2011-minimally,mirza-tonelli-2014-analysis,dunietz-etal-2017-corpus,hosseini2021predictting}, question answering~\citep{oh2016semi}, or commonsense reasoning~\citep{sap2019atomic,bosselut-etal-2019-comet}.
For a recent survey, we refer to~\citet{feder2021causal}.

\paragraph{Usage of MDL in NLP}
Although MDL has been used for causal discovery for low-dimensional data~\citep{budhathoki2017mdl,mian2021discovering,marx2021formally}, only very few studies adopt MDL on high-dimensional NLP data. Most existing uses of MDL on NLP are for probing and interpretability: e.g.,
\citet{voita2020information} use it for probing of a small Bayesian model and network pruning, based on the method proposed by~\citet{blier2018description} to calculate MDL for deep learning.
We are not aware of existing work using MDL for causal discovery, or to verify causal concepts such as ICM in the context of NLP.

\paragraph{Existing Discussions on SSL and DA in NLP}
SSL and DA has long been used in NLP, as reviewed by~\citet{sogaard2013semi} and~\citet{ramponi-plank-2020-neural}. However, there have been a number of studies that report negative results for SSL~\citep{clark-etal-2003-bootstrapping,steedman-etal-2003-bootstrapping,reichart-rappoport-2007-self,abney2007semisupervised,spreyer-kuhn-2009-data,sogaard-rishoj-2010-semi} and DA~\citep{plank2014importance}. 
Our works constitutes the first explanation of  the ineffectiveness of SSL and DA on certain NLP tasks from the perspective of causal and anticausal learning.

\section{Conclusion}
This work presents the first effort to use causal concepts such as the ICM principle and the distinction between causal and anticausal learning to shed light on some commonly observed trends in NLP.
Specifically, we provide an  explanation of observed differences in SSL~(\cref{icm:tab:cipher,icm:tab:meta_ssl}) and DA~(\cref{icm:tab:da}) performance on a number of NLP tasks: DA tends to work better for causal learning tasks, whereas SSL typically only works for anticausal learning tasks, as predicted by the ICM principle.
These insights, together with our categorization of common NLP tasks~(\cref{icm:tab:nlp_task_class}) into causal and anticausal learning, may prove useful for future NLP efforts.
Moreover, we empirically confirm using MDL that the description of data is typically shorter in the causal than in the anticausal direction~(\cref{icm:tab:mdl_res}), suggesting that a causal footprint can also be observed for text data. This has interesting potential implications for discovering causal relations between different types of NLP data.

\section*{Ethical Considerations}

\paragraph{Use of Data} 
This paper uses two types of data, a subset of an existing machine translation dataset, and synthetic decipherment data. As far as we know, there are no sensitive issues such as privacy regarding the data usage.

\paragraph{Potential Stakeholders}
This research focuses on meta properties of two commonly applied methodologies, SSL and DA in NLP. Although this research is not directly connected to specific applications in society, the usage of this study can benefit future research in SSL and DA.

\mainchapter{On the Causal Nature of Sentiment Analysis}\label{ch:psychcausal}

Sentiment analysis (SA) aims to identify the sentiment expressed in a text, such as a product review. Given a review and the sentiment associated with it, this work formulates SA as a combination of two tasks: (1) a causal discovery task that distinguishes whether a review ``primes'' the sentiment (Causal Hypothesis C1), or the sentiment ``primes'' the review (Causal Hypothesis C2); and (2) the traditional prediction task to model the sentiment using the review as input.
Using the peak-end rule in psychology, we classify a sample as C1 if its overall sentiment score approximates an average of all the sentence-level sentiments in the review, and C2 if the overall sentiment score approximates an average of the peak and end sentiments.
For the prediction task, we use the discovered causal mechanisms behind the samples to improve LLM performance by proposing \textit{causal prompts} that give the models an inductive bias of the underlying causal graph, leading to {substantial improvements} by up to 32.13 F1 points on zero-shot five-class SA.
Our code is available at \url{https://github.com/cogito233/causal-sa}.

\section{Introduction}
Sentiment analysis (SA) is the task of identifying the sentiment $y$ given a piece of text $x$.
The field has a rich history originating from subjectivity analysis \citep{wiebe1994tracking,hatzivassiloglou2000effects}, and developed rapidly with the availability of large opinionated online data such as reviews with ratings \cite[\textit{inter alia}]{turney2002thumbs,nasukawa2003sentiment,zhang2015character,keung-etal-2020-multilingual}. 

\begin{figure*}[t]
    \centering
    \includegraphics[width=\textwidth]{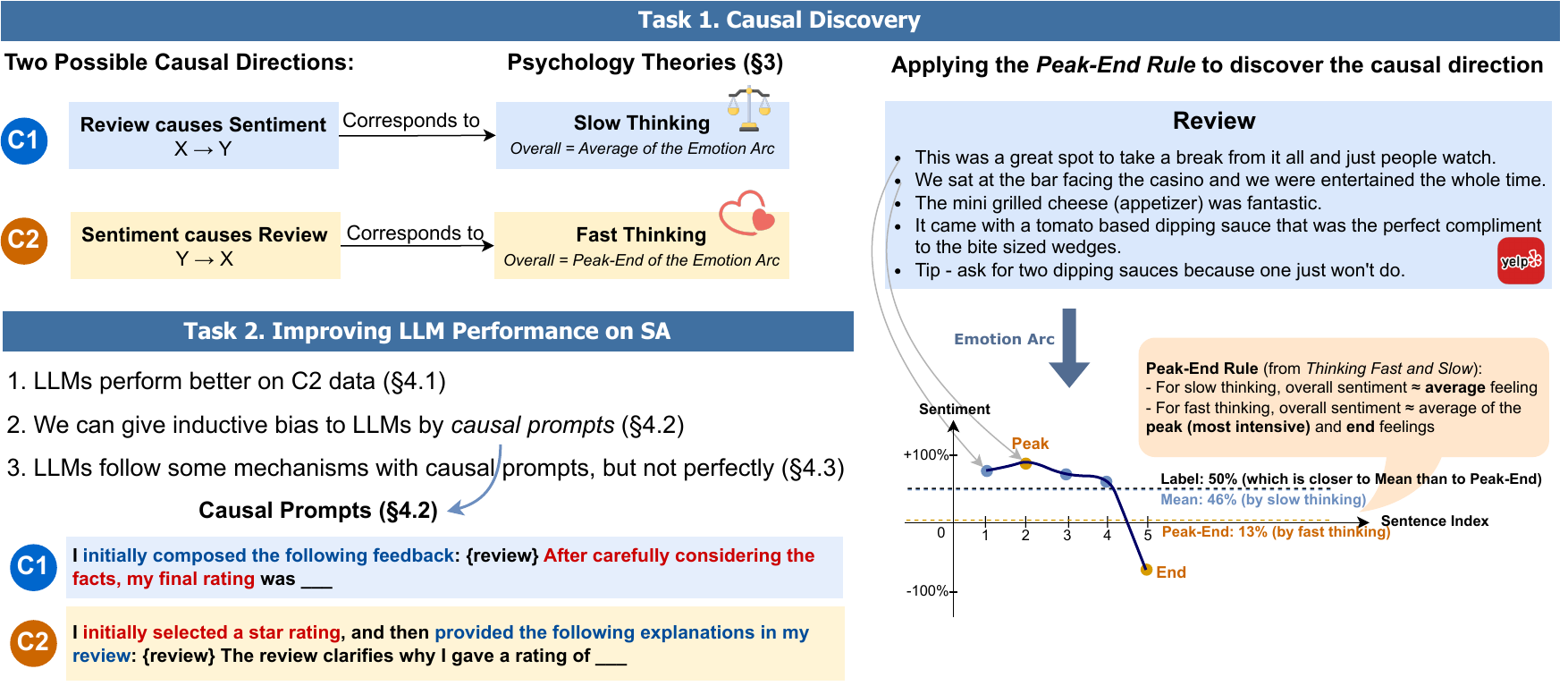}
    \caption{
    An overview of the paper structure, where we first investigate the causal discovery task, and then use it to improve LLM performance. For each document-level text review, we parse its \textit{emotion arc} consisting of the sentiment of each sentence in the review, and then use the peak-end rule \citep{kahneman1993more,kahneman2011thinking} to identify whether the overall sentiment is an average of the arc (corresponding to \textit{Slow Thinking}), or an average of the peak and end sentiments (corresponding to \textit{Fast Thinking}).
    }
    \label{psychcausal:fig:structure}
\end{figure*}

Despite recent advances in large language models (LLMs), 
it is still challenging to address the fine-grained five-class SA (which corresponds to the five star ratings in most datasets) for document-level classification \citep{Choi_2020,fei-etal-2023-reasoning,math9212722}, due to the subtle nature of the task including aspects such as inter-aspect relations, commonsense reasoning, 
among others 
\citep{poria2023beneath,venkit-etal-2023-sentiment}.

In this paper, we propose a causally-informed solution for the SA task. Different from the approach of naïvely applying up-to-date LLMs, we leverage insights from causal inference to
propose a reformulation for SA into two tasks, as in \cref{psychcausal:fig:structure}: (1) a causal discovery task to identify the cause-effect relation between the review $X$ and the sentiment $Y$, and (2) the traditional prediction task $f: {x} \mapsto y$ to model the sentiment using the review as input.

We first look into the causal discovery task.
In the study of affect science \citep{salovey2004emotional,barrett2006solving,feinstein2013lesion}, language can be the cause of emotion \citep{satpute2013functional,kassam2013effects} -- namely a review priming the following sentiment, i.e., the Causal Hypothesis C1 of $X\rightarrow Y$); or emotion can affect the use of language \citep{barrett2006solving} -- namely sentiment priming the review as an ad-hoc justification for the emotion, i.e., the Causal Hypothesis C2 of $Y\rightarrow X$.
These two processes might arise from the data annotation process \citep{jin-etal-2021-causal}, but is hard to discover post-hoc in existing datasets.

Given the possibility of both causal directions $X\rightarrow Y$ or $Y\rightarrow X$ in the SA data, we identify the actual underlying mechanism based on insights from psychology \citep{kahneman2011thinking,epstein1994integration}. Specifically, we identify the correspondence of the above two causal mechanisms with the \textit{Fast} and \textit{Slow Thinking} systems \citep{kahneman2011thinking}: (1) a review-driven sentiment (as in C1) largely resembles the Slow Thinking process applying reasoning based on evidence, and (2) the process of first coming up with the sentiment and then justifying it by a review (as in C2) conforms to Fast Thinking.
Given this correspondence, we apply the peak-end rule from psychology \citep{kahneman1993more,kahneman2011thinking}. As shown in the right part of \cref{psychcausal:fig:structure}, we classify a sample as C1 if its overall sentiment score approximates an average of all the sentence-level sentiments in the review, and as C2 if the overall sentiment score approximates an average of the peak and end sentiments.

Based on the identified causal mechanism behind SA data from the causal discovery task, we further explore how it can improve prediction performance in the era of LLMs. Existing literature highlights ``causal alignment,'' namely to align the prediction direction along the underlying causal direction \citep{jin-etal-2021-causal,scholkopf2022causality,scholkopf2021towards}, but to our knowledge we are the first to explore how causal alignment improves model performance of SA in the era of LLMs. Specifically, we answer three subquestions: (Q1) If using the standard SA prompt, do models perform differently on C1/C2 data? (Q2) Does it help if we make the prompt aware of the underlying causality, i.e., use \textit{causal prompts}? And (Q3) When prompted causally, 
do LLMs mechanistically understand the corresponding causal processes?

Our empirical results show that under the standard prompt, LLMs perform better on data corresponding to the C2 causal process. Moreover, causal prompts aligned with the causal direction of the data can substantially improve the performance of zero-shot SA. Finally, we apply mechanistic interpretability methods to probe the models, and find that there is still improvement space for LLMs to correctly grasp the essence of 
the two causal processes. 
In summary, the contributions of this paper are as follows:
\begin{enumerate}
    \item 
    We propose the dual nature of SA as a combination of two tasks: a causal discovery task, and a prediction task.
    \item 
    For causal discovery, we ground the two possible causal processes in psychology, and use the peak-end rule to identify them.
    \item 
For the prediction task, we inspect existing LLMs' performance on data corresponding to the two underlying causal processes, and design \textit{causal prompts} to improve model performance by up to 32.13 F1 points.
\end{enumerate}

\section{Problem Formulation of SA}
In this section, we formulate SA as a combination of two tasks: the traditional prediction task in NLP and the causal discovery task in statistics, which we will introduce in the following.

\subsection{The Prediction Task (in NLP)}\label{psychcausal:sec:pred}
SA is a prediction task to identify the sentiment $y$ given a piece of text $x$.
We adopt the setup in most existing SA datasets \citep{maas2011learning,zhang2015character,keung-etal-2020-multilingual}, where the text $x$ is a review consisting of $n$ sentences $(t_1, \dots, t_n)$, and the label $y$ is a sentiment score corresponding to the star rating of the review in 1 (most negative), 2, \dots, 5 (most positive).

\subsection{The Cause-Effect Discovery Task (in Statistics)}\label{psychcausal:sec:discovery}

As a separate problem, there is an established task in causal discovery, the causal-effect problem \cite[see the review by][]{janzing2019causeeffect}, which aims to tell the cause from effect using only observational data. Its formal formulation is as follows:
Suppose we have an i.i.d. dataset $\mathcal{D} :=\{(x_i, y_i)\}_{i=1}^n$ containing $n$ observational data pairs of the two variables, $X$ and $Y$. The task is to \textit{infer whether $X$ causes $Y$ (i.e., $X \rightarrow Y$), or $Y$ causes $X$ (i.e., $Y \rightarrow X$)}, if one out of the two is true. In causality, ``$\rightarrow$'' indicates the directional causal relation between two variables. The two hypotheses can also be expressed in their equivalent structural causal models \cite[SCMs;][]{pearl2009causality} as introduced in \citet{peters2017elements}:
\begin{align}
&\text{Causal Hypothesis 1 (C1): } 
 X \rightarrow Y  
\\
& \Leftrightarrow 
Y: = f_Y(X, N_Y) \text{ with } N_Y \perp X~,
\\\nonumber
\\
&\text{Causal Hypothesis 2 (C2): } 
 Y \rightarrow X  
\\
& \Leftrightarrow 
X: = f_X(Y, N_X) \text{ with } N_X \perp Y~,
\end{align}
where $N_{i}$ is an unobserved noise term orthogonal to the input distribution.

\subsection{Causality and NLP Model Performance}\label{psychcausal:sec:causality_helps_nlp}

For many years, causality and machine learning have been two separate domains on their own.
Recently, researchers started to think about how the causal knowledge of the data can improve machine learning performance on the prediction task, especially for the two variable cause-effect case \citep{schoelkopf2012causal,jin-etal-2021-causal,ni-etal-2022-original}.
The essence of this line of research is that causality makes the two learning tasks $x \mapsto y$ and $y \mapsto x$ asymmetric, as one function's prediction direction \textit{aligns with} the ground-truth causal direction behind the two random variables, and another contradicts.
We call this phenomenon ``\textit{causal alignment},'' or ``direction match,'' of the prediction task and the causality.

To contrast the contribution of our work, we review the 
previous literature on causal alignment, which only shows its effect on the performance of trained-from-scratch machine learning models, without any indications in the era of LLMs:

\begin{enumerate}
    \item 
Causal alignment makes a model more robust against \textbf{covariant shifts} \cite[\textit{inter alia}]{jin-etal-2021-causal,scholkopf2022causality,schott2018towards} 
\item 
Semi-supervised learning (SSL) only works under causal misalignment, as the cause variable contains no information about the mechanism, but the effect variable does. So in the misaligned case, additional $P_X$ (i.e., the effect variable) helps \textbf{SSL} \citep{schoelkopf2012causal,jin-etal-2021-causal}.

\item 
Learning a causally-aligned model induces \textbf{less Kolmogorov complexity} (a more minimal description length) than the causally-misaligned model on the same $X$-$Y$ data \citep{jin-etal-2021-causal,janzing2010causal}
\item
Causal alignment significantly affects model performance in \textbf{supervised learning}, in the case of machine translation \citep{ni-etal-2022-original}.
\end{enumerate}

All the above findings are drawn under the training condition that we can isolate the training data to be only of one causal direction.
In the era of LLMs, we have seen substantial differences: (1) the training data can be a mixture of both causal directions, (2) the operationalization of the prediction task is through prompting, but no longer a separate model for each direction, and,  (3) in general, research has shifted to designing better prompts for already pre-trained models in their inference mode. 

Given these changes, we use the rest of the paper to address the following research questions:
\begin{enumerate}
    \item What is the causal direction in SA? (\cref{psychcausal:sec:psych})
    \item Can causal alignment help us improve SA prompts in the era of LLMs? (\cref{psychcausal:sec:improve})
\end{enumerate}

\section{Causal Discovery of Sentiment and Review}

\label{psychcausal:sec:psych}

\subsection{Problem Setup
}\label{psychcausal:sec:paradigm}
As mentioned previously, the setup of the bivariate causal discovery problem is to infer whether $X$ causes $Y$ (C1), or $Y$ causes $X$ (C2), based on a dataset $\mathcal{D} :=\{(x_i, y_i)\}_{i=1}^n$ containing only observational data of the joint distribution.

\paragraph{Challenges}
The common paradigm to check causal discovery results is to generate simulated data, of which the ground truth causal graph is known \citep{zhang2009causality,spirtes2016causal}.
However, in the context of the established SA datasets, such as Yelp \citep{zhang2015character}, Amazon \citep{keung-etal-2020-multilingual}, and App Review \citep{app_review}, we would not be able to track each individual user and survey their original causal process when composing the review and the rating. 
Another solution would also be difficult, as it would require SA to abandon all the above well-established datasets, and meticulously collect new data while surveying the users' underlying causal process.

\paragraph{Our Approach}
In the context of our work, we propose that there are still rich findings that we could derive from the observation-only data in the existing datasets, without interviewing or conducting new costly data collection.

The key to our approach is the psychology theories of the two causal processes, as the relation between sentiment and text has been well-studied and verified by randomized control trials (RCTs), among many other experiments.
In the rest of the section, we first introduce in \cref{psychcausal:sec:psych_theory} the psychology theories of fast and slow thinking, followed by the Peak-End Rule as the quantitative signal. Then, we operationalize the theory with computational techniques in \cref{psychcausal:sec:emotion_arc}, and the report findings on three different SA datasets in \cref{psychcausal:sec:psych_exp}.

\subsection{Psychological Processes Underlying Sentiment Processing}\label{psychcausal:sec:psych_theory}

\paragraph{Two Systems of Emotional Responses}
In psychology, the bifurcation into System 1 and System 2 in human decision-making, including sentiment processing, has garnered substantial empirical support \citep{kahneman2011thinking,epstein1994integration}. 

System 1, or the ``\textit{Fast Thinking}'' system, operates involuntarily, effortlessly, and without conscious awareness. It is often optimized in evolution to provide rapid responses to environmental stimuli \citep{ledoux1998emotional}, and guides most of our daily cognitive processing \citep{kahneman2011thinking}, and emotional responses such as fear or joy \citep{zajonc1980feeling}.

Conversely, System 2, often termed as ``\textit{Slow Thinking},'' is deliberate, slower, and more rational, requires more conscious effort \citep{kahneman2011thinking}, and allows for self-regulation and thoughtful consideration before making decisions \citep{baumeister1998ego}.
The interplay between these systems influences everything from mundane to critical decisions, highlighting the complexity of human emotional and cognitive processing \citep{kahneman2002representativeness,kahneman2011thinking}.

\paragraph{Correspondence to the Two Causal Processes}
There is a nice correspondence between the fast/slow thinking systems and our two causal hypotheses. As mentioned previously, the Causal Hypothesis 2 (C2) posits $Y \rightarrow X$, where the sentiment $Y$ causes the review $X$, which aligns well with the \textit{Fast Thinking} system \citep{kahneman2011thinking,ledoux1998emotional}, as it rapidly generates an emotional reaction $Y$, and then writes text to justify it $Y$. 
On the other hand, the Causal Hypothesis 1 (C1) refers to the case where $X \rightarrow Y$, namely the review $X$ causing the sentiment $Y$. It is an instance of \textit{Slow Thinking} \citep{baumeister1998ego,kahneman2002representativeness}, which deliberately uses conscious efforts to list out the up- and downsides of an experience in the review $X$, and come up with a thoughtful final decision as the rating $Y$.

\paragraph{Quantitative Signals of the Two Processes}
In sentiment processing, an evidence for the two processes is the famous \citet{kahneman1993more} study illustrating the \textit{Peak-End Rule} of
how individuals recall and evaluate past
emotional experiences, which we show in \cref{psychcausal:fig:structure}. 
As we know, fast thinking is prone to systematic biases and errors in the judgment \citep{tversky1974judgment}, and the \citet{kahneman1993more} study provides important quantitative results showing that,
in the Fast Thinking system, people's emotional memories of an experience are disproportionately influenced by its most intense point (the ``peak'') and its conclusion (the ``end''), rather than by the average experience as in the Slow Thinking system. The important role of peak and end for the fast thinking system implies that
it is the intensity of specific moments that dominate memory and judgment.

\begin{table*}[t]
    \centering \small
    \setlength\tabcolsep{3pt}
    \begin{tabular}{lccccccccc}
\toprule

& \multicolumn{3}{c}{Yelp} & \multicolumn{3}{c}{Amazon} & \multicolumn{3}{c}{App Review}
\\
\cmidrule(lr){2-4} \cmidrule(lr){5-7} \cmidrule(lr){8-10}
& All & C1 & C2 & All & C1 & C2 & All & C1 & C2 \\
\midrule
\# Samples & 34,851 & 19,557 (56\%) & 15,294 (44\%) & 2,582 & 1,393 (54\%) & 1,189 (46\%) & 9,696 & 3,809 (39\%) & 5,887 (61\%)\\
\# Sents/Review & 11.11  & 11.30 & 10.87 & 6.70 & 6.62 & 6.80 & 6.34 & 6.33 & 6.35\\ 
\# Words/Sent & 15.53 & 15.55  & 15.49 & 11.04 & 11.28 & 10.77 & 10.53 & 10.90 & 10.29\\
Vocab Size & 64,864 & 48,889 & 44,826 & 10,271  & 7,609  & 7,049 & 20,248 & 12,773 & 15,400\\
Avg Sentiment & 2.93 & 2.74 & 3.18 & 2.94 & 2.82 & 3.07 & 2.9 & 2.72 & 3.02\\ \hline
Avg $\lambda_1$ & 3.78 & 2.97 &  4.83 & 3.77 & 3.10 & 4.55 &  6.03 & 5.18 & 6.58 \\
Avg $\lambda_2$ &  4.48 & 6.05 & 2.48 & 4.21 & 5.72 & 2.45 & 5.10 & 7.96 & 3.26\\

\bottomrule
    \end{tabular}
    \caption{Statistics of the entire datasets and their C1 and C2 subsets for Yelp, Amazon, and App Review. We can see that a roughly balanced number of reviews aligning with the C1 and C2 processes.
    }
    \label{psychcausal:tab:psych}
\end{table*}
\subsection{Operationalizing the Theory}

We summarize the previous psychological insights in the upper left part of \cref{psychcausal:fig:structure}, where the Causal Hypothesis 1 corresponds to taking the average of all emotional experiences mentioned in the review $X$ for the sentiment $Y$, and the Causal Hypothesis 2 uses the peak and end emotions in the review $X$ to derive the sentiment $Y$.
In this section, we introduce a formalization of the theory, and suggest signals to distinguish the two causal hypotheses.

\paragraph{Emotion Arc}\label{psychcausal:sec:emotion_arc}
To capture the aforementioned trajectory of emotional experiences, we use the concept of the \textit{emotion arc} \citep{reagan2016emotional}, an example of which we visualize in \cref{psychcausal:fig:structure}.
Contextualizing it in the task of SA, we formally define an emotion arc of the review as follows.
Given a review $x$ consisting of $n$ sentences $(t_1, \dots, t_n)$, we identify the sentiment for each of them, thus obtaining a series of sentiment labels $(s_1, \dots, s_n)$.
We denote this series as the emotion arc $\bm{e}:=(s_1, \dots, s_n)$ of the review.

\paragraph{The Two Causal Processes}
Provided the notion of the emotion arc $\bm{e}:=(s_1, \dots, s_n)$ for a review $x$,
we formulate the sentiment labels corresponding to the two causal processes as follows:
\begin{align}
 \nonumber
    \text{Slow Thinki} &\text{ng (Causal Process 1): }
    \\
    \hat{y}_{\mathrm{avg}} &= \frac{1}{n} \left( s_1+ \dots+s_n \right)
    ~,
    \\
    \lambda_1 &= |y - \hat{y}_{\mathrm{avg}}|
    ~,
    \\
\nonumber
    \text{Fast Thinki} &\text{ng (Causal Process 2): }
    \\
\hat{y}_{\mathrm{peakEnd}} &= \frac{1}{2} \left( \mathrm{Peak} (s_1, \dots,s_n) + s_n \right)
    ,
\end{align}
\begin{align} 
    \lambda_2 &= |y - \hat{y}_{\mathrm{peakEnd}}|
    ~,
    \end{align}
where $\lambda_i$ indicates the alignment of the actual sentiment $y$ with the Causal Process $i$, and $\mathrm{Peak}(\cdot)$ 
selects the sentiment with the strongest intensity by its distance from the neutral sentiment 3, which is the middle point among the sentiment range 1--5, i.e.,
$\mathrm{Peak} (s_1, \dots,s_n) := 
s_{\argmax_i |s_i - 3|}
$.

\begin{table}[t]
    \centering \small
    \begin{tabular}{p{.9\linewidth}lllllll}
\toprule
\multicolumn{1}{c}{\textbf{Example of a C1-Dominant Review}}
\\
$\bm{\cdot}$ This was a great spot to take a break from it all and just people watch. $s_1=4.57$
\\
$\bm{\cdot}$ We sat at the bar facing the casino and we were entertained the whole time.$s_2=4.67$\\
$\bm{\cdot}$ The mini grilled cheese (appetizer) was fantastic. $s_3 = 4.53$\\
$\bm{\cdot}$ It came with a tomato based dipping sauce that was the perfect compliment to the bite sized wedges. $s_4 = 4.20$\\
$\bm{\cdot}$ Tip - ask for two dipping sauces because one just won't do. $s_5 = 1.60$\\

\textbf{Stars $y$:} 4 \\
\textbf{Psychology Scores ($\downarrow$):} $\lambda_1 = 0.0884  < \lambda_2 = 0.8683$ \\
\midrule
\multicolumn{1}{c}{\textbf{Example of a C2-Dominant Review}}
\\
$\bm{\cdot}$ I read the reviews and should have steered away... but it looked interesting. $s_1 = 3.72$
 \\
$\bm{\cdot}$ Salad was wilted, menus are on the wall, with no explanation so you are ordering blind, service was NOT with a smile from the bartender to the waitress, to the server who helped the waitress, and the waitress never checked back to see how everything is. $s_2 = 2.20$\\
$\bm{\cdot}$ Terribly overpriced for what you get, and as an Italian, this does not even pass for a facsimile thereof! $s_3 = 1.45$\\
$\bm{\cdot}$ Stay away for sure. $s_4 = 1.85$\\
$\bm{\cdot}$ I only gave them one star, as I had to fill something in, they should get no stars! $s_5 = 1.32$\\

\textbf{Stars $y$:} 1 \\
\textbf{Psychology Scores ($\downarrow$):} $\lambda_1 = 1.1827 > \lambda_2 = 0.3647$ \\
\bottomrule
    \end{tabular}
    \caption{Examples of C1- and C2-dominant reviews.}
    \label{psychcausal:tab:psych_examples}
\end{table}

Here, we interpret $\lambda_i$ as an indicator for each causal process, where a small value (with the best value being zero) implies the alignment with the process $i$.
We show two examples in \cref{psychcausal:tab:psych_examples}, one aligning well with the Causal Process C1 with a small $\lambda_1$, and another aligning well with the Causal Process C2 with a small $\lambda_2$.

\begin{table*}[t]
    \centering \small
    \setlength\tabcolsep{4pt}
    \begin{tabular}{llccccccccc}
    \toprule
 && Random & GPT-2 XL & LLaMa-7B & Alpaca-7B & GPT-3 & GPT-3.5 & GPT-4\\ \midrule
\multirow{3}{*}{F1} & Overall & 19.82 {\tiny$\pm$2.07} & 10.23 {\tiny$\pm$4.12} & 31.78 {\tiny$\pm$5.32} & 46.01 {\tiny$\pm$5.35} & 52.71 {\tiny$\pm$1.73} & 57.98 {\tiny$\pm$5.11} & 59.54 {\tiny$\pm$4.69}
    \\
& C1 Subset  & 21.36 {\tiny$\pm$2.26} & 5.80 {\tiny$\pm$3.11} & 27.30 {\tiny$\pm$4.73} & 37.77 {\tiny$\pm$7.66} & 43.96 {\tiny$\pm$2.93} & 58.64 {\tiny$\pm$1.48} & 58.62 {\tiny$\pm$2.54}  \\
& C2 Subset &  20.43 {\tiny$\pm$2.95} & \textbf{16.37} {\tiny$\pm$5.33} & \textbf{37.66} {\tiny$\pm$7.86} & \textbf{55.82} {\tiny$\pm$4.02} & \textbf{65.40} {\tiny$\pm$1.37} & \textbf{59.09} {\tiny$\pm$9.13} & \textbf{62.57} {\tiny$\pm$6.85} \\ \hline
\multirow{3}{*}{Acc} & Overall &19.78 {\tiny$\pm$2.07} & 23.06 {\tiny$\pm$2.10} & 39.28 {\tiny$\pm$5.07} & 47.72 {\tiny$\pm$4.19} & 53.22 {\tiny$\pm$1.35} & 58.36 {\tiny$\pm$4.13} & 59.84 {\tiny$\pm$4.17}
\\
& C1 Subset & 20.61 {\tiny$\pm$2.23} & 16.18 {\tiny$\pm$1.59} & 36.55 {\tiny$\pm$4.05} & 42.14 {\tiny$\pm$5.26} & 43.61 {\tiny$\pm$2.89} & \textbf{59.89} {\tiny$\pm$1.09} & 59.62 {\tiny$\pm$1.96} \\
& C2 Subset &  18.86 {\tiny$\pm$2.78} & \textbf{30.79} {\tiny$\pm$2.68} & \textbf{42.33} {\tiny$\pm$7.24} & \textbf{53.93} {\tiny$\pm$3.91} & \textbf{63.93} {\tiny$\pm$1.28} & 56.66 {\tiny$\pm$8.08} & \textbf{60.08} {\tiny$\pm$7.05}\\

    \bottomrule
    \end{tabular}
    \caption{Performance of different models on the five-class classification of Yelp-5. We use five paraphrases for the prompt (in \cref{psychcausal:appd:prompt_we_use}), and report the average performance with the standard deviation.
    }
    \label{psychcausal:tab:prompt_c0}
\end{table*}
\begin{table*}[t]
    \centering \small
    \setlength\tabcolsep{3pt}
    \begin{tabular}{llcccccccccccc}
\toprule
&&Random & GPT-2 XL & LLaMa-7B & Alpaca-7B & GPT-3 & GPT-3.5 & GPT-4 \\ \midrule
\multirow{4}{*}{F1} 
& D=C1, P=C1 &20.47 {\tiny$\pm$2.47} & 6.12 {\tiny$\pm$2.77} & 55.16 {\tiny$\pm$7.16} & 52.74 {\tiny$\pm$5.04} & 38.36 {\tiny$\pm$6.66} & 60.62 {\tiny$\pm$3.24} & 54.23 {\tiny$\pm$4.17} \\
& D=C1, P=C2 &20.26 {\tiny$\pm$2.31} & 15.98 {\tiny$\pm$3.59} & 36.22 {\tiny$\pm$8.95} & 35.10 {\tiny$\pm$5.40} & 54.44 {\tiny$\pm$1.64} & 52.85 {\tiny$\pm$6.01} & 58.58 {\tiny$\pm$3.33}\\
& D=C2, P=C1 & 22.35 {\tiny$\pm$3.02} & 31.98 {\tiny$\pm$8.66} & 56.69 {\tiny$\pm$8.46} & 54.74 {\tiny$\pm$14.26} & 74.64 {\tiny$\pm$3.06} & \textbf{78.18} {\tiny$\pm$1.21} & 72.52 {\tiny$\pm$3.68} \\
& D=C2, P=C2 & 20.35 {\tiny$\pm$2.18} & \textbf{48.50} {\tiny$\pm$7.66} & \textbf{66.82} {\tiny$\pm$7.90} & \textbf{71.22} {\tiny$\pm$3.99} & \textbf{77.09} {\tiny$\pm$1.33} & 78.16 {\tiny$\pm$1.81} & \textbf{76.80} {\tiny$\pm$1.36}\\ \hline

\multirow{4}{*}{Acc}
& D=C1, P=C1 &19.60 {\tiny$\pm$2.67} & 12.96 {\tiny$\pm$2.91} & 58.23 {\tiny$\pm$5.30} & 55.81 {\tiny$\pm$4.00} & 43.16 {\tiny$\pm$6.20} & 60.39 {\tiny$\pm$3.25} & 54.06 {\tiny$\pm$3.97}  \\
& D=C1, P=C2 & 19.68 {\tiny$\pm$2.46} & 22.61 {\tiny$\pm$5.73} & 43.07 {\tiny$\pm$8.18} & 41.45 {\tiny$\pm$4.06} & 56.36 {\tiny$\pm$1.94} & 53.11 {\tiny$\pm$5.89} & 58.68 {\tiny$\pm$3.45}\\
& D=C2, P=C1 & 20.97 {\tiny$\pm$3.19} & 43.51 {\tiny$\pm$6.30} & 57.54 {\tiny$\pm$7.21} & 54.82 {\tiny$\pm$12.59} & 76.60 {\tiny$\pm$3.23} & 77.38 {\tiny$\pm$1.43} & 70.69 {\tiny$\pm$3.93}\\
& D=C2, P=C2 & 19.03 {\tiny$\pm$2.18}& \textbf{51.59} {\tiny$\pm$5.09} & \textbf{68.16} {\tiny$\pm$9.24} & \textbf{71.83} {\tiny$\pm$4.33} & \textbf{76.70} {\tiny$\pm$1.35} & \textbf{78.92} {\tiny$\pm$1.31} & \textbf{75.38} {\tiny$\pm$1.70} \\
\bottomrule
    \end{tabular}
    \caption{
    Performance on Yelp using the \textit{causal prompts} on the two causal subsets. 
We experiment all combinations of data nature (``D'') and causal prompt type (``P''), and report the average performance across the five paraphrases for each prompt, with the standard deviation.
    }
    \label{psychcausal:tab:prompt_c12}
\end{table*}

\subsection{Findings on SA Datasets}\label{psychcausal:sec:psych_exp}
\paragraph{Dataset Setup}
We adopt three commonly used datasets in SA: Yelp \citep{zhang2015character}, Amazon \citep{keung-etal-2020-multilingual}, and App Review \citep{app_review}.
For the Amazon data, we concatenate each review's title with its text.
Since the model performance on many binary classification datasets is saturated \citep{poria2023beneath,yang2019xlnet}, we use the 5-way classification version of the SA datasets when applicable.

Since we need to utilize the emotion arc, we keep only reviews with at least five sentences, after sentence tokenization using the Spacy package \citep{spacy2}.
We apply this filtering above on the test set of the Yelp dataset, the English test of Amazon, and the unsplit entire dataset of App Review. We report the statistics of remaining samples in \cref{psychcausal:tab:psych}.

To obtain the emotion arcs, we calculate the sentiment score of each sentence by the \texttt{sentiment-analysis} pipeline\footnote{\href{https://huggingface.co/distilbert-base-uncased-finetuned-sst-2-english}{https://huggingface.co/distilbert-base-uncased-finetuned-sst-2-english}. We map its output value in 0--1 to the 5-class labels, by converting a value in 0--0.2 to the class label of 1, 0.2--0.4 to label 2, ..., and 0.8--1 to label 5.} from Huggingface \citep{wolf-etal-2020-transformers}.

\paragraph{Causal Discovery}
For each input sample, we process them as in \cref{psychcausal:tab:psych_examples}, namely first obtaining the sentence-level sentiments to form the emotion arc, and then calculating the alignment scores $\lambda_1$ and $\lambda_2$ for each causal process, respectively.
We consider an example as \textit{dominated} by the causal process $C_i$ if the alignment score $\lambda_i$ is more optimal than the other.

We report the resulting statistics in \cref{psychcausal:tab:psych}. For each dataset, we describe their overall statistics, as well as the statistics of data with the underlying causal process of C1, and that of C2. We can see that Yelp and Amazon have an almost balanced split of C1 and C2, while App Review has 61\% C2 data compared to 39\% C1 data. See \cref{psychcausal:appd:distr_plot} for an additional visualization of the $\lambda_1$-$\lambda_2$ distribution across the 1K data points.

\section{Improving Sentiment Classifiers with Causal Alignment}
\label{psychcausal:sec:improve}
Using our proposed causal discovery method, we have identified two distinct subsets with their corresponding causal processes C1 and C2.
Now, we address the last practical question proposed in \cref{psychcausal:sec:causality_helps_nlp}:
\begin{quote}
    \textit{Can causal alignment help us improve SA in the era of LLMs?}
\end{quote}

Specifically, we take the commonly used approach in the era of LLMs, i.e., prompting pre-trained LLMs for the SA task, and look into how alignment with the underlying causal process could help SA performance. We answer the following three subquestions in this section:
\begin{enumerate}[label=Q\arabic*.]
    \item 
    Using the standard SA prompt, do models perform differently on C1/C2 data? (\cref{psychcausal:sec:q1})
    \item 
    Does it help if we make the prompt aware of the underlying causality, i.e., use causal prompts? (\cref{psychcausal:sec:q2})
    \item 
    When prompted causally, 
    do LLMs really understand the causal processes? (\cref{psychcausal:sec:q3})
\end{enumerate}

\subsection{Q1: Do Models Perform Differently on C1/C2 Data?}\label{psychcausal:sec:q1}

\paragraph{Experimental Setup}
The first question is whether models perform differently on data with the causal nature of C1 or C2. We use the subsets identified by our psychologically-grounded causal discovery, and test a variety of available autoregressive LLMs, including the open-weight GPT-2 \citep{radford2019language},
LLaMa \citep{touvron2023llama}, and
Alpaca \citep{alpaca}; as well as the closed-weight models with OpenAI API, the instruction-tuned GPT-3 ({text-davinci-002}) \citep{brown2020gpt3,ouyang2022instructGPT}, GPT-3.5 ({gpt-3.5-turbo-0613}), GPT-4 ({gpt-4-0613}) \citep{openai2023gpt4}.
We also add a random baseline which uniformly samples the label space for each input.

We use the standard prompt formulation for SA in the format of ``\texttt{[Instruction] Review Text: \{$x$\}$\backslash$n Label:}''. The experiments are on a set of randomly selected 1K samples from the test set of Yelp-5 \citep{zhang2015character}, due to the time- and cost-expensive inference of the above LLMs.
(E.g., LLaMa/Alpaca takes 96 GPU hours to run.)
See more experimental details in
\cref{psychcausal:appd:models}.

\paragraph{Results}
We show the performance of the six LLMs in \cref{psychcausal:tab:prompt_c0}, and report the F1 and accuracy across the five-class classification on Yelp-5.
We can see that the existing LLMs perform the best on the subset with the causal process C2, implying that the decision pattern of LLMs is closer to the Fast Thinking system, which takes the peak-end average of the emotion arc.

\subsection{Q2: Do Causal Prompts Help?}\label{psychcausal:sec:q2}

\paragraph{Designing \textit{Causal Prompts}}
Inspired by the fact that models perform differently on C1/C2 data, our next question is, will it help if we directly give a hint to the LLMs about the underlying causal graph?

\begin{table}[ht]
    \centering \small
    \begin{tabular}{m{0.6cm}p{.85\linewidth}}
\toprule
& Prompt Design \\ \midrule
    C1     & 
As a customer writing a review, I initially \textit{composed} the following feedback: ``\texttt{[review]}'' 
\\ &
\textit{After carefully considering the facts}, I selected a star rating from the options of ``1'', ``2'', ``3'', ``4'', or ``5''. My final rating was:
\\\hline
    C2 & 
As a customer writing a review, I initially \textit{selected} a star rating from the options ``1'', ``2'', ``3'', ``4'', and ``5'', and then provided the following explanations in my review: ``\texttt{[review]}'' 

\\ &
The review \textit{clarifies} why I gave a rating of
\\
\bottomrule
    \end{tabular}
    \caption{Causally-aware prompts describing the SA task in contexts with the C1 and C2 causal graphs.}
    \label{psychcausal:tab:prompt_c12_design}
\end{table}

To this end, we propose the idea of \textit{causal prompts}, which are prompts that describe the causal story behind the input and output variables. We list our designed prompts for the C1 and C2 stories in \cref{psychcausal:tab:prompt_c12_design}. 

\paragraph{Results}
We report the performance for all combinations of the dataset natures and prompt natures in \cref{psychcausal:tab:prompt_c12}, where we find that the most-performant setting uses Prompt C2 on the data subset with the same causal nature, C2. This alignment leads to the best performance across almost all models by both F1 and accuracy. 
On the C2 data, we also see that Prompt C2 outperforms the standard SA prompt in \cref{psychcausal:tab:prompt_c0} by a substantial margin, such as 32.13 F1 points increase for GPT-2, and 14.23 F1 points increase for GPT-4.

However, although Prompt C2 shows a strong performance, the other causal prompt, i.e., Prompt C1, does not always help the data subset C1 in all cases, from which we raise a further question -- how well do LLMs really mechanistically understand our prompts? We explore this question in the next section.

\subsection{Q3: Can LLMs Correctly Capture the Causal Stories in the Prompts?}\label{psychcausal:sec:q3}

Although the proposal of the two causal prompts is intuitive for humans, we still need to inspect whether LLMs are able to understand them correctly.

\paragraph{Method}
Mechanistically, for a model to solve SA for the causal process C1 correctly, it needs to treat the sentence-level sentiments across all sentences \textit{equally}; and for a model to solve SA for the causal process C2 correctly, it needs to pay \textit{more} attention to the peak and end sentiments on the emotion arc.

Targeting the two mechanisms, we use causal tracing \citep{meng2022locating} to attribute the final sentiment prediction to the source sentences in the input. Briefly, causal tracing uses causal mediation analysis \citep{pearl2001direct} to quantify the causal contribution of the internal neuron activations of a model to its final prediction \citep{vig2020causal}. 
We use causal tracing to inspect the causal effects of the hidden states on the model prediction, using the open-weight models, LLaMa and Alpaca. We use the causal effects of the first-layer neurons for each sentence, which we aggregate to obtain the final prediction.
See implementation details in \cref{psychcausal:appd:causal_tracing}.

\begin{figure}[t]
    \centering
    \includegraphics[width=0.6\columnwidth]{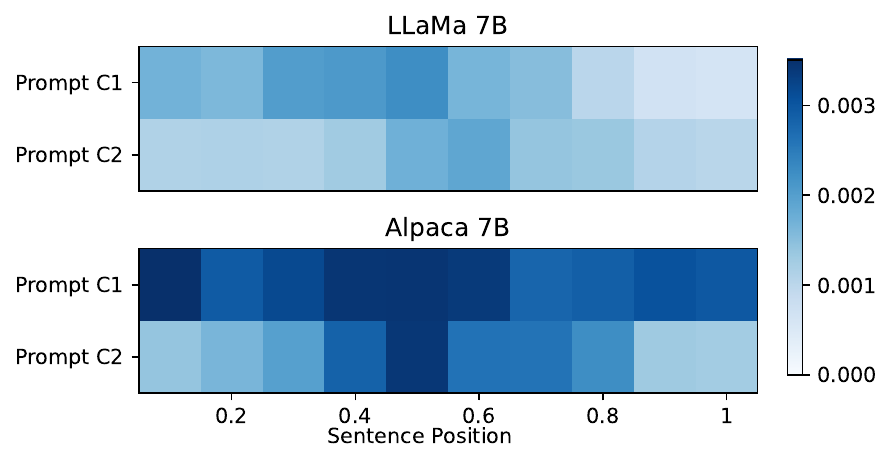}
    \caption{Causal attribution in LLaMa-7B and Alpaca-7B, showing how much each sentence contributes to the prediction probability. 
}
    \label{psychcausal:fig:causal_trace}
\end{figure}

\paragraph{Results}
We plot the causal attribution results of how much each sentence contributes to the final prediction
in \cref{psychcausal:fig:causal_trace}.
Here, the ideal behavior of the models is that Prompt C1 should trigger \textit{uniform} attention over all the sentences, which is roughly observed through the more even shades of color of the ``Prompt C1'' row than the ``Prompt C2'' row in \cref{psychcausal:fig:causal_trace} in the row of ``Prompt C1''. On the other hand, Prompt C2 should trigger \textit{more} attention to the sentences corresponding to the peak and end sentiments. For this, we see the models have high attention to the middle sentence, as in the ``Prompt C2'' row in \cref{psychcausal:fig:causal_trace}. Additionally, the average causal effect of the peak sentence on predictions under Prompt C2 is 0.0013 for LLaMa and 0.0029 for Alpaca, quantitatively aligning with our expectation that the peak sentence would show a high contribution.
Nonetheless, note that no model sufficiently attends to the end sentence under Prompt C2. This implies that they do not fully grasp the expected contribution pattern of the peak-end rule, missing the significant role of the end sentence.

\section{Related Work}

\paragraph{SA}
The task of SA aims to identify the sentiment given a piece of text.
It has a rich history originating from subjectivity analysis \citep{wiebe1994tracking,hatzivassiloglou2000effects}, and developing rapidly with the availability of large opinionated online data such as reviews with star ratings \cite[\textit{inter alia}]{turney2002thumbs,nasukawa2003sentiment,zhang2015character,keung-etal-2020-multilingual}. Most literature on SA focuses on building computational models, from using traditional linguistic rules \citep{hatzivassiloglou1997predicting,DBLP:conf/emnlp/ChoiC08}, to the application of machine learning methods, from traditional naive bayes and support vector machines \citep{pang2002thumbs,moraes2013document,DBLP:conf/ecir/TanCWX09}, to early deep learning models
\citep{socher2013recursive,kim2014convolutional,xing-etal-2020-tasty}, and finally entering the era of LLMs \citep{DBLP:conf/nodalida/HoangBR19,raffel2020exploring,yang2019xlnet}.

\paragraph{Psychology and Affective Science}
In the study of emotion, or affect science \citep{salovey2004emotional,barrett2006solving,feinstein2013lesion},
previous work
finds that not only the emotion people perceive influences or prime 
how they communicate in the
moment \citep{barrett2006solving}, but language can also influence
emotion, which can be observed in functional magnetic neuroimaging \citep{satpute2013functional}, and also experiments showing the act of self-reporting the emotion in writing can change the physical reaction to the emotion \citep{kassam2013effects}.
In his seminal work, \citet{kahneman2011thinking} uses the two systems of thinking to reveal the mechanisms of how people come up with their sentiment, where fast thinking conforms to the peak-end rule \citep{kahneman1993more}, and slow thinking is more reflective of the overall sentiment.

\paragraph{Cause-Effect Distinction}
Distinguishing the cause from effect based on observational data is a long-standing and fundamental problem in causality \citep{hoyer2008nonlinear,zhang2009causality,janzing2019causeeffect}. Existing methods to address this problem are based on statistics \citep{hoyer2008nonlinear,peters2010identifying,shajarisales2015telling,mooij2014distinguishing}, physics \citep{janzing2007causally,janzing2016algorithmic}, information theory \citep{janzing2012informationgeometric,chaves2014inferring,mejia2022obtaining}, and algorithmic complexity \citep{janzing2010causal,jin-etal-2021-causal}. 
However, we are the first to look at the rich nature of NLP datasets, and directly approach the difference in the causal and anticausal mechanisms grounded in interdisciplinary insights. 

As for our causal prompts, the most similar studies are the non-causally-grounded explorations for prompt tuning, such as by varying the patterns of masked language modeling \citep{schick-schutze-2022-true} and using the noisy channel method \citep{min-etal-2022-noisy}. However, these studies are not aware of the underlying causal processes, thus neglecting the connection of prompts with the causal nature of data, and also the explicit causal story of the sentiment-review relation.

\section{Conclusion}
In conclusion, we have formulated the task of SA into a prediction problem and a causal discovery problem. We first identified the cause-effect relation among existing SA datasets, namely 
whether the review primes the rating, or the sentimental judgment primed the review writing process.
To achieve this causal discovery, we obtain insights from existing psychology studies, namely aligning the above two causal processes with the famous Fast Thinking and Slow Thinking systems, with their distinct qualitative signals.
Given the causal understanding of the dataset, we further improve the performance of LLMs on SA using our proposed causal prompts. Our research paves the way for more causally-aware future research in SA.

\section*{Limitations and Future Work}

This study has several limitations. First, the rapid progression of LLMs makes it challenging to keep up with all newly proposed models and architectures. Since our work covers only a set of recent LLMs at the time of this study, we encourage future research to apply our methods to additional LLMs and other SA datasets.

Although our study is grounded in well-established psychological theories, there remains the possibility that new theories could emerge, necessitating updates to the calculation of the $\lambda$ values for the two causal processes. However, the causal processes identified in this work appear plausible, as evidenced by the effectiveness of the causally aligned prompts in improving language model performance.

Regarding the causal graph formulation, we focus on basic bivariate causal graphs, but future work could include more variables, such as confounders, mediators, and colliders. 

The nature of this work is to introduce a paradigm shift for SA, and formulate the task differently. 
Therefore, we see lots of space for future extensions, such as to explore the causal nature of SA in different settings, different languages, and also aspect-based sentiment analysis \citep{DBLP:conf/semeval/PontikiGPPAM14,xing-etal-2020-tasty,hua2023systematic}.

\section*{Ethical Considerations}
Regarding data concerns and user privacy, our study employs several established NLP datasets, and the examples we cite do not include sensitive user information.

Concerning potential stakeholders and misuse, this research primarily introduces a new perspective on the SA task. A possible negative impact concerns the general application of SA, which could be used to analyze user mentality for surveillance or fraudulent purposes. We acknowledge that studies on SA inherently involve these risks, and we firmly oppose the misuse of SA models in such contexts.

\part{Causality for Text-Based Computational Social Science}\label{part:4}

\mainchapter{Mining the Cause of Political Decision-Making from Social Media: A Case Study of COVID-19 Policies across the US States}\label{ch:covidtwitter}

Mining the causes of political decision-making is an active research area in the field of political science. In the past, most studies have focused on long-term policies that are collected over several decades of time, and have primarily relied on surveys as the main source of predictors. However, the recent COVID-19 pandemic has given rise to a new political phenomenon, where  political decision-making consists of frequent short-term decisions, all on the same controlled topic---the pandemic. In this paper, we focus on the question of how public opinion influences policy decisions, while controlling for confounders such as COVID-19 case increases or unemployment rates. Using a dataset consisting of Twitter data from the 50 US states, we classify the sentiments toward governors of each state, and conduct controlled studies and comparisons.  Based on the compiled samples of sentiments, policies, and confounders, we conduct causal inference to discover trends in political decision-making across different states.
Our code and data are publicly available at \url{https://github.com/zhijing-jin/covid-twitter-and-policy}.

\section{Introduction}
Policy responsiveness is the study of the factors that policies respond to \citep{stimson1995dynamic}. One major direction is that politicians tend to make policies that align with the expectations of their constituents, in order to run successful re-election in the next term \citep{canes2002out}.

An overview of existing studies on policy responsiveness reveals several patterns, summarized in \cref{covidtwitter:tab:intro}. First, most work focuses on the \textit{long-term} setting, where the policies are collected over a span of several decades, e.g., \citet{caughey2018policy}'s collection of public opinion surveys and state policymaking data over 1936-2014, and \citet{lax2009gay}'s collection of public opinion polls and gradual policy changes over 1999-2008. Second, the data sources of existing studies are mostly surveys and polls, which can be time-consuming and expensive to collect \citep{lax2012democratic}. Third, the resulting data are often of relatively small sizes, for both the number of policies and the number of public opinion. 

\begin{table}[t]
    \centering
    \small
    \begin{tabular}{lp{2.3cm} p{2.3cm}}
    \toprule
    & \textbf{Previous Work} & \textbf{This Work} \\ \midrule
    \textbf{Policy Type} & Long-term, gradu-al (over decades) & Short-term (weekly/monthly) \\ \hline
    \textbf{Policy Sparsity} & Less policies on the same topic & Many policies on the same topic across states \\ \hline
    \textbf{Data Source} & Surveys & Trillions of tweets \\ \hline
    \textbf{Data Collection} & -- & NLP \& Causality \\
    \bottomrule
    \end{tabular}
    \caption{Comparison of the characteristics and paradigms of existing work versus our work.}
    \label{covidtwitter:tab:intro}
\end{table}

Different from previous work on long-term policies, our work focuses on the special case of COVID pandemic, during which political leaders make a number of frequent, short-term policies on the same topic: social distancing. Moreover, instead of collecting surveys, we use Twitter to collect public opinion, which is instant, costless, and massive, e.g., trillions of data points. We limit our scope to US policies because the 50 states provide abundant policy data, and a good background for both controlled groups and comparative studies. %

\begin{figure*}[t]
    \centering
    \includegraphics[width=0.75\paperwidth]{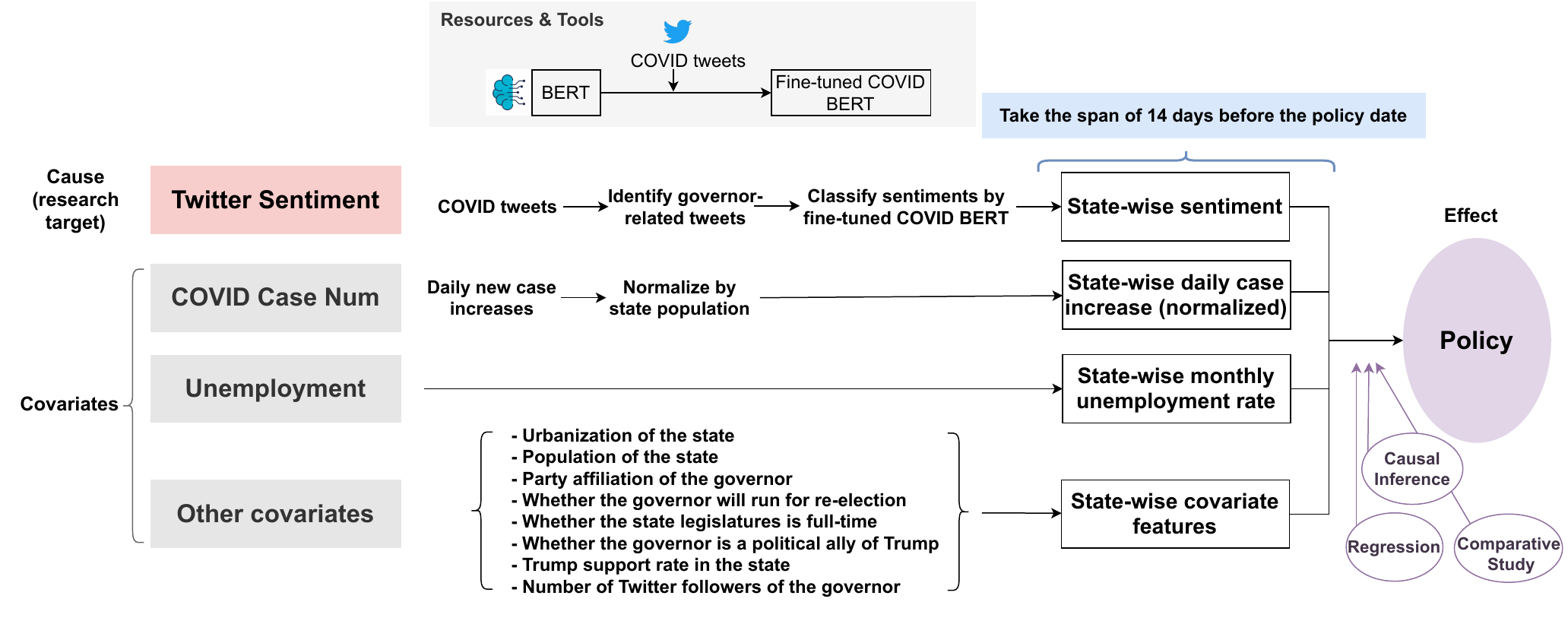}
    \caption{The data collection pipeline and architecture of our system to predict the state-wise COVID policies.}
    \label{covidtwitter:fig:data_pipeline}
\end{figure*}

We present one of the first efforts to address policy responsiveness for short-term policies, namely the causal impact of public Twitter sentiments on political decision-making. This is distinct from existing studies on COVID policies that mostly explore the impact of policies, such as predicting public compliance  \citep{grossman2020political,allcott2020polarization,barrios2020risk,gadarian2021partisanship,defranza2020religion}. 
Specifically, since governors have legislative powers through executive orders, we focus our study on each state governor's decisions and how public opinion towards the governor impacts their decisions. For example, governors that optimize short-term public opinion are more likely to re-open the state even when case numbers are still high.

Our workflow is illustrated in \cref{covidtwitter:fig:data_pipeline}.
We start by collecting 10.4M governor-targeted COVID tweets, which we annotate for sentiment with a BERT-based classifier. Next, we annotate 838 social distancing policies and collect data on ten potential confounders such as average daily case increases or unemployment rates.
Finally, we conduct multiple analyses on the causal effect of Twitter sentiment on COVID policies. For interpretability, we first use a multivariate linear regression to  identify correlations of sentiments and policies, in addition to considering all the confounders. We also use do-calculus \citep{pearl1995causal} to quantify the causal impact of Twitter sentiment on policies. We also conduct cross-state comparisons, cross-time period analysis, and multiple other analyses.

The main contributions of our work are as follows. 
First, we compile a dataset of public opinion targeted at governors of the 50 US states with 10.4M tweets. 
Second, we annotate a dataset of 838 COVID policy changes of all 50 states, along with data of ten confounders of each state. 
Third, we conduct regression analyses and causal analyses on the effect of Twitter sentiment on policies. 
Finally, we implement additional fine-grained analyses such as cross-state comparisons, cross-time period analysis, and multiple other analyses.

\section{Related Work}
\paragraph{Policy Responsiveness}
Policy responsiveness (i.e., public opinion$-$\textit{causes}$\rightarrow$policies) is an active research field in political science, where people study how policies respond to different factors \citep{stimson1995dynamic}. Studies show that policy preferences of the state public can be a predictor of future state policies \citep{caughey2018policy}.
For example, \citet{lax2009gay} show that more LGBT tolerance leads to more pro-gay legislation in response.
Most policies and public opinion studied in existing literature are often long-term and gradual, taking several decades to observe \citep{lax2009gay,lax2012democratic,caughey2018policy}.

\paragraph{Crisis Management Policies}
Another related topic is crisis management policies, where most studies focus on the reverse causal problem of our study -- how crisis management policies impact public opinion (i.e., policies$-$\textit{causes}$\rightarrow$public opinion). A well-known phenomenon is the rally ``round the flag'' effect, which shows that during a crisis, there will be an increased short-run public support for the political leader \citep{mueller1970presidential,mueller1973war,baum2002constituent}, due to patriotism
(Mueller, 1970; Parker, 1995), lack of opposing views or criticism \citep{brody1989reconsideration}, and traditional media coverage \citep{brody1991assessing}. 

To the best of our knowledge, there is not much research on how public opinion influence policies (i.e., public opinion$-$\textit{causes}$\rightarrow$policies) during a crisis. Our work is one of the few to address this direction of causality.

\paragraph{COVID Policies}
There are several different causal analyses related to COVID-19 policies, although different from our research theme. Existing studies focus on how social distancing policies mitigate COVID spread (i.e., policies$-$\textit{causes}$\rightarrow$pandemic spread) \citep{kraemer2020effect}, what features in public attitudes impact the compliance to COVID policies (i.e., public attitudes/ideology$-$\textit{causes}$\rightarrow$policy compliance) \citep{grossman2020political,allcott2020polarization,barrios2020risk,gadarian2021partisanship}, how polices change the public support of leaders (i.e., policy$-$\textit{causes}$\rightarrow$public support). \citet{bol2021effect,ajzenman2020more}, how pandemic characteristics affect Twitter sentiment \citep{gencoglu2020causal}, and how political partisanship impacts policies (i.e., partisanship$-$\textit{causes}$\rightarrow$policy designs) \citep{adolph2021pandemic}. 
However, there is no existing work using public sentiments (e.g., from social media) to model COVID policies.

\paragraph{Opinion Mining from Social Media}
Social media, such as Twitter, is a popular source to collect public opinions 
\citep{thelwall2011sentiment,paltoglou2012twitter,pak-paroubek-2010-twitter,rosenthal2015semeval}. 
\citet{arunachalam-sarkar-2013-new} suggest that Twitter can be a useful resource for governments to collect public opinion. Existing usage of Twitter for political analyses mostly targets at election result prediction \citep{beverungen2011evaluating,mohammad2015sentiment,tjong-kim-sang-bos-2012-predicting}, and opinion towards political parties \citep{pla-hurtado-2014-political} and presidents \citep{marchetti-bowick-chambers-2012-learning}. To the best of our knowledge, this work is one of the first to use Twitter sentiment for causal analysis of policies.

\section{Governor-Targeted Public Opinion}
To investigate the causality between public opinion and each state governor's policy decisions, we first describe how we mine public opinion in this Section; we then  describe the process we use to collect policies and other confounders in \cref{covidtwitter:sec:policy_and_confounder_collection}.
\begin{table*}[t]
    \centering
    \small
    \begin{tabular}{p{1.2cm}p{4.7cm}p{4.5cm}p{4.2cm}}
    \toprule
        & \multicolumn{1}{c}{\textbf{Positive}} & \multicolumn{1}{c}{\textbf{Neutral}} & \multicolumn{1}{c}{\textbf{Negative}} \\ \midrule
\textbf{Percentage} & \multicolumn{1}{c}{15.8\%} & \multicolumn{1}{c}{36.5\%} & \multicolumn{1}{c}{47.7\%} \\ \hline
\textbf{Length} & \multicolumn{1}{c}{15.51} & \multicolumn{1}{c}{12.21} & \multicolumn{1}{c}{16.39} \\ \hline
\textbf{Topics} & we, support, thank, great, governors, covid, action 
& people, masks, covid, cases, state, today, total
& cases, state, covid, close, deaths, people, trump \\ \hline
\textbf{4-Grams}
    & - great governors responded executive
    \newline
    -~responded~executive~action
    promptly
    \newline
    - quickly , support americans
    & - positive patients nursing homes
\newline
- governors ordered covid positive
\newline
- today 's update numbers
    & - covid patients nursing homes
    \newline
    - america 's governors forced
    \newline
    - covid patients nursing homes
\\ \hline
\textbf{Example} & "I am a small business owner, we kept health insurance for the furloughed staff of my two restaurants, month after month, even while one restaurant was closed and the other only has limited service.
Why? Because I have a conscience. We are in a pandemic." 
& "Today: 
@GovInslee
 3 pm news conference on WA’s coronavirus response.
Inslee to be joined by state schools chief.
Your daily \#covid19 updates via 
@seattletimes"
& "And the politicians that are doing the conditioning are out, maskless, celebrating with their family and friends...
@GavinNewsom
Glad I never once fell for it. Covid-19 was always just a power-grab for politicians"\\
    \bottomrule
    \end{tabular}
    \caption{Label distribution (Percentage), average number of words per tweet (Length), topics extracted by LDA topic modeling \citep{blei2003latent}, top 4-grams, and examples of positive, neutral, and negative tweets.}
    \label{covidtwitter:tab:twit_example}
\end{table*}

We collect governor-targeted public opinion in two steps: (1) retrieve governor-related COVID tweets (\cref{covidtwitter:sec:gov_tweet}), and (2) train a sentiment classification model for the COVID tweets and compile sentiments towards governors (\cref{covidtwitter:sec:classifier}).

\subsection{Retrieve Governor-Related COVID Tweets}\label{covidtwitter:sec:gov_tweet} 

We use the COVID-related tweet IDs curated by \citet{chen2020tracking}.\footnote{COVID-related Tweet IDs: \url{https://github.com/echen102/COVID-19-TweetIDs}} \citet{chen2020tracking} identified these tweets by tracking COVID-related keywords and accounts. We provide the list of keywords and accounts they used in \cref{covidtwitter:sec:twitter_keyword_chen}. 
We hydrate the tweet IDs to obtain raw tweets using an academic Twitter Developer account. This  process took several months to complete, and resulted in a dataset of 1.01TB.
The retrieved 1,443,871,617 Tweets span from January 2020 to April 2021.

Since this study focuses on governor's policy decision-making process, we focus on the public opinion that are more directly related to the governors. %
Specifically, we focus on tweets that tagged, replied to, or retweeted state governors.
We obtain 10,484,084 tweets by this filter. On average, each of the 50 states has about 209K tweets that address the state governor. The rationale of this filter is that the governors and their teams are likely to have directly seen (a portion of) these tweets, since they showed up in governor's Twitter account.

\subsection{Classify
Sentiments towards Governors}\label{covidtwitter:sec:classifier}

Existing studies on COVID Twitter sentiment analysis  \citep{manguri2020twitter,kaur2020twitter,vijay2020sentiment,chakraborty2020sentiment,singh2021sentiment} mostly use TextBlob \citep{loria2018textblob}, or some simple supervised models \citep{machuca2021twitter,kaur2021proposed,mansoor2020global}.

For our study, we use the state-of-the-art BERT model pretrained on COVID tweets by \citet{muller2020covid}.\footnote{\url{https://huggingface.co/digitalepidemiologylab/covid-twitter-bert-v2}} We finetune this pretrained COVID BERT on the Twitter sentiment analysis data from SemEval 2017 Task 4 Subtask A \citep{rosenthal-etal-2017-semeval}. Given tweets collected from a diverse range of topics on Twitter, the model learns a three-way classification (positive, negative, neutral). In the training set, there are 19,902 samples with positive sentiments, 22,591 samples with neutral sentiments, and 7,840 samples with negative sentiments.

We tokenize the input using the BERT tokenizer provided by the Transformers Python package \citep{wolf-etal-2020-transformers}. We add [CLS] and [SEP] tokens at start and end of the input, respectively. The input is first encoded by the pretrained COVID BERT. Then, we use the contextualized vector $C$ of the [CLS] token as the aggregate sentence representation. The model is finetuned on the classification task by training an additional feed-forward layer $\log(\mathrm{softmax}(CW))$ that assigns the softmax probability distribution to each sentiment class.

Prior to training, we preprocess the tweets by deleting the retweet tags, and pseudonymising each tweet by replacing all URLs with a common text token. We also replace all unicode emoticons with textual ASCII representations. During training, we use a batch size of 32 and fine-tune for 5 epochs.
We use a dropout of 0.1 for all layers, and the Adam optimizer \citep{kingma2017adam} with a learning rate of 1e-5. Additionally, due to the specific nature of our classification task (i.e., mining opinion towards the governor), we add a post-processing step to classify a tweet as supportive of a governor (i.e., positive) if the tweet retweets a tweet from the governor's official account.

\vskip 0.1in
\noindent {\bf Model Performance.} We evaluate our model accuracy on two test sets. First, on the test set of SemEval 2017, our finetuned model achieves 79.22\% accuracy and 79.29\% F1.
Second, we also evaluate our model performance on our own test set. Since the features of general tweets provided in SemEval 2017 might differ from COVID-specific tweets, we extracted 500 random tweets from the Twitter data we collected in \cref{covidtwitter:sec:gov_tweet}. We asked a native English speaker in the US to annotate the Twitter sentiment with regard to the state governor that the tweet addresses. The annotator has passed a small test batch before annotating the entire test set. 

We use the TextBlob classifier as our baseline, since it is the most commonly used in existing COVID Twitter sentiment analysis literature. On our test set's three-way classification, the TextBlob baseline has 23.35\% accuracy and 16.67\% weighted F1.
Our finetuned BERT classifier has 60.23\% accuracy and 62.31\% weighted F1. Detailed scores per class is in \cref{covidtwitter:appd:bert_accuracy}. When applying the sentiment classifier, we care more about whether the average sentiment over a time period is accurate, so we also turn the test set into groups of tweets each containing 20 random samples. The average mean squared error (MSE) for the average sentiment of each group is 0.03889 for the BERT model, and 0.22749 for the TextBlob model. 
We apply the finetuned COVID BERT classifier on the governor-related tweets we extracted previously.
As listed in \cref{covidtwitter:tab:twit_example}, among 10.4M tweets, 15.8\% are positive, 36.5\% neutral, and 47.7\% negative.\footnote{Note that label imbalance is commonly observed on Twitter data \citep{guerra2014sentiment}.} 

We use Latent Dirichlet Allocation (LDA) topic modeling \citep{blei2003latent} to extract key topics of each category. Typical topic words in positive tweets include ``we,'' ``support,'' ``thank,'' ``great,'' and ``governors,'' while negative tweets tend to mention more about ``america's governors forced ...'' and support Trump, perhaps Trump's tweets on ``liberation.''

\section{Collection of Policies and Confounders }\label{covidtwitter:sec:policy_and_confounder_collection}

We focus on state-wide social distancing policies, and collect 838 social distancing policies from 50 states over the period January 2020 -- April 2021 (described in \cref{covidtwitter:sec:policy_collection}).

Since we want to focus on the causal effect of public sentiment on policy, we must control for possible confounding factors. In particular, case numbers and unemployment rates are potentially the most important confounders, the collection of which is introduced in \cref{covidtwitter:sec:case_collection}. In addition, we also collect eight other potential confounders suggested by political science experts (described in \cref{covidtwitter:sec:other_collection}).
The collection process is illustrated in \cref{covidtwitter:fig:data_pipeline}.

\subsection{Social Distancing Policy Annotation}\label{covidtwitter:sec:policy_collection}

We annotate the social distancing policies related to COVID for each of the 50 states in the US. For each state, the annotators are asked to go through the entire list of COVID-related executive orders from January 2020 to April 2021. In cases where the states do not use executive orders for COVID regulations, we also consider proclamations and state guidance on social distancing.

The policies are rated on a scale of 0 (loosest) - 5 (strictest). We provide guidance as to the level of strictness that each number indicates, as detailed in \cref{covidtwitter:appd:annot_policy}. Four annotators are asked to conduct the ratings.
Since the annotation is very tedious, taking up to 3 hours per state, we do not conduct double annotations. Instead, given our original annotations (for which we score each policy based on its official legal document in PDF), we did a quick second pass by confirming that our scores roughly match the succinct 1$\sim$2-sentence textual summary of each policy
provided by the Johns Hopkins Coronavirus Resource Center.\footnote{Social distancing policy summaries: \url{https://coronavirus.jhu.edu/data/state-timeline}}

\subsection{Key Confounders: State-Level Case Numbers and Unemployment Rates}\label{covidtwitter:sec:case_collection}
We collect COVID daily new confirmed case numbers from the open-source COVID database\footnote{COVID case number data: \url{https://github.com/KFFData/COVID-19-Data}} curated by the Kaiser Family Foundation. For a fair comparison across states, we normalize the case numbers by the population of the state.
We retrieve the seasonly adjusted data of monthly unemployment rates for each state from the U.S. Bureau of Labor Statistics.\footnote{Monthly unemployment data: \url{https://www.bls.gov/web/laus/ststdsadata.zip}}

\subsection{Additional Confounders}\label{covidtwitter:sec:other_collection}

For additional confounders, we collect both state data as well as governor features.
\vskip 0.1in
\noindent{\bf State Features.} For state features, we collect the population\footnote{Population data: \url{https://www.census.gov/programs-surveys/decennial-census/data/tables.2010.html}} and urbanization rate from US 2010 Census \citep{us2012united}.\footnote{Urbanization data: \url{https://www.icip.iastate.edu/tables/population/urban-pct-states}.} In addition, we also collect the last US presidential election returns of each state.\footnote{Presidential election return data: \url{https://www.nytimes.com/elections/2016/results/president}} Note that it is necessary to use pre-policy data, so we collect the presidential election returns from 2016 but not from 2020. For the presidential election returns, we obtain the percentage of votes for Donald Trump to indicate Trump's support rate.

\vskip 0.1in
\noindent{\bf Governor Features.} 
For each governor, we collect their party affiliation, whether the governor will run for the next gubernatorial election,\footnote{For simplicity, we collect the pre-COVID data at the time point of January 2020, and do not consider the change of governorships in two states in early 2021.} and whether the state legislatures are full-time or not, collected from National Conference of State Legislatures.\footnote{\url{https://www.ncsl.org/}} In addition, we also annotate whether the governor is a political ally of Trump or not. We conduct the annotation based on the background and past news reports of each governor. For corner cases, we quote additional evidence in our annotation, e.g., for republican governors who do not support Trump, and democratic governors who support Trump.
We also collect the number of Twitter followers for each governor, since it might be correlated with how much attention the governor pays to the twitter reactions.

\cref{covidtwitter:tab:data_stats} lists the statistics of the confounder data we collected.
\begin{table}[ht]
    \centering
    \small
    \resizebox{\columnwidth}{!}{%
    \begin{tabular}{llll}
    \toprule
    \multicolumn{4}{c}{\textbf{Numerical Features}} \\
    & Mean ($\pm$std) & Min & Max \\ \hline
    Daily Case Increase (\%) & 0.02 ($\pm$0.02) & 0.0 & 0.45 \\
    Unemployment Rate (\%) & 5.51 ($\pm$3.25) &
2.0 & 
29.5 \\
    Urbanization (\%) & 73.58
($\pm$14.56)
& 38.7
& 95\\
    Population (M) & 12.94
($\pm$45.68)
& 0.57
& 325.38\\
    Trump's Support Rate (\%) & 48.29
($\pm$11.93)
& 4
& 68\\ 
    \# Twitter Followers (K) & 237
($\pm$458)
& 7
& 2596\\
\midrule
    \multicolumn{4}{c}{\textbf{Binary Features}} \\
    & Yes & No \\\hline
    Gov Is Republican & 26 & 24 \\
    Will Run for Re-election & 39 & 11 \\
    Full-Time Legislatures & 10 & 40 \\
    Trump's Political Ally & 22 & 28 \\
    \bottomrule
    \end{tabular}
    }
    \caption{Statistics of the ten confounders collected for policy prediction task.}
    \label{covidtwitter:tab:data_stats}
\end{table}

\section{Mining Decisive Factors of COVID Policies}
Since we are interested in discovering the key factors that changes the decisions of policy-makers, we focus on the change of  policies (e.g., changing from complete close down to reopening K-12 schools) rather than absolute values of the policy strictness.
For each policy in state $s$ on date $t$, we calculate the change $\Delta \mathrm{policy}$ as the difference of this policy from the previous policy that was issued. 

Since sentiment may change rapidly and many policies are updated frequently during COVID, for each policy change $\Delta \mathrm{policy}$, we focus on the average sentiment over the time span $(t - \Delta t, t)$ from $\Delta t$ days prior to the policy date $t$. Here, we set $\Delta t=14$ since many epidemiology reports are based on 14-day statistics, e.g., the 14-day notification rate.

When building the policy prediction model, we also need to account for confounders. For the confounders, most are static over time for a given state, except  for the daily case increases and the unemployment rates that change over time, for which we take the average over the 14-day time span.

Based on the data above, we seek to answer the following questions: (Q1) What variables are indicative of policy changes?, and (Q2) What causal impact does sentiment have on the policies?

\subsection{Q1: What Variables Are Indicative of Policy Changes?}\label{covidtwitter:sec:linear_reg}

To aim for interpretability, we choose a multivariate linear regression as our model, which is commonly used in political science literature on COVID policies \citep{grossman2020political,allcott2020polarization,barrios2020risk,gadarian2021partisanship}. 
Specifically, we model the policy change $\Delta \mathrm{policy}$ as a function of all variables, including our main focus -- Twitter sentiments -- and all the confounders, which form in total 11 variables.\footnote{For each input variable, we first normalize by adjusting mean to zero and standard deviation to 1.}

\paragraph{Sentiment, Case Numbers, Unemployment Are Important}
The first experiment is to compare how well different combinations of input variables fit the policy change. We use mean squared error (MSE) as the measure of model capability.

When taking into consideration all variables, the model has an MSE score of 0.368. As a further step, we test whether a smaller number of inputs can achieve similar results. We find that when only taking three variables as inputs, the MSE is 0.369, which is 0.001 from the model taking in all variables. Among all combinations of three variables, the proposed three key variables, sentiment, case numbers, and unemployment rates, achieve the best performance of 0.369.

Note that it is reasonable that with rational decision-making, politicians consider the case numbers and unemployment rates when making COVID policies. The focus of this study is to show the \textit{additional} effect of sentiment, the role of which is not explicitly pointed out in previous COVID policy research.

\paragraph{The Role of Non-Sentiment Variables}
First, given the presence of the sentiment variable in the model, we test the additional effect of non-sentiment variables. 
As shown in \cref{covidtwitter:tab:keep_sentiment}, case numbers and unemployment rate both lead to non-trivial improvement of the models, and unemployment is more important.

\begin{table}[ht]
    \centering
    \begin{tabular}{llcl}
    \toprule
    \textbf{Additional Non-Sentiment Variables} & \textbf{MSE ($\downarrow$)} \\ \hline
    Sentiment-only & 0.618 \\
      + Case & 0.532 \\
    + Unemp & 0.407 \\
    + Case, Unemp & 0.369 \\
    + Case, Unemp, Others & 0.368 \\
    \bottomrule
    \end{tabular}
    \caption{The MSE of models taking as input the additional non-sentiment variables, such as case increases (Case), unemployment (Unemp), and other confounders (Others).}
    \label{covidtwitter:tab:keep_sentiment}
\end{table}

\paragraph{The Role of Sentiment}
Second, we look into the role of sentiment. We take the optimal 11-variable, 3-variable, and 2-variable models, and conduct ablation studies to inspect how much does sentiment contribute exclusively in \cref{covidtwitter:tab:ablation}.

We show that for each model, sentiment has a crucial impact of more than 0.032 on the model performance. Note that in linear regression, we do not need to explicitly disentangle the correlations within sentiments and other confounders -- in \cref{covidtwitter:tab:ablation}, the effect of sentiment is demonstrated in addition to fitting all other variables that may contain correlations. 

\begin{table}[ht]
    \centering
    \begin{tabular}{lr}
    \toprule
    \textbf{Model} & \textbf{MSE ($\downarrow$)} \\ \hline
    11-Variable model & 0.368 \\ %
    \quad $-$Senti & Deterioration of 0.032 \\ %
    \hline
    3-Variable model & 0.369 \\ %
    \quad $-$Senti & Deterioration of 0.032 \\ %
    \hline
    2-Variable model & 0.407 \\ %
    \quad $-$Senti & Deterioration of 0.034 \\ %
    
    \bottomrule
    \end{tabular}
    \caption{Ablation study of sentiment for the optimal 11-, 3-, 2-variable models. Note that the 11-variable model is the full model taking in all variables.}
    \label{covidtwitter:tab:ablation}
\end{table}

\subsection{Q2: What Causal Impact Does Sentiment Have on the Policies?}\label{covidtwitter:sec:do_calc}

In the previous section, we investigated the most indicative variables of policies. The experiments indicate how important each variable is to the regression target, i.e., how well they serve as a predictor, although such \textit{correlation} does not necessarily capture \textit{causation}. 
In this section, we are interested in the causal impact of sentiment on policies, and we use causal inference methods to quantify the impact.

\begin{figure}[ht]
    \centering
    \includegraphics[width=.8\columnwidth]{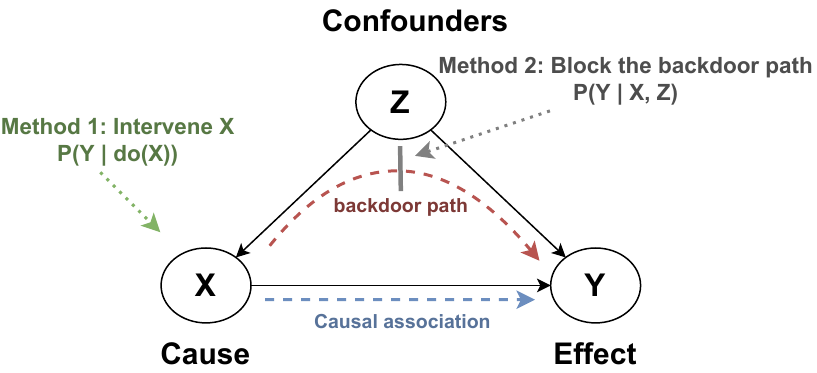}
    \caption{Backdoor Adjustment.}
    \label{covidtwitter:fig:backdoor}
\end{figure}

\paragraph{Formulation by Do-Calculus}
Formally, we are interested in the effect of a cause $X$ (i.e., Twitter sentiment) on the outcome $Y$ (i.e., policy change) in the presence of the confounder $Z$ (i.e., case numbers, unemployment, etc.), as shown in \cref{covidtwitter:fig:backdoor}.

To formulate the causal impact, \citet{pearl1995causal} defines a language for causality called do-calculus, by which the causal impact of $X$ on $Y$ is formulated as the interventional distribution:
\begin{align}
    P(Y|\mathrm{do}(X) )
    ~,
\end{align}
where $\mathrm{do}(X)$ refers to an intervention on the cause $X$.

Note that the interventional distribution $P(Y|\mathrm{do}(X) )$ may be different from the observational distribution $P(Y|X)$ in the presence of the confounder $Z$. Specifically, in the above \cref{covidtwitter:fig:backdoor}, there are two ways how $X$ correlates with $Y$. The first is the causal path $X\rightarrow Y$, and the second is the backdoor path $X \leftarrow Z \rightarrow Y$.

There are two ways to account for the backdoor path: Method 1 needs to intervene on $X$, e.g., create a counterfactual situation where all confounders are the same but the Twitter sentiment can be set to negative vs. positive. In our study of Twitter opinion on COVID policies, this is not a feasible experiment to conduct, due to the fundamental problem of causal inference \citep{rubin1974estimating,holland1986statistics} (namely, for each sample $i$, we are usually only able to observe one value of $X$ but not both). The other method, backdoor adjustment, circumvents the problem, which will be introduced in the following.

\paragraph{Backdoor Adjustment}
The key challenge in the above causal inference is that we need to account for the confounder $Z$. 
Backdoor adjustment \citep{pearl1995causal} presents an approach to estimate the causal impact of $X$ on $Y$ by using only \textit{observational} data. Basically, we need to block all backdoor paths by conditioning on nodes that can break the unwanted connections between $X$ and $Y$. Moreover, these nodes should not contain any descendants of $X$. In our case, we condition on the confounder $Z$, and turn the interventional distribution into the observational distribution:
\begin{align}
    P(Y|\mathrm{do}(X) ) = \sum_Z P(Y|X, Z) P(Z)
    ~.
\end{align}

The causal impact of $X$ (i.e., positive or negative sentiment) on $Y$ (i.e., policy change) becomes
\resizebox{0.95\columnwidth}{!}{
\centering
  \begin{minipage}{\linewidth}
  \begin{align}
\begin{split}
    \beta &= \mathbb{E}[Y|\mathrm{do}(X=1) ] - \mathbb{E}[Y|\mathrm{do}(X=-1) ] \\
    &= \sum_Z ( \mathbb{E}[Y|X=1, Z ] - \mathbb{E}[Y|X=-1, Z  ]) P(Z)\\
    &= \mathbb{E}_Z [ \mathbb{E}[Y|X=1, Z ] - \mathbb{E}[Y|X=-1, Z ] ]
    ~.
\end{split}
\label{covidtwitter:eq:backdoor}
\end{align}
  \end{minipage}
}

\paragraph{Results}
We apply \cref{covidtwitter:eq:backdoor} to all states using a 10-dim vector $Z$ that encodes all confounders.\footnote{Due to length restrictions, please refer to the arXiv version of our paper for additional implementation details of the backdoor adjustment.} Then we rank the states by $\beta$ values, which represents the causal impact of sentiment on the state policies.

\begin{table}[ht]
    \centering 
    \begin{tabular}{lc|lc}
    \toprule
    \multicolumn{2}{l|}{\textbf{Top 5 States with Large $\beta$}} & \multicolumn{2}{l}{\textbf{Top 5 States with Small $|\beta|$}} \\
    {State} & $\beta$ {Value} & {State} & $\beta$ {Value} \\ \hline
Colorado & 4.292 & Arizona & 0.053 \\
Massachusetts & 1.157 & West Virginia & 0.030 \\
Florida & 1.124 & Pennsylvania & 0.023 \\
Texas & 1.095 & Nebraska & -0.001 \\
South Dakota & 1.057 & Alabama & -0.065 \\

\bottomrule
    \end{tabular}
    \caption{Top five states with the largest $\beta$ values, and the $\beta$ values that are closest to zero.}
    \label{covidtwitter:tab:backdoor}
\end{table}

In \cref{covidtwitter:tab:backdoor}, we show the top five states with highest $\beta$ values, and five states with $\beta$ values that are the closest to zero. The higher the $\beta$ value, there exists more alignment between people's sentiment and the state policy strictness in the state.

There are some associations between our results and real-world patterns. 
For instance, among the top five states in \cref{covidtwitter:tab:backdoor}, Colorado’s high $\beta$ value reflects  its Democratic governor's large net favorable rating compared to the Republican politicians.\footnote{For example, see \href{https://web.archive.org/web/20210522173231/https://www.denverpost.com/2019/07/03/jared-polis-cory-gardner-poll/}{this poll result}  by Colorado Poll reported by Denver Post.}
Massachusetts also has a high governor approval rate, and most people support the COVID policies. The three Republican states, South Dakota, Texas, and Florida, also have high $\beta$, but they are in a different scenario. The loose policies in all these states are in line with general sentiment across the states to refuse restrictions.

\section{Fine-Grained Analyses}
\subsection{Early-Stage vs. Late-Stage Decisions}

Since the COVID pandemic is an unprecendented situation, it is likely that in early stages of the pandemic, politicians tend to rely on their pre-judgements, and as time goes on, they form a better understanding of the situation and adjust their reaction towards the public opinion. We compare the causal impact of sentiment on policies in the first three months of the outbreak (i.e., from March to June 1, 2020) and afterwards (i.e., from June 1, 2020 to now). \cref{covidtwitter:tab:change} shows that the states with the most changes in $\beta$ are Montana, Washington, Georgia, Tennessee, and Indiana.

\begin{table}[ht]
    \centering
    \begin{tabular}{lcl}
    \toprule
     \textbf{State} & \textbf{Change in $\beta$ before and after June 1} \\ \hline
         Montana & +9.39 \\
         Washington & +4.03 \\
         Georgia & +3.15 \\
         Tennessee & +2.94 \\
         Indiana & +2.53 \\
    \bottomrule
    \end{tabular}
    \caption{Top 5 states with the most change in the causal impact of sentiment on policies from March to June 1, 2020, versus from June 1, 2020 to April, 2021.}
    \label{covidtwitter:tab:change}
\end{table}

\subsection{Cross-State Comparison}

For cross-state comparison, we identify states that are similar in terms of the confounders, and then compare how different policies are a result of different public sentiments. For simplicity, we consider the two most important confounders, case numbers and unemployment rates. We evaluate the similarity matching on the two time series across different states by the dynamic time warping algorithm \citep{berndt1994using}, and extract state pairs that are the most similar in terms of the confounders.

\begin{figure}[t]
\centering
\begin{subfigure}{.32\columnwidth}
  \centering
  \includegraphics[width=\linewidth]{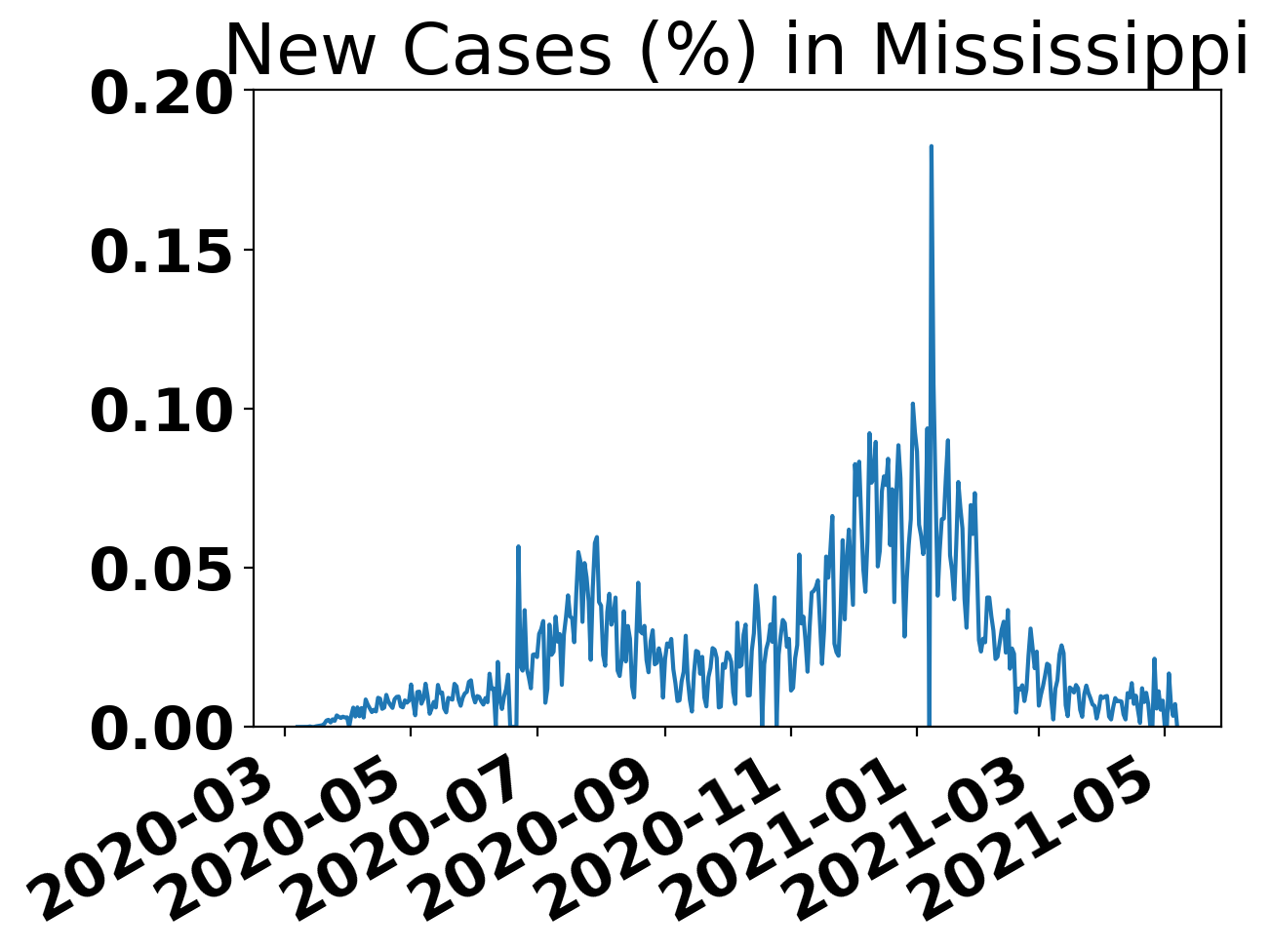}
  \caption{Cases in MS.}
  \label{covidtwitter:fig:sub1}
\end{subfigure} 
\hfill
\begin{subfigure}{.32\columnwidth}
  \centering
  \includegraphics[width=\linewidth]{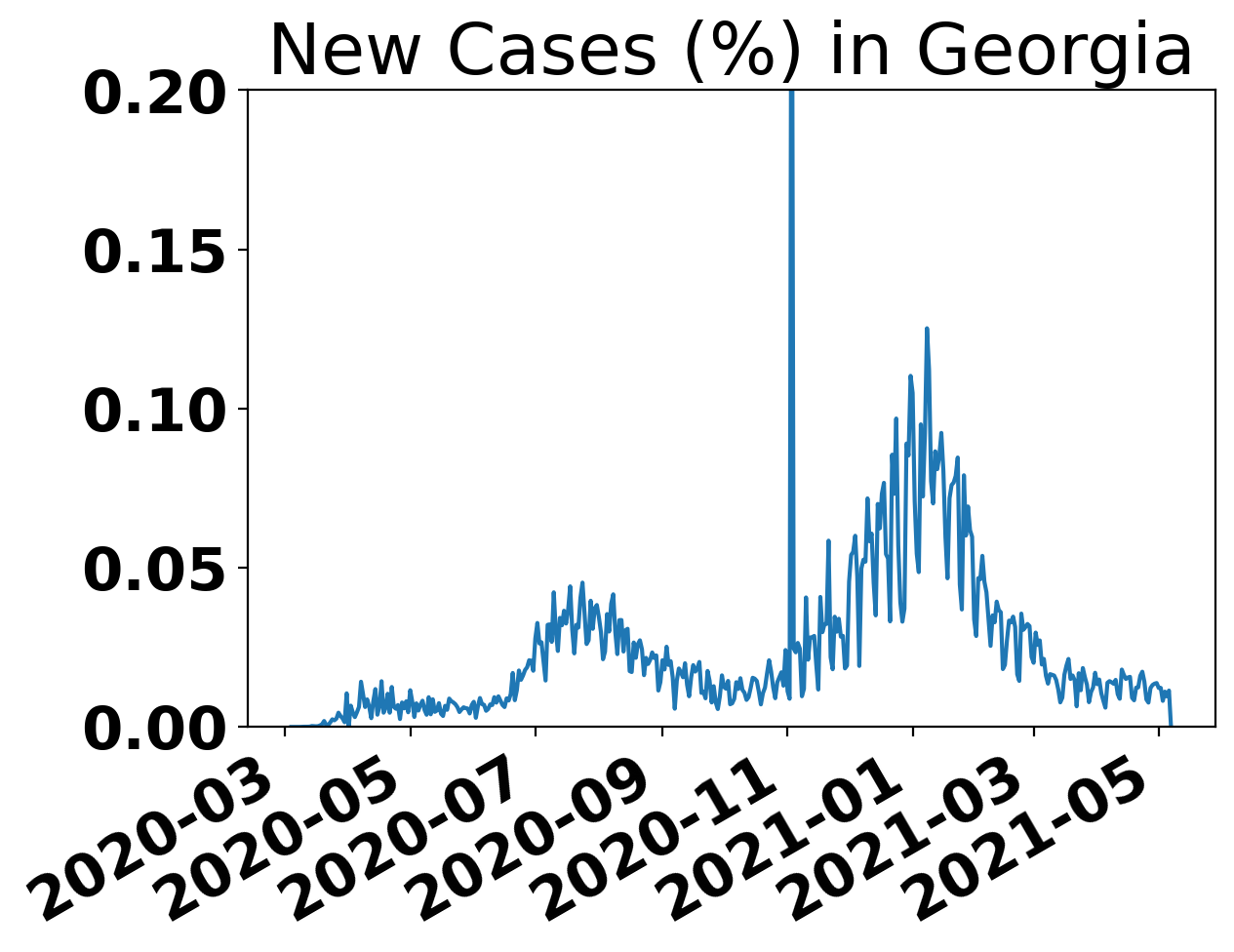}
  \caption{Cases in GA.}
  \label{covidtwitter:fig:sub2}
\end{subfigure}
\hfill
\begin{subfigure}{.32\columnwidth}
  \centering
  \includegraphics[width=\linewidth]{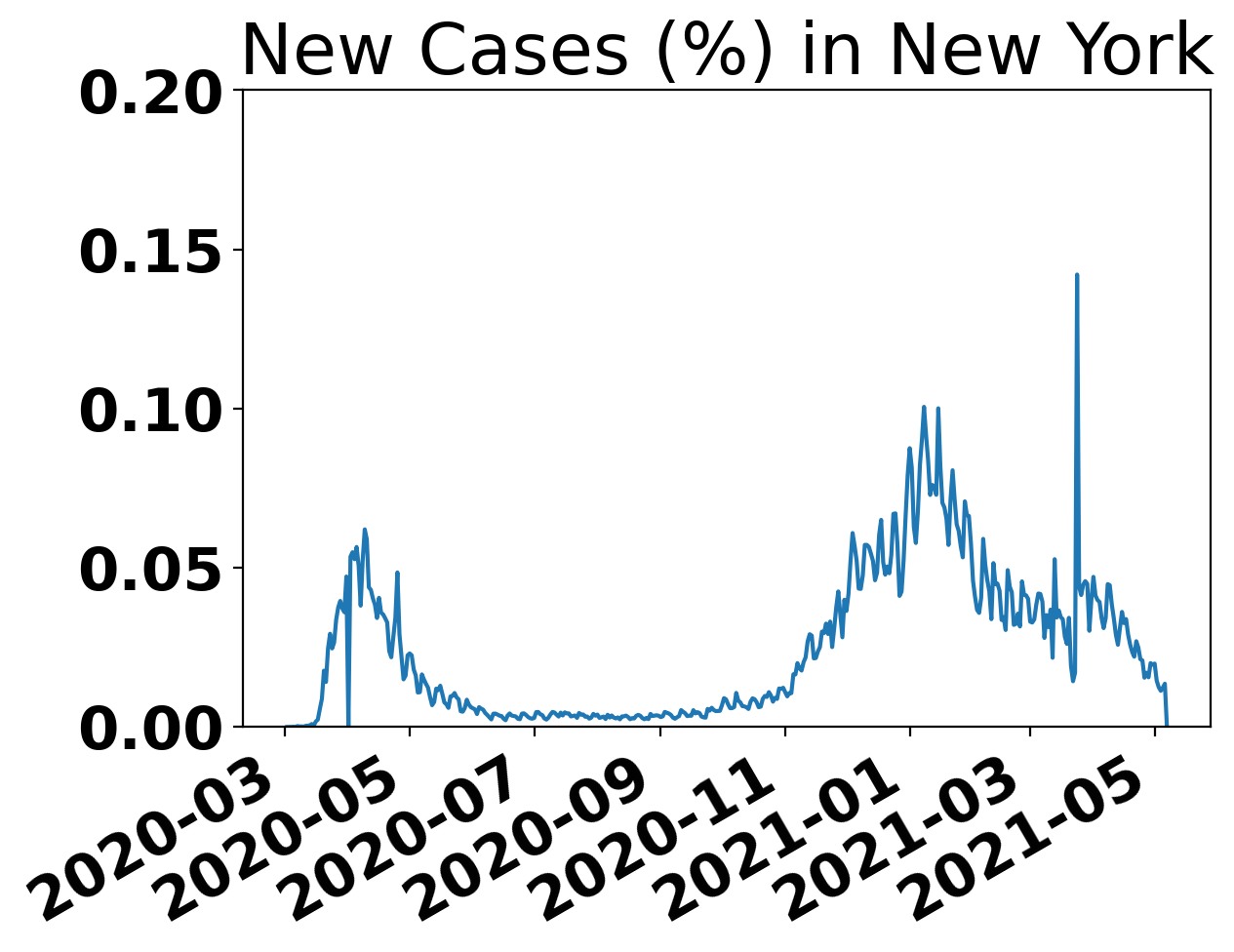}
  \caption{Cases in NY.}
  \label{covidtwitter:fig:sub3}
\end{subfigure}
\begin{subfigure}{.32\columnwidth}
  \centering
  \includegraphics[width=\linewidth]{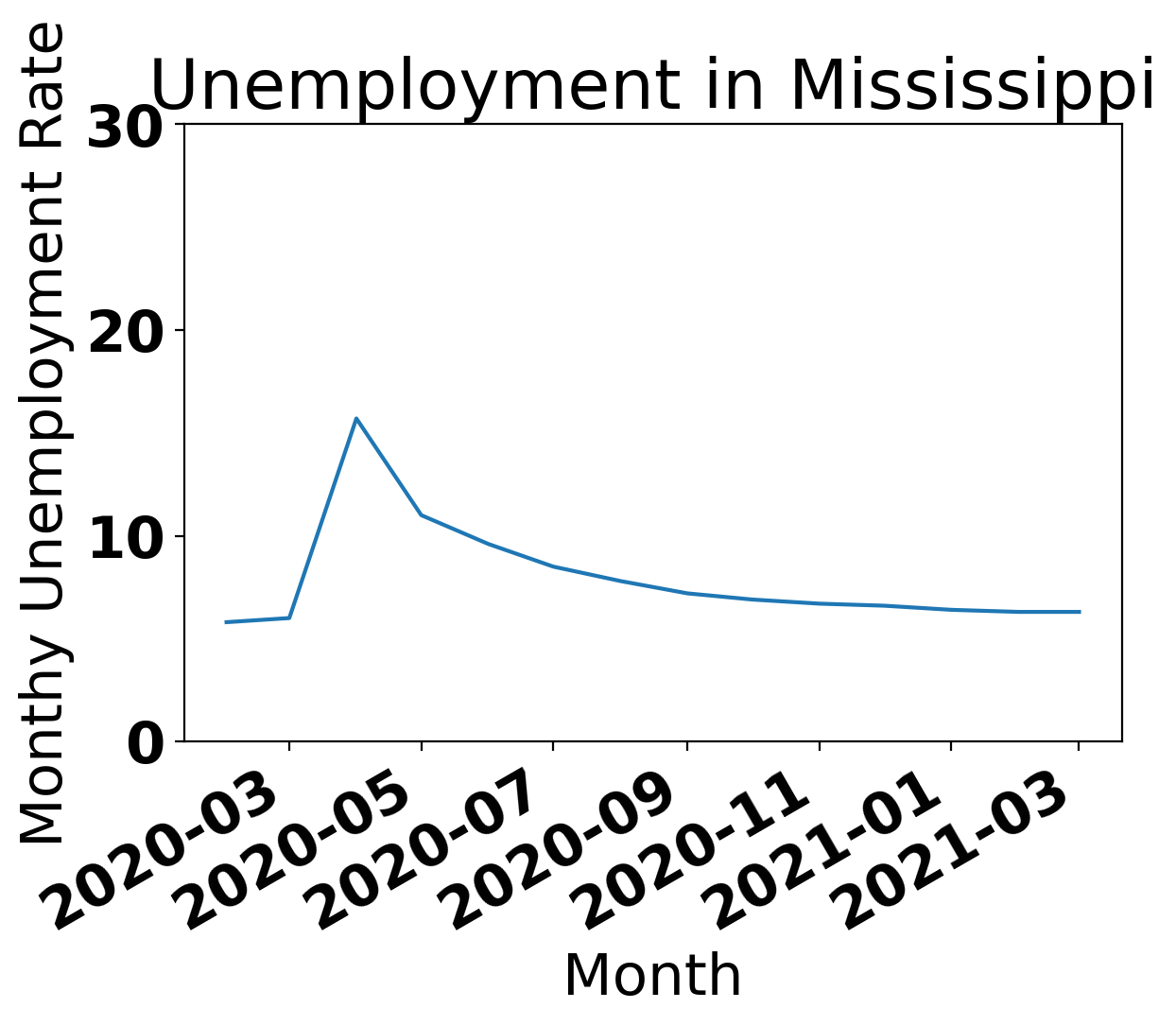}
  \caption{Unemployment in MS.}
\end{subfigure} 
\hfill
\begin{subfigure}{.32\columnwidth}
  \centering
  \includegraphics[width=\linewidth]{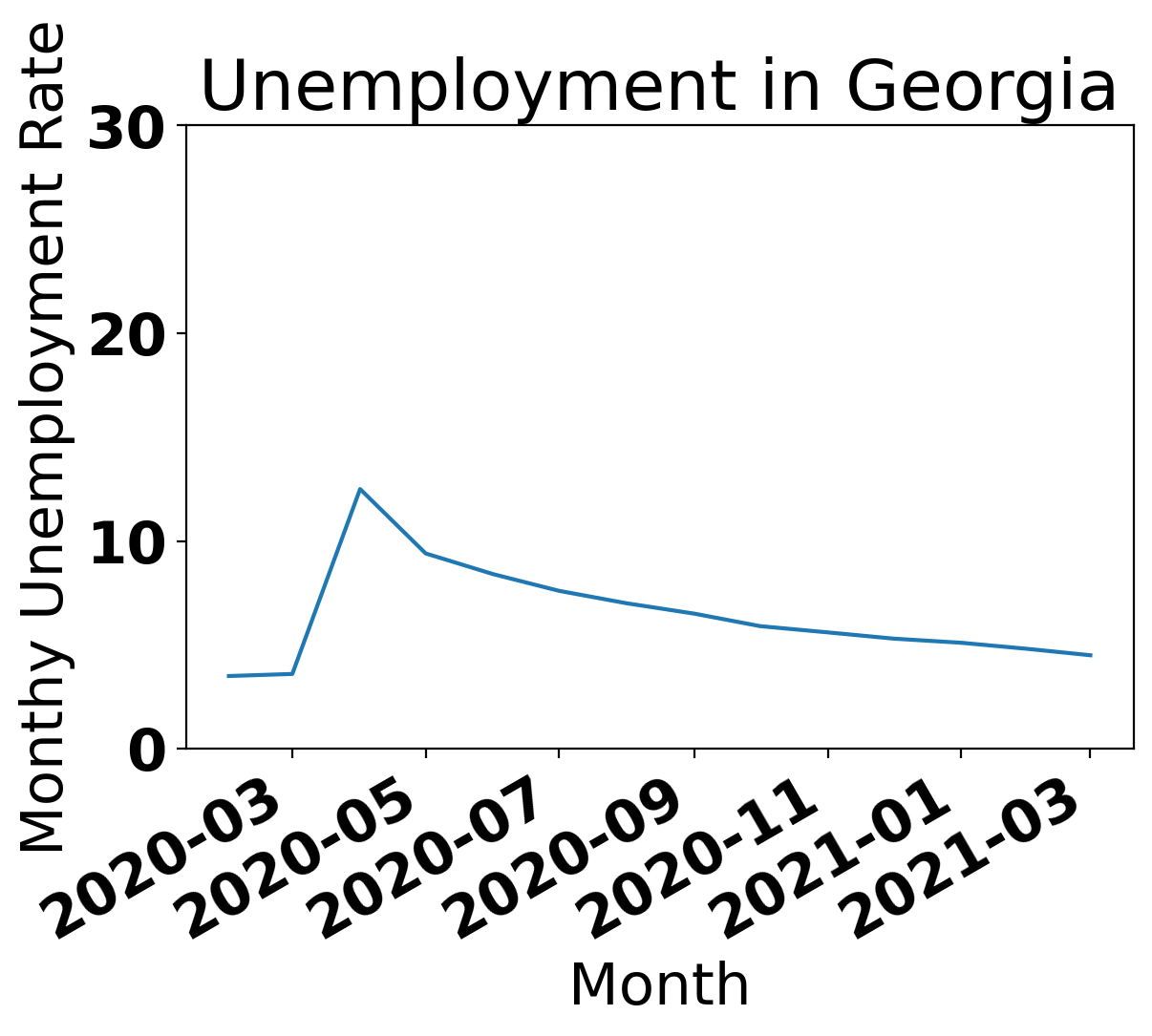}
  \caption{Unemployment in GA.}
\end{subfigure}
\hfill
\begin{subfigure}{.32\columnwidth}
  \centering
  \includegraphics[width=\linewidth]{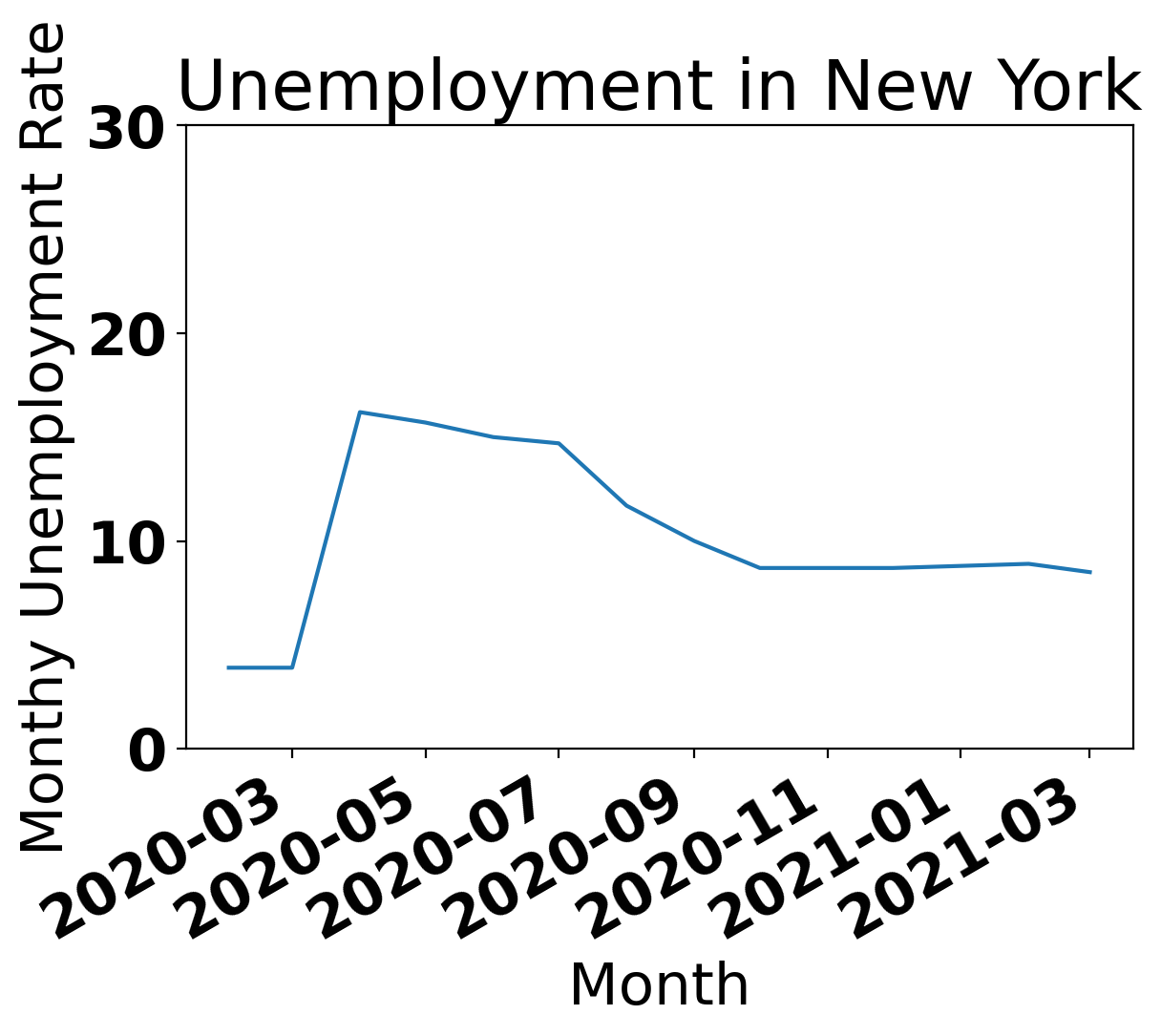}
  \caption{Unemployment in NY.}
\end{subfigure}
\begin{subfigure}{.32\columnwidth}
  \centering
  \includegraphics[width=\linewidth]{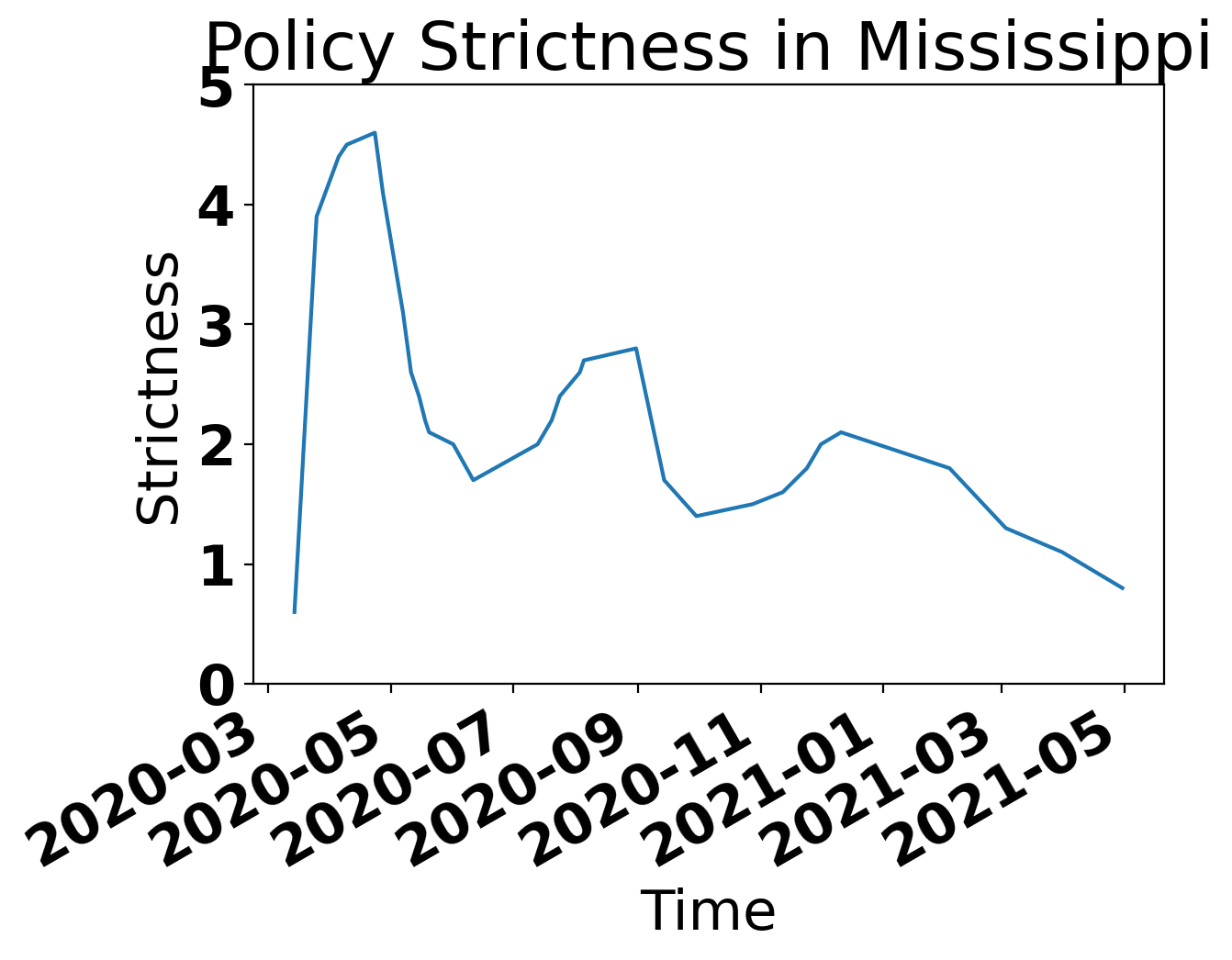}
  \caption{Policy of MS.}
\end{subfigure} 
\hfill
\begin{subfigure}{.32\columnwidth}
  \centering
  \includegraphics[width=\linewidth]{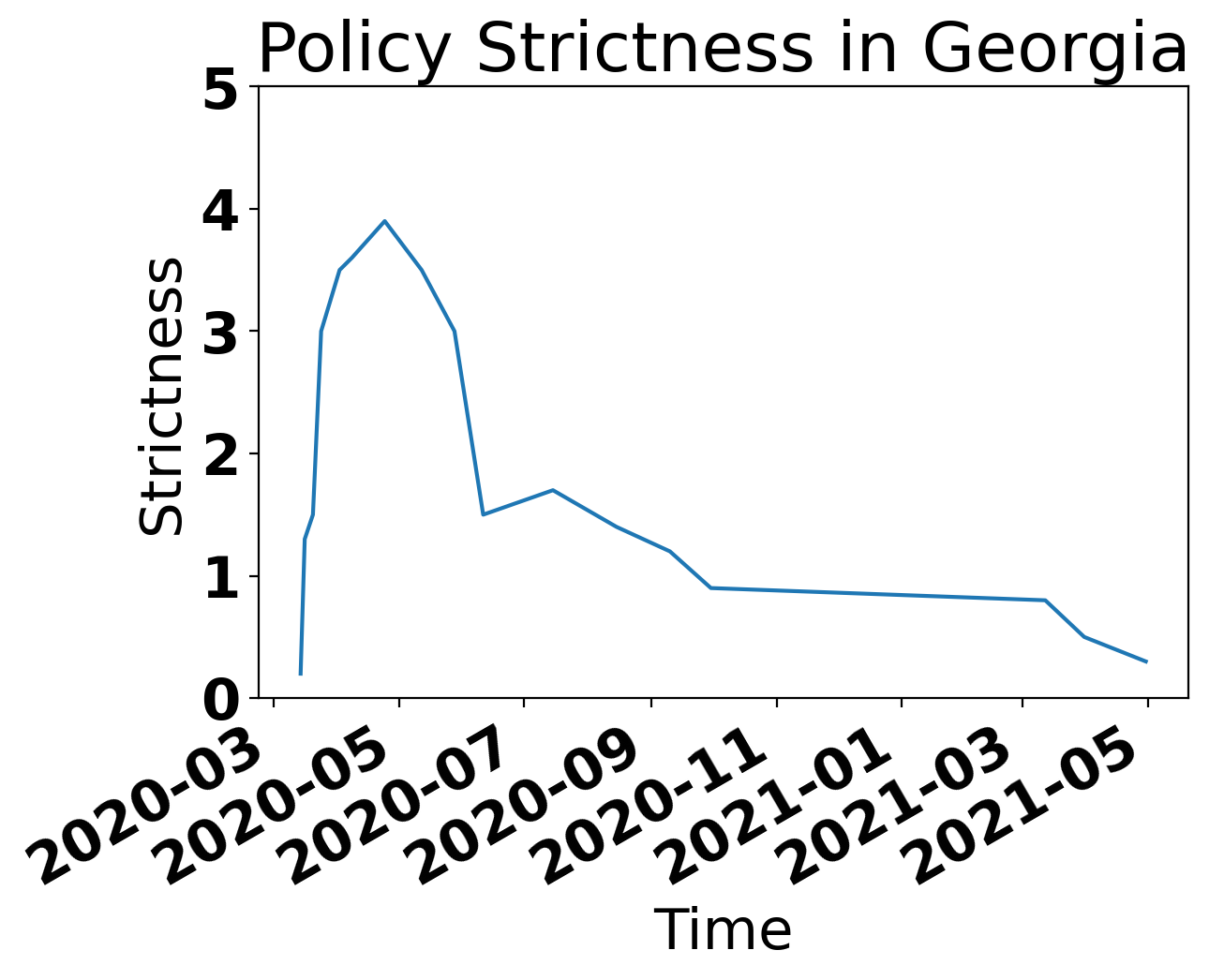}
  \caption{Policy of GA.}
\end{subfigure}
\hfill
\begin{subfigure}{.32\columnwidth}
  \centering
  \includegraphics[width=\linewidth]{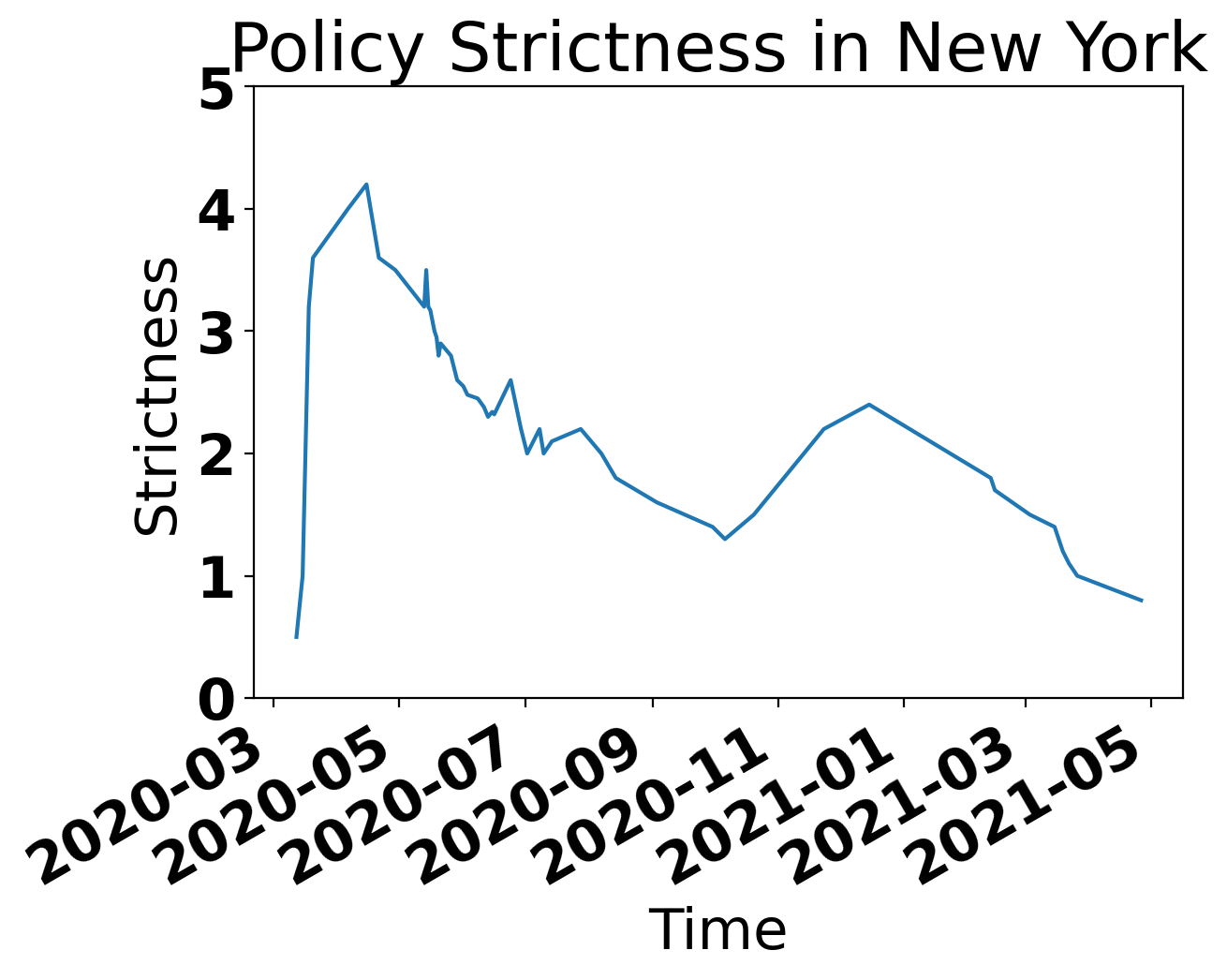}
  \caption{Policy of NY.}
\end{subfigure}
\caption{Comparative study of states. MS and GA is a pair of states with the most similar confounders, and NY is an irrelevant state to contrast how different MS and GA are from other states. Note that unemployment data is only available until March 2021.}
\label{covidtwitter:fig:comparative_states}
\end{figure}

In \cref{covidtwitter:fig:comparative_states}, we show an example pair of states, Mississippi (MS) and Georgia (GA), which have highly similar case numbers and unemployment rates at most time steps. Note that we use the New York (NY) state to show in contrast how the above pair is different from another unrelated state.

In the comparative study of MS and GA, they can be considered as counterfactuals for each other. In their policy curves, the policy strictness in MS responds to the COVID case numbers (e.g., the policies are stricter on the rising slope of case numbers), but the policies in GA remain loose even during the rising trends in July -- August 2020, and November 2020 -- January 2021. We look into the sentiment differences across the two states: For example, during November 2020 -- January 2021, GA experienced a very low average sentiment of -0.58 in the [-1, 1] scale, whereas MS experienced a milder sentiment of -0.04. By the controled comparison, the more negative sentiment is the potential cause for looser policies in GA.

\section{Additional Discussions}
\paragraph{Fine-Grained Opinions behind the Sentiments.}
To further interpret why positive tweets usually lead to stricter social distancing policies (and negative tweets lead to looser policies), we look into the correlation of Twitter sentiment and the user's opinion towards social distancing policies. Note that usually it is not easy to directly get an unsupervised intent classifier on COVID specific tweets. Hence, we ask the annotators to classify the opinion on social distancing for the 500 tweets in our test set as supportive, against, and not related to social distancing. Among the tweets about social distancing with positive sentiment, 95.13\% support social distancing. Among the tweets about social distancing with negative sentiment, 69.38\% are against social distancing and ask for the reopening of the state.

\paragraph{Additional Analyses.}
We put our additional analyses in \cref{covidtwitter:appd:analyses}, including correlation across all variables, and alternative causal analysis models such as difference-in-differences \citep{abadie2005semiparametric}, and continuous-valued propensity score matching \citep{hirano2004propensity,bia2008stata}.

\paragraph{Limitations.}
There are several limitations of this study. For example, a common limitation of many causal inference settings is the uncertainty of hidden confounders. In our study, we list all the variables that we believe should be considered, but future studies can investigate the effect of other confounders.

Another limitation is the accuracy of the Twitter sentiment classifier. Since the Twitter sentiment during COVID is very task-specific, modeling the sentiments can be very challenging. For example, our model often misclassifies ``increased positive cases'' as a positive sentiment. Another challenge is that some tweets refer to a url. These cases are difficult to deal with, and might be worth more detailed analyses in future studies.

In the data setting, one limitation is that for causal inference, modeling the whole time series is extremely challenging, so we empirically take the 14-day time span, which is a commonly used time span for many other COVID measures.

\paragraph{Future Work.}
This work is the first work to use NLP and causal inference to address policy responsiveness, and we explicitly measure the alignment of government policies and people's voice. This signal can be very important for the government and decision-makers.

In future work, a similar approach can  be used together with other variables (e.g., economic growth, participation in health/vaccination campaigns, well-being) to determine to which extent such people-government alignment relates to societal outcomes.

\section{Conclusion}
In this paper, we conducted multi-faceted analyses on the causal impact of Twitter sentiment on COVID policies in the 50 US states. To enable our study, we compile a large dataset of over 10 million governor-targeted COVID tweets, we annotate 838 state-level policies, and we collect data ten potential confounders such as daily COVID cases and unemployment rates. We use a multivariate linear regression and do-calculus to quantify both the correlation of Twitter sentiment as well as its causal impact  on policies, in the presence of other confounders. To our knowledge, this is one of the first studies to utilize massive social media data on crisis policy responsiveness, and lays the foundation for future work at the intersection of NLP and policy analyses.

\section*{Ethical Considerations}
\paragraph{Use of Data}
For the data used in this study, the COVID-related tweets are a subset of the existing dataset provided by \citet{chen2020tracking}. Following the data regulations of Twitter, we will not publicize the raw tweet text. If necessary, we can provide the list of tweet IDs to future researchers. For the policy strictness we annotated, we will open-source it since it is public information that can benefit societies affected by the pandemic, and has no privacy or ethical issues. For other confounding variables, the data are also public information.

\paragraph{Potential Stakeholders}
This research can be used for policy-makers or political science researchers. The research on causality between public opinion and political decision-making helps make policies more interpretable. One potential caveat is that there might be parties that maliciously manipulate sentiments on Twitter to affect politicians. A mitigation method is to control the flow of misinformation, terrorism and violent extremism on social media. The ideal use of the study is to reflect the process how a democracy system surveys the opinion from people, and makes policies that best balances people's long-term and short-term interests.

\mainchapter{CausalCite: A Causal Formulation of Paper Citations}\label{ch:causalcite}

\renewcommand{\ourmodel}{{\modelfont{TextMatch}}\xspace}
\newcommand{\ourname}{\textsc{CausalCite}\xspace}
\newcommand{\pciavg}{ACI\xspace}
\newcommand{\blackgreen}[1]{\textbf{\textcolor{mygreen2}{#1}}}
\graphicspath{ {/fig/causalcite/} }

Citation count of a paper is a commonly used proxy for evaluating the significance of a paper in the scientific community. Yet citation measures are widely criticized for failing to accurately reflect the true impact of a paper.
Thus, we propose \ourname, a new way to measure the significance of a paper by assessing the causal impact of the paper on its follow-up papers.
\ourname\ is based on a novel causal inference method, \textit{\ourmodel}, which adapts the traditional matching framework to high-dimensional text embeddings.  \ourmodel\ encodes each paper using text embeddings from large language models (LLMs), extracts similar samples by cosine similarity, and synthesizes a counterfactual sample as the weighted average of similar papers according to their similarity values. 
We demonstrate the effectiveness of \ourname\ on various criteria, such as high correlation with paper impact as reported by scientific experts on a previous dataset of 1K papers, (test-of-time) awards for past papers, and its stability across various subfields of AI. We also provide a set of findings that can serve as suggested ways for future researchers to use our metric for a better understanding of the quality of a paper.
Our code is available at \url{https://github.com/causalNLP/causal-cite}.

\section{Introduction}
Recent years have seen explosive growth in the number of scientific publications, making it increasingly challenging for scientists to navigate the vast landscape of scientific literature.
Therefore, identifying a good paper has become a crucial challenge for the scientific community, not only for technical research purposes, but also for making decisions, such as funding allocation \citep{carlsson2009allocation}, research evaluation \citep{moed2006citation}, recruitment \citep{holden2005bibliometrics}, and university ranking and evaluation \citep{piro2016how}. 

\begin{figure}[ht]
    \centering
    \includegraphics[width=.7\linewidth]{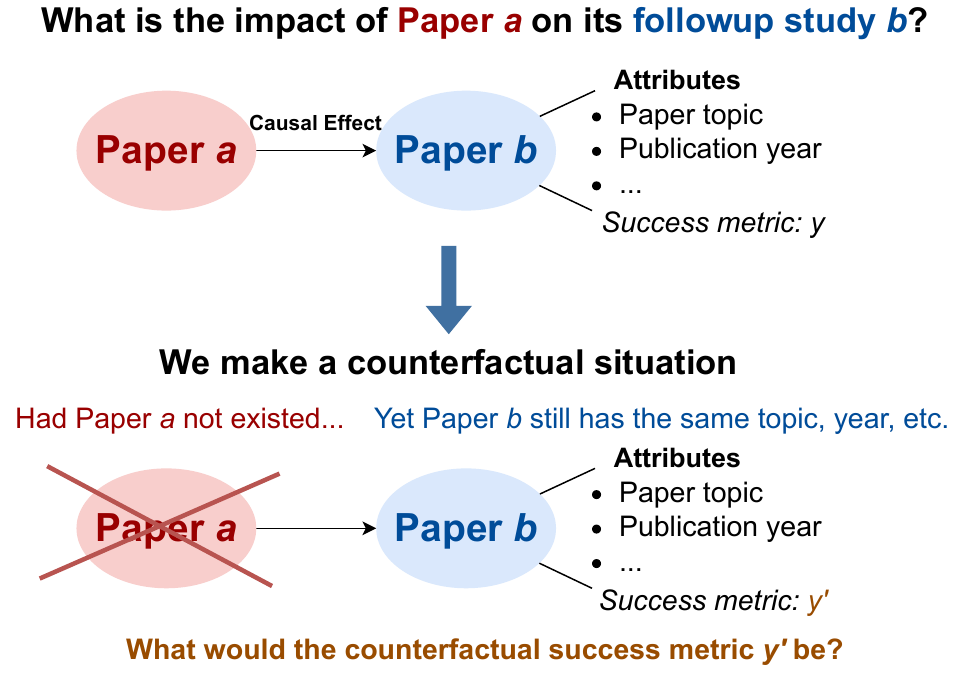}
    \caption{An overview of our research question.}
    \label{causalcite:fig:intro}
\end{figure}

A traditional approach to recognize paper quality is peer review, a mechanism that requires large efforts, and yet has inherent randomness and flaws \citep{cortes2021inconsistency, rogers-etal-2023-report, shah2021survey, prechelt2018community, resnik2008perceptions}.
Moreover, the number of papers after peer review is still overwhelmingly large for researchers to read, leaving the challenge of identifying truly impactful research unaddressed.
Another commonly used metric is citations. However, this metric faces criticism for biases, such as a preference for survey, toolkit, and dataset papers \citep{zhu, valenzuela2015identifying}.
Together with altmetrics \citep{wilsdon2016metric}, which incorporates social media attention to a paper, both metrics also bias towards papers from major publishing countries \citep{rungta-etal-2022-geographic,gomez2022leading}, with extensive publicity and promotion, and authored by established figures.

To provide a more equitable assessment of paper quality, we employ the causal inference framework \citep{hernan2010causal} to quantify a paper's impact by how much of the academic success in the follow-up papers should be \textit{causally attributed} to this paper.
We introduce \ourname, an enhanced citation based metric that poses the following \emph{counterfactual} question (also shown in \cref{causalcite:fig:intro}): ``\textit{had this paper never been published, what would have happened to its
follow-up studies?}''
To compute the causal attribution of each follow-up paper, we contrast its citations (the treatment group) with citations of papers that address a similar topic, but are not built on the paper of interest (the control group).

Traditionally, this problem is  solved by using the matching method \citep{rosenbaum1983central} in causal inference, which discretizes the value of the confounder variable, and compares the treatment and control groups with regard to each discretized value of the confounder variable. However, this approach does not apply when the confounder variable is
high-dimensional, e.g., text data, such as the content of the paper. %
Thus, we improve the matching method to adapt for textual confounders, by marrying recent
advancement of large language models (LLMs) with traditional causal inference.
Specifically, we propose \ourmodel, which uses LLMs to encode an academic paper as a high-dimensional text embedding to represent the confounders, and then, instead of iterating over discretized values of the confounder, we match each paper in the treatment group with papers from the control group with high cosine similarity by the text embeddings.

\ourmodel makes contributions in three different aspects: (1) it relaxes the previous constraint that the confounder variable should be binned into a limited set of intervals, and makes the matching method applicable for high-dimensional continuous variable type for the confounder; (2) since there are millions of papers, we enable efficient matching via a matching-and-reranking approach, %
first using information retrieval (IR) \citep{cosineSim} to extract a small set of candidates, and then applying semantic textual similarity (STS) \citep{majumder2016semantic,chandrasekaran2022evolution} for fine-grained reranking; and (3) we enable a more stable causal effect estimation by
leveraging all the close matches to synthesize the \textit{counterfactual citation score} by a weighted average according to the similarity scores of the matched papers. 

\ourname quantifies scientific impact via a causal lens, offering an alternative understanding of a paper's impact within the academic community.
To test its effectiveness, we conduct extensive experiments using the Semantic Scholar corpus \citep{lo-etal-2020-s2orc,Kinney2023TheSS}, comprising of $206$M papers and $2.4$B citation links.
We empirically validate \ourname by showing higher predictive accuracy of paper impact (as judged by scientific experts on a past dataset of 1K papers \citep{zhu}) compared to citations and other previous impact assessment metrics.
We further show a stronger correlation of the metric with the test-of-time (ToT) paper awards.
We find that, unlike citation counts, our metric exhibits a greater balance across various research domains in AI, e.g., general AI, NLP, and computer vision (CV). While citation numbers for papers in these domains vary significantly -- for example, while an average CV paper has many more citations than an average NLP paper, \ourname scores papers across AI sub-fields more similarly.

After demonstrating the desirable properties of our metric, we also present several case studies of its applications.
Our findings reveal that the quality of conference best papers is noisier on average than that of ToT papers (\cref{causalcite:sec:best}). %
We then showcase and present \ourname for several well-known papers (\cref{causalcite:sec:famous_papers}) and utilize \ourname to identify high-quality papers that are less recognized by citation counts (\cref{causalcite:sec:outlier}).

In conclusion, our contributions are as follows:
\begin{enumerate}%
    \item We introduce \ourname, a counterfactual causal effect-based formulation for paper citations.
    \item We develop \ourmodel, a new method that leverages LLMs and causal inference to estimate the counterfactual causal effect of a paper.
    \item We conduct comprehensive analyses, including various performance evaluations and present new findings using our metric.
\end{enumerate}

\section{Problem Formulation}
Our problem formulation involves a citation graph and a causal graph. We use lowercase letters for specific papers and uppercase for an arbitrary paper treated as a random variable.

\paragraph{Citation Graph}
In the citation graph $\mathbb{G} := (\mathbb{P}, \mathbb{L})$, $\mathbb{P}$ is a set of papers, and each edge $\ell_{i,j} \in \mathbb{L}$ indicates that an earlier paper ${p}_i$ influences (i.e., is cited by) a follow-up paper ${p}_j$.
To obtain the citation graph, we use the Semantic Scholar Academic Graph dataset \citep{Kinney2023TheSS} with 206M papers and 2.4B citation edges.
\ifperfect \red{We focus on non-trivial influence, limiting citation links to those identified as highly influential by Semantic Scholar \citep{valenzuela2015identifying}.}\fi

\begin{figure}[ht]
    \centering
    \includegraphics[width=.8\linewidth]{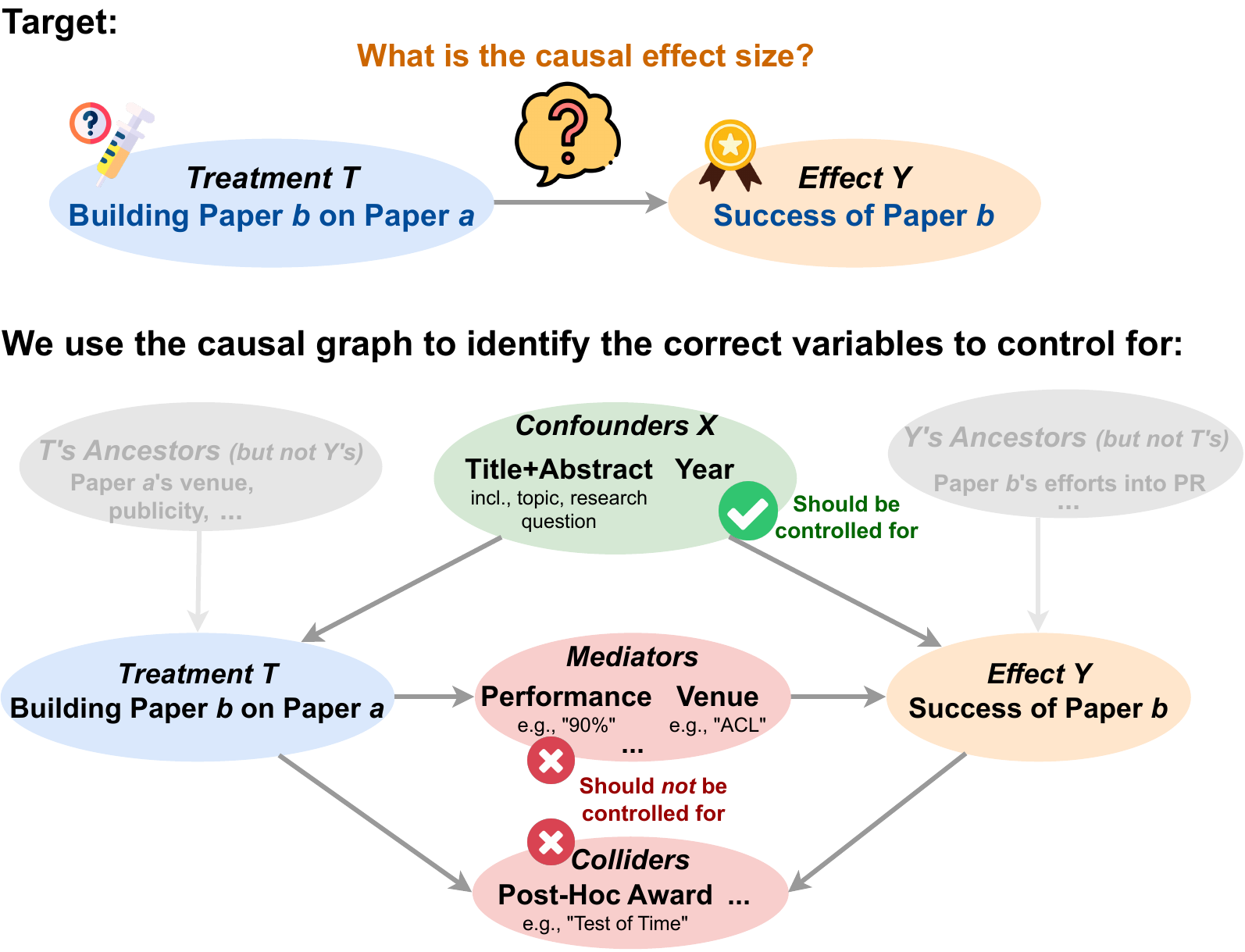}
    \caption{The causal graph of our study. 
}

    \label{causalcite:fig:graph}
\end{figure}

\paragraph{Causal Graph.}
The causal graph, shown in \cref{causalcite:fig:graph}, highlights the contribution of a paper $a$ to a follow-up paper $b$.
We use a binary variable $T$ to indicate if $a$ influences $b$ and an effect variable $Y$ to represent the success of $b$.
We use $\log_{10}$ of citation counts to quantify $Y$, although other transformations can also be used.
We introduce two sets of variables in this causal graph:
(i) The set of confounders, which are the common causes of $T$ and $Y$.
For instance, the research area of $b$ impacts both the likelihood of a paper citing $a$ and its own citation count.
(ii) Descendants of the treatment, comprising mediators (e.g., paper $a$ influencing the quality of paper $b$ and subsequently influencing its citations) and colliders (e.g., both the influence from $a$ and the citations of $b$ influencing later awards received by $b$).

\subsection{\ourname Indices}
In this section, we introduce various indices that measure the causal impact of a paper.

\paragraph{Two-Paper Interaction: Pairwise Causal Impact (PCI).}
To examine the causal impact of a paper $a$ on a follow-up paper $b$, we define the pairwise causal impact $\mathrm{PCI}(a, b)$ by unit-level causal effect:
\begin{align}
    \mathrm{PCI}(a, b) := y^{t=1} - y^{t=0}
    ~,
\end{align}
where we compare the outcomes $Y$ of the paper $b$ had it been influenced by paper $a$ or not, denoted as the actual $y^{t=1}$ and the counterfactual $y^{t=0}$, respectively.
Note that the counterfactual
$y^{t=0}$ can never be observed, but only estimated by statistical methods, as we will discuss in \cref{causalcite:sec:prev_method}.
\ifperfect \textcolor{gray}{
We take the commonly adopted stable unit treatment value assumption (SUTVA) \citep{rubin1980}, where there is no interaction among the units, such as peer effects for students assigned in the same class.
\fi

\paragraph{Single-Paper Quality Metrics: Total Causal Impact (TCI) and Average Causal Impact (ACI).}
Let $\bm{S}$ denote the set of all follow-up studies of paper $a$.
We define total causal impact $\mathrm{TCI}(a)$ as the sum of the pairwise causal impact index $\mathrm{PCI}(a, b)$ across all $b \in \bm{S}$.
That is,
\begin{equation}
    \mathrm{TCI}(a) := \sum_{b \in \bm{S}} \mathrm{PCI}(a, b)~.
    \label{causalcite:eq:tci}
\end{equation}
This definition provides an aggregated measure of a paper's influence across all its follow-up papers.

As the causal inference literature is usually interested in the average treatment effect, we further define the average causal impact (ACI) index as the average per paper PCI:
\begin{equation}
    \mathrm{ACI}(a) := \frac{\mathrm{TCI}(a)}{|\bm{S}|} = \frac{1}{|\bm{S}|} \sum_{b \in \bm{S}} \left( y^{t=1} - y^{t=0}\right)
    ~.
    \label{causalcite:eq:aci}
\end{equation}
\ifarxiv 
We note that $\mathrm{ACI}(a)$ is equal to the \textbf{a}verage \textbf{t}reatment effect on the \textbf{t}reated (ATT) of paper $a$ \citep{pearl2009causality}.
\fi

\section{The \ourmodel Method}\label{causalcite:sec:method}

As illustrated in \cref{causalcite:fig:intro}, the objective of our study is to quantify the causal effect of the treatment $T$ (i.e., whether paper $b$ is built on paper $a$) on the effect $Y$ (i.e., the outcome of paper $b$).
To approach this, we envision a counterfactual scenario: what if paper $a$ had never been published, yet certain key characteristics of paper $b$ remain unchanged?
The critical question then becomes: which key characteristics of paper $b$ should be \emph{controlled} for in this hypothetical situation?

\subsection{What Does Causal Inference Tell Us about What Variables to Control for, and What Not?}
    In causal inference, selecting the appropriate variables for control is a delicate and crucial process that affects the accuracy of the analysis.
    Pearl's seminal work on causality guides us in differentiating between various types of variables \citep{pearl2009causality}.
    
    Firstly, we must control for \emph{confounders} – variables that influence both the treatment and the outcome.
    Confounders can create spurious correlations; if not controlled, they can lead us to mistakenly attribute the effect of these external factors to the treatment itself.
    For example, in assessing the impact of one paper on another, if both papers are in a trending research area, the apparent influence might be due to the popularity of the topic rather than the papers’ content.
    
    However, not all variables warrant control.
    Mediators and colliders should be explicitly avoided in control.
    Mediators are part of the causal pathway between the treatment and outcome.
    By controlling them, we would block the very effect we are trying to measure.
    Colliders, affected by both the treatment and the outcome, can introduce bias when controlled.
    Controlling a collider can inadvertently create associations that do not naturally exist.
    In general, this also includes not controlling for the descendants of the treatment, as it could obscure the direct impact we intend to study.
    
    Lastly, variables that do not share a causal path with both the treatment and outcome, known as \emph{unshared ancestors}, are less critical in our analysis.
    They do not contribute to or confound the causal relationship we are exploring, and thus, controlling for them does not add value to our causal understanding.

\subsection{Can Existing Causal Inference Methods Handle This Control?}\label{causalcite:sec:prev_method}

Several causal inference methods have been proposed to address the problem of estimating treatment effects while controlling for confounders.
Next, we will discuss the workings and limitations of three classical methods.

\paragraph{Randomized Control Trials (RCTs) Assumes Intervenability.}
The ideal way to obtain causal effects is through randomized control trials (RCTs).
For example, when testing a drug, we randomly split all patients into two groups, the control group and the treatment group, where the random splitting ensures the same distribution of the confounders across the two groups
such as gender and age.
However, RCTs are usually not easily achievable, in some cases too expensive (e.g., tracking hundreds of people's daily lives for 50 years), and in other cases unethical (e.g., forcing a random person to smoke), or infeasible (e.g., getting a time machine to change a past event in history).

For our research question on a paper's impact, utilizing RCTs is impractical as it is infeasible to randomly divide researchers into two groups, instructing one group to base their research on a specific paper $a$ while the other group does not, and then observe the citation count of their papers years later.

\paragraph{Ratio Matching Iterates over Discretized Confounder Values.}
In the absence of RCTs, matching is as an alternate method for determining causal effects from observational data.
In this case, we can let the treatment assignment happen naturally, such as taking the naturally existing set of papers and running causal inference by adjusting for the variables that block all paths.
Given a set of naturally observed papers, one of the most commonly used causal inference methods is ratio matching \citep{rosenbaum1983central}, whose basic idea is to iterate over all possible values $\bm{x}$ of the adjustment variables $\bm{X}$ and obtain the difference between the treatment group $\mathcal{T}$ and control group $\mathcal{C}$:
\begin{equation}
    \widehat{\mathrm{ACI}}(a) 
    = \sum_{\bm{x}} P(\bm{x}) \left( \frac{1}{|\mathcal{T}_{\bm{x}}|} \sum_{i \in \mathcal{T}_{\bm{x}}} y_i - \frac{1}{|\mathcal{C}_{\bm{x}}|} \sum_{j \in \mathcal{C}_{\bm{x}}} y_j
    \right)
    ~,
\end{equation}
where for each value $\bm{x}$, we extract all the units corresponding to this value in the treatment and control sets, compute the average of the effect variable $Y$ for each set, and obtain the difference.

While ratio matching is practical when there is a small set of values for the adjustment variables to sum over, its applicability dwindles with high-dimensional variables like text embeddings in our context. This scenario may generate numerous intervals to sum over, presenting numerical challenges and potential breaches of the positivity assumption.

\paragraph{One-to-One Matching Is Susceptible to Variance.}
To handle high-dimensional adjustment variables,
one possible way is to avoid pre-defining all their possible intervals, but, instead, iterating over each unit in the treatment group to match for its closest control unit \cite[e.g.,][]{mcgue2010causal,sato2022twin}.
Consider a given follow-up paper $b$, and a set of candidate control papers $\bm{C}$, where each paper $c_i$ has
a citation count $y_i$, and vector representation $\bm{t}_i$ of the confounders  (e.g., research topic).
One-to-one matching estimates PCI as
\begin{equation}
\begin{split}
\widehat{\mathrm{PCI}}(a, b) 
    &= y_b - y_{\argmax_{c_i \in \bm{C}} m_i}
     \\
    &= y_b - y_{\argmax_{c_i \in \bm{C}}\mathrm{sim}(\bm{t}_b, \bm{t}_i
)}
    ~,
\end{split}
\label{causalcite:eq:match_score}
\end{equation}
where we approximate the counterfactual sample by the paper $c_i\in \bm{C}$ which is the most similar to paper $b$ by the matching score $m_i$, which is obtained by the cosine similarity $\mathrm{sim}$ of the confounder vectors.
A limitation of the
one-to-one matching method is that it
might induce large  instability in the result, as only taking one paper 
with similar contents
may have a large variance in citations when the matched paper slightly differs.

\subsection{How Do We Extending Causal Inference to Text Variables?}

\subsubsection{Theoretical Formulation of \ourmodel: Stabilizing Text Matching by Synthesis}
To fill in the aforementioned gap in the existing matching methods, we propose \ourmodel, which mitigates the instability issue of one-to-one matching by replacing it with
a convex combination of a set of matched samples to form a synthetic counterfactual sample.
Specifically, we identify a set of papers $c_i \in \bm{C}$ with high matching scores $m_i$ to the paper $b$, and synthesize the counterfactual sample by an interpolation of them:
\begin{align}
    \widehat{\mathrm{PCI}}(a,b) = y_b - \sum_{c_i \in \bm{C}} w_i y_i
     = y_b - \sum_{c_i \in \bm{C}} \frac{m_i}{\sum_{c_i \in \bm{C}} m_i} y_i
     ~,
     \label{causalcite:eq:syn}
\end{align}
where the weight $w_i$ of each paper $c_i$ is proportional to the matching score $m_i$ and normalized.

The contributions of our method are as follows:
(1) we adapt the traditional matching methods from low-dimensional covariates to any high-dimensional variables such as text embeddings;
(2)
different from the ratio matching, we do not stratify the covariates, but synthesize a counterfactual sample for each observed treated units;
(3) due to this iteration over each treated unit instead of taking the population-level statistics, we closely control for exogenous  variables for the ATT estimation, which circumvents that need for 
the structural causal models;
(4) we further stabilize the estimand by a convex combination
of a set of similar papers. Note that the contribution of \cref{causalcite:eq:syn} might seem to bear similarity with synthetic control \citep{abadie2003economic,abadie2010synthetic}, but they are fundamentally different, 
in that synthetic control runs on time series, and fit for the weights $w_i$ by linear regression between the time series of the treated unit and a set of time series from the control units, using each time step's values in the regression loss function.

\subsubsection{Overall Algorithm}
To operationalize our theoretical formulation above, we introduce our overall algorithm
in \cref{causalcite:alg:getPCI}. We briefly give an overview of the the algorithm with more details to be elaborated in later sections. We use the weighted average of the matched samples following our \ourmodel method in \cref{causalcite:eq:syn} through
\cref{causalcite:line:eq_prep1,causalcite:line:eq_prep2,causalcite:line:eq_prep3,causalcite:line:eq_prep4,causalcite:line:eq1,causalcite:line:eq2,causalcite:line:eq3,causalcite:line:eq4,causalcite:line:eq5,causalcite:line:eq_final}. In our experiments, we use the interpolation of up to top 10 matched papers. We encourage future work to explore other hyperparameter settings too. Given the PCI estimation, the main spirit of the $\textsc{GetACIandTCI}(a)$ function is to average or sum over all the follow-up studies of paper $a$, following the theoretical formulation in \cref{causalcite:eq:tci,causalcite:eq:aci}
and implemented in our algorithm through \cref{causalcite:line:add1,causalcite:line:add2,causalcite:line:add3,causalcite:line:add4,causalcite:line:add5,causalcite:line:add6}.

\begin{algorithm}[ht]
\footnotesize
    \caption{Get causal impact indices $\mathrm{ACI}$ and $\mathrm{TCI}$}
    \begin{algorithmic}[1]
    
        \State \textbf{Input}: Paper $a$.

        \Procedure{GetACIandTCI}{$a$}
            \State $\bm{D} \gets \mathrm{GetDesc}(a)$ \Comment{Get descendants by DFS}
            \State $\bm{B} \gets \mathrm{GetChildren}(a)$
            \State $\bm{B}' \gets \mathrm{SampleSubset}(\bm{B})$
            \Comment{See \cref{causalcite:sec:sampling}}
            \State $\bm{C} \gets \mathrm{EntireSet} \backslash \{\bm{D} \cup \{a\}\}$
            \Comment{Get non-descendants}
            \State $\mathrm{ACI} \gets 0$
            \label{causalcite:line:add1}
            \For {each $b_i$ in $\bm{B}'$}
            \label{causalcite:line:add2}
                \State $I_i \gets \textsc{GetPCI}(a, b_i, \bm{C})$
            \label{causalcite:line:add3}
                \State $\mathrm{ACI} \gets \mathrm{ACI}+ \frac{1}{|\bm{B}'|} \cdot I_i$
            \label{causalcite:line:add4}
            \EndFor
            \label{causalcite:line:add5}
            \State $\mathrm{TCI} \gets \mathrm{ACI} \cdot |\bm{B}|$
            \label{causalcite:line:add6}
            \State \textbf{return} $\mathrm{ACI}$ and $\mathrm{TCI}$
    \EndProcedure
        \Statex
        \Procedure{GetPCI}{$a, b, \bm{C}$}
            \State $\bm{C}_{\mathrm{sameYear}} \gets \mathrm{FilterByYear}(\bm{C}, b_{\mathrm{year}})$
            \label{causalcite:line:year}
            \For {each $p_i$ in $\bm{C}_{\mathrm{sameYear}} \cup \{b\}$}\label{causalcite:line:remove1}
                \State $\bm{t}_i \gets \mathrm{RemoveMediator}(\mathrm{TitleAbstract}_i)$
                \label{causalcite:line:remove2}
            \EndFor\label{causalcite:line:remove3}
            \State $\bm{C}_{\mathrm{coarse}} \gets \mathrm{BM25}(b, \bm{C}_{\mathrm{sameYear}}, \text{topk}=100)$ \label{causalcite:line:bm}
            \For {each $c_i$ in $\bm{C}_{\mathrm{coarse}}$} \label{causalcite:line:sim1}
                \State $m_i \gets \mathrm{Sim}(\bm{t}_b, \bm{t}_{i})$
                \label{causalcite:line:sim}
            \EndFor \label{causalcite:line:sim2}
            \State $\bm{C}_{\mathrm{top10}} \gets \mathrm{argmax10}_m(\bm{C}_{\mathrm{coarse}})$ 
            \Statex
            \State $M \gets 0$ \label{causalcite:line:eq_prep1}
            \For {each $c_i$ in $\bm{C}_{\mathrm{top10}}$} \Comment{For the normalization later}\label{causalcite:line:eq_prep2}
                \State $M \gets M + m_i$\label{causalcite:line:eq_prep3}
            \EndFor\label{causalcite:line:eq_prep4}
            \State $\hat{y}^{t=0} \gets 0$ 
            \label{causalcite:line:eq1}
            \For {each $c_i$ in $\bm{C}_{\mathrm{top10}}$}
            \label{causalcite:line:eq2}
                \State $w_i \gets \frac{m_i}{M} $
                \label{causalcite:line:eq3}
                \State $\hat{y}^{t=0} \gets \hat{y}^{t=0} + w_i \cdot y_i$
                \Comment{Apply \cref{causalcite:eq:syn}}
                \label{causalcite:line:eq4}
            \EndFor            \label{causalcite:line:eq5}
            \State \textbf{return} $y_b - \hat{y}^{t=0}$ \label{causalcite:line:eq_final}
        \EndProcedure
    \end{algorithmic}\label{causalcite:alg:getPCI}
\end{algorithm}

\subsubsection{Key Challenges and Mitigation Methods}
We address several technical challenges below.

\subsubsubsection{Confounders of Various Types}
First, as we mentioned in the causal graph in \cref{causalcite:fig:graph}, the confounder set consists of a text variable (title and abstract concatenated together) and an ordinal variable (publication year). Therefore, the similarity operation $\mathrm{Sim}$ between two papers should be customized. For our specific use case, we first filter by the publication year in \cref{causalcite:line:year}, as it is not fair to compare the citations of papers published in different years. Then, we apply the cosine similarity method paper embeddings as in \cref{causalcite:line:sim}. As a general solution, we recommend to separate hard logical constraints, and soft matching preferences, where the hard constraints should be imposed to filter the data first, and then all the rest of the variables can be concatenated to apply the similarity metric on.

\subsubsubsection{Excluding the Mediators from Confounders%
}
\label{causalcite:sec:adjust}
Another key challenge to highlight is that the text variable we use for the confounder might accidentally include some mediator information. For example, the quality or performance of a paper could be expressed in the abstract, such as ``we achieved 90\% accuracy.''
Therefore, we conduct a specific preprocessing procedure before feeding the text variable to the similarity function. For the $\mathrm{RemoveMediator}$ function in \cref{causalcite:line:remove2}, we exclude all numerical expressions such as percentage numbers, as well as descriptions such as ``state-of-the-art.'' For generalizability, the essence of this step is a entanglement action to separate the confounder variable (in this case, the research content) and all the descendants of the treatment variable (in this case, mentions of the performance). For more complicated cases in future work, we recommend a separate disentanglement model to be applied here.

\subsubsubsection{Efficient Matching-and-Reranking Method
}
Since we use one of the largest available paper databases, the Semantic Scholar dataset \citep{Kinney2023TheSS} containing 206M papers, we need to optimize our algorithm for large-scale paper matching.
For example, after we filter by the publication year, the number of candidate papers $\bm{C}_{\mathrm{sameYear}}$ could be up to 8.8M.
In order to conduct text matching across millions of papers, we use a \textit{matching-and-reranking} approach, by combining
two NLP tasks, information retrieval (IR) \citep{cosineSim} and semantic textual similarity (STS) \citep{majumder2016semantic,chandrasekaran2022evolution}. 

Specifically, we first run large-scale matching to obtain 100 candidates papers (\cref{causalcite:line:bm}) using the common IR method,
BM25 \citep{bm25}.
Briefly, BM25 is a bag-of-words retrieval function that uses term frequencies and document lengths to estimate relevancy between two text documents. Deploying this method, we can find a set of candidate papers for, for example, {two} million papers, at a speed {250}x faster than the text embedding cosine similarity matching.
Then, we conduct
a fine-grained reranking using cosine similarity 
(\cref{causalcite:line:sim1,causalcite:line:sim,causalcite:line:sim2}). In the cosine similarity matching process, 
we use the MPNet model \citep{song2020mpnet} to encode the text of each paper $c_i$ into an embedding $\bm{t}_i$, with which we get the matching score $m_i$ according to \cref{causalcite:eq:match_score} in \cref{causalcite:line:sim}, and the normalized weight $w_i$ by \cref{causalcite:eq:syn} in \cref{causalcite:line:eq3}.

\subsubsubsection{Numerical Estimation
}\label{causalcite:sec:sampling}
Given the large number of papers, it is numerically challenging to aggregate the TCI from individual PCIs,
because the number of follow-up papers for a study can be up to tens of thousands, such as the 57,200 citations by 2023 for the ImageNet paper \citep{deng2009imagenet}.
To avoid extensively running PCI for all follow-up papers, we propose a new numerical estimation method using a carefully designed random paper subset.

A naive way to achieve this aggregation is Monte Carlo (MC) sampling. However, %
unfortunately, MC sampling requires very large sample sizes when it comes to estimating long-tailed distributions, which is the usual case of citations. Since citations are more likely to be concentrated in the head part of the distribution, we cannot afford the computational budget for huge sample sizes that cover the tails of the distribution. Instead, we propose a novel numerical estimation method for sampling the follow-up papers, inspired by importance sampling \citep{Singh2014SamplingT,Kloek1976BayesianEO}.

\ifperfect
\begin{align}
    \mathrm{TCI}(a) := |\bm{S}| \cdot \frac{1}{N} \sum_{i=1}^{N} \mathrm{PCI}(a, b_i) 
    ~.
\end{align}
\fi

Our numerical estimation method works as follows: First, we propose the formulation that the relation between ACI and TCI is an integral over all possible paper $b$'s. Then, we formulated the above sampling problem as integral estimation or area-under-the-curve estimation. 
We draw inspiration from Simpson's method, which estimates integrals by binning the input variable into small intervals. Analogously, although we cannot run through all PCIs, we use citations as a proxy, bin the large set of follow-up papers according to their citations into $n$ equally-sized intervals, and perform random sampling over each bin, which we then sum over. In this way, we make sure that our samples come from all parts of the long-tailed distribution and are a more accurate numerical estimate for the actual TCI.

\section{Performance Evaluation}
    The contribution of a paper is inherently multi-dimensional, making it infeasible to encapsulate its richness fully through a scalar.
    Yet the demand for a single, comprehensible metric for research impact persists, fueling the continued use of traditional citations despite their known limitations.
    In this section, we show how our new metrics significantly improve upon traditional citations by providing quantitative evaluations comparing the effectiveness of citations, Semantic Scholar's highly influential (SSHI) citations \citep{valenzuela2015identifying}, and our \ourname metric.
\subsection{Experimental Setup}
\paragraph{Dataset} We use the Semantic Scholar dataset \citep{lo-etal-2020-s2orc,Kinney2023TheSS}\footnote{\href{https://api.semanticscholar.org/api-docs/datasets}{https://api.semanticscholar.org/api-docs/datasets}} which includes a corpus of 206M scientific papers, and a citation graph of 2.4B+ citation edges. For each paper, we obtain the title and abstract for the matching process. We list some more details of the dataset in \cref{causalcite:appd:data}, such as the number of papers reaching 8M per year after 2012.

\paragraph{Selecting the Text Encoder}
When projecting the text into the vector space, we need a text encoder with a strong representation power for scientific publications,
and is sensitive towards two-paper similarity comparisons regarding their abstracts containing key information such as the research topics.
For the representation power for scientific publications, instead of general-domain models such as BERT \citep{devlin-etal-2019-bert} and RoBERTa \citep{liu2019roberta}, 
we consider LLM variants\footnote{Note that we follow the standard notion by \citet{Yang2023HarnessingTP} to refer to BERT and its variants as LLMs.} pretrained on large-scale scientific text, such as SciBERT \citep{beltagy-etal-2019-scibert}, SPECTER \citep{specter2020cohan}, and MPNet \citep{song2020mpnet}.

To check the quality of two-paper similarity measures, 
we conduct a small-scale empirical study comparing human-ranked paper similarity and model-identified semantic similarity in \cref{causalcite:appd:emb}, according to which MPNet outperforms the other two models.

\paragraph{Implementation Details}\label{causalcite:sec:impl}
We deploy the \textit{all-mpnet-base-v2} checkpoint of the MPNet using the \textit{transformers} Python package \citep{wolf-etal-2020-transformers}, and set the batch size to be 32.
For the set of matched papers, we consider papers with cosine similarity scores higher than 0.81, which we optimize empirically on 100 random paper pairs. We take the top ten most similar papers above the threshold. 
In special cases where there is no matched paper above the threshold, it means that no other paper works on the same idea as Paper $b$, and we make the counterfactual citation number to be zero, which also reflects the quality of Paper $b$ as its novelty is high.

To enable efficient operations on the large-scale citation graph, we use the Dask framework,\footnote{\href{https://dask.org/}{https://dask.org/}} which optimizes for data processing and distributed computing.
We optimize our program to take around 100GB RAM, and on average 25 minutes for each $\mathrm{PCI}(a,b)$ after matching against up to millions of candidates. More implementation details are in \cref{causalcite:appd:impl}. 
For the estimation of TCI, we empirically select the sample size to be 40, which is a balance between the computational time and performance, as found in \cref{causalcite:appd:sample_size}.

\subsection{Author-Identified Paper Impact}
    In this experiment, we follow the evaluation setup in \citet{valenzuela2015identifying} to use an annotated dataset \citep{zhu} comprised of 1,037 papers, annotated according to whether they serve as significant prior work for a given follow-up study.
    Although paper quality evaluation can be tricky,
    this dataset was cleverly annotated by first collecting a set of follow-up studies and letting one of the authors of each paper go through the references they cite and select the ones that significantly impact their work. In other words, for a given paper $b$, each reference $a$ is annotated as whether $a$ has significantly impacted $b$ or not.

\cref{causalcite:tab:zhu} reports the accuracy of our \ourname metric, together with two existing citation metrics: citations, and SSHI citations \citep{valenzuela2015identifying}. See the detailed derivation of the accuracy scores in \cref{causalcite:appd:zhu}.
From this table, we can see that our \ourname metric achieves the highest accuracy, 80.29\%, which is 5 points higher than SSHI, and 9 points higher than the traditional citations.

\subsection{Test-of-Time Paper Analysis}

\begin{figure*}[th!]
  \begin{minipage}{0.47\textwidth}
    \centering
    \small 
    \centering
\begin{tabular}{lcl}
            \toprule
            \textbf{Metric}  & \textbf{Accuracy} \\
            \midrule
            Citations & 71.33  \\ 
            SSHI Citations & 75.25\\
            \ourname  & \textbf{80.29} \\ 
            \bottomrule
        \end{tabular}
        \captionof{table}{Accuracy of all three citation metrics.
        }
        \label{causalcite:tab:zhu}
  \end{minipage}
  \hfill
  \begin{minipage}{0.47\textwidth}
    \centering
    \small 
    \begin{tabular}{lccccll}
        \toprule
        \textbf{Metric}  & \textbf{Corr. Coef.} \\
        \midrule 
        Citations & 0.491  %
        \\ 
        SSHI Citations & 0.317\\
        TCI       & \textbf{0.640} %
        \\
        \bottomrule
    \end{tabular}
    \label{causalcite:tab:fullComparison}
    \captionof{table}{Correlation coefficients of each metric and ToT paper award by Point Biserial Correlation \citep{tate1954correlation}.
    }\label{causalcite:tab:tot}
  \end{minipage}
  \begin{minipage}{0.3\textwidth}
    \centering
    
        \includegraphics[width=\linewidth]{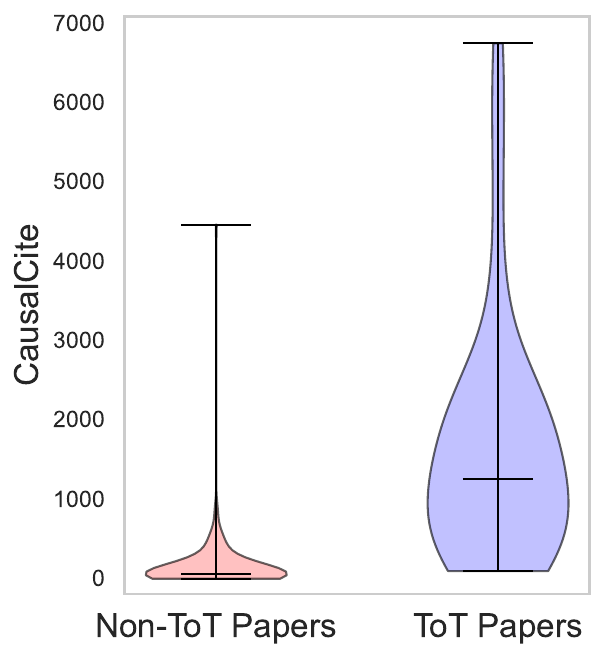}
        \caption{Distributions of ToT (mean: 142) and non-ToT papers (mean: 1,623). 
        }
        \label{causalcite:fig:ToTViolin} \label{causalcite:fig:tot}

  \end{minipage}\hfill
  \begin{minipage}{0.55\textwidth}
    \centering
    \includegraphics[width=\linewidth]{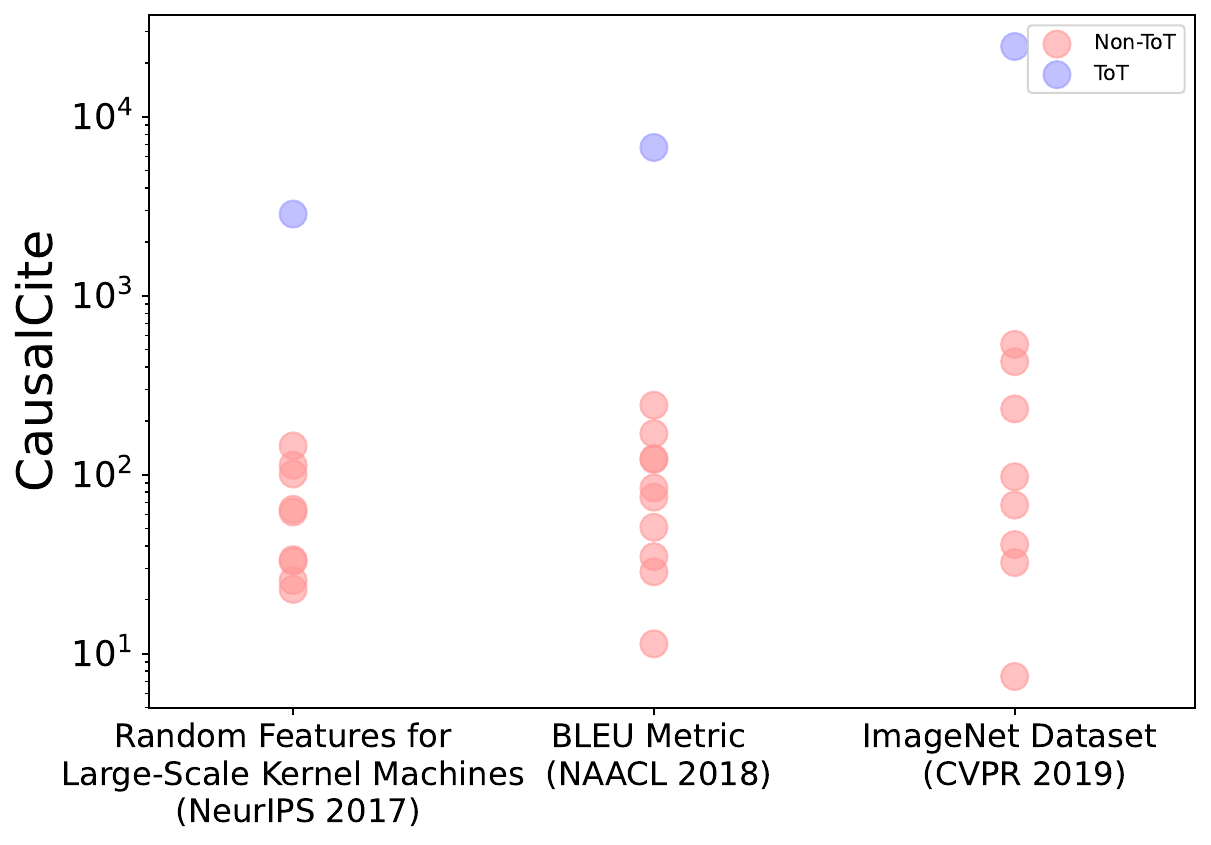}
        \caption{The \ourname values of three example ToT papers from general AI, NLP, and CV. 
        }
        \label{causalcite:fig:ToTMain}\label{causalcite:fig:totexample}
  \end{minipage}\hfill
\end{figure*}

    The test-of-time (ToT) paper award is a prestigious honor bestowed upon papers that have made substantial and enduring impacts in their field.
    In this section, we collect a dataset of {$792$} papers, including $72$ ToT papers, and a control group of $10$ randomly selected non-ToT papers from the same conference and year as each ToT paper.
    To collect this ToT paper dataset, we look into ten leading AI conferences spanning general AI (NeurIPS, ICLR, ICML, and AAAI), NLP (ACL, EMNLP, and NAACL), and CV (CVPR, ECCV, and ICCV), for which we go through each of their websites to identify all available ToT papers.\footnote{We get this list by selecting the top conferences on Google Scholar using the h5-Index ranking in each of the above domains: general AI (\href{https://scholar.google.com/citations?view_op=top_venues&vq=eng_artificialintelligence}{link}), CV (\href{https://scholar.google.com/citations?view_op=top_venues&vq=eng_computervisionpatternrecognition}{link}), and NLP (\href{https://scholar.google.com/citations?view_op=top_venues&vq=eng_computationallinguistics}{link}).}

In \cref{causalcite:tab:tot}, we show the correlations of various metrics with the ToT awards.
    In this table, \ourname achieves the highest correlation of 0.639, which is +30.14\% better than that of citations.
    Furthermore, we visualize the correspondence of our metric and ToT, and observe a substantial difference between
    the
    \ourname distributions of ToT vs. non-ToT papers
    in \cref{causalcite:fig:ToTViolin}. We also show three examples of ToT papers in \cref{causalcite:fig:totexample}, where the ToT papers differ from the non-ToT papers by one or two orders of magnitude.

\subsection{Topic Invariance of \ourname}
    \begin{table}[ht]
        \small 
        \centering
        \begin{tabular}{lcccccc}
            \toprule
            \textbf{Research Area}
            & \textbf{\pciavg} & \textbf{Citations} & \textbf{SSHI} \\
            \midrule
            General AI (n=16) & 0.748
            & 2,024  & 267\\
            CV (n=36) & 0.734
            & 7,238  & 1,088  \\
            NLP (n=20)    & 0.763 
            & 1,785 & 461    \\ 
            \bottomrule
        \end{tabular}
        \caption{The average of each metric by research area on our collected set of 72 ToT papers.}
\label{causalcite:tab:topicAgnostic}
    \end{table}

A well-known issue with citations is their inconsistency across different fields.
What might be considered a large number of citations in one field might be seen as average in another.
In contrast, we show that our ACI index does not suffer from this issue.
We show this using our ToT dataset, where we control for the quality of the papers to be ToT but vary the domain by the three fields: general AI, CV, and NLP.
We observe in \cref{causalcite:tab:topicAgnostic} that even though some domains have significantly more citations (for instance, CV ToT papers have, on average, $4.05$ times more citations than NLP), the ACI remains consistent across various fields.

\section{Findings}
Having demonstrated the effectiveness of our metrics, we now explore some open-ended questions:
(1) Do best papers have high causal impact? (\cref{causalcite:sec:best})
(2) How does the \ourname value distribute across papers?
(\cref{causalcite:sec:curve}) 
(3) What is the impact of some famous papers evaluated by \ourname? (\cref{causalcite:sec:famous_papers})
(4) Can we use this metric to correct for citations? (\cref{causalcite:sec:outlier}).

\ifperfect \zhijing{Reviewer 2 of ICLR suggests that we have a stronger head-to-head comparison with citations, highlighting (1) criticisms of the existing citation-count system, (2) How does CAUSALCITE solve the claimed limitations.}
\fi

\subsection{Do Best Papers Have High Causal Impact?} \label{causalcite:sec:best}
Selecting best paper awards is an arguably much harder task than ToT papers, as it is difficult to predict of the impact of a paper when it is just newly published.
Therefore, we are interested in the actual causal impact of best papers. Similar to our study on ToT papers, we collect a dataset of {$444$} papers including $74$ best papers and a control set of random $5$ non-best papers from the same conference in the same year, using the same set of the top ten leading AI conferences.
We find that the correlation of the \ourname metric with best papers is $0.348$, which is very low compared to the $0.639$ correlation with the ToT papers.
This shows that the best papers do not necessarily have a high causal impact.
One interpretation can be that the best paper evaluation is a forecasting task, which is much more challenging than the retrospective task of ToT paper selection.

\subsection{What Is the Nature of the \ourname Distribution?}\label{causalcite:sec:curve}
\begin{figure}[ht]
    \centering
    \includegraphics[width=.8\linewidth]{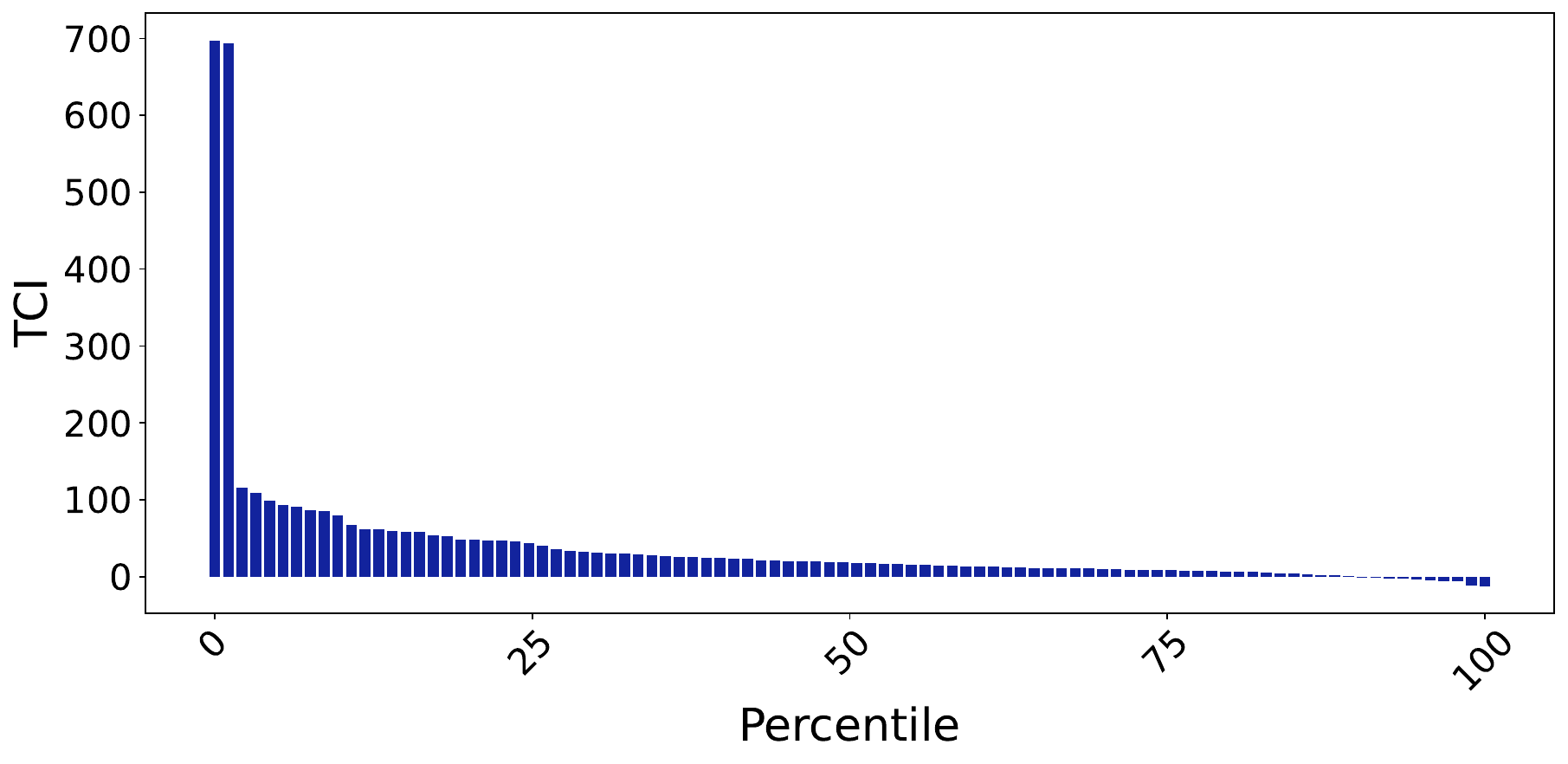}
        \caption{The distribution of TCI values by percentile of 100 random papers, which shows a long tail indicating that high impact is concentrated in a relatively small portion of papers.
        }
    \label{causalcite:fig:curve}
\end{figure}

We explore how the \ourname scores are distributed across papers in general. We plot \cref{causalcite:fig:curve} using a random set of 100 papers from the Semantic Scholar dataset, which is a reasonably large size given the computation budget mentioned in \cref{causalcite:sec:impl}.
From this plot, we can see 
a power law distribution with a long tail, echoing with the common belief that the paper impact follows the power law, with high impact concentrated in a relatively small portion of papers.

\subsection{Selected Paper Case Study} \label{causalcite:sec:famous_papers}
\begin{table}[ht]
    \small 
    \centering
        \begin{tabular}{lccc}
            \toprule
            Paper Name & TCI    & Citations & \pciavg \\ \midrule
            Transformers  & 52,507 & 68,064    & 0.771                         \\
            BERT       & 40,675 & 59,486    & 0.683                         \\
            RoBERTa    & 6,932  & 14,434    & 0.480                         \\
            
            \bottomrule
        \end{tabular}
        \caption{Case study of some selected NLP
        papers.}
        \label{causalcite:tab:empirical}

\end{table}    
In addition to the shape of the overall distribution, we also look at our metric's correspondence to some selected papers shown in
\cref{causalcite:tab:empirical}. For example, we know that the Transformer paper \citep{vaswani2017attention} is a more foundational work than
its follow-up work BERT \citep{devlin-etal-2019-bert}, and BERT is more foundational than its later variant, RoBERTa \citep{liu2019roberta}. This monotonic trend is confirmed in their TCI and ACI values too. Again, this is a preliminary case study, and we welcome future work to cover more papers.

\subsection{Discovering Quality Papers beyond
Citations}
\label{causalcite:sec:outlier}

Another important contribution of our metric is that it can help discover papers that are traditionally overlooked by citations.
To achieve the discovery, we formulate the problem as outlier detection, where we first use a linear projection to handle the trivial alignment of citations and \ourname,
and then analyze the outliers using the interquartile range (IQR) method \citep{outliers}. See the exact calculation in \cref{causalcite:appd:outlier}.
We show the three subsets of papers in \cref{causalcite:tab:outlier}, where the two outlier categories, the overcited and undercited papers, correspond to the false positive and false negative oversight by citations, respectively. 
An additional note is that, when we look into some characteristics of the three categories, we find that the citation frequency in result section, i.e., the percentage of times they are cited in results section compared to all the citations, correlates with these categories.
Specifically, we find that the undercited papers tend to have more of their citations concentrated in the results section, which usually indicates that this paper constitutes an important baseline for a follow-up study, while the overcited papers tend to be cited out of the results section, which tends to imply a less significant citation.

\begin{table}[ht]
\small 
\centering
\begin{tabular}{lcccccc}
\toprule
Paper Category & Result Citations & Residual %
\\ \midrule

Overcited Papers ({7.04}\%) & 1.26 & -1.792 \\
Aligned Papers ({91.20}\%) & 1.51 & 0.118 \\
Undercited Papers ({1.76}\%) & 1.90  & 1.047 \\
\bottomrule
\end{tabular}
\caption{We use our \ourname metric to discover outlier papers that are overlooked by citations. 
For each paper category, we include their portion relative to the entire population, the percentage of citations occurred in the result section (Result Citations), and average residual value by linear regression.
}\label{causalcite:tab:outlier}
\end{table}

\section{Related Work}
    The quantification of scientific impact has a rich history and continuously evolves with technology.
    Bibliometric analysis has been largely influenced by early methods that relied on citation counts \citep{garfield1964use, garfield1972citation, garfield1964science}.
    \citet{hou2017exploration} investigate the evolution of citation analysis, employing reference publication year spectroscopy (RPYS) to trace its historical development in scientometrics.
    \citet{donthu2021conduct} provide practical guidelines for conducting bibliometric analysis, focusing on robust methodologies to analyze scientific data and identify emerging research trends.

    Indices such as the h-index, introduced by \citet{hirsch2005index}, are established tools for measuring research impact.
    The more recent Relative Citation Ratio (RCR), developed by \citet{hutchins2016relative}, provides a field-normalized alternative to traditional metrics.
    \citet{valenzuela2015identifying} introduced SSHI, an approach to identify meaningful citations in scholarly literature.
    However, these metrics are not without limitations.
    As \citet{wroblewska2021research} discussed, conventional citation-based metrics often fail to capture the multidimensional nature of research impact.
    In this context, \citet{elmore2018altmetric} discussed the Altmetric Attention Score, which evaluates the broader societal and online impact of research.
    
    With the increasing availability of large datasets and the advent of digital technologies, new opportunities for bibliometric analysis have emerged.
    \citet{iqbal2021decade} highlighted the role of NLP and machine learning in enhancing in-text citation analysis.
    Similarly, \citet{umer2021scientific} explored the use of textual features and SMOTE resampling techniques in scientific paper citation analysis.
    \citet{jebari2021use} analyzed citation context to detect research topic evolution, showcasing data analysis for scientific discourse.
    \citet{chang2023citesee} explored augmenting citations in scientific papers with historical context, offering a novel perspective on citation analysis.
    \citet{manghi2021new} introduced scientific knowledge graphs, an innovative method for evaluating research impact.
    \citet{bittmann2021applied} explored statistical matching in bibliometrics, discussing its utility and challenges in post-matching analysis.
    The use of AI in bibliometric analysis is highlighted in research by \citet{chubb2022speeding} and the systematic review of AI in information systems by \citet{collins2021artificial}.
    Network analysis approaches, as discussed by \citet{chakraborty2020patent} in the context of patent citations and by \citet{dawson2014current} in learning analytics, further illustrate the diverse applications of advanced methodologies in understanding citation patterns.

\ifperfect

Citet papers:

\textbf{Citation Analysis Techniques and Metrics}
A decade of in-text citation analysis based on natural language processing and machine learning techniques: an overview of empirical studies \citep{iqbal2021decade} \\
How to conduct a bibliometric analysis: An overview and guidelines \citep{donthu2021conduct}\\
Scientific papers citation analysis using textual features and SMOTE resampling techniques \citep{umer2021scientific} \\
Research impact evaluation and academic discourse \citep{wroblewska2021research}\\
New trends in scientific knowledge graphs and research impact assessment \citep{manghi2021new} \\
Exploration into the evolution and historical roots of citation analysis by referenced publication year spectroscopy \citep{hou2017exploration}\\
The use of citation context to detect the evolution of research topics: a large-scale analysis \citep{jebari2021use}\\
Relative citation ratio (RCR): a new metric that uses citation rates to measure influence at the article level \citep{hutchins2016relative}\\
The Altmetric attention score: what does it mean and why should I care? \citep{elmore2018altmetric}\\
Applied usage and performance of statistical matching in bibliometrics: The comparison of milestone and regular papers with multiple measurements of disruptiveness as an empirical example \citep{bittmann2021applied}\\ 
Speeding up to keep up: exploring the use of AI in the research process \citep{chubb2022speeding} \\
Artificial intelligence in information systems research: A systematic literature review and research agenda \citep{collins2021artificial} \\ 
Patent citation network analysis: A perspective from descriptive statistics and ERGMs \citep{chakraborty2020patent}\\ 
Current state and future trends: A citation network analysis of the learning analytics field \citep{dawson2014current} \\
CiteSee: Augmenting Citations in Scientific Papers with Persistent and Personalized Historical Context \citep{chang2023citesee} \\

\subsection{citation analysis}

Citations are important for direct research impact and quality indicator \citep{Aksnes},
And as performance indicators for decisions such as
tenure evaluation, grant assessment, evaluate to the impact of grants.
use of citation indicators in evaluation of the scientific performance of research groups, departments, and institutions (Moed, 2005); evaluation of research proposals (Cabezas-Clavijo, Robinson-Garcia, Escabias, Jimenez-Contreras, 2013); allocation of research funding (Carlsson, 2009); and hiring of academic personnel (Holden, Rosenberg, Barker, 2005). 

Pros and cons:
Arguably, citations might reflect scientific impact and relevance -Aksnes

acknowledged flaws, solidity/plausibility, originality, and societal value

An important study might be underestimated because it was
not published.
Simply counting citations, meanwhile, fails to capture the idea that articles should be judged relative to similar papers: an algebra paper with a few dozen citations, for example, may have a greater impact in mathematics than a widely cited cancer study would have in oncology

https://arxiv.org/abs/2305.12920

\paragraph{Matching, Synthetic Control, ...}

\paragraph{Network Annotation}
network interference (JHU)

\subsection{Different Measures}

Scholar-centric indices:

- Total citations

- h-index

Paper-centric indices:

- Citations

- Highly influential citations

Paper-to-paper indices:

- Whether the paper-to-paper citation is a highly influential citation (binary)

\subsection{Direct Citations}
\subsection{Highly Influential Citations}
\subsection{Shortcomings of Existing Metrics}

\fi
\section{Conclusion}
In this study, we propose \ourname, a novel causal formulation for paper citations. Our method combines traditional causal inference methods with the recent advancement of NLP in LLMs to provide a new causal outlook on paper impact by answering the causal question: ''Had this paper never been published, what would be the impact on this paper’s current follow-up studies?''.
With extensive experiments and analyses using expert ratings and test-of-time papers as criteria for impact, our new \ourname metric demonstrates clear improvements over the traditional citation metrics. Finally, we use this metric to investigate several open-ended questions like ``Do best papers have high causal impact?'', conduct a case study of famous papers, and suggest future usage of our metric for discovering good papers less recognized by citations for the scientific community.

\section*{Limitations and Future Work}
There are several limitations for our work. For example, as mentioned previously, our metric has a high computational budget. %
Future work can explore more efficient optimization methods. Also, we model the content of the paper by its title and abstract, it could also be possible for future work to benefit from modeling the full text, given appropriate license permissions.

As for another limitation, our study is based on data provided by the Semantic Scholar corpus. This corpora has certain properties such as being more comprehensive with computer science papers, but less so in other disciplines. Its citation data also has a delay compared to Google Scholar, so for the newest papers, the citation score may not be accurate, making it more difficult to calculate our metric.

Additionally, our study provides a general framework for causal inference given a causal graph that involves text. It is totally possible that for a more fine-grained problem, the causal graph will change, in which case, we undersuggest future researchers to derive the new backdoor adjustment set, and then adjust the algorithm accordingly. An example of such a variable could be the author information, which might also be a confounder. 

Finally, since quality evaluation of a paper is a multi-faceted task, theoretically, a single number can never give more than a rough approximation, because it collapses multiple dimensions into one and loses information. Our argument in this paper is just to show that our formulation is theoretically more accurate than the citation formulation. We take one step further, instead of solving the quality evaluation problem which is much more nuanced. Some intrinsic problems in citations that we can also not solve (because our metrics still rely on using citations, just contrasting them in the right away) include (1) if a paper is newly published, with zero citations, there is no way to obtain a positive causal index, and (2) we do not solve the fair attribution problem when multiple authors share credit of a paper, as our metric is not sensitive towards authors.

\section*{Ethical Considerations}

\textbf{Data Collection and Privacy}
The data used in this work are all from Open Source Semantic Scholar data, with no user privacy concerns. The potential use of this work is for finding papers that are unique and innovative but do not get enough citations due to lack of popularity or awareness of the field. This metric can act as an aid when deciding impact of papers, but we do not suggest its usage without expert involvement. Through this work, we are not trying to demean or criticize anyone's work we only intend to find more papers that have made a valuable contribution to the field.

\textbf{CS-Centric Perspective}
The authors of this paper work in Computer Science (mostly Machine Learning) hence a lot of analysis done on the quality of papers that required sanity checks are done on ML papers. The conferences selected for doing the ToT evaluation were also CS Top conferences, hence they might have induced some biases. The metric in general has been created generically and should be applicable to other domains as well, the Author Identified Most Influential Papers study is also done on a generalized dataset, but we encourage readers in other disciplines to try out the metric on papers from their field.

\mainchapter{Conclusion \& Future Work}

\renewcommand{\ourmodel}{{\textsc {CausalCoT}}\xspace}

This dissertation explored causal methods for NLP. We first started with the question of whether LLMs can do causal reasoning. In \cref{part:1}, we build formal causal reasoning benchmarks covering two key skills, causal discovery (\cref{ch:corr2cause}) and causal effect reasoning (\cref{ch:cladder}), which existing models struggle to address. We develop \ourmodel to enhance model performance, inspired by a combination of symbolic and LLM-driven steps.
A natural question next is to interpret how models make their decisions, which leads to our exploration in \cref{part:2} to use causal methods for intrinsic (\cref{ch:compmech}) and behavioral interpretability analysis (\cref{ch:mathrobust}). 

Apart from causal inference to improve performance and to interpret the models, we also look into the causality among the variables in the dataset in \cref{part:3}. We find that whether the variables hold a causal or an anticausal relation has large impact on their performance in different settings (\cref{ch:icm}). We further extend this exploration to cases where the variable causal relations are not evident \textit{a priori}, but discovered using interdisciplinary insights (\cref{ch:psychcausal}). Based on the variable relations we also suggest causal prompts to make use of those relations to improve LLMs' decisions.

Lastly, we also demonstrate social applications using causal methods and NLP together in \cref{part:4}. Our first study utilizes NLP for sentiment classification on social media text, and perform causal effect estimation between sentiment and social policies (\cref{ch:covidtwitter}). Our second study looks into causal analysis for paper citations, which is a more technically challenging case to adjust for textual confounders, where we use LLMs to encode the text to high-dimensional vector space, and perform a text-based matching algorithm (\cref{ch:causalcite}).

Building on the insights gained from all the above research, we outline a few open challenges and future directions:

The first direction to highlight is to build a more comprehensive framework of causal reasoning. My previous work distinguishes two types of causal understanding in LLMs: knowledge-based causal understanding \citep{cui2024odyssey}, and knowledge-independent formal reasoning \citep{jin2023cladder,jin2024large}. Looking ahead, there is a large need to systematize reasoning as an interplay of both. To achieve it, one way is to develop more sophisticated models and training strategies that enhance the causal inference abilities of LLMs. This includes exploring novel architectures, fine-tuning methods, and datasets that better capture causal relationships.
Another way is to advance tool-augmented LLMs connecting different subskills into an overall pipeline.

Since reasoning does not necessarily need to be constrained to passively processing the text, it is also a promising direction to explore cause of reasoning in interactive systems, such as conversational agents and decision support tools. Research in this area can focus on integrating causal inference with real-time interactions and feedback mechanisms.
In addition to the interaction setup, some other modalities of data can also be included as inputs for causal inference, which can generalize text-only reasoning to image, video, audio, structured information, and many others.

In parallel to improving model performance, we can also advance the use of causal methods for interpretable, robust, and fair NLP models. Interpretability, at its essence, is a causal discovery problem, where we aim to understand what elements or mechanisms in the model contribute to the final prediction. The key technical challenges are to scale up causal inference to a large number (e.g., trillions) of neurons in the model, and to obtain human-understandable high-level interpretation. For robustness and fairness, my existing work uses SCMs to formulate them as causal alignment, between the desired decision-making mechanism, and the model-learned mechanisms. Further work can formalize more evaluation pipelines in causal terms, and apply our framework for robustness and fairness assessments.

Finally, with all the technological advancements in LLMs, we are in an era to see the rise of cross-disciplinary applications to connect Causal NLP with fields such as healthcare, economics, and education. We look forward to collaborative research that leverages domain-specific knowledge with Causal NLP methods, leading to impactful advancements in these areas.

If these efforts above are successful, they could lead to the development of more intelligent and robust NLP systems capable of sophisticated causal reasoning. Such advancements have the potential to transform numerous fields. I believe that the continuous improvement and application of causal methods for LLMs will be a critical driver of progress in AI, inspiring future research and opening new avenues for interdisciplinary collaboration.

\newpage
\thispagestyle{empty}
\mbox{}
\newpage

\appendix
\chapter{Appendix}
\section{Additional Materials for \cref{ch:corr2cause}}

\subsection{Implementation Details}\label{corr2cause:appd:implementation}
When finetuning on our data,  for GPT-based models, we use the default settings of the OpenAI finetuning API; and for BERT-based models, 
we use the \texttt{transformers} library \citep{wolf-etal-2020-transformers} and train the models on a server with an NVIDIA Tesla A100 GPU with 40G of memory. To fit for the GPU memory, we set the batch size to be 8. We use the validation set to tune the learning rate, which takes value in \{2e-6,  5e-6, 1e-5, 2e-5, 5e-5\}; dropout rate, which takes value in \{0, 0.1, 0.2, 0.3\}; and weight decay, which takes value in \{1e-4, 1e-5\}. We train the models until convergence, which is usually around ten epochs.

\paragraph{Prompts}
When querying the autoregressive LLMs, we formulate the prompt as follows:

\textit{Question: \texttt{[premise]}}

\textit{Can we deduct the following: \texttt{[hypothesis]}? Just answer "Yes" or "No."}

\textit{Answer:}

\subsection{Generating Natural Stories}\label{corr2cause:appd:story}

To generate the natural stories based on our symbolic expressions, we utilize the state-of-the-art LLM, GPT-4, which is very good at story generation. We design detailed instructions in the prompt, and generate around 200 stories in our case study. We show two examples stories in \cref{corr2cause:tab:story}, and the report the overall statistics in \cref{corr2cause:tab:story_stats}.
\begin{table}[ht]
    \centering
    \begin{tabular}{p{1cm}p{6cm}p{6cm}l}
    \toprule
& Example 1 (Label=Negative) & Example 2 (Label=Positive) \\ \midrule
Symbolic Form        
& \textit{Premise:} Suppose there is a closed system of 2 variables, A and B. All the statistical relations among these 2 variables are as follows: A correlates with B. 
\newline
\textit{Hypothesis:} A directly affects B.  
\newline
\textit{Relation between the promise and hypothesis:} The premise does not a necessary condition for the hypothesis.
& \textit{Premise:} Suppose there is a closed system of 3 variables, A, B and C. All the statistical relations among these 3 variables are as follows: A correlates with C. B correlates with C. However, A is independent of B. 
\newline
\textit{Hypothesis:} A directly affects C.
\newline
\textit{Relation between the promise and hypothesis: }The premise is a necessary condition for the hypothesis. So if the premise is true, the hypothesis must be true. \\ \hline
Natural Story         
& \textit{Premise:} Suppose there is a closed system of 2 variables, ice cream sales and swimming pool attendance. All the statistical relations among these 2 variables are as follows: ice cream sales correlate with swimming pool attendance.
\newline
\textit{Hypothesis:} Ice cream sales directly affect swimming pool attendance.
\newline
\textit{Relation between the premise and hypothesis:} The premise does not provide a necessary condition for the hypothesis. The correlation between ice cream sales and swimming pool attendance could be due to a third variable, such as hot weather, which increases both ice cream sales and swimming pool attendance. Therefore, it is not necessarily true that ice cream sales directly affect swimming pool attendance.
& 
\textit{Premise:} Let's consider three factors: eating junk food (A), obesity (C), and watching television (B). There is a correlation between eating junk food and obesity, and between watching television and obesity. However, eating junk food and watching television are independent from each other.
\newline
\textit{Hypothesis:} Eating junk food directly affects obesity.
\newline
\textit{Relation between the premise and hypothesis:} The premise provides the necessary conditions for the hypothesis. It establishes the independent variables A (eating junk food) and B (watching television) and their correlations with obesity. Given that these are true, it supports the hypothesis that eating junk food directly affects obesity.
\\
         \bottomrule
    \end{tabular}
    \caption{Examples of natural stories generated based on the symbolic form in our \ourdata dataset, showing the broad application value of our dataset as the starting point for various verbalizations of the correlation-to-causation inference task.}
    \label{corr2cause:tab:story}
\end{table}
\begin{table}[ht]
    \centering\small
    \begin{tabular}{lcccccc}
    \toprule
Test Set Size  & 102 
\\
Dev Set Size & 102 
\\
\# Tokens/Premise & 64.88 
\\
\# Tokens/Hypothesis & 13.54 
\\
\# Tokens/Explanation & 64.66 
\\
\% Positive Labels & 1.67 
\\
\bottomrule
    \end{tabular}
    \caption{Statistics of our generated natural stories. We report the number of samples in the test and development sets; number of tokens per premise (\# Tokens/Premise), hypothesis (\# Tokens/Hypothesis), and explanation (\# Tokens/Explanation); and percentage of the positive labels (\% Positive Labels).}\label{corr2cause:tab:story_stats}
\end{table}

For more information, the exact prompt we use is ``\textit{Here is a causal inference rule: \texttt{[symbolic form]} Please provide a real-world example instantiating this phenomenon. Format it also as "Premise:", "Hypothesis:", and "Relation between the promise and hypothesis:".}''

\subsection{Templates and Paraphrases}\label{corr2cause:appd:templates}
We use the verbalization templates in \cref{corr2cause:tab:paraphrase} to compose the hypotheses for all six causal relations.
\begin{table}[ht]
    \centering \small
    \begin{tabular}{ll}
\toprule
\textbf{Causal Relation} & \textbf{Hypothesis Template} \\\midrule
Is-Parent & \texttt{\{Var i\}} directly causes \texttt{\{Var j\}}. \\
Is-Ancestor & \texttt{\{Var i\}} causes something else which causes \texttt{\{Var j\}}. \\
Is-Child & \texttt{\{Var j\}} directly causes \texttt{\{Var i\}}. \\
Is-Descendant & \texttt{\{Var j\}} is a cause for \texttt{\{Var i\}}, but not a direct one. \\
Has-Collider & There exists at least one collider (i.e., common effect) of \texttt{\{Var i\}} and \texttt{\{Var j\}}. \\
Has-Confounder & There exists at least one confounder (i.e., common cause) of \texttt{\{Var i\}} and \texttt{\{Var j\}}. \\
\midrule
\textbf{\textit{Paraphrases}} \\

Is-Parent & \texttt{\{Var i\}} directly affects \texttt{\{Var j\}}. \\
Is-Ancestor & \texttt{\{Var i\}} influences \texttt{\{Var j\}} through some mediator(s). \\
Is-Child & \texttt{\{Var j\}} directly affects \texttt{\{Var i\}}. \\
Is-Descendant & \texttt{\{Var j\}} influences \texttt{\{Var i\}} through some mediator(s). \\
Has-Collider & \texttt{\{Var i\}} and \texttt{\{Var j\}} together cause some other variable(s). \\
Has-Confounder & Some variable(s) cause(s) both \texttt{\{Var i\}} and \texttt{\{Var j\}}. \\
\bottomrule
    \end{tabular}
    \caption{Templates and their paraphrases for each causal relation in the hypothesis. We use \texttt{\{Var i\}} and \texttt{\{Var j\}} as placeholders for the two variables.}
    \label{corr2cause:tab:paraphrase}
\end{table}

\subsection{Change Log for the Dataset Version Update}\label{corr2cause:appd:version}

\begin{table}[ht]
    \centering \small
    \begin{tabular}{lp{3.3cm}llll}
    \toprule
    Two Equivalent Forms & Duplication Property   &  De-Duplication Method \\ \midrule
\tworowbrace { }     Is-Parent(\texttt{i}, \texttt{j}) & \multirow{2}{*}{Two exact same strings} & \multirow{2}{*}{Keep only one, by forcing $i<j$} \\
{ }{ } Is-Child(\texttt{j}, \texttt{i})\\
 
\tworowbrace { }     Is-Ancestor(\texttt{i}, \texttt{j}) (Original)  & Two different strings, but  & \multirow{2}{*}{Randomly sample one out of the two} \\
{ }{ } Is-Descendent(\texttt{j}, \texttt{i}) (Original)  &  semantically equivalent    \\
 
\tworowbrace { }         Is-Ancestor(\texttt{i}, \texttt{j}) (Paraphrased) & \multirow{2}{*}{Two exact same strings} & \multirow{2}{*}{Keep only one, by forcing $i<j$}\\
{ }{ } Is-Descendent(\texttt{j}, \texttt{i}) (Paraphrased) \\

\tworowbrace { }     Has-Collider(\texttt{i}, \texttt{j})  & Two different strings, but  & \multirow{2}{*}{Randomly sample one out of the two} \\
{ }{ } Has-Collider(\texttt{j}, \texttt{i})  &  semantically equivalent    \\
 
\tworowbrace { }     Has-Confounder(\texttt{i}, \texttt{j})  & Two different strings, but  & \multirow{2}{*}{Randomly sample one out of the two} \\
{ }{ } Has-Confounder(\texttt{j}, \texttt{i})  &  semantically equivalent    \\
    \bottomrule
    \end{tabular}
    
    \caption{De-duplication methods for the six causal relation types and their verbalizations.}
    \label{corr2cause:tab:version}
\end{table}
\paragraph{De-Duplication Strategy}
As mentioned in \cref{corr2cause:sec:stats} in the main paper, our original dataset (v1.0) has duplication due to symmetric relations and verbalizations. 
We introduce in \cref{corr2cause:tab:version} several reasons for why duplicated hypotheses exist in our original data. One typical reason is symmetric relations such as Is-Parent(A, B) and Is-Child(B, A), and, similarly, the paraphrased version of Is-Ancestor(A, B) and Is-Descendent(B, A).
Another typical reason is the semantic equivalence in the verbalization templates, which applies to the Has-Collider and Has-Confounder relations. For example, the verbalized texts of Has-Collider(A, B) and Collider(B, A) are ``There exists at least one collider (i.e., common effect) of \{A and B, B and A\},'' respectively, which are semantically-equivalent paraphrases of each other, so we randomly keep one out of the two.

\paragraph{Resulting Dataset Statistics after De-Duplication}

Since the reason for duplication in the first place is due to symmetry in the causal relation, or verbalization, the resulting new data, \ourdata v2.0, is exactly a half of the original data. As we reported previously in \cref{corr2cause:tab:stats} of \cref{corr2cause:sec:stats}, the total number of samples cuts down to half, while the label distribution and all other properties are the same. To compose each split, we apply the same de-duplication method for the test, train, and development sets. We notice that some duplicates are across the splits, so we prioritize keeping the test and training sets untouched (to minimally affect the experimental results), and then reduce the development set by removing the cross-split duplicates, namely:
\begin{itemize}
    \item test\_2.0 = deduplicate(test\_1.0)
    \item train\_2.0 = deduplicate(train\_1.0)
    \item dev\_2.0 = deduplicate(dev\_1.0) $\backslash$ \{test\_2.0, train\_2.0\}
\end{itemize}
We expect minimal or almost no change to the experimental results. In case of the slight possibility that this change in the development set might affect the model selection in the training process, future work can feel free to re-train the models and update the exact performance number.

\subsection{Spurious Correlation Analysis}\label{corr2cause:appd:spurious}
The inspirations of our two robustness tests (paraphrasing and variable refactorization) come from our data analysis. 
We check for spurious correlations in the data by reporting in \cref{corr2cause:tab:ngram} the point-wise mutual information (PMI) between the label and any n-gram with no more than four tokens.
In addition, we also report the difference of the PMI with the two labels in the $|\text{Diff}|$ column of \cref{corr2cause:tab:ngram}, and report the top 10 n-grams.

The design spirit for our robustness test is that if
the models' correct judgment relies on exploiting these spurious correlations, then such reliance will be broken in our perturbations.

\begin{table}[ht]
    \centering \small
\begin{tabular}{lcccc}
\toprule
N-Gram &  PMI w/ Non-Ent. Label &  PMI w/ Ent. Label &    $|\text{Diff}|$ \\
\midrule
a cause            &  1.692209 & -1.025611 &  2.717820 \\
a cause for        &  1.663640 & -0.983790 &  2.647430 \\
A causes           &  1.640679 & -0.951610 &  2.592289 \\
A causes something &  1.621820 & -0.926075 &  2.547895 \\
a direct           &  1.606052 & -0.905316 &  2.511369 \\
a direct one       &  1.592673 & -0.888107 &  2.480781 \\
for D              &  1.584826 & -0.878180 &  2.463006 \\
for D but          &  1.583897 & -0.877014 &  2.460911 \\
for E              &  1.582980 & -0.875864 &  2.458844 \\
for E but          &  1.582074 & -0.874728 &  2.456802 \\

\bottomrule
\end{tabular}

    \caption{PMI between the labels and n-grams. The labels include non-entailment (Non-Ent.) and entailment (Ent.). And the n-grams include all with no more than four words. The $|\text{Diff}|$ column shows the absolute value of the difference between the PMIs with two labels.
    We show the top 10 n-grams with the largest differences of their PMIs with the two classes in the $|\text{Diff}|$ column.
    }
    \label{corr2cause:tab:ngram}
\end{table}

We can see that some spurious correlations are rooted in the framing of the hypothesis, such as ``a cause (for)'', and ``a direct (one)'' (which we use the paraphrasing task to break), and others are connected to the variable names, such as ``for D (but)'' and ``for E (but)'' (which we use the variable refactorization to break).

\subsection{Fine-Grained Error Analysis}\label{corr2cause:appd:error}
In addition to the fine-grained analysis by causal relation type in \cref{corr2cause:tab:finetune_by_class} for fine-tuned models, we also report such error analysis for non-finetuned models in \cref{corr2cause:tab:by_class}.
\begin{table}[ht]
\centering \small
\begin{tabular}{llcccccccc}
\toprule
{Selected Models}      & {Relation Type} & {F1} & {Precision} & {Recall} & {Accuracy} \\ \midrule
GPT-3.5             & All                    & 21.69       & 17.79      & 27.78      & 69.46        \\
GPT-3.5             & Is-Parent              & 8.82        & 100        & 4.62       & 83.47        \\
GPT-3.5             & Is-Ancestor            & 0           & 0          & 0          & 90.67        \\
GPT-3.5             & Is-Child               & 9.84        & 100        & 5.17       & 85.33        \\
GPT-3.5             & Is-Descendant          & 14.29       & 11.9       & 17.86      & 84           \\
GPT-3.5             & Has-Collider           & 34.24       & 25.51      & 52.07      & 35.12        \\
GPT-3.5             & Has-Confounder         & 15.33       & 8.86       & 56.76      & 37.8         \\ \hline
GPT-4               & All                    & 29.08       & 20.92      & 47.66      & 64.6         \\
GPT-4               & Is-Parent              & 0           & 0          & 0          & 82.67        \\
GPT-4               & Is-Ancestor            & 30.77       & 31.25      & 30.3       & 88           \\
GPT-4               & Is-Child               & 0           & 0          & 0          & 84.53        \\
GPT-4               & Is-Descendant          & 26.98       & 17.35      & 60.71      & 75.47        \\
GPT-4               & Has-Collider           & 44.1        & 30.18      & 81.82      & 32.71        \\
GPT-4               & Has-Confounder         & 20.67       & 11.53      & 100        & 23.86        \\ \hline
RoBERTa MNLI        & All                    & 22.79       & 34.73      & 16.96      & 82.5         \\
RoBERTa MNLI        & Is-Parent              & 0           & 0          & 0          & 82.67        \\
RoBERTa MNLI        & Is-Ancestor            & 0           & 0          & 0          & 91.2         \\
RoBERTa MNLI        & Is-Child               & 0           & 0          & 0          & 84.53        \\
RoBERTa MNLI        & Is-Descendant          & 0           & 0          & 0          & 92.53        \\
RoBERTa MNLI        & Has-Collider           & 43.45       & 39.73      & 47.93      & 59.52        \\
RoBERTa MNLI        & Has-Confounder         & 0           & 0          & 0          & 84.45        \\ \bottomrule
\end{tabular}
\caption{Fine-grained evaluation results for some selected non-fine-tuned models.}
\label{corr2cause:tab:by_class}
\end{table}

These results are particularly revealing, showing how off-the-shelf models perform in recognizing specific relations. Specifically, GPT-3.5 cannot recognize ancestor relations, whereas GPT-4 fails at all direct causation recognition with parents and children. And RoBERTa MNLI only did collider relation relatively correctly. Note that, when the F1 score is zero, the accuracy number is a result of always predicting the negative class of that relation.

\subsection{LLM Performance Optimization}
\label{corr2cause:appd:optimization}

Since our experiments in \cref{corr2cause:sec:0shot} are based on plain, zero-shot prompts, we explore whether better prompting strategies could improve the performance. We enhance the query prompt by incorporating several strategies: (1) Utilizing a system prompt that specifies the model's expertise (``You are a highly intelligent question-answering bot with profound knowledge of causal inference.''); (2) Including a pair of few-shot examples, one positive and one negative; (3) Implementing chain-of-thought prompting with ``Let's think step by step.'' to encourage the language model to generate step-by-step reasoning. In \cref{corr2cause:tab:optimize_res}, we present the evaluation results on the relatively affordable model, GPT-3.5, where the optimized prompt leads to a 4-point improvement in F1 over the original performance. However, we can see that despite the deployment of all three strategies, the model continues to struggle with this challenging task.

\begin{table}[ht]
    \centering \small
    \begin{tabular}{lccccc}
    \toprule
    & F1 & Precision & Recall
    & Accuracy\\ \midrule
    GPT-3.5 (plain query; original) & 21.69 & 17.79 & 27.78 & 69.46 \\

    GPT-3.5 (enhanced query) & 25.44 & 17.29 & 48.11 & 52.01 \\
    
    \bottomrule
    \end{tabular}
    \caption{Performance of GPT-3.5 with different queries. We quote the original performance from \cref{corr2cause:tab:res}.
    }
    \label{corr2cause:tab:optimize_res}
\end{table}

\newpage

\section{Additional Materials for \cref{ch:cladder}}
\subsection{Supplementary for Dataset Generation}

\subsubsection{List of References for Causal Inference}
\label{cladder:appd:books}
When collecting the causal graphs, query types, and commonsensical stories for our dataset, we took our examples from the following books (sorted by year):
\begin{enumerate}
    \item Causality \citep{pearl2009causality}
    \item Causal inference in statistics: {A} Primer \citep{glymour2016causal}
    \item Elements of Causal Inference \citep{peters2017elements}
    \item The Book of Why \citep{pearl2018book}
    \item Introduction to Causal Inference \citep{neal2020introduction}
    
\end{enumerate}

And the following papers:
\begin{enumerate}
    \item Causes and Explanations: A Structural-Model Approach. Part I: Causes \citep{halpern2005causespartI}
    \item Causes and Explanations: A Structural-Model Approach. Part II: Explanations \citep{halpern2005causespartII}
    \item Causality and Counterfactuals in the Situation Calculus \citep{hopkins2007situationcalculus}
    \item Causal inference in statistics: An overview \citep{pearl2009overview}

\end{enumerate}
\begin{figure}[ht]
    \centering
    \includegraphics[width=0.6\textwidth]{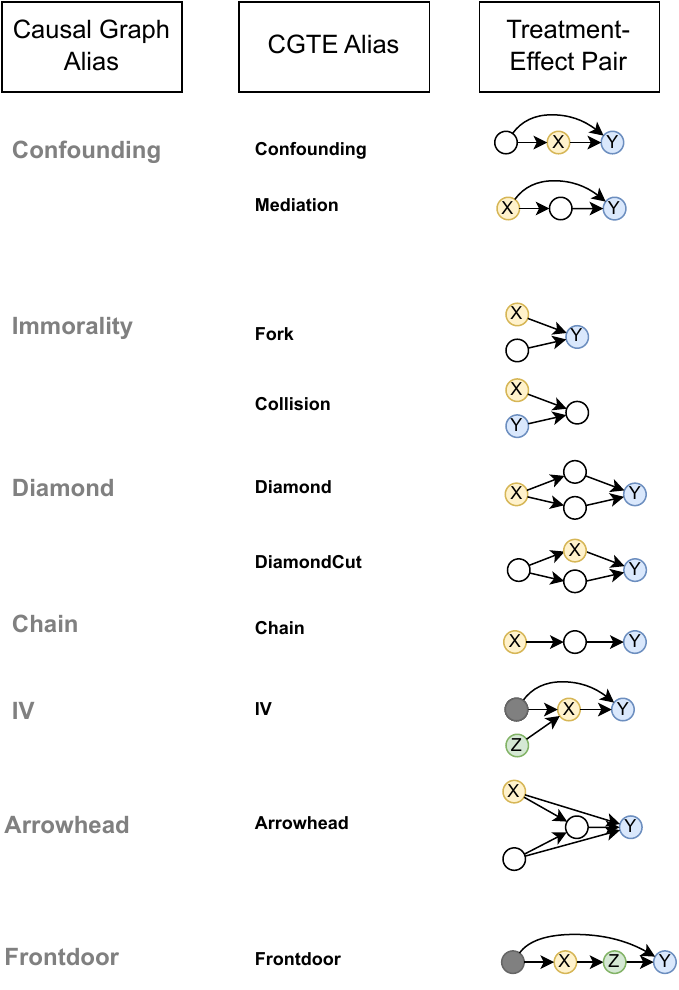}
    \caption{List of all ten causal graphs with treatment-effect pairs (CGTEs). We omit CGTEs that trivially resemble existing ones.}
    \label{cladder:fig:causal_graphs}
\end{figure}
\subsubsection{Formulation of the Query Types}\label{cladder:appd:query_type2estimand}
Here, we introduce all the query types included in our dataset.

\paragraph{Rung-1 Queries: Marginal and Conditional Probabilities.} For marginal probabilities, we ask questions about the overall distribution of a variable. %
For conditional probabilities, we ask whether conditioning on one variable increases or decreases the likelihood of another variable. %
For the explaining away questions, we condition on a collider node and ask how that affects the correlation between the two parents.

\paragraph{Rung-2 Queries: ATE and Adjustment Set.}
For ATE questions, we ask whether the treatment ($X=1$) increases or decreases the likelihood of the effect variable $Y=y$.
For adjustment set questions, we ask whether a set of variables should be adjusted for when estimating the causal effect between treatment and effect. By adjusting, we aim to blocked the non-causal paths from the treatments to effect, and hence eliminate spurious correlation. For example, to query whether the set {gender} is an adjustment set for the effect of a treatment on recovery, we ask \textit{"To estimate the effect of the treatment on recovery, should we directly look at how the treatment correlates with recovery, or should we look at gender-specific correlation?"}
In the collider bias questions, similarly to the explaining away questions, we condition on a collider variable and ask about how an intervention on one of the parents (treatment $X$) affects the other parent (outcome $Y$). However since by construction $X$ and $Y$ do not have common causes, the answer to this question is always ``no''.

\paragraph{Rung-3 Queries: Counterfactual Probability, ATT, NDE, and NIE.}
For counterfactual probability, we ask about what would have been the likelihood of $Y=y$, if the treatment variable $X$ had been $x$, given sufficient evidence $e$ such that the query is identifiable. For ATT, we ask how the likelihood of $Y=y$ would change for those who received treatment ($X=1$) if there had been no treatment ($X=0$). For NDE, we ask whether the $X=1$ directly increases or decreases the likelihood of the $Y=y$, not through any mediators. For NIE, we ask whether the treatment (setting $X=1$) increases or decreases the likelihood of $Y=y$ through mediators, not directly.

\subsubsection{Collection of Causal Graphs}\label{cladder:appd:causal_graphs}
We include all the ten causal graphs with treatment-effect pairs (CGTEs) in \cref{cladder:fig:causal_graphs}.

Note that one causal graph can have several different CGTEs, such as the confounding structure, which has three CGTEs: confounding, mediation, and collision in the triangle form. To generate all the causal graphs and CGTEs here, 
we iterate all commonly used ones within four nodes in the CI books, and omit CGTEs whose solution by CI methods trivially resembles existing ones.

\subsubsection{Data Coverage}
\label{cladder:appd:querycoverage}

Starting from the full set of 12 distinct causal graphs and 10 query types, there are a few combinations that must be omitted as the ground truth answer would be trivial or ill-defined. For example, in the ``Immorality'' graph, the treatment ``X'' and outcome ``Y'' are by construction statistically independent, so there correlation is necessarily 0. Similarly, there are several graphs where certain causal queries are ill-defined or don't make sense to ask. Specifically: 

\begin{enumerate}

    \item For the Natural Direct Effect, we only include questions on the ``IV'', ``Arrowhead'', ``Confounding'', ``Mediation'' and ``DiamondCut'' graphs. 
    \item For the Natural Indirect Effect, we only include questions on the ``Mediation'', ``Frontdoor'', ``Arrowhead'', ``Diamond'' and ``Chain'' graphs.
    \item For the Collider Bias and Explaining Away effect, we only include questions on the ``Collision'' graph.

    \item For the Average Treatment Effect, we include questions on all graphs except ``Collision''.
    \item For the (deterministic) Counterfactuals, we include questions on all graphs except ``Collision''.
    \item For the Average Treatment Effect on the Treated (ATT), we include questions on all graphs except ``Collision'' and ``IV''.

\end{enumerate}

The ``balanced'' benchmark (main benchmark in v1.5), containing 10,112 questions split between all stories, graphs, query types, and commonsensicalness, is balanced such that there are roughly the same number of questions for each distinct story-graph-query combination (ranging from 50-100 per combination) across the different variants: commonsense, anticommonsense, and nonsense. Furthermore, we balance the distribution of correct answers so that there are the same number of ``yes''s and ``no''s.

The ``aggregate'' variant (main benchmark in v1.0) contains 10,560 questions and is primarily balanced across all stories. However since the number of stories for each variant (commonsense, anticommonsense, and nonsense) varies significantly, the results in an unbalanced benchmark in terms of sensicalness.

\subsubsection{Query Form and Text Templates}\label{cladder:appd:query}

We provide in \cref{cladder:tab:query} the text templates we use for each query type.
\begin{table}[ht]
    \centering \small
    \begin{tabular}{p{2cm}p{2.4cm}p{8cm}llll}
\toprule
Query Type & Symbolic Expression & Natural Language Question Template
\\ \midrule
\multicolumn{3}{l}{\textbf{\textit{Rung 1: Association}}} \\
Marg. Prob. & $P(Y)$ & Is the overall likelihood of \texttt{\{$v_{\mathrm{noun}}(X=1)$\}} greater than chance?\\
Cond. Prob. & $P(Y|X)$ & Is the chance of \texttt{\{$v_{\mathrm{noun}}(Y=1)$\}} larger when observing \texttt{\{$v_{\mathrm{noun}}(X=1)$\}}? \\
\midrule
\multicolumn{3}{l}{\textbf{\textit{Rung 2: Intervention}}} \\
ATE & $\mathbb{E}[Y|do(\mathrm{X=1})]-\mathbb{E}[Y|do(\mathrm{X=0})]$ & Will \texttt{\{$v_{\mathrm{noun}}(X=1)$\}} increase the chance of \texttt{\{$v_{\mathrm{noun}}(Y=1)$\}}? \\
Adjust. Set & If $\bm{S}$ opens a backdoor path & To understand how \texttt{\{$v_{\mathrm{overall}}(X)$\}} affects \texttt{\{$v_{\mathrm{overall}}(Y=1)$\}}, should we look directly at how \texttt{\{$v_{\mathrm{overall}}(X)$\}} correlates with \texttt{\{$v_{\mathrm{overall}}(Y)$\}} in general, or this correlation case by case according to \texttt{\{$v_{\mathrm{overall}}(\bm{S})$\}}? \\
\midrule
\multicolumn{3}{l}{\textbf{\textit{Rung 3: Counterfactuals}}} \\
Counterf. Prob. & $P(Y_x=y) $ & Can we infer that \texttt{\{$v_{\mathrm{sent}}(Y=1)$\}} had it been that \texttt{\{$v_{\mathrm{cond}}(X=1)$\}} instead of X=0? \\
ATT & $\mathbb{E}[Y_1-Y_0|\mathrm{X=1}]$ & For \texttt{\{$v_{\mathrm{attr}}(X=1)$\}}, would it be more likely to see \texttt{\{$v_{\mathrm{noun}}(Y=1)$\}} \texttt{\{$v_{\mathrm{cond}}(X=0)$\}}?\\
NDE & $\mathbb{E}[Y_{1,M_0}-Y_{1,M_0}]$ & If we disregard the mediation effect through \texttt{\{$v_{\mathrm{overall}}(Y=1)$\}}, would \texttt{\{$v_{\mathrm{noun}}(X=1)$\}} still positively affect \texttt{\{$v_{\mathrm{noun}}(Y=1)$\}}?
\\
NIE & $\mathbb{E}[Y_{0,M_1}-Y_{0,M_0}]$ & Does \texttt{\{$v_{\mathrm{overall}}(X)$\}} affect \texttt{\{$v_{\mathrm{overall}}(Y)$\}} through \texttt{\{$v_{\mathrm{overall}}(\mathrm{OtherVars})$\}}?
\\
\bottomrule
    \end{tabular}
    \caption{Example natural language templates for each query type.}
    \label{cladder:tab:query}
\end{table}

\subsubsection{Nonsensical Stories}\label{cladder:appd:stories}
To come up with a collection of nonsensical variable names, we use GPT-4 to generate some meaningless words.
Specifically, we use the prompt: ``Create 100 non-existent words that are short, i.e., within 5-characters.'', with temperature=0 with the OpenAI interface.
The collection of nonsensical words we later use as variable names are as follows: ziblo, truq, fyze, glimx, jorv, wexi, snov, yupt, kraz, qixy, vubr, chiz, pliv, moxa, fygo, rukz, tasp, xevo, jyke, wibl, zorf, quzy, nyrp, gwex, smez, vytz, hupx, cwoj, lirf, ovka, pexu, yigz, twaz, kwox, zuph, fraq, jyxo, swoy, uvzi, nekl, gyzp, rixq, vwem, xyfu, blyz, qwip, zeku, tijv, yomx, hwaz, czix, plof, muvy, fyqo, rujz, tasb, xevi, jyka, wibm, zorx, quzw, nyro, gwet, smeu, vyta, hupz, cwoi, lirg, ovki, pexy, yigw, twac, kwoz, zupj, fraq, jyxi, swoq, uvzo, nekm, gyzl, rixw, vwen, xyfo, blyx, qwiu, zeky, tijw, yomz, hwax, czir, ploz, muvq, fyqi, rujx, tasn, xevu, jyko, wibp, zory, and quzt.

\subsubsection{Anti-Commonsensical Stories} \label{cladder:appd:anti}

For the anti-commonsensical stories, we randomly do one of the actions: 

\begin{enumerate}
    \item Replace the effect variable $Y$ with an attribute that would not be an effect variable in any of the stories. Such replacement variables include: ``lip thickness'', ``earthquakes'', ``lactose intolerance'', ``rainfall'', ``is allergic to peanuts'', ``brown eyes'', ``curly hair'', ``black hair'', ``foot size'', ``freckles''
    \item Create an irrelevant treatment variable $X$ that does not play a causal role in any of our commonsensical stories. Such as: ``can swim'', ``is religious'', ``has a brother'', ``has visited England'', ``likes spicy food'', ``is vegetarian'', ``speaks english'', ``drinks coffee'', ``plays card games'', ``listens to jazz'', ``solar eclipse'', ``has a sister'', ``full moon''
\end{enumerate}

To transform a commonsensical story into an anti-commonsensical story, we apply one of these replacements sampled uniformly, resulting in stories such as:

\begin{itemize}
    \item Ability to swim has a direct effect on studying habit and exam score. Studying habit has a direct effect on exam score.
    \item Gender has a direct effect on department competitiveness and peanut allergy. Department competitiveness has a direct effect on peanut allergy.
    \item Liking spicy food has a direct effect on relationship status. Appearance has a direct effect on relationship status.
    \item Playing card games has a direct effect on diabetes and lifespan. Smoking has a direct effect on diabetes and lifespan. Diabetes has a direct effect on lifespan. Smoking is unobserved.
\end{itemize}

For a full list of the replacements and how the replacements are made, check out the code.

\subsubsection{Explanation Template}\label{cladder:appd:expl}
Step \stepone{} Extract the causal graph: The causal graph expressed in the context is: "$\mathcal{G}$". \\
\\
Step \steptwo{} Identify the query type: The query type of the above question is "$\mathrm{query\_type}$".\\
\\
Step \stepthree{} Formulate  the query to its symbolic form: The formal form of the query is "$\mathrm{symbolic\_expression}$".\\
\\
Step \stepfour{} Collect all the available data: The available data are: "$\bm{d}$". \\
\\
Step \stepfive{} Derive the estimand: Based on the graph structure and causal query, the question can be simplified into estimand "$\mathrm{est}$".\\ 
\\
Step \stepsix{} Solve for the estimand: Plug in the available data "$\bm{d}$" into "$\mathrm{est}$". \\
$\mathrm{est}(\bm{d})$ \\
$\approx \mathrm{float}(a)$ \\
\\
Since the estimate for the estimand is $\mathrm{float}(a)$, the overall answer to the question is $\mathrm{bool}(a)$. \\
\subsection{Experimental Details}
\subsubsection{\ourmodel Prompt}\label{cladder:appd:prompt}

Q: \texttt{[question from the dataset]}

Guidance: Address the question by following the steps below:

Step 1) Extract the causal graph: Identify the causal graph that depicts the relationships in the scenario. The diagram should simply consist of edges denoted in "var1 -> var2" format, separated by commas.

Step 2) Determine the query type: Identify the type of query implied by the main question. Choices include "marginal probability", "conditional probability", "explaining away effect", "backdoor adjustment set", "average treatment effect", "collider bias", "normal counterfactual question", "average treatment effect on treated", "natural direct effect" or "natural indirect effect". Your answer should only be a term from the list above, enclosed in quotation marks.

Step 3) Formalize the query: Translate the query into its formal mathematical expression based on its type, utilizing the "do(·)" notation or counterfactual notations as needed.

Step 4) Gather all relevant data: Extract all the available data. Your answer should contain nothing but marginal probabilities and conditional probabilities in the form "P(...)=..." or "P(...|...)=...", each probability being separated by a semicolon. Stick to the previously mentioned denotations for the variables.

Step 5) Deduce the estimand using causal inference: Given all the information above, deduce the estimand using skills such as do-calculus, counterfactual prediction, and the basics of probabilities. Answer step by step.

Step 6) Calculate the estimand: Insert the relevant data in Step 4 into the estimand, perform basic arithmetic calculations, and derive the final answer. There is an identifiable answer. Answer step by step.

A: \texttt{[LLM previous response]}

Q: Based on all the reasoning above, output one word to answer the initial question with just "Yes" or "No".

A: \texttt{[LLM final answer]}
\subsection{Additional Technical Background for Preliminaries}
\label{cladder:app:more_preliminaries}

\subsubsection{Graphical Models}

We adopt the causal inference framework described in~\citep{pearl2009causality}. A causal graph $G:=(\bm{V}, \bm{E})$ consists of a set of $k$ vertices $\bm{V}:\{V_1, \dots, V_k\}$ and directed edges $\bm{E}:=\{e_{ij} \}$, where the existence of each $e_{ij}$ means that there is a direct causation from $V_i$ to $V_j$, also denoted as $V_i\rightarrow V_j$. We also introduce some notations to describe the relative positions among the nodes. Following a standard assumption in causality (but see, e.g.,~\citep{bongers2021foundations}), we will assume that  $\mathcal{G}$ is a direct acyclic graph (DAG), where we denote the \textit{parents}  of a node $V_i$ as $\PA(V_i) := \{V_j | e_{ij} \in \bm{E} \}$. 
We denote \textit{descendants} $\DE(V_i) := \{V_j | V_j \rightarrow \dots \rightarrow V_i \in \bm{E} \}$ of a node $V_i$ as all the nodes that have at least one direct path leading to a node. 
We call a node $V_k$ as a \textit{confounder} (i.e., common cause) of the other two nodes $V_i$ and $V_j$ if $e_{ki}, e_{kj} \in \bm{E}$; a \textit{collider} (i.e., common effect) if $e_{ik}, e_{jk} \in \bm{E}$; and a \textit{mediator} if $e_{ik}, e_{kj} \in \bm{E}$.

Among all the variables in $\bm{V}$, we use $X$ and $Y$ to denote two special variables, the treatment and effect, respectively.

\subsubsection{Illustration of the Three Rungs of the Causal Ladder}\label{cladder:appd:ladder}

\begin{figure}[ht]
    \centering
    \includegraphics[width=\textwidth]{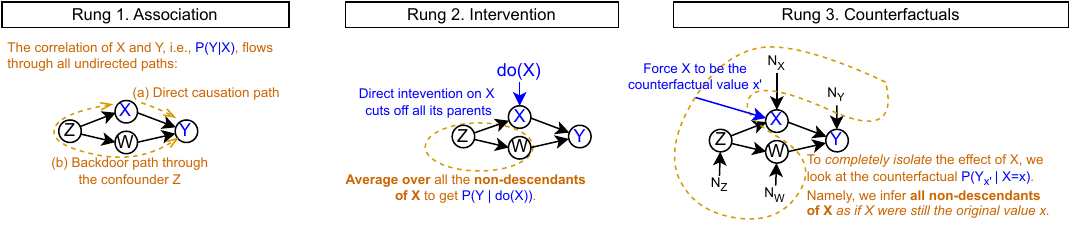}
    \caption{
    The Causal Ladder consists of three rungs: association, intervention and counterfactuals. We color in blue the treatment $X$ and effect $Y$, as well as the actions on $X$. We color in orange words about how to get the estimand, and we use the orange circle to include all the non-descendants of $X$.}
    \label{cladder:fig:ladder}
\end{figure}

In \cref{cladder:fig:ladder}, we illustrate the difference among the three rungs by enumerating what actions are performed on the variables other than target variables $X$ and $Y$.

\subsubsection{Causal Inference Methods}\label{cladder:appd:ci_engine}
We introduce do-calculus which can downgrade the Rung-2 queries to Rung-1 quantities when it is applicable, and counterfactual predictions which downgrade the Rung-3 queries.

\subsubsubsection{Do-Calculus}
\paragraph{Do-Operator as a Notation} 
As mentioned in Rung 2, the $\docal$-operator is a convenient notation to represent an intervention on a variable. For example, $\docal(X = x)$ sets the value of variable $X$ to $x$.

\paragraph{Three Inference Rules for Climbing the Ladder}

{Do-calculus} is a set of rules that allows us to answer higher-rung questions using lower-rung quantities, such as probability distributions of Rung 1.
Given a causal graphical model with and four disjoint sets of variables $X$, $Y$, $Z$, and $W$, and a joint probability distribution that is Markov and faithful to the graph,
do-calculus contains the following three rules:

\noindent \textit{Rule 1 (Insertion/deletion of observations):}
\begin{equation}
    P(Y | \docal(X), Z, W) = P(Y | \docal(X), W)~,
\end{equation}
if $Y$ and $Z$ are d-separated by $X \cup W$ in $G^*$, the graph obtained from  $\mathcal{G}$ by removing all arrows pointing into variables in $X$.

\noindent \textit{Rule 2 (Action/observation exchange):}
\begin{equation}
    P(Y | \docal(X), \docal(Z), W) = P(Y | \docal(X), Z, W)~,
\end{equation}
if $Y$ and $Z$ are d-separated by $X \cup W$ in $G^{\dagger}$, the graph obtained from  $\mathcal{G}$ by removing all arrows pointing into variables in $X$ and all arrows pointing out of variables in $Z$.

\noindent \textit{Rule 3 (Insertion/deletion of actions):}
\begin{equation}
    P(Y | \docal(X), \docal(Z), W) = P(Y | \docal(X), W)~,
\end{equation}
if $Y$ and $Z$ are d-separated by $X \cup W$ in $G^{\ddagger}$, the graph obtained from  $\mathcal{G}$ by first removing all arrows pointing into variables in $X$ (thus creating $G^*$) and then removing all arrows pointing into variables in $Z$ that are not ancestors of any variable in $W$ in $G^*$.

These rules are sound and complete \citep{shpitser2006}. Namely, iff we have all the terms on the right hand side, then the causal term on the left hand side is identifiable.

\paragraph{Example Application of Do-Calculus}
Taking the example in \cref{cladder:fig:generator}, $g_1$ maps the query type ATE to its symbolic expression $\mathbb{E}[Y | \docal(X=1)] - \mathbb{E}[Y | \docal(X=0)]$. 

Next, $g_2$ further simplifies the estimand given the confounding graph, as in the flow chart in the middle of \cref{cladder:fig:generator}:
\begin{align}
    \mathrm{ATE} :=&~ \mathbb{E}[Y | \docal(X=1)] - \mathbb{E}[Y | \docal(X=0)] \\
    =& \sum_{z} P(Z = z) [\mathbb{E}(Y | X = 1, Z = z) - \mathbb{E}(Y | X = 0, Z = z)]
    ~, \label{cladder:eq:backdoor}
\end{align}
which which resolves all the $\docal(\cdot)$ terms to probability terms. This example shows the famous backdoor adjustment in do-calculus \citep{pearl1995causal}.

\subsubsubsection{Three Steps for Counterfactual Prediction}
Given a SCM $M$, distribution on the exogenous variables $P(u)$, and evidence $e$ from the model $\langle M,P(u)\rangle$, the probability of the counterfactual "if $X$ had been $x$ then $Y$ would have been y, given we observed $e$,'' denoted $P(Y_{x}=y|e)$, can be evaluated using the following three steps \citep{pearl2009causality}:

\noindent \textit{\textbf{Abduction:}}
Update the probability distribution $P(u)$ by the evidence $e$ to obtain $P(u|e)$

\noindent \textit{\textbf{Action:}}
Modify $M$ by the action $do(X=x)$, i.e. replace $X$ with $X=x$ in the structural equations, to obtain the modified SCM $M_x$

\noindent \textit{\textbf{Prediction:}}
Use the modified model $\langle M_x, P(u|e)\rangle$, to compute the probability of $Y=y$.

\subsection{Previous Results on \ourdata v1.0}\label{cladder:appd:v1}

\subsubsection{Dataset Statistics for v1.0}

\begin{figure}[ht]
  \centering
  \begin{minipage}[b]{0.6\textwidth}
    \centering \small
    \begin{tabular}{lc|ccccc}
\toprule
& Total & Rung 1 & Rung 2 & Rung 3 \\ \hline
Size &  & \\
\quad \# Samples & 10,560 & 3,288 & 3,288 & 3,984 \\
Question & \\ 
\quad \# Sentences/Sample & 6.85 & 6.00 & 7.00 & 7.25 \\
\quad \# Words/Sample & 94.47 & 76.41 & 96.84 & 103.42 \\
\quad \# Nodes/Graph & 3.54 & 3.54 & 3.55 & 3.54 \\
\quad \# Edges/Graph & 3.44 & 3.41 & 3.43 & 3.46\\
Answer & \\
\quad Positive Class (\%) & 50 & 50 & 50 & 50 \\
Explanations & \\
\quad \# Sentences/Sample & 13.11 & 12.04 & 13.76 & 13.83 \\
\quad \# Words/Sample & 146.82 & 141.88 & 147.88 & 151.30 \\
\bottomrule
    \end{tabular}
    \captionof{table}{Statistics of our \ourdata data v1.0.}
    \label{cladder:tab:stats_v1}
  \end{minipage}
  \hfill
  \begin{minipage}[b]{0.25\textwidth}
    \centering
    \includegraphics[width=\textwidth]{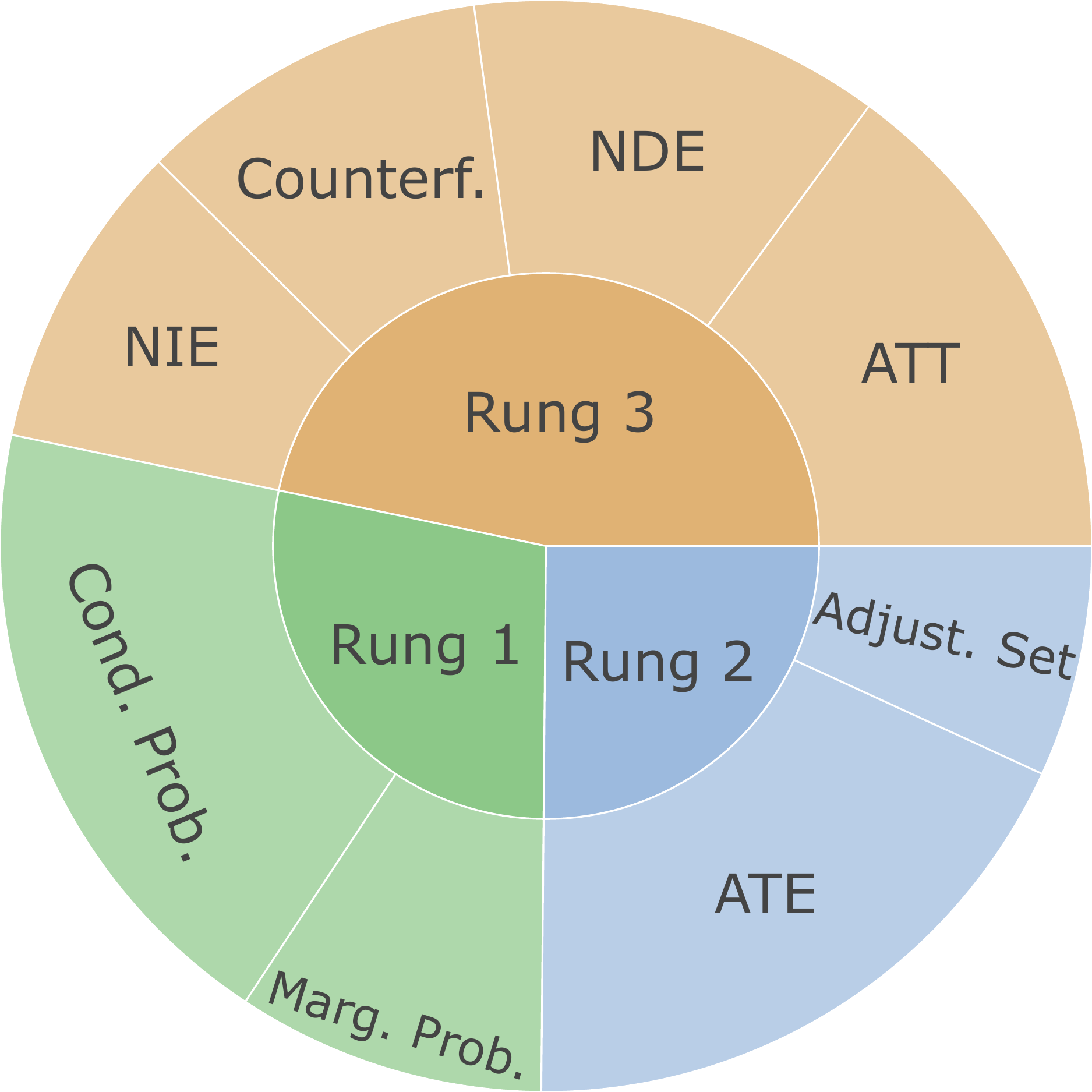}
    \captionof{figure}{Distributions of query types in our dataset v1.0. 
    }\label{cladder:fig:query_v1}
  \end{minipage}
\end{figure}
Our data-generating procedure has the potential to algorithmically generate very large amounts of questions. 
In practice, we pick a dataset size that is large enough to be representative, and at the same time not too large to be problematic given the expensive inference costs of LLMs. We therefore set our dataset size to be 10K. 
We report the statistics of our dataset %
in \cref{cladder:tab:stats_v1}.

The dataset roughly balanced across the query types, graph structures, stories, and ground-truth answers (as seen in \cref{cladder:fig:query_v1}). 
Note that there are some slight adjustments such as more samples for ATE because it allows us to test various techniques, including backdoor and front door adjustments. 
More details on our design choices can be found in~\cref{cladder:appd:querycoverage}.

\subsubsection{Main Results on v1.0}
\begin{table}[ht]
    \centering \small
    \setlength\tabcolsep{5pt}
    \begin{tabular}{lc|ccc|ccccccccccc}
\toprule
& \multirow{2}{*}{Overall Acc.} & \multicolumn{3}{c|}{Acc. by Rung} & \multicolumn{3}{c}{Acc. by Empirical Alignment} \\
&& 1 & 2 & 3 & Anti-C. & Nonsens. & Comm. \\
\hline
Random & 49.27 & 50.28 & 48.40 & 49.12 & 49.69 & 49.01 & 49.12\\
LLaMa & 45.22 & 63.33 & 31.10 & 41.45 & 45.31 & 45.21 & 45.12\\
Alpaca & 45.54 & 63.33 & 31.57 & 41.91 & 45.94 & 45.21 & 45.49\\
GPT-3 Non-Instr. (davinci) & 
47.42 & 63.88 & 32.99 & 44.89 & 47.0 & 48.28   & 46.97\\
GPT-3 Instr. (text-davinci-001) & 
57.07
& 63.95 & 63.63 & 48.04 & 59.12 & 57.81   & 54.28\\
GPT-3 Instr. (text-davinci-002) & 
56.24
& 46.03 & 69.55 & 55.04 & 54.75 & 59.65   & 54.31\\
GPT-3 Instr. (text-davinci-003) & 
62.69
& 58.0 & 80.83 & 54.52 & 63.93 & 62.09  & 62.05\\
GPT-3.5 (queried in May 2023) & 
61.71 
& \ul{65.12} & 69.9 & 54.11 & 65.43 & 55.15   & 64.55\\
GPT-4 (queried in May 2023) & 
64.28 & 53.94 & 81.87 & \ul{63.11} & 65.75 & 60.87  & 66.21\\
+ \ourmodel & 
\ul{66.64} & 61.67 & \ul{86.13} & 58.23 & \ul{69.32} & \ul{63.02}  & \ul{67.60}\\
\bottomrule
    \end{tabular}
    \caption{Performance of all models on our \ourdata dataset v1.0. We report the overall accuracy (Acc.), and also fine-grained accuracy by rung and by empirical alignment.}
    \label{cladder:tab:main_res_v1}
\end{table}
We compare the performance of all models in \cref{cladder:tab:main_res_v1}. First, 
we can see that the causal reasoning task in \ourdata is in general very challenging for all models. And models such as the earlier, non-instruction-tuned GPT-3 and both LLaMa and Alpaca are no better than random performance.
With instruction-tuning, models start to show some improvement. And amongst all, our \ourmodel achieves the highest performance of 66.64\%, which is 2.36 points better than vanilla GPT-4.

Moreover, from the accuracy by empirical alignment level in \cref{cladder:tab:main_res_v1}, we can see that the original GPT-4 model performs the best on commonsensical data, but 5.34 points worse on nonsensical data. However, our \ourmodel enhances the reasoning ability across all levels, with  substantial improvement on anti-commonsensical data and nonsensical data, indicating that \ourmodel is particularly beneficial on unseen data.
\subsubsection{Ablation Study on v1.0}
\begin{wraptable}{r}{3.5cm}
    \centering \small
    \begin{tabular}{lc}
    \toprule
    & Acc. \\
    \ourmodel & 66.64 \\
    w/o Step \stepone{}     & 64.54 \\
    w/o Step \steptwo{}     & 63.74\\
    w/o Step \stepthree{} & 63.43 \\
    w/o Step \stepfour{} & 64.47\\
    \bottomrule
    \end{tabular}
    \caption{Ablation study.}
    \label{cladder:tab:ablation}
\end{wraptable}
We conduct an ablation study for our multi-step \ourmodel. We ablate each of the four subquestions, and observe in \cref{cladder:tab:ablation} that classifying the query type and formalizing it has the most effect on the model's performance, which might be because that they are the crucial formalization step in order to do the causal inference correctly. Meanwhile, removing Steps \stepone{} and \stepfour{}, which are mostly about parsing the prompt correctly, have the least impact on performance.

\subsection{More Experiments}

\subsubsection{Details of Our Error Analysis}\label{cladder:appd:caption_error}
For Step 2 about the query type prediction, we report the overall F1 classification score, and also F1 by rungs. For the rest of the steps, we manually annotate the correctness of 100 samples of \ourmodel. We report the correctness of $\mathrm{est}$ by accuracy, and the correctness of the predicted set of available data by taking the F1 with the ground-truth $\bm{d}$. For Step 5, we report the accuracy of whether the model simplifies the estimand correctly to $\mathrm{est}'$ using causal inference, and also arithmetic correctness (Arith.).

\subsubsection{ROSCOE Evaluation} \label{cladder:par:roscoe}
\begin{figure}[ht]
    \centering
    
    \includegraphics[width=\linewidth]{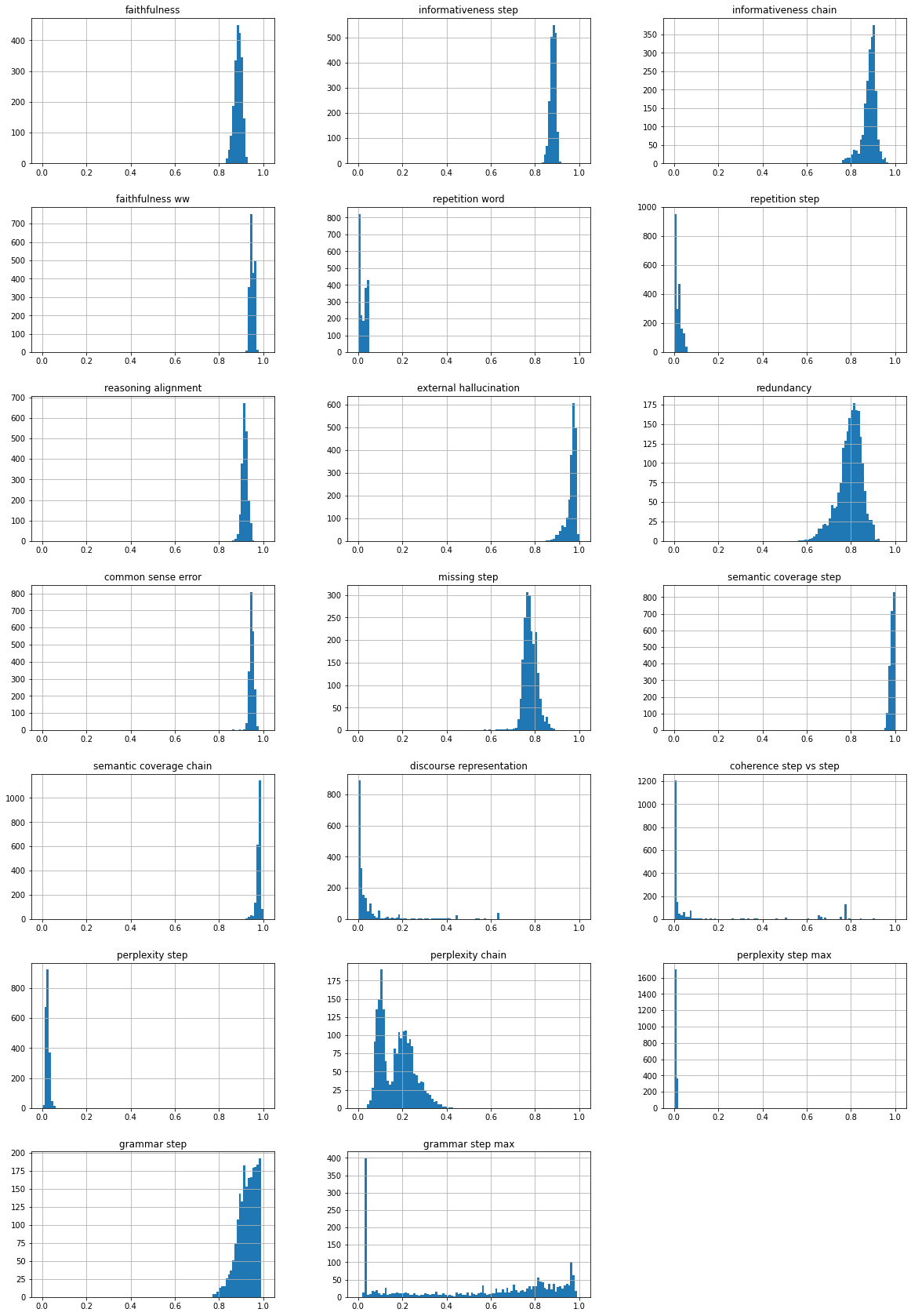}
    \caption{ROSCOE scores of answers from \ourmodel on 2,000 randomly sampled questions from our dataset.}
    \label{cladder:fig:roscoe_scores}
\end{figure}

We employed the ROSCOE suite of evaluation metrics on step-by-step text reasoning, as introduced by \citep{golovneva2022roscoe}, to automate the evaluation of the outputs from \ourmodel on 2,000 randomly sampled questions from our dataset. Differing from conventional metrics, ROSCOE is specifically designed to scrutinize the quality of large language model outputs, focusing on aspects such as semantic consistency, logicality, informativeness, fluency, and factuality, all evaluated within the context of step-by-step reasoning, rather than solely the final response. This allows for a more objective and comprehensive assessment of a model's output, greatly aiding in the verification of its interpretability. The results of this evaluation can be found in \cref{cladder:tab:roscoe_scores} and \cref{cladder:fig:roscoe_scores}. We consider the model's performance as unsatisfying if it falls out of the top quantile, namely
receiving a score $s \in [0, 1]$ smaller than 0.25 when the score should be minimized, or greater than 0.75 when it should be maximized.

We can see in the plot that the good-performing aspects are 
faithfulness to the original question, reasoning alignment with the ground truth, and absence of external hallucinations, which are consistently within the top quantile. This suggests that the model carries out accurate reasoning within the constraints of the fictitious world introduced in each question.

However, there are some performance dips in redundancy, perplexity chain, and missing step metrics. The first two could potentially be attributed to complex elements such as graph notation, while the relatively lower ``missing step'' score warrants further investigation. Despite these observations, this analysis largely aligns with our qualitative understanding of the models' good response ability in answering causal questions in our dataset.

\begin{table}[ht]
    \centering \small
    \begin{tabular}{lrrrrrrr}
\toprule
{} &  Mean &   Std &   Min &   25\% &   50\% &   75\% &   Max \\
\midrule
Faithfulness             &  0.89 &  0.02 &  0.83 &  0.88 &  0.89 &  0.90 &  0.93 \\
Informativeness Step     &  0.88 &  0.01 &  0.83 &  0.87 &  0.88 &  0.89 &  0.92 \\
Informativeness Chain    &  0.88 &  0.03 &  0.76 &  0.87 &  0.89 &  0.90 &  0.96 \\
Faithfulness Word          &  0.95 &  0.01 &  0.92 &  0.94 &  0.95 &  0.96 &  0.97 \\
Repetition Word          &  0.02 &  0.02 & -0.00 &  0.00 &  0.02 &  0.04 &  0.05 \\
Repetition Step          &  0.02 &  0.01 & -0.00 &  0.00 &  0.01 &  0.03 &  0.06 \\
Reasoning Alignment      &  0.92 &  0.01 &  0.86 &  0.91 &  0.92 &  0.93 &  0.95 \\
External Hallucination   &  0.97 &  0.02 &  0.84 &  0.96 &  0.97 &  0.98 &  0.99 \\
Redundancy               &  0.80 &  0.05 &  0.56 &  0.77 &  0.80 &  0.83 &  0.92 \\
Common Sense Error       &  0.95 &  0.01 &  0.86 &  0.94 &  0.95 &  0.96 &  0.98 \\
Missing Step             &  0.78 &  0.03 &  0.58 &  0.76 &  0.78 &  0.80 &  0.88 \\
Semantic Coverage Step   &  0.99 &  0.01 &  0.95 &  0.98 &  0.99 &  0.99 &  1.00 \\
Semantic Coverage Chain  &  0.98 &  0.01 &  0.93 &  0.98 &  0.98 &  0.99 &  0.99 \\
Discourse Representation &  0.06 &  0.13 &  0.00 &  0.01 &  0.01 &  0.05 &  0.67 \\
Coherence Step Vs Step   &  0.14 &  0.27 &  0.00 &  0.00 &  0.01 &  0.07 &  0.94 \\
Perplexity Step          &  0.02 &  0.01 &  0.00 &  0.02 &  0.02 &  0.03 &  0.07 \\
Perplexity Chain         &  0.17 &  0.07 &  0.05 &  0.11 &  0.17 &  0.23 &  0.42 \\
Perplexity Step Max      &  0.00 &  0.00 &  0.00 &  0.00 &  0.00 &  0.01 &  0.02 \\
Grammar Step             &  0.93 &  0.04 &  0.77 &  0.90 &  0.93 &  0.96 &  0.99 \\
Grammar Step Max         &  0.53 &  0.35 &  0.02 &  0.12 &  0.65 &  0.85 &  0.99 \\
\bottomrule
\end{tabular}
    \caption{Statistics of ROSCOE scores evaluated on answers from \ourmodel on 2,000 randomly sampled questions from our dataset. }
    \label{cladder:tab:roscoe_scores}
\end{table}

\clearpage
\subsection{Comparison with Existing Causality-Related Datasets}\label{cladder:appd:related}
We show in \cref{cladder:tab:data_comparison} the distinction of our work from all existing causality-related datasets that address either the causality-as-knowledge task, or the causality-as-language-comprehension task.

\begin{table}[ht]
    \centering \small
    \setlength\tabcolsep{3pt}
    \begin{tabular}{lccccC{1.5cm}cC{1.5cm}C{1.5cm}C{1.5cm}cccccccccccccc}
    \toprule

& \multicolumn{3}{c}{\textbf{Question Types}}
& \multicolumn{4}{c}{\textbf{Skill Types}}
\\ \cline{5-8}
& Assoc. & Interv.
& Counterf.
& CI Method & Formalization of Causal Queries & Causal RE & Qualitative Reasoning 
\\ \midrule
\multicolumn{8}{l}{\textbf{\textit{Datasets for Causality as Knowledge (Commonsense Causality)}}} \\
COPA \citeyearpar{gordon-etal-2012-semeval} & \xmark & \cmark & \xmark & \xmark & \xmark & \xmark & \xmark 
\\
Event2Mind \citeyearpar{rashkin-etal-2018-event2mind} & \xmark & \cmark & \xmark & \xmark & \xmark & \xmark & \xmark 
\\
ATOMIC \citeyearpar{sap2019atomic} & \xmark & \cmark & \xmark & \xmark & \xmark & \xmark & \xmark 
 \\
SocialIQA \citeyearpar{Sap2019SocialIQA} & \xmark & \cmark & \xmark & \xmark & \xmark & \xmark & \xmark 
\\
TimeTravel \citeyearpar{qin-etal-2019-counterfactual} & \xmark & \cmark & \xmark & \xmark & \xmark & \xmark & \xmark 
\\
Goal-Step \citeyearpar{zhang-etal-2020-reasoning} & \xmark & \cmark & \xmark & \xmark & \xmark & \xmark & \xmark 
\\
Abductive (ART) \citeyearpar{bhagavatula2020abductive} & \xmark & \cmark & \xmark & \xmark & \xmark & \xmark & \xmark 
\\
Com2Sense \citeyearpar{singh-etal-2021-com2sense} & \xmark & \cmark & \xmark & \xmark & \xmark & \xmark & \xmark 
\\
CRASS \citeyearpar{frohberg-binder-2022-crass} & \xmark & \xmark & \cmark & \xmark & \xmark & \xmark & \xmark 
\\
\multicolumn{8}{l}{\textbf{\textit{Datasets for Causality as Language Comprehension (Causal Relation Extraction)}}} \\
SemEval2021 Task8 \citeyearpar{hendrickx-etal-2010-semeval} & \xmark & \xmark & \xmark & \xmark & \xmark & \cmark & \xmark 
\\
EventCausality \citeyearpar{do-etal-2011-minimally} & \xmark & \xmark & \xmark & \xmark & \xmark & \cmark & \xmark 
\\
Causal-TimeBank \citeyearpar{mirza-etal-2014-annotating} & \xmark & \xmark & \xmark & \xmark & \xmark & \cmark & \xmark 
\\
CaTeRS \citeyearpar{mostafazadeh-etal-2016-caters} & \xmark & \xmark & \xmark & \xmark & \xmark & \cmark & \xmark 
\\
BECauSE \citeyearpar{dunietz-etal-2017-corpus} & \xmark & \xmark & \xmark & \xmark & \xmark & \cmark & \xmark 
\\
TellMeWhy \citeyearpar{lal-etal-2021-tellmewhy} & \xmark & \xmark & \xmark & \xmark & \xmark & \cmark & \xmark 
\\
\hline

\multicolumn{8}{l}{\textbf{\textit{Datasets for Formal Causal Reasoning}}} \\
Corr2Cause \citep{jin2024large} & \xmark & \cmark & \xmark & \cmark & \cmark & \xmark & \xmark 
\\
\ourdata (Ours) & \mybigtext{\cmark} & \mybigtext{\cmark} & \mybigtext{\cmark} & \mybigtext{\cmark} & \mybigtext{\cmark} & \mybigtext{\cmark} & \mybigtext{\cmark} 
\\
    \bottomrule
    \end{tabular}
    \caption{Comparison of our dataset and existing causal or reasoning datasets. The aim of our dataset is to test the pure reasoning ability of LLMs on causal questions. 
    For each dataset, we first identify whether its question types cover the three rungs: association (Assoc.), intervention (Interv.), and counterfactuals (Counterf.). We also check what skill types the dataset tests: the application of causal inference methods (CI Method), formalization of causal queries, causal relation extraction from the given text (Causal RE), and qualitative reasoning.
    }
    \label{cladder:tab:data_comparison}
\end{table}

\newpage
\section{Additional Materials for \cref{ch:compmech}}

\subsection{Experiments for Pythia-6.9b}
\label{appendix:Pythia}
This section extends the experimental analysis conducted on GPT-2 to Pythia-6.9b. The goal is to replicate the prior methodology and compare the outcomes across the two different models, thus contributing to a broader understanding of model behaviors under similar conditions.

\subsubsection{Macroscopic Inspection across Layers and Token Positions}
\label{app-subsec:residual_stream_pythia}
\begin{figure}[ht]
    \centering
    \begin{subfigure}{.4\textwidth}
        \centering
        \includegraphics[width=\linewidth]{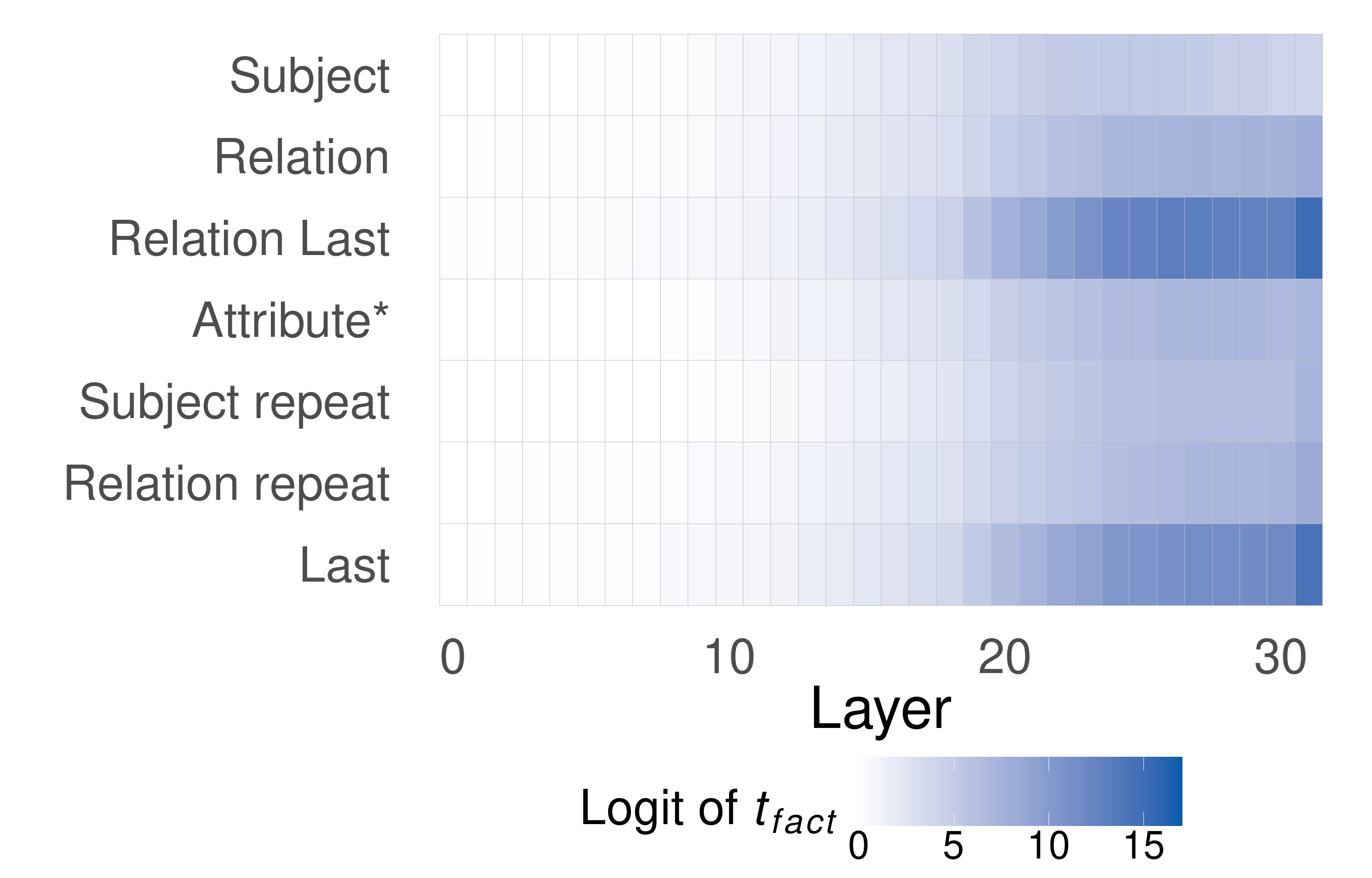}
    \end{subfigure}%
    \hspace{40pt}
    \begin{subfigure}{.4\textwidth}
        \centering
        \includegraphics[width=\linewidth]{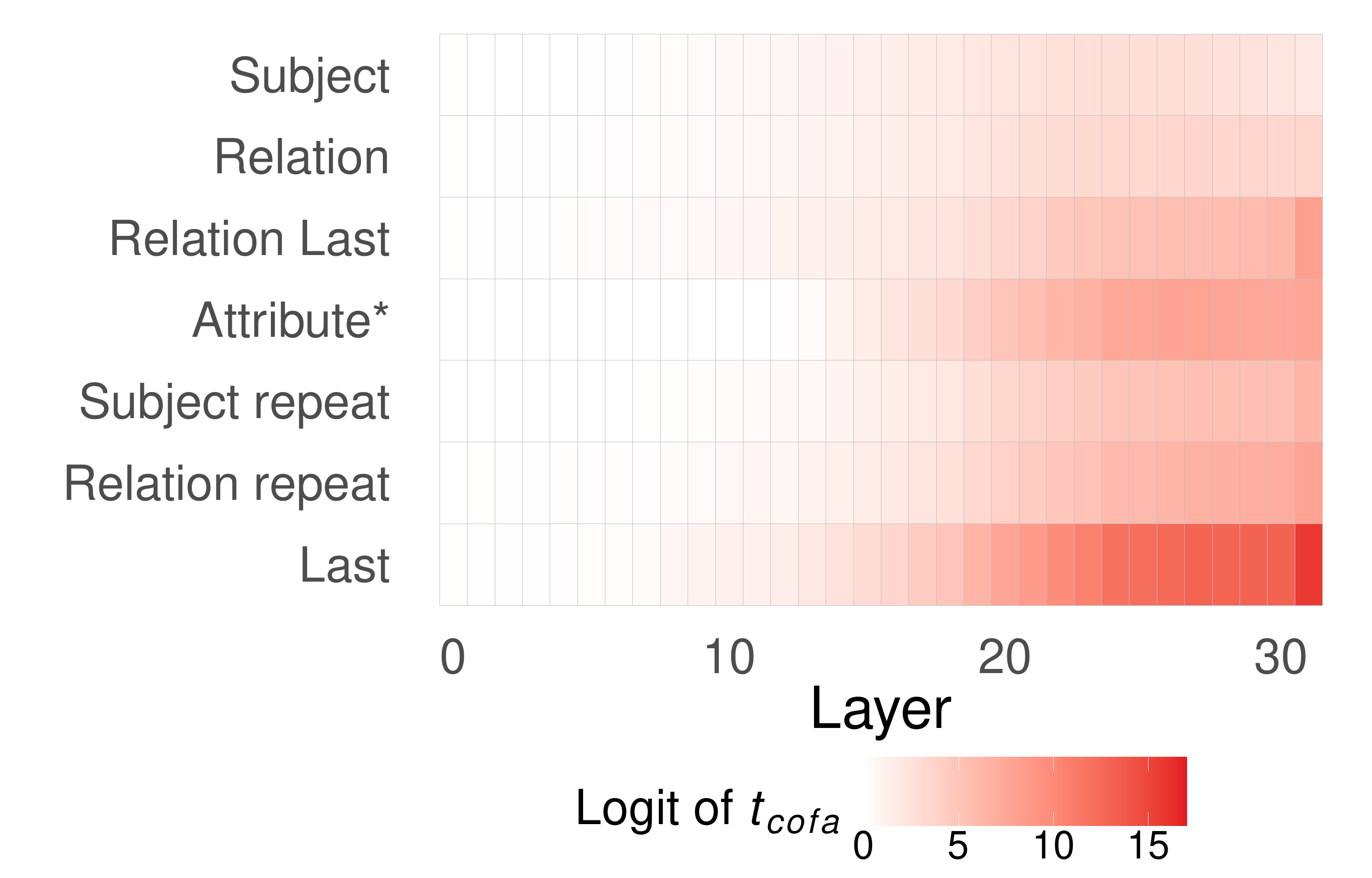}
    \end{subfigure}
    \caption{\textbf{Layer-wise Position Analysis of Relevant Tokens in GPT-2-small}. The figure presents the logit values for two pertinent tokens across various positions and layers. The left panel illustrates the logit values for the factual token $\tfact$, while the right panel illustrates the logit values for the counterfactual token $\talt$. }
    \label{fig:app_residual_stream_pythia}
\end{figure}

\cref{fig:app_residual_stream_pythia} provides a comparative analysis of the logit values for two specific tokens, labeled as factual and counterfactual, across various positions and layers in Pythia-6.9b.

\subsubsection{Intermediate Inspection of Attention and MLP Blocks}
\label{app-subsec:blocks_pythia}
\begin{figure}[ht]
    \centering
    \begin{subfigure}{.46\textwidth}
        \includegraphics[width=\linewidth]{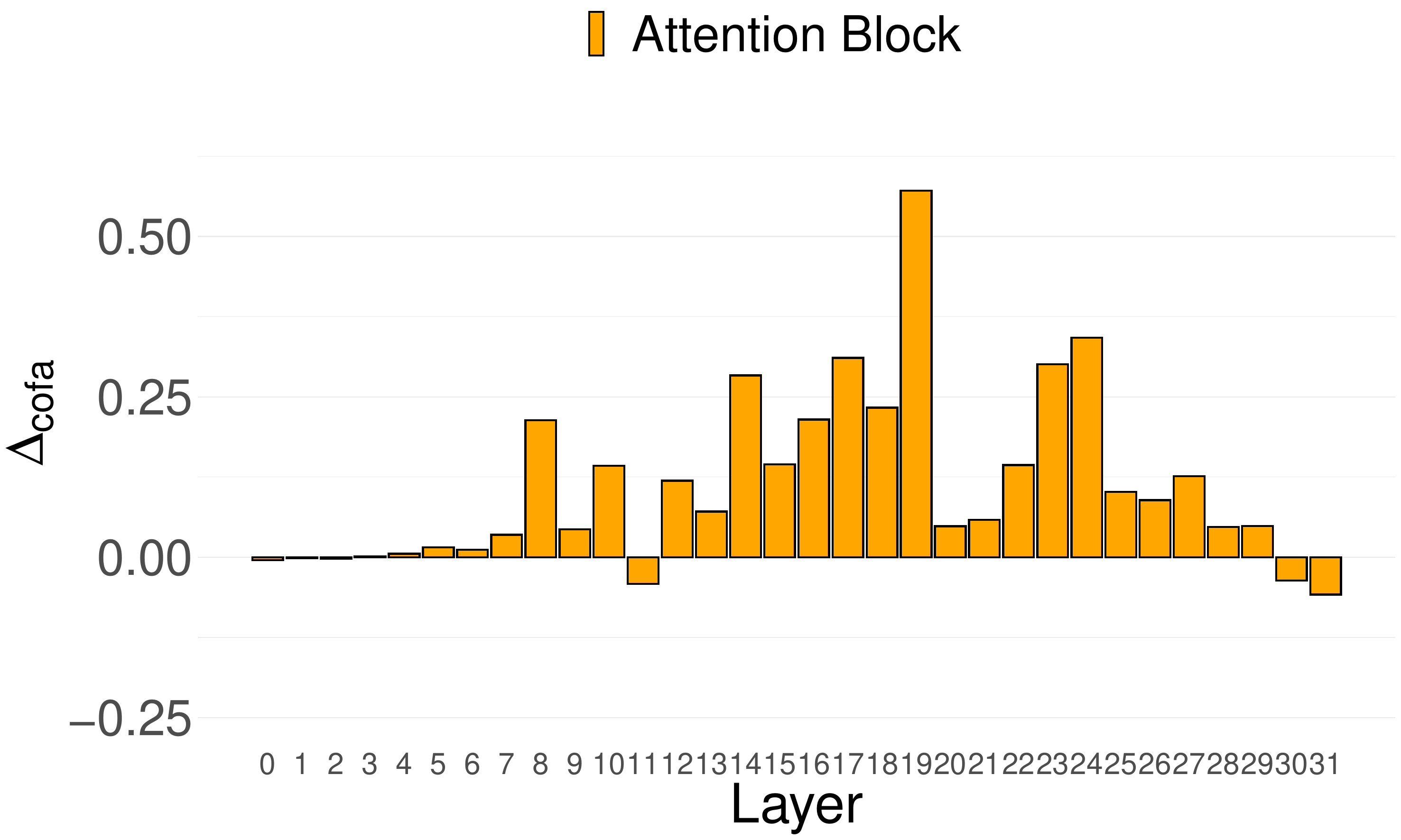}
    \end{subfigure}
    \begin{subfigure}{.46\textwidth}
        \centering
        \includegraphics[width=\linewidth]{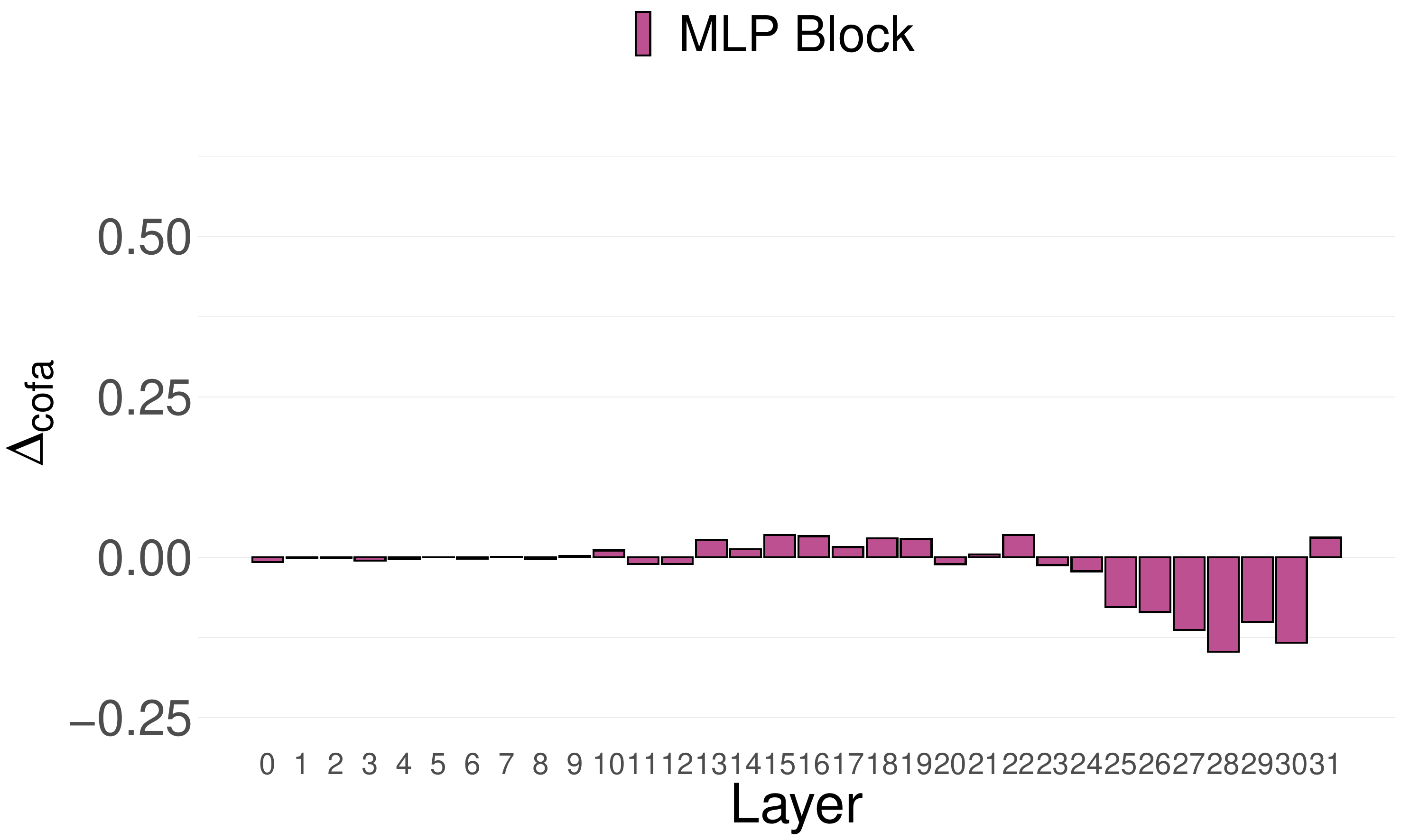}
    \end{subfigure}%
    \caption{\textbf{Attribution of Logit Differences to Attention and MLP Blocks.} delineates the differential impact of Attention and MLP Blocks on logit values at the terminal input position. The attention mechanism is shown to predominantly influence early layer processing in the left panel, while the right panel details the increased contribution of MLP Blocks to the factual token's logits in the concluding layers, illustrating the dynamic interplay between these fundamental neural network elements.}
    \label{fig:app_attn_mlp_contribution_pythia}
\end{figure}

This subsection exposes the contributions of Attention and MLP Blocks to the differences in logit values across layers within Pythia-6.9b. \cref{fig:app_attn_mlp_contribution_pythia} explores how these components influence the computation of logits for two tokens, represented as the difference $\dalt = \text{Logit}(\talt) - \text{Logit}(\tfact)$ at the final position of the input. The analysis specifically highlights the distinct effects of these blocks at different stages of the model's operation.

\subsubsection{Microscopic Inspection of Individual Attention Heads}
\label{app-subsec:attention_heads_pythia}

\begin{figure*}[ht]
\centering

        \includegraphics[width=0.7\linewidth]{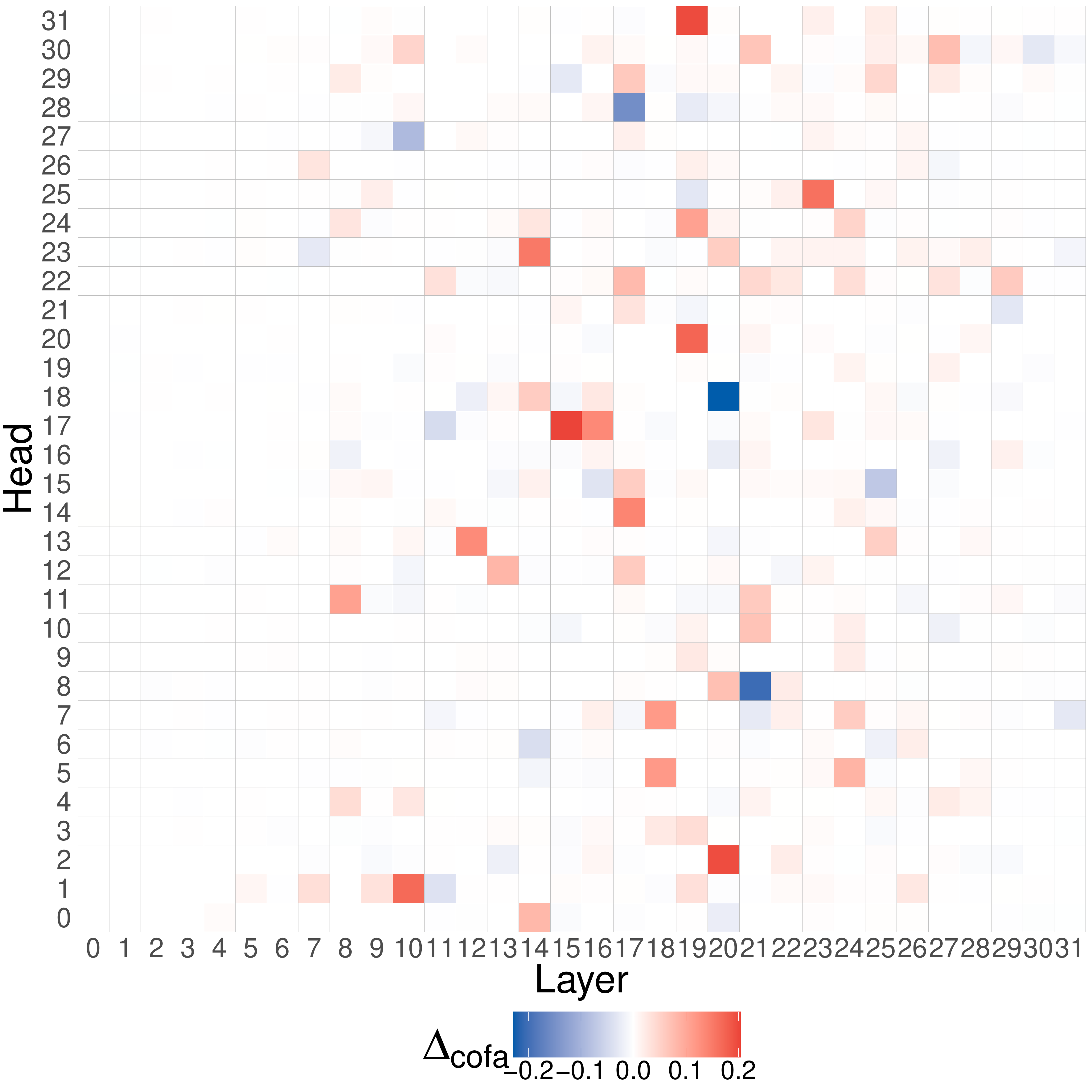}

\caption{\textbf{Direct Contribution of Attention Heads.} The figure displays the direct contribution of all heads in Pythia-6.9b to the logit difference $\Delta_{\text{cofa}}$ with heads favoring $\tfact$ highlighted in blue and those favoring $\talt$ in red.}
\label{fig:app_relevant_attention_heads_pythia}
\end{figure*}

Figure \ref{fig:app_relevant_attention_heads_pythia} quantifies the direct contributions of all attention heads to the difference in logit values, labeled as $\Delta_{\text{cofa}}$. It specifically identifies heads that preferentially enhance the logits for $\tfact$ (shown in blue) versus those favoring $\talt$ (depicted in red), offering insights into how attention mechanisms differentially prioritize token attributes.

\begin{figure}[ht]
\centering
        \includegraphics[width=.5\linewidth]{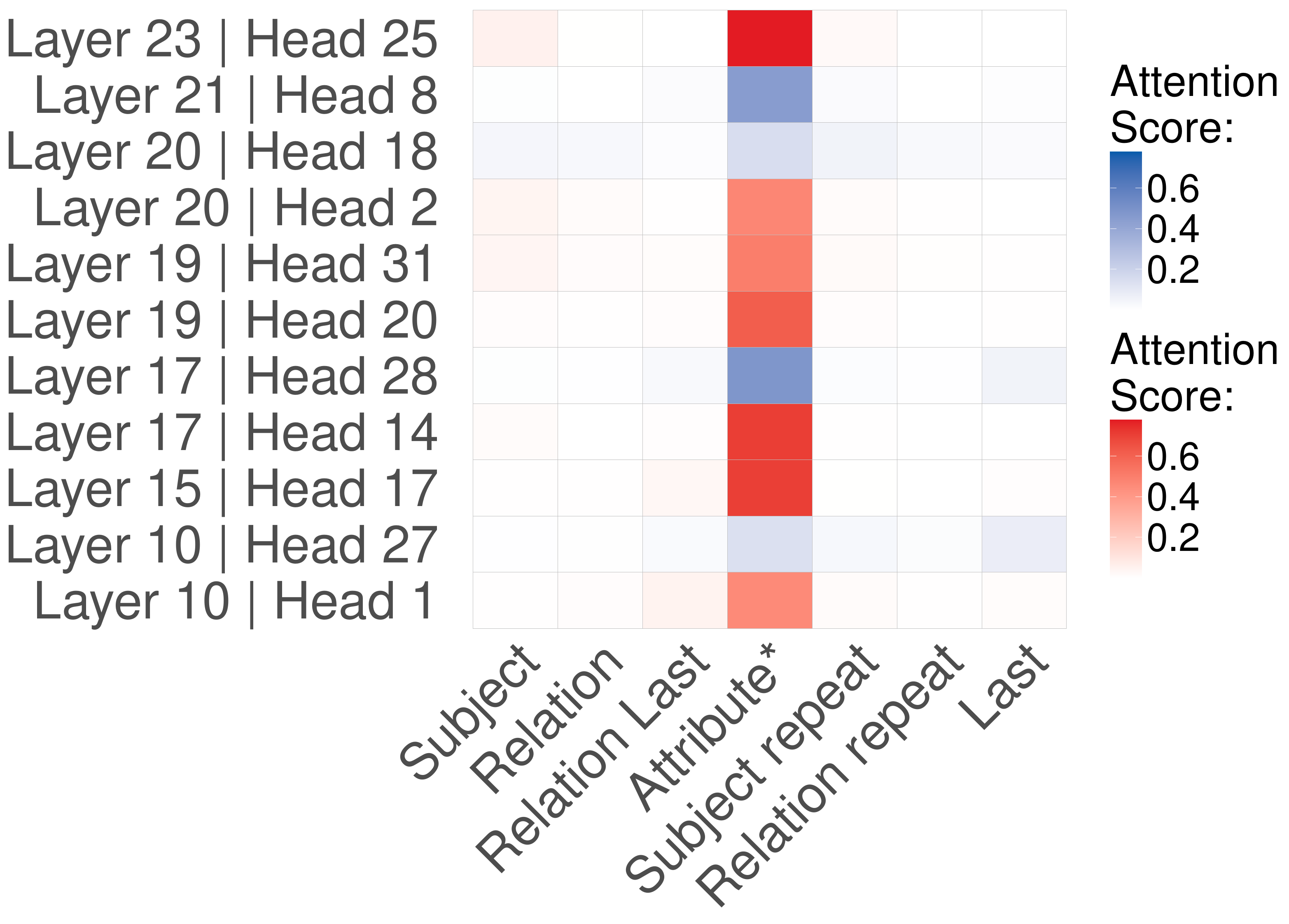}
    \caption{\textbf{Attention Pattern for Relevant Attention Heads.} The panel illustrates the attention patterns of relevant heads for the last position, demonstrating consistent attention to the attribute position by both red and blue heads.  }
    \label{fig:app_attention_heads_analysis_pythia}
\end{figure}

\cref{fig:app_attention_heads_analysis_pythia} presents the attention patterns of the relevant attention heads at the last token position. It shows the consistent pattern of the relevant heads, with a consistent focus on the attribute position.

\subsection{Other Experiment for GPT-2}
\label{appendix:GPT-2}
\subsubsection{Ranks Analysis in the Last Position}
\label{app-subsec:rank_gpt2}
\begin{figure}[ht]
    \centering
        \includegraphics[width=.5\linewidth]{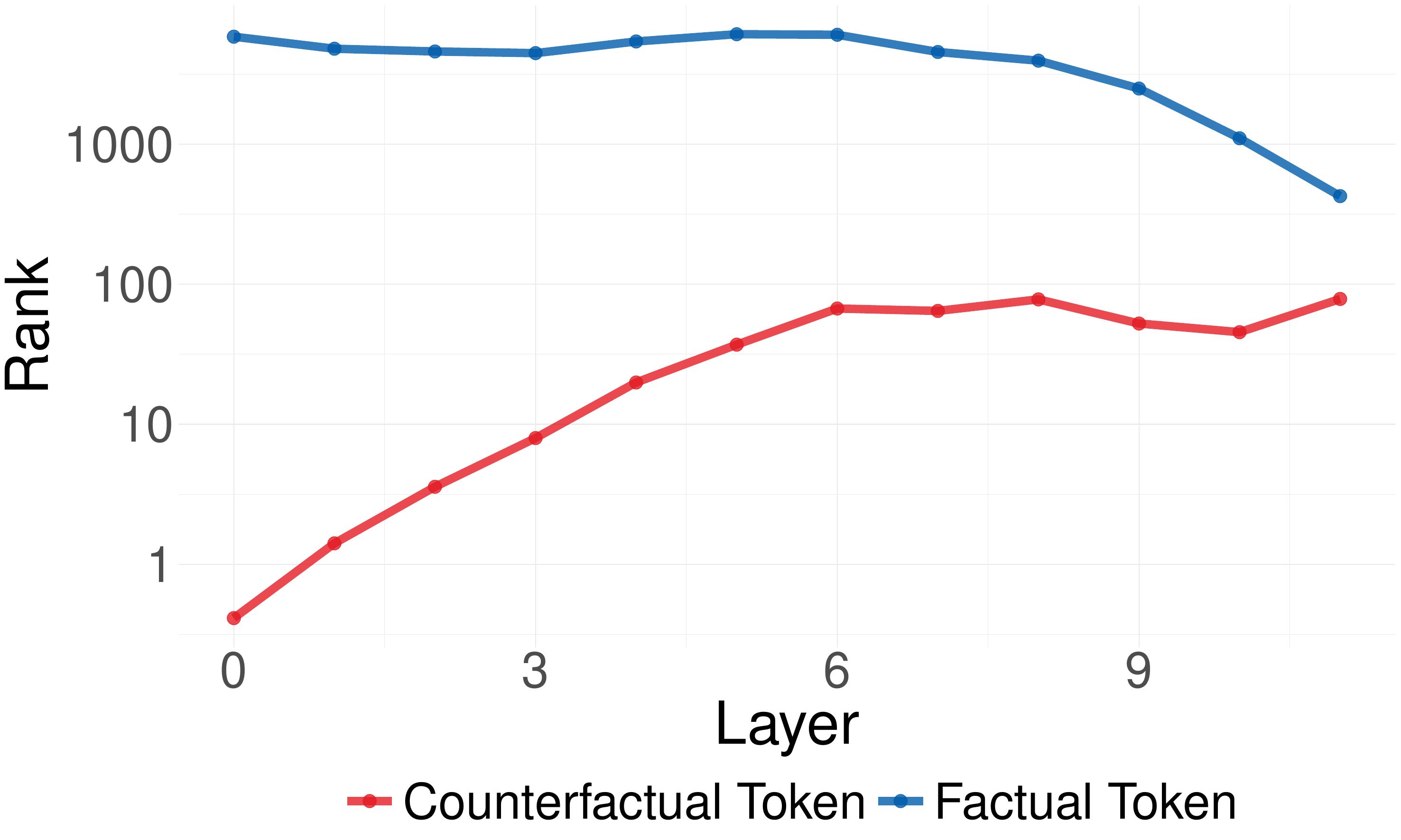}
    \caption{\textbf{Rank of Target Tokens for Attribute Position Across Layers in GPT-2.} This figure depicts the trend where the logit rank for the factual token $\tfact$ decreases while the rank for the counterfactual token $\talt$ increases at the attribute position. In the concluding layers, this pattern is evident as $\tfact$ typically secures a lower rank, in contrast to $\talt$, which shows an upward trajectory in rank. However, it is important to note that $\talt$'s rank consistently remains lower than that of $\tfact$.}
    \label{fig:app_rank_logit_GPT-2}
\end{figure}

We provide additional information in \cref{fig:app_rank_logit_GPT-2} mapping the logits to ranks of the tokens, and find that 
the rank of $\talt$ in the projected logit distribution remains very low: $\talt$ is among the 20 most likely tokens in the first five layers and between the 20th and the 70th in the last part of the network.
\subsubsection{Attention Pattern of Relevant Attention Heads}
\label{app-subsec:full_attn_pattern_gpt2}
\begin{figure}[ht]
\centering
     \begin{subfigure}{.7\textwidth}
        \centering
        \includegraphics[width=.7\linewidth]{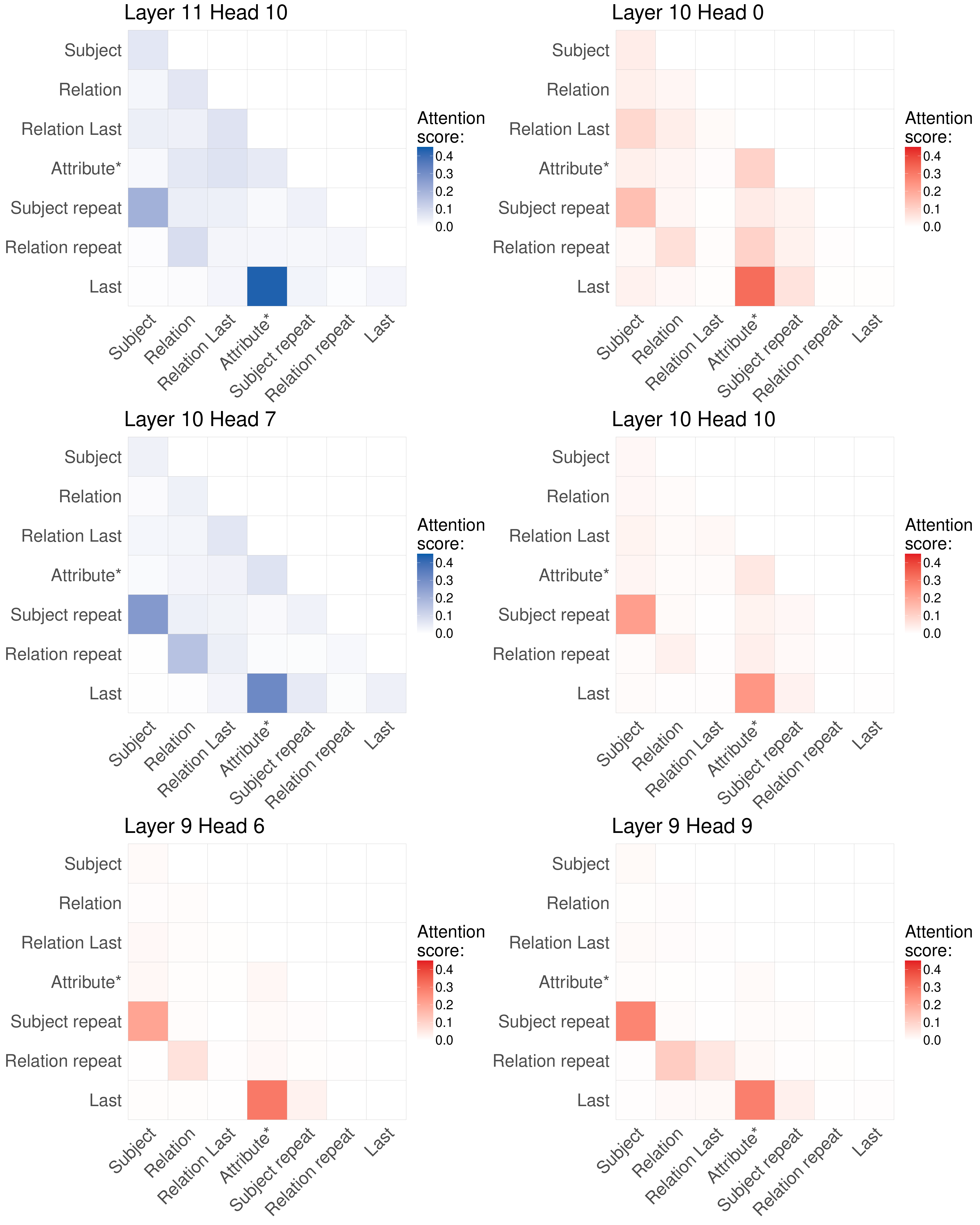}
    \end{subfigure}%
    \caption{\textbf{Attention Pattern of Significant Heads.} This figure illustrates the comprehensive attention pattern of heads substantially influencing $\Delta_{\talt}$. Notably, a similar pattern emerges for both heads favoring $\talt$ (depicted in red) and those favoring $\tfact$ (illustrated in blue), particularly in the attention edge between the attribute and the final position.}
    \label{app:fig_full_head_pattern_GPT-2}
\end{figure}

\cref{app:fig_full_head_pattern_GPT-2} shows the full attention pattern for the relevant attention heads, as identified in \cref{sec:results}. It is show as the attention pattern is similar between all the relevant attention heads, independently if the heads is favoring $\tfact$ or $\talt$. 

\newpage
\section{Additional Materials for \cref{ch:mathrobust}}

\subsection{Creation of the Prompts}
\label{mathrobust:appd:prompt_creation}

We consider MWP examples from the union of the three datasets SVAMP, ASDiv-A, and MAWPS. The textual template $\bm{t}$ of a problem consists of a context (describing a real-world state and/or actions) and a question.
In order to obtain suitable prompts for the models, we convert the problems' questions into statements where the result of the problem is expected to be the first token after the prompt.
E.g., in the example in \cref{mathrobust:sec:problem_setup}, \emph{how many trees will he have?} is converted into \emph{the number of trees that he will have is \_}.
From the MWP templates of the SVAMP/ASDiv-A/MAWPS collection (we consider all splits), we filter out the templates whose questions do not start with \emph{How many...}, and we use spaCy\footnote{\url{https://spacy.io}} to identify the subject, the object and the verbs in the sentence. This allows us to convert the last sentence of the template from \emph{The number of... is}. This way, we obtain 437 statement-based MWP templates for two-operand problems and 307 for three-operand problems. We manually checked a subset of the templates to identify possible mistakes in the conversion procedure.

\subsection{Frequently Asked Questions}

\subsubsection{How do the intervention data look like?}
\label{mathrobust:appd:examples}
In \cref{mathrobust:tab:AppendixExamples} we report examples of MWP pairs representing different types of intervention.

\subsubsection{What is the accuracy of the evaluated models on the generated problems?}
\label{mathrobust:appd:accuracy}
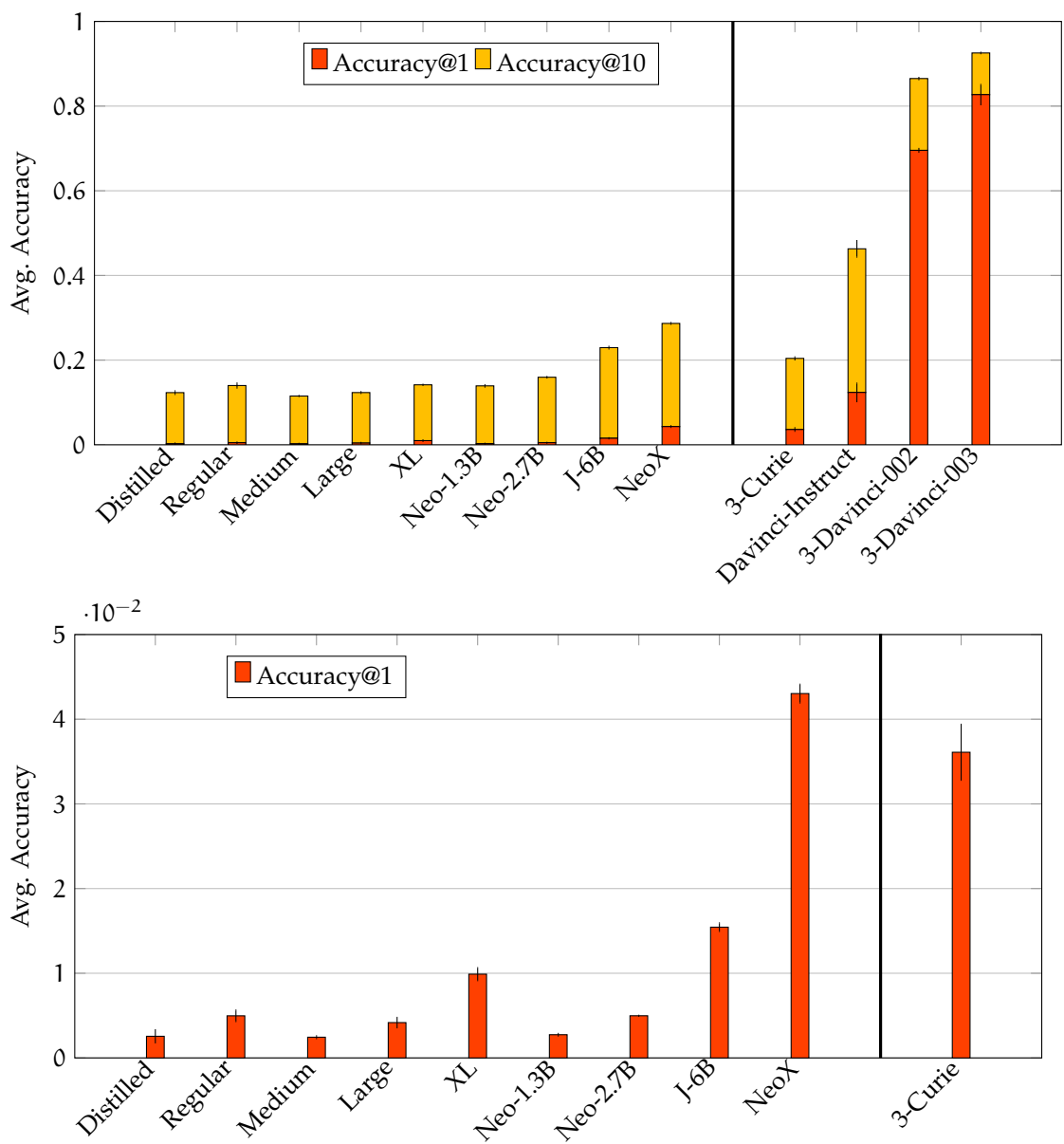
\begin{figure}\small
\begin{tikzpicture}
\begin{axis}
[
  xtick=data,
  ymax=1,
  ymin=0,
  ylabel= Avg. Accuracy,
  height=0.5\columnwidth,
  width=\columnwidth,
  symbolic x coords={Distilled, Regular, Medium, Large, XL, Neo-1.3B, Neo-2.7B, J-6B, NeoX, \ , 3-Curie, Davinci-Instruct, 3-Davinci-002, 3-Davinci-003, A},
  enlarge y limits=0.0,
  enlarge x limits=0.1,
  legend style={at={(0.40,0.95)},
  anchor=north,legend columns=-1},
  ybar stacked,
  bar width=7pt,
  grid=major,
    xmajorgrids=false,
    x tick label style={rotate=45,anchor=east}
]
\addplot+ [
   black, fill=orange!50!red,
   error bars/.cd,
   y dir=both,
   y explicit,
        error mark options={
      rotate=90,
      mark size=0pt,
    }
]coordinates {
(3-Davinci-002, 0.6955377574370709) +- (-, 0.003546910755148736)
(J-6B, 0.015441647597254006) +- (0, 0.00045308924485125846)
(Neo-2.7B, 0.004979405034324943) +- (-, 9.153318077802838e-06)
(Neo-1.3B, 0.00274370709382151) +- (-, 0.00010297482837528594)
(XL, 0.009892448512585812) +- (-, 0.0007162471395881012)
(Large, 0.00417162471395881) +- (-, 0.0005697940503432495)
(Medium, 0.002446224256292906) +- (-, 0.00011670480549199084)
(Regular, 0.00497254004576659) +- (-, 0.0006292906178489701)
(Distilled, 0.002556064073226545) +- (-, 0.0007254004576659038)
(3-Curie, 0.03609839816933638) +- (-, 0.00326086956521739)
(NeoX, 0.04302974828375286) +- (-, 0.00105720823798627)
(Davinci-Instruct, 0.12368421052631579) +- (-, 0.020938215102974826)
(3-Davinci-003, 0.8271167048054919) +- (-, 0.022997711670480536)
};

\addplot+ [
   black, fill=orange!50!yellow,
   error bars/.cd,
   y dir=both,
   y explicit,
        error mark options={
      rotate=90,
      mark size=0pt,
    }
]coordinates {
(3-Davinci-002, 0.16956521739130437) +- (0, 0.0017162471395880674)
(J-6B, 0.21408237986270023) +- (0, 0.002155606407322652)
(Neo-2.7B, 0.15461098398169335) +- (0, 0.0007665903890160114)
(Neo-1.3B, 0.13646681922196793) +- (0, 0.0019244851258581191)
(XL, 0.13179176201373) +- (0, 0.00034782608695652084)
(Large, 0.11921510297482839) +- (0, 0.0011945080091533208)
(Medium, 0.11280778032036615) +- (0, 0.00027917620137300064)
(Regular, 0.13502974828375286) +- (0, 0.004995423340961105)
(Distilled, 0.12067963386727688) +- (0, 0.003290617848970251)
(3-Curie, 0.16802059496567506) +- (0, 0.0026315789473684292)
(NeoX, 0.24368878718535467) +- (0, 0.0010892448512585806)
(Davinci-Instruct, 0.3392448512585812) +- (0, 0.018535469107551494)
(3-Davinci-003, 0.09839816933638446) +- (0, 0.0008009153318077611)

};
\draw[very thick] (axis cs:\ ,0) -- (axis cs:\ ,1.2);,

\legend{Accuracy@1, Accuracy@10}
\end{axis}
\end{tikzpicture}

\begin{tikzpicture}
\begin{axis}
[
  xtick=data,
  ymax=0.05,
  ymin=0,
  ylabel= Avg. Accuracy,
  height=0.5\columnwidth,
  width=\columnwidth,
  symbolic x coords={Distilled, Regular, Medium, Large, XL, Neo-1.3B, Neo-2.7B, J-6B, NeoX, \ , 3-Curie, A},
  enlarge y limits=0.0,
  enlarge x limits=0.1,
  legend style={at={(0.25,0.95)},
  anchor=north,legend columns=-1},
  ybar stacked,
  bar width=7pt,
  grid=major,
    xmajorgrids=false,
    x tick label style={rotate=45,anchor=east}
]
\addplot+ [
   black, fill=orange!50!red,
   error bars/.cd,
   y dir=both,
   y explicit,
        error mark options={
      rotate=90,
      mark size=0pt,
    }
]coordinates {
(J-6B, 0.015441647597254006) +- (0, 0.00045308924485125846)
(Neo-2.7B, 0.004979405034324943) +- (-, 9.153318077802838e-06)
(Neo-1.3B, 0.00274370709382151) +- (-, 0.00010297482837528594)
(XL, 0.009892448512585812) +- (-, 0.0007162471395881012)
(Large, 0.00417162471395881) +- (-, 0.0005697940503432495)
(Medium, 0.002446224256292906) +- (-, 0.00011670480549199084)
(Regular, 0.00497254004576659) +- (-, 0.0006292906178489701)
(Distilled, 0.002556064073226545) +- (-, 0.0007254004576659038)
(3-Curie, 0.03609839816933638) +- (-, 0.00326086956521739)
(NeoX, 0.04302974828375286) +- (-, 0.00105720823798627)
};
\draw[very thick] (axis cs:\ ,0) -- (axis cs:\ ,1.1);,

\legend{Accuracy@1}
\end{axis}
\end{tikzpicture}

\caption{Average accuracy of the models on the generated instances of MWPs. Results are averaged over two sets consisting of 500 problem instances generated for each template. The lower figure shows a zoomed-in visualization of the accuracy at 1.}
\label{mathrobust:fig:accuracy}
\end{figure}

We report the accuracy of the models considered for our main evaluation in terms of accuracy at 1 and accuracy at 10. Results are displayed in \cref{mathrobust:fig:accuracy}.
The accuracy of the LLaMA models is 11.1\%, 25.7\%, 32.8\%, and 13.0\% respectively for the 7B, 13B, 30B, and Alpaca versions. The accuracy of the GPT-3 Davinci models on the three-operand problems is 2\%, 11\%, and 15\% for the Instruct, Davinci-002, and Davinci-003 versions, respectively.

\begin{table*}
    \resizebox{\textwidth}{!}{
    \centering
    \begin{tabularx}{\textwidth}{ lX c}
    \toprule[0.1em]
    \multirow{4}{*}{$\mathrm{TCE}(\bm{N} \rightarrow R)$ } & Ruby has 87 candies. If she shares the candies among 29 friends, the number of candies that each friend gets is & \multirow{2}{*}{$g = 87 / 29 = 3$} \\
    \cmidrule{2-3}
    & Ruby has 35 candies. If she shares the candies among 5 friends, the number of candies that each friend gets is & \multirow{2}{*}{$g = 35 / 5 = 7$} \\
    \midrule[0.1em]
    \multirow{4}{*}{$\mathrm{DCE}(\bm{N} \rightarrow R)$ } & The school is composed of 13 buildings each having 10 classrooms. The number of classrooms that the school has is  & \multirow{2}{*}{$g = 10 \times 13 = 130$} \\
    \cmidrule{2-3}
    & The school is composed of 65 buildings each having 2 classrooms. The number of classrooms that the school has is & \multirow{2}{*}{$g = 65 \times 2 = 130$} \\
    \midrule[0.1em]
    \multirow{6}{*}{$\mathrm{DCE}(S \rightarrow R)$ } &  The razorback t-shirt shop ordered 6 cases of t-shirts. If each case contains 17 t-shirts the number of t-shirts that they ordered is  & \multirow{3}{*}{$g = 17 \times 6 = 102$} \\
    \cmidrule{2-3}
    & The roller coaster at the state fair costs 6 tickets per ride. If 17 friends were going to ride the roller coaster the number of tickets that they would need is  & \multirow{3}{*}{$g = 17 \times 6 = 102$} \\
    \midrule[0.1em]
    \multirow{5}{*}{$\mathrm{TCE}(\bm{T} \rightarrow R)$ } & Sean has 23 whistles. He has 6 more whistles than Charles. The number of whistles that Charles has is  & \multirow{2}{*}{$g = 23 - 6 = 17$} \\
    \cmidrule{2-3}
    &  Jovana filled her bucket with 23 pounds of shells. If she adds 6 more pounds of shell to fill her bucket, the number of pounds that she has is  & \multirow{3}{*}{$g = 23 + 6 = 29$} \\

 \bottomrule[0.1em]
\end{tabularx}
}
    \caption{For each of the causal effects measured (left column), we report a pair of MWPs illustrating the intervention performed (center), along with their respective ground-truth result (left column).}
    \label{mathrobust:tab:AppendixExamples}
\end{table*}

\subsubsection{What is the relation between accuracy and the RCC metric?}

We examine the relationship between performance and robustness, computing the Pearson correlation coefficient between accuracy (accuracy@10) and the relative confidence change (RCC) metric. On a per-template basis (500 instances for each template), we found accuracy to be positively correlated with $\mathrm{TCE}(\bm{N} \text{ on } R)$ and $\mathrm{TCE}(T \text{ on } R)$  (0.24 and 0.49, respectively) and negatively correlated with $\mathrm{DCE}(\bm{N} \rightarrow R)$and $\mathrm{DCE}(S \rightarrow R)$ (-0.26 and -0.36, respectively). We see these results as a quantitative validation of the intuition behind our framework: the better the model’s performance, the more the model tends to correctly adjust its prediction after a result-altering intervention (higher sensitivity) and to correctly not change its prediction after a result-preserving intervention (higher robustness).

Moreover, we conduct an additional sanity check as in \citet{patel-etal-2021-nlp}: removing the question from the MWP templates, we observe a sensitivity-robustness degradation to random guessing (i.e., TCE $\simeq$ DCE). This indicates that the measurement of the causal effects within our framework is not affected by patterns in the templates that might have been picked up or memorized by large models.

\subsection{Computation of Causal Effects for GPT-3}
\label{mathrobust:appd:gpt3_approx}
We access GPT-3 through the OpenAI APIs, which allow a user to prompt the model and obtain the probabilities assigned by the model to the $k$-th most likely vocabulary entries, for each token generated.
To overcome this limitation, we approximate the relative probability change $\delta_{\mathrm{rcc}}$ as follows, depending on the kind of effect measured.

The limit for $k$ is set by OpenAI to 5. However, for our main set of experiments (i.e., computing the causal effects of $\bm{N}$, $S$, and $\bm{T}$) we were granted an increased limit of $k$ to 100. This allowed us to obtain reasonable estimates for the causal effects, as the number of cases in which $P(g)$ is not defined is less than $10\%$ of the number of examples that we consider.

\begin{algorithm}
\caption{Computation of $\delta_{\mathrm{rcc}}$ for GPT-3}\label{mathrobust:alg}
\begin{algorithmic}[1]
\State $\bm{Q} = (\bm{t}, \bm{n}, g)$
\State $\bm{Q}' = (\bm{t}', \bm{n}', g')$

\If{$P(g)$ is defined}
    \If{$P'(g)$ is defined}
        \State $ \Delta = \frac{P(g) - P'(g)}{P'(g)}$
    \Else
        \State $\hat{P}' \gets P'(k\text{-th most likely token})$
        \State $ \Delta = \frac{P(g) -\hat{P}'}{\hat{P}'}$
    \EndIf
\Else
    \State $\Delta = 0$
\EndIf

\If{$P'(g')$ is defined}
    \If{$P(g')$ is defined}
        \State $ \Delta' = \frac{P'(g') - P(g')}{P(g')}$
    \Else
        \State $\hat{P} \gets P(k\text{-th most likely token})$
        \State $ \Delta' = \frac{P'(g') -\hat{P}}{\hat{P}}$ 
    \EndIf
\Else
    \State $\Delta' = 0$ 
\EndIf
\State $ \delta_{\mathrm{rcc}} = \frac{1}{2} (\Delta + \Delta')$

\end{algorithmic}
\end{algorithm}

\subsubsection{TCE(\textbf{N} { on } R) and TCE(\textbf{T} { on } R) }
In cases when $P(g)$ is defined (i.e. when $g$ appears in the top $k$ token predictions) and $P'(g)$ is not defined, we compute a lower bound on the relative change using the upper bound on $P'(g)$ given by the probability of the $k$-th most likely token. This gives us a conservative estimate of $\Delta$. For cases in which $P(g)$ is not defined, we cannot say anything about the relative change, and we set $\Delta = 0$. The same applies when swapping $P$ and $P'$. This procedure is illustrated by \cref{mathrobust:alg}.

\subsubsection{DCE(\textbf{N} -> R) and DCE(S  ->  R)}
In this case, we simply discard the examples for which $P(g)$ is not defined or $P'(g)$ are not defined. In that is not the case, then we compute  $\delta_{\mathrm{rcc}}$ as in \cref{mathrobust:sec:distr_diff_metrics}.

\subsubsection{Heatmap Illustration}
The heatmap for GPT-3 displayed in \cref{mathrobust:fig:heatmaps} was computed by taking the raw probability score produced by the model over the whole vocabulary, as the limit on the available top predicted tokens makes it impossible to normalize it over the set $\{0,\dots,300\}$, as done for the other models. The probability was set to 0 when $g$ did not appear in the model's top 5 predictions for the next token after the prompt.

\subsection{Computing Infrastructure and Inference Details}
\label{mathrobust:appd:computing_infrastructure}
To run our experiments, we used a single NVIDIA TITANRTX with 24GB of memory for all the versions of GPT-2 and GPT-Neo. We used a single NVIDIA A100 with 40GB of memory for GPT-J-6B and a single NVIDIA A100 with 80GB of memory for GPT-NeoX and the LLaMA models (two for the 30B version). We accessed GPT-3 using the OpenAI APIs. The longest run (GPT-J) on the four kinds of experiments corresponding to the four kinds of effects measured took $\sim$12 hours, using 500 MWP instances for each of the 437 templates. Due to budget and resource constraints, the experiments on GPT-3, GPT-NeoX, and LLaMA were carried out using 20 examples generated for each template and took $\sim$7 hours.
Experiment tracking was carried out using Weights \& Biases\footnote{\url{http://wandb.ai/}}.

\newpage
\section{Additional Materials for \cref{ch:icm}}

\subsection{Meta Study Settings of SSL and DA}\label{icm:appd:meta_ssl}
For the meta study of SSL, we covered but are not limited to all relevant papers cited by the review on NLP SSL by \citet{sogaard2013semi}. We went through the leaderboard of many NLP tasks and covered the SSL papers listed on the leaderboards.
The papers covered by our meta study are available on our GitHub.

For supervised DA, we searched papers with the keyword domain adaptation and task names from a wide range of tasks that use supervised DA.

Note that for fair comparison, we do not consider papers without a comparable supervised baseline corresponding to the SSL, or a comparable unadapted baseline corresponding to the DA.
We do not consider MT DA which tackles the out-of-vocabulary (OOV) problem because $P(E|C)$ may be different for OOV \citep{habash2008four,daume2011domain}.

\subsection{Experimental Details of Minimum Description Length}\label{icm:appd:mdl}
We calculate the MDL(X) and MDL(Y) by a language model, and obtain MDL(X|Y) and MDL(Y|X) using translation models. For language model, we use the autoregressive GPT2 \citep{radford2019language}, and for the translation model, we the Marian Neural Machine Translation model \citep{mariannmt} trained on the OPUS Corpus \citep{tiedemann-nygaard-2004-opus}. Both these models use the layers from the transformer model \citep{vaswani2017attention}. The autoregressive language model consists only of decoder layers, whereas the translation model used six encoder and six decoder layers. Both of these models have roughly the same number of parameters. We used the huggingface implementation \citep{wolf-etal-2020-transformers} of these models for their respective set of languages.

\newpage
\section{Additional Materials for \cref{ch:psychcausal}}

\subsection{Implementation Details}

\subsubsection{Model Details}\label{psychcausal:appd:models}
\paragraph{Using Closed-Weight Models}
For the use of GPT model series, we use the OpenAI API,\footnote{ \url{https://openai.com/api/}} with a text generation temperature of 0.
We spent around 400 USD across around 20-30K single API calls.

\paragraph{Using Open-Weight Models}\label{psychcausal:appd:exp_details}
For reproducibility, we set the generation temperature to 0 for all the models used in our work. For the open-weight models, GPT2-XL, LLaMa-7B and Alpaca-7B, it took around 24 hours on 4 GPUs RTX 2080 to generate their predictions on 1K data points for the 5 paraphrases of the causally-neutral prompt (denoted as C0), and on 500 data points for the 5 paraphrases of the C1 prompt, and 5 paraphrases of the C2 prompt.
The causal tracing experiments with LLaMa-7B and Alpaca-7B on 100 data points took around 24 hours each using one GPU V100.

\subsubsection{Implementation Details for Causal Tracing}
\label{psychcausal:appd:causal_tracing}

We introduce the workings of the causal tracing method  \citep{meng2022locating} as follows.
First, we compute the hidden states of the residual stream of LLaMa-7B's layers for two inputs, (1) the original input: the prompt+review, and (2) the corrupted input: prompt + a corrupted version of the review by adding random noise immediately after the token embeddings. Then, we restore one by one the clean state of the residual stream into the corrupted version and measure the effect of the clean state on the probability of the originally predicted token for each token sequence and layer position. 

Since this process is highly time-consuming, taking around 12 hours for 50 samples even using the smallest LLaMa model with 7B parameters, we do a case study on the 7B LLaMa and Alpaca using 100 random samples from the 1K test set. 
For these experiments, we follow the idea of APE \citep{zhou2023large} to use the best-performing prompts on the 1k test set for C1 and C2.
\subsubsection{Prompts}\label{psychcausal:appd:prompt_we_use}

\subsubsubsection{Prompts to Get paraphrases}
Since we need to report the average performance across five paraphrases of the same prompt, for each original prompt, we call GPT to generate the four paraphrases.

Below is the prompt that we used for this paraphrase generation process:
\begin{quote}
You are an expert in prompt engineering for large language models (LLMs). And you are also a native English speaker who writes fluent and grammatically correct text.

Given the following prompt for NLP sentiment analysis, you provide four alternative prompts.

\#\#\#\#\#\#\# Original Prompt \#\#\#\#\#\#\#
\texttt{[Our original prompt]}

\#\#\#\#\#\#\# Alternative Prompt 1 \#\#\#\#\#\#\#

... (Then, we let the model to generate all the way to ``Alternative Prompt 4''.)

\end{quote}

We queried the GPT-4 model with temperature 0\ifarxiv ~on June 8, 2023\fi.

\subsubsubsection{Neutral Prompt}
In addition to the standard prompt to query LLMs in the main paper, we show its four paraphrases in \cref{psychcausal:tab:prompts_paraphrases_c0}.

\begin{table}[ht]
    \centering \small
    \begin{tabular}{p{.9\linewidth}}
\toprule
Prompt Design \\ \midrule 
As a proficient data annotator in natural language processing (NLP), your responsibility is to determine the sentiment of the given review text. Please assign a sentiment value from ``1'' (very negative) to ``5'' (very positive).\newline Review Text: ``\texttt{[review]}'' \newline
Sentiment Score:
\\\hline
As a skilled data annotator in the field of natural language processing (NLP), your task is to evaluate the sentiment of the given review text. Please classify the sentiment using a scale from ``1'' (highly negative) to ``5''  (highly positive).\newline Review Text: ``\texttt{[review]}'' \newline
Sentiment Rating:
\\\hline
As an expert data annotator for NLP tasks, you are required to assess the sentiment of the provided review text. Kindly rate the sentiment on a scale of ``1'' (extremely negative) to ``5'' (extremely positive).\newline Review Text: ``\texttt{[review]}'' \newline
Sentiment Score:
\\\hline
As a proficient data annotator in natural language processing (NLP), your responsibility is to determine the sentiment of the given review text. Please assign a sentiment value from ``1'' (very negative) to ``5'' (very positive).\newline Review Text: ``\texttt{[review]}'' \newline
Sentiment Assesment:
\\
\bottomrule
    \end{tabular}
    \caption{Four additional paraphrases of the neutral prompt (C0) generated with GPT-4.}
    \label{psychcausal:tab:prompts_paraphrases_c0}
\end{table}

\subsubsubsection{Causal Prompts}
In addition to the standard C1 and C2 prompts in the main paper, we show the four paraphrases for each of them in  \cref{psychcausal:tab:prompts_paraphrases_c1,psychcausal:tab:prompts_paraphrases_c2}, respectively.

\begin{table}[ht]
    \centering \small
    \begin{tabular}{p{.9\linewidth}}
\toprule
Prompt Design \\ \midrule 
As a customer sharing my experience, I crafted the following review: ``\texttt{[review]}'' \newline
Taking into account the details of my experience, I chose a star rating from the available options of ``1'',``2'', ``3'', ``4'', or ``5''. My ultimate rating is:
\\\hline
As a client providing my opinion, I penned down the subsequent evaluation: ``\texttt{[review]}'' \newline
Upon thorough reflection of my encounter, I picked a star rating among the choices of ``1'', ``2'', ``3'', ``4'', or ``5''. My conclusive rating stands at:
\\\hline
As a patron expressing my thoughts, I drafted the ensuing commentary: ``\texttt{[review]}'' \newline
After meticulously assessing my experience, I opted for a star rating from the range of ``1'',``2'', ``3'', ``4'', or ``5''. My definitive rating turned out to be:
\\\hline
As a consumer conveying my perspective,  I authored the following assessment: ``\texttt{[review]}'' \newline
By carefully weighing the aspects of my interaction, I determined a star rating from the possibilities of ``1'',``2'', ``3'', ``4'', or ``5''. My final verdict on the rating is:
\\
\bottomrule
    \end{tabular}
    \caption{Four additional paraphrases of the causal prompt C1 generated with GPT-4.}
    \label{psychcausal:tab:prompts_paraphrases_c1}
\end{table}

\begin{table}[ht]
    \centering \small
    \begin{tabular}{p{.9\linewidth}}
\toprule
Prompt Design \\ \midrule
As a customer sharing my experience, I first chose a star rating from the available choices of ``1'',``2'', ``3'', ``4'', or ``5'', and subsequently elaborated on my decision with the following statement: ``\texttt{[review]}'' \newline
The review elucidates the reasoning behind my assigned rating of
\\\hline
As a client providing my opinion, I initially picked a star rating from the range of ``1'' to ``5'', and then proceeded to justify my selection with the following commentary: ``\texttt{[review]}'' \newline
The review sheds light on the rationale for my given rating of
\\\hline
As a patron expressing my thoughts, I started by selecting a star rating from the scale of ``1'' to ``5'', and then offered an explanation for my choice in the following review text: ``\texttt{[review]}'' \newline
The review expounds on the basis for my designated rating of
\\\hline
As a consumer conveying my perspective,  I began by opting for a star rating within the ``1'' to ``5'' spectrum, and then detailed my reasoning in the subsequent review passage: ``\texttt{[review]}'' \newline
The review delineates the grounds for my conferred rating of
\\
\bottomrule
    \end{tabular}
    \caption{Four additional paraphrases of the causal prompt C2 generated with GPT-4.}
    \label{psychcausal:tab:prompts_paraphrases_c2}
\end{table}

\subsection{Additional Experimental Results}
\subsubsection{Few-Shot Results}
For reproducibility and controllability, we use the zero-shot prompting setting across the experiments in the main paper, to avoid randomness in few-shot prompting according to which examples are selected as the few shots, and the order of the examples.

As a supplementary information in case this is of some readers' interest, we provide the few-shot prompting results in \cref{psychcausal:tab:prompt_c0_fewshot,psychcausal:tab:prompt_c12_fewshot}.
\begin{table}[ht]
    \centering \small
    \begin{tabular}{llccccccccc}
    \toprule
 && Random 
 & GPT-3 Few-Shot\\ \midrule
\multirow{3}{*}{F1} & Overall & 19.82 {\tiny$\pm$2.07}
& 63.35 {\tiny$\pm$0.80}
    \\
& C1 Subset  & 21.36 {\tiny$\pm$2.26} 
& 54.44 {\tiny$\pm$1.24}  \\
& C2 Subset &  20.43 {\tiny$\pm$2.95} 
& 75.65 {\tiny$\pm$0.45} \\ \hline
\multirow{3}{*}{Acc} & Overall &19.78 {\tiny$\pm$2.07} 
& 64.14 {\tiny$\pm$0.86}
\\
& C1 Subset & 20.61 {\tiny$\pm$2.23} 
& 54.22 {\tiny$\pm$1.28} \\
& C2 Subset &  18.86 {\tiny$\pm$2.78} 
& 75.18 {\tiny$\pm$0.53}\\

    \bottomrule
    \end{tabular}
    \caption{Few-shot performance of the standard SA prompts on Yelp-5. We use five paraphrases for the prompt, and report the average performance with the standard deviation.
    }
    \label{psychcausal:tab:prompt_c0_fewshot}
\end{table}
\begin{table}[ht]
    \centering \small
    \setlength\tabcolsep{4.8pt}
    \begin{tabular}{llcccccccccccc}
\toprule
&&Random 
& GPT-3 Few-Shot\\ \midrule
\multirow{4}{*}{F1} 
& Data=C1, Prompt=C1 &20.47 {\tiny$\pm$2.47} 
& 49.18 {\tiny$\pm$0.76} \\
& Data=C1, Prompt=C2 &20.26 {\tiny$\pm$2.31} 
& 52.79 {\tiny$\pm$2.64}\\
& Data=C2, Prompt=C1 & 22.35 {\tiny$\pm$3.02} 
& 80.46 {\tiny$\pm$1.29} \\
& Data=C2, Prompt=C2 & 20.35 {\tiny$\pm$2.18} 
& 75.88 {\tiny$\pm$1.86}\\ \hline

\multirow{4}{*}{Acc}
& Data=C1, Prompt=C1 &19.60 {\tiny$\pm$2.67} 
& 50.12 {\tiny$\pm$0.71}  \\
& Data=C1, Prompt=C2 & 19.68 {\tiny$\pm$2.46} 
& 53.83 {\tiny$\pm$2.46}\\
& Data=C2, Prompt=C1 & 20.97 {\tiny$\pm$3.19} 
& 81.21 {\tiny$\pm$1.17}\\
& Data=C2, Prompt=C2 & 19.03 {\tiny$\pm$2.18}
& 76.36 {\tiny$\pm$1.53}\\
\bottomrule
    \end{tabular}
    \caption{
    Few-shot performance on Yelp using two different causal prompts on the two causal subsets. We use five paraphrases for each prompt, and report the mean performance with the standard deviation.
    }
    \label{psychcausal:tab:prompt_c12_fewshot}
\end{table}

\subsubsection{$\lambda_1$-$\lambda_2$ Distribution Plot}\label{psychcausal:appd:distr_plot}

\begin{figure}[ht]
    \centering
    \includegraphics[width=0.45\textwidth]{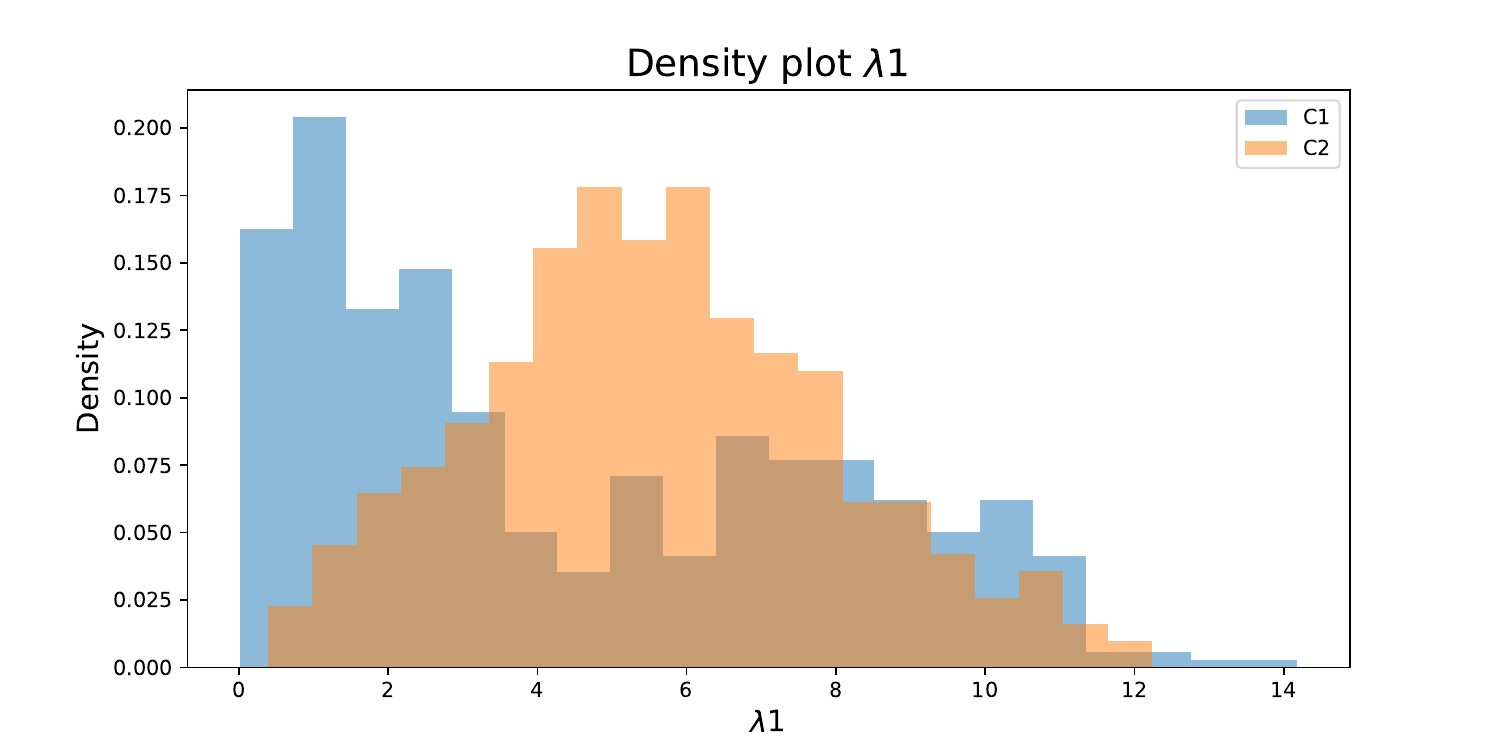}
    \\
    \includegraphics[width=0.45\textwidth]{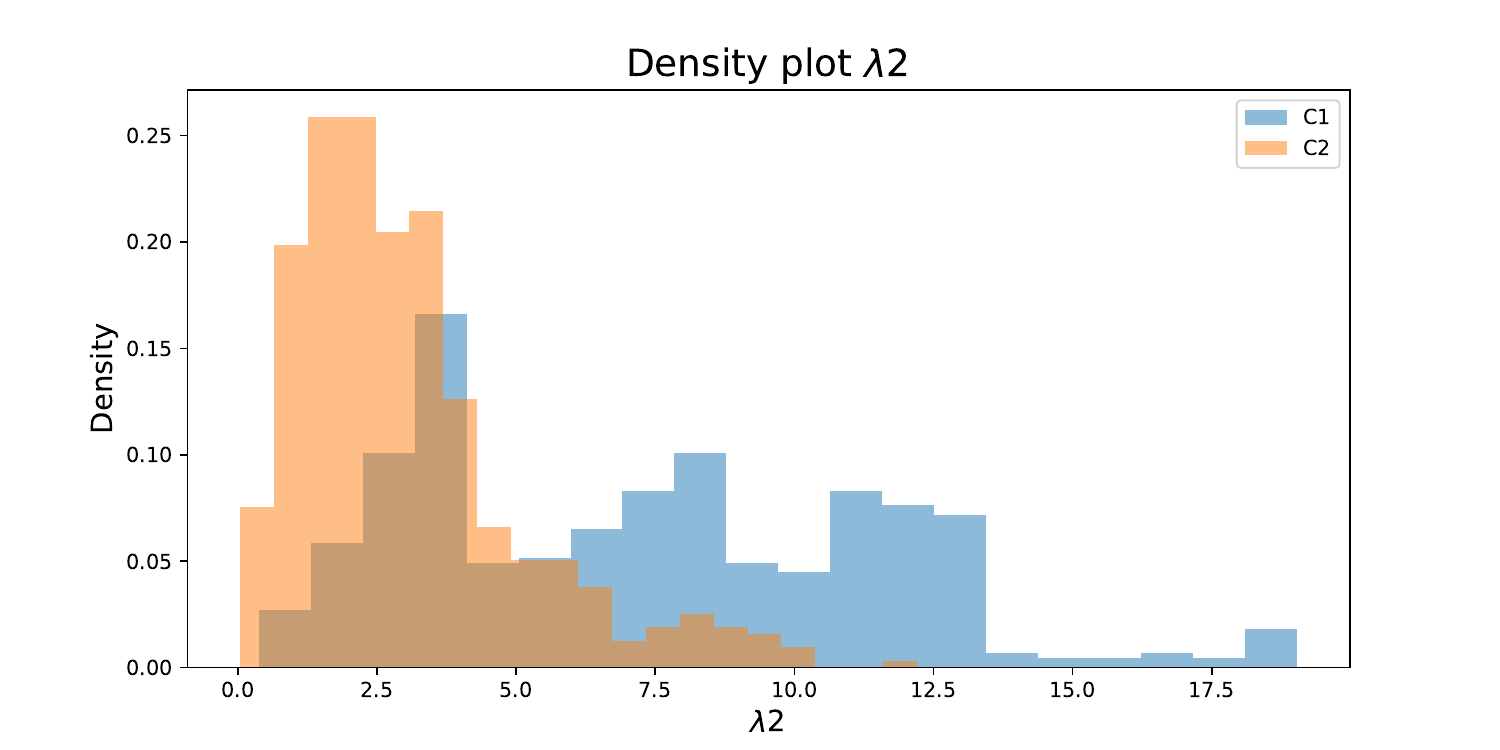}
    \caption{The $\lambda_1$-$\lambda_2$ density plots of C1 (above) and C2 (below). 
    }
    \label{psychcausal:fig:density_lambda}
\end{figure}

To provide a clear understanding of the distributions of \(\lambda_1\) and \(\lambda_2\), we include their density plots of the causal processes C1 and C2 in \cref{psychcausal:fig:density_lambda}.
The mean values of \(\lambda_1\) and \(\lambda_2\) for each group are in \cref{psychcausal:tab:mean}.
\begin{table}[ht]
\centering
\begin{tabular}{|c|c|c|}
\hline
 & \textbf{C1} & \textbf{C2} \\ 
\hline
$\mu(\lambda_1)$ & 4.48 & 5.62 \\ 
\hline
$\mu(\lambda_2)$ & 7.31 & 3.02 \\ 
\hline
\end{tabular}
\caption{Mean values of the lambdas for C1 and C2.}
\label{psychcausal:tab:mean}
\end{table}

Further, we performed the Mann-Whitney U rank test to determine if the underlying distributions of \(\lambda_1\) and \(\lambda_2\) for groups C1 and C2 are the same. The results are as follows:
\begin{itemize}
    \item For \(\lambda_1\), the p-value is \(8.4572 \times 10^{-71}\), leading us to reject the null hypothesis that the two groups come from the same distribution.
    \item For \(\lambda_2\), the p-value is \(1.36138 \times 10^{-11}\), also leading us to reject the null hypothesis that the distributions are the same.
\end{itemize}

These statistical results indicate significant differences between the distributions of \(\lambda_1\) and \(\lambda_2\) across the causal process groups, which indicate distinct underlying characteristics in the sentiment dynamics of the two groups.
\begin{figure}[ht]
    \centering
    \includegraphics[width=0.3\textwidth]{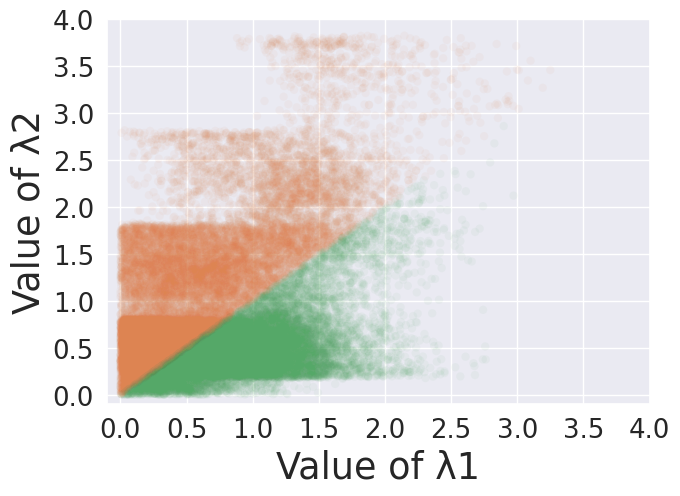}
\hfill
    \includegraphics[width=0.3\textwidth]{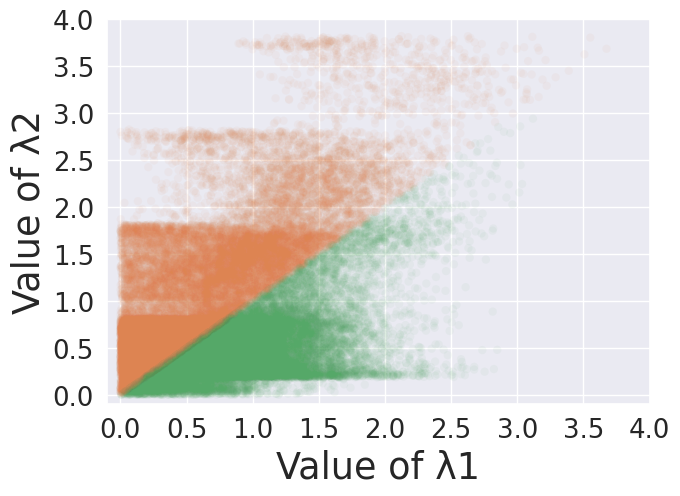}
\hfill
    \includegraphics[width=0.3\textwidth]{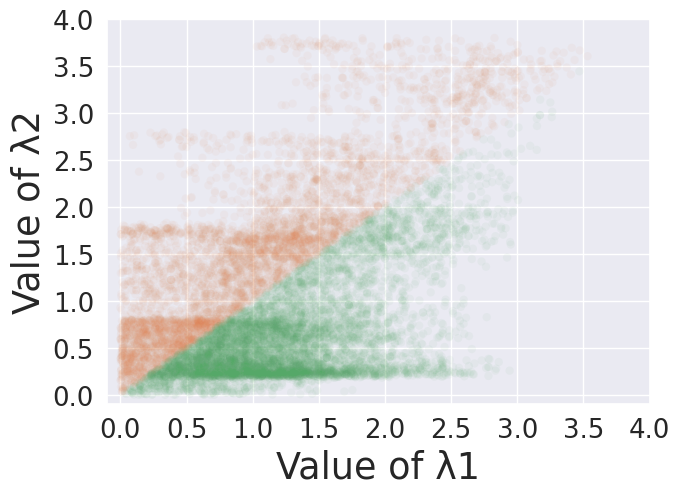}
    \caption{The $\lambda_1$-$\lambda_2$ plot on Yelp-5 (left), Amazon  (middle), and App Review (right). 
    We draw the $y=x$ diagonal line, and 
    the orange dots in the upper-left triangle represent the C1-dominant subset,
    and green dots in the lower-right triangle are the C2-dominant subset.
    }
    \label{psychcausal:fig:lambda_plot}
\end{figure}

\ifarxiv

\subsection{Emotion Arc Clustering}
We analyze the emotional arc patterns of Yelp reviews. \citet{reagan2016emotional} identified 6 basic emotional arc shapes in stories. However, reviews are usually shorter and therefore present fewer variations. We take each sentence of the review and predict its sentiment. Then we divide the review into ten bins and compute an average sentence sentiment for each decile to make reviews with different lengths comparable. Reviews shorter than 10 sentences generate null values for some deciles which we fill with the information of the next decile.
In \cref{psychcausal:fig:clusters}, we illustrate the 4 clusters found, they have the following characteristics:

\paragraph{Positive + Early Rise:}
This cluster primarily comprises highly positive reviews, where customers express satisfaction and praise for their overall experience. Interestingly, 21.1\% of these reviews begin with a negative first sentence, which often indicates initially low expectations or a negative first impression. However, despite the initial negativity, the reviews tend to turn positive as customers elaborate on their positive experiences.

\paragraph{Negative + Early Fall:}
This cluster mainly consists of predominantly negative reviews. Similarly to the Positive cluster, some reviews (28.14\%) start with a sentence with the opposite sentiment, usually indicating high expectations followed by disappointment.

\paragraph{Rise:}
The main characteristic of this cluster is the positive ending of the review, despite the initial negativity observed in the first half, with an average sentiment of -1.63. An important fraction of the reviews in this cluster (52.49\%) start with a positive comment as a summary, but then proceed to highlight the negative aspects of the experience. Despite the initial criticisms, the reviews conclude with positive points, suggesting that the overall experience was still satisfactory.

\paragraph{Fall:}
In contrast to the previous cluster, the Fall cluster is characterized by a negative ending of the review, despite a generally positive first half with an average sentiment of  2.18. An important proportion (36\%) of the reviews in this cluster begin with a negative comment as a summary, but then proceed to describe the positive aspects before eventually highlighting the negative ones. This cluster showcases a shift in sentiment from positive to negative, indicating a decline in satisfaction as the review progresses.
\begin{table}[t]
    \centering \small
    \begin{tabular}{p{.95\textwidth}}
    \toprule
    \textit{\textbf{Positive + Early Rise}} \\
    \textbf{Review:} \textit{Was there last Friday. Seats right in front if the stage. The show was good. The headliner, while a bit long, was good. Fantastic service from our waitresses. Will definitely go back.} \\
    \textbf{Review:} \textit{This is by far my favorite Panera location in the Pittsburgh area. Friendly, plenty of room to sit, and good quality food \& coffee. Panera is a great place to hang out and read the news - they even have free WiFi! Try their toasted sandwiches, especially the chicken bacon dijon.} \\
    \hline
    \textit{\textbf{Negative + Early Fall}} \\
    \textbf{Review:} \textit{Pass on this place, there are better restaurants mere feet away.\newline The menu here is too large, which is a sure sign none of the food is going to be good.  And, its not good.  Some of the salads are alright, but its just not good food. \newline The service is friendly and prompt, but the beer is over priced. They do have a good selection though. \newline This place is open late if you need a bite to eat, but there are so much better options out there.} \\
    \textbf{Review:} \textit{Wings are overpriced. And the quality of them are bad. They were tough and greasy. The staff are pleasant but then over all experience was too expensive for a sports bar.} \\
    \hline
    \textit{\textbf{Rise}} \\
    \textbf{Review:} \textit{To be honest, I feel that this is one of the most overpriced restaurants in the entire city. The food is average to good, the place is beautiful with outdoor seating, but in my opinion the price is just not worth it. They have a really good happy hour, so I would definitely recommend going to that and maybe trying an appetizer or two.} \\
    \textbf{Review:} \textit{The first time I came here, I waited in line for 20 minutes.  When it was my turn, I realized I left my wallet in the car.  It hurt so bad, I didn't come back for a year. \newline I can walk to this place from my house- which is dangerous because those biscuits are just OH SO DREAMY.  I can't describe them. Just get some.\newline Do I feel guilty about noshing on fabulous Strawberry Napoleons and Jewish Pizza (kind of like a modified, yet TOTALLY delicious fruitcake bar) at 10:15am?  Hecks, naw... But they do have quiche and some other breakfast-y items for those who prefer a more traditional approach to your stomach's opening ceremony. \newline Just go early :)  They open at 10 on Saturdays.  And bring cash...it's easier that way.} \\
    \hline
    \textit{\textbf{Fall}} \\
    \textbf{Review:} \textit{It's cheap, I'll say that, but otherwise it's bland food served by workers who mostly don't seem to notice they're working, and when they do, only respond snarkily. There are many better vegetarian and vegan options to choose from} \\
    \textbf{Review:} \textit{I do like my Mad Mex, however predictable and non-authentic it may be.  The portion sizes are mammoth and I come away with a satisfied sense of regret.  Their beer menu is happily extensive. Charging me \$9 for chips and salsa is a bit of crime, wouldn't ya say though!?!  I mean, c'mon!  Our service has most times been lacking--a bit rushed and on the inattentive side.  Also, why do you require your wait staff to not servestraws/ lemons/etc unless asked by cusotmers---weirdness-cut out these odd cost-cutting, anti-service friendly measures please} \\
    \bottomrule
    \end{tabular}
    \caption{Example reviews for each emotion arc cluster.}
    \label{psychcausal:tab:cluster_examples}
\end{table}
\FloatBarrier

\begin{figure}[ht]
    \centering
    \includegraphics[width=\columnwidth]{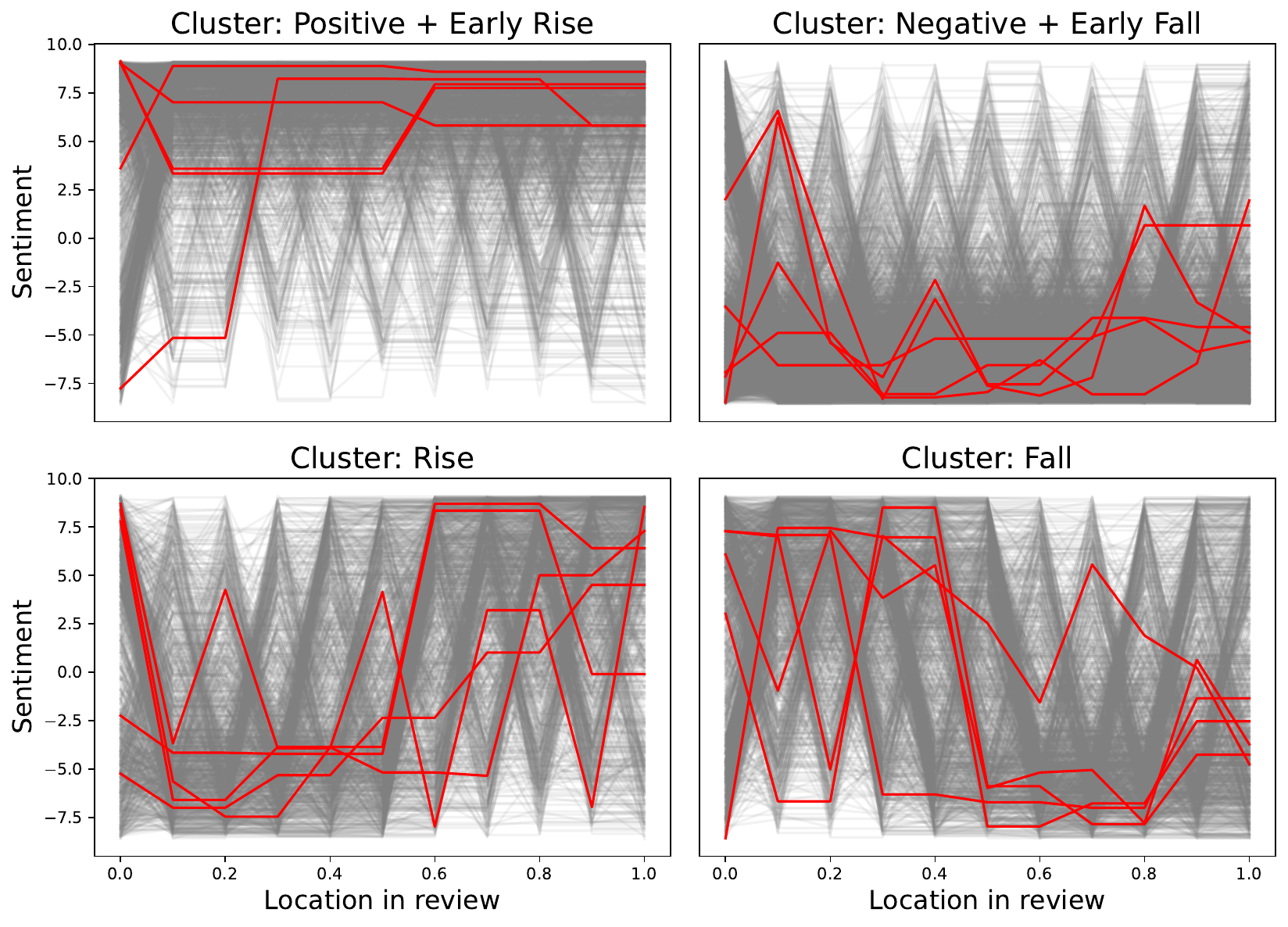}
    \caption{Four emotion arc clusters.}
    \label{psychcausal:fig:clusters}
\end{figure}

\fi
\subsection{Additional Interpretability by Shapley Values}
\begin{figure}[ht]
    \centering
    \includegraphics[width=\linewidth]{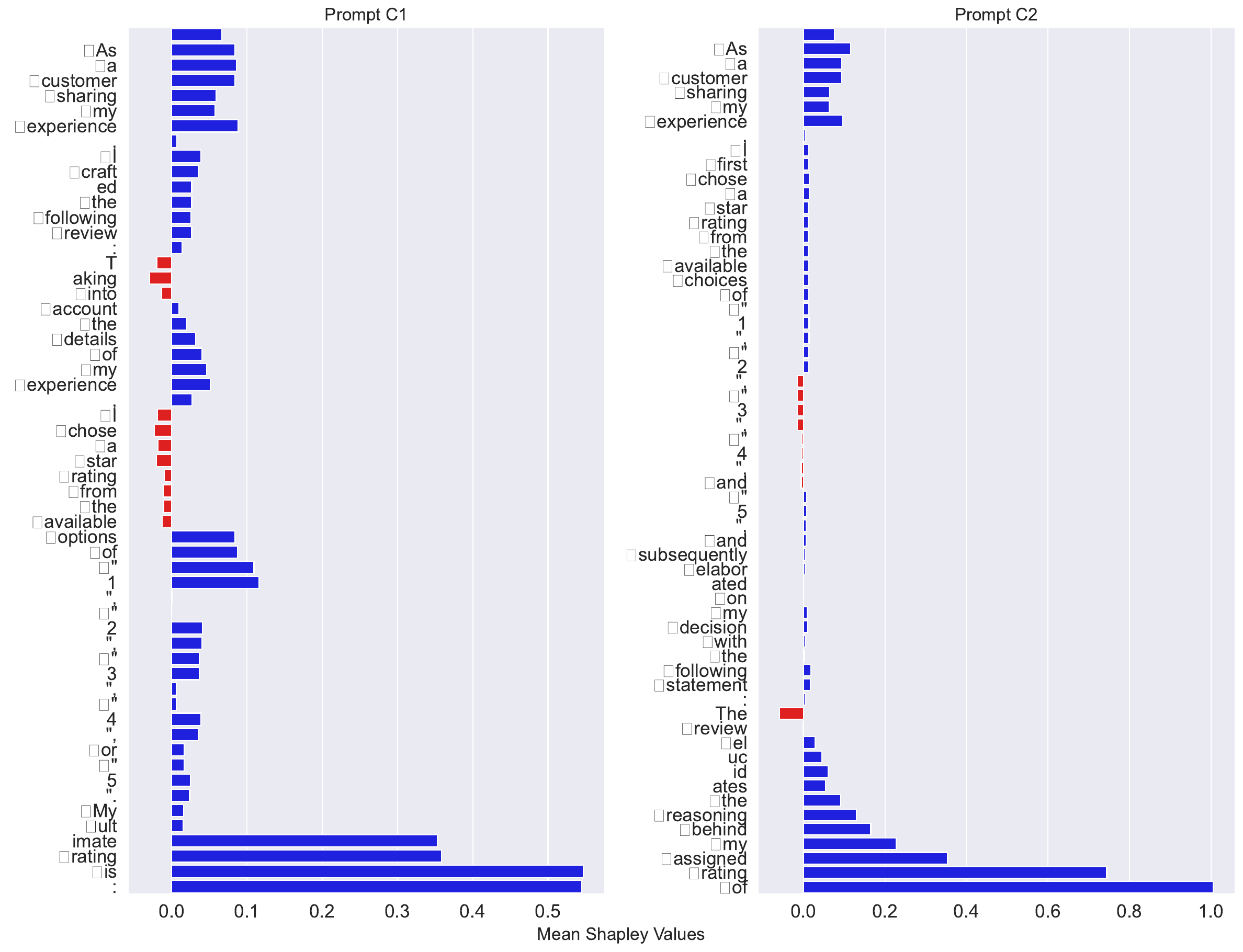}
    \caption{Shapley values for the two types of prompts.}
    \label{psychcausal:fig:shapley_values}
\end{figure}

\begin{figure}[ht]
    \centering
    \includegraphics[width=\linewidth]{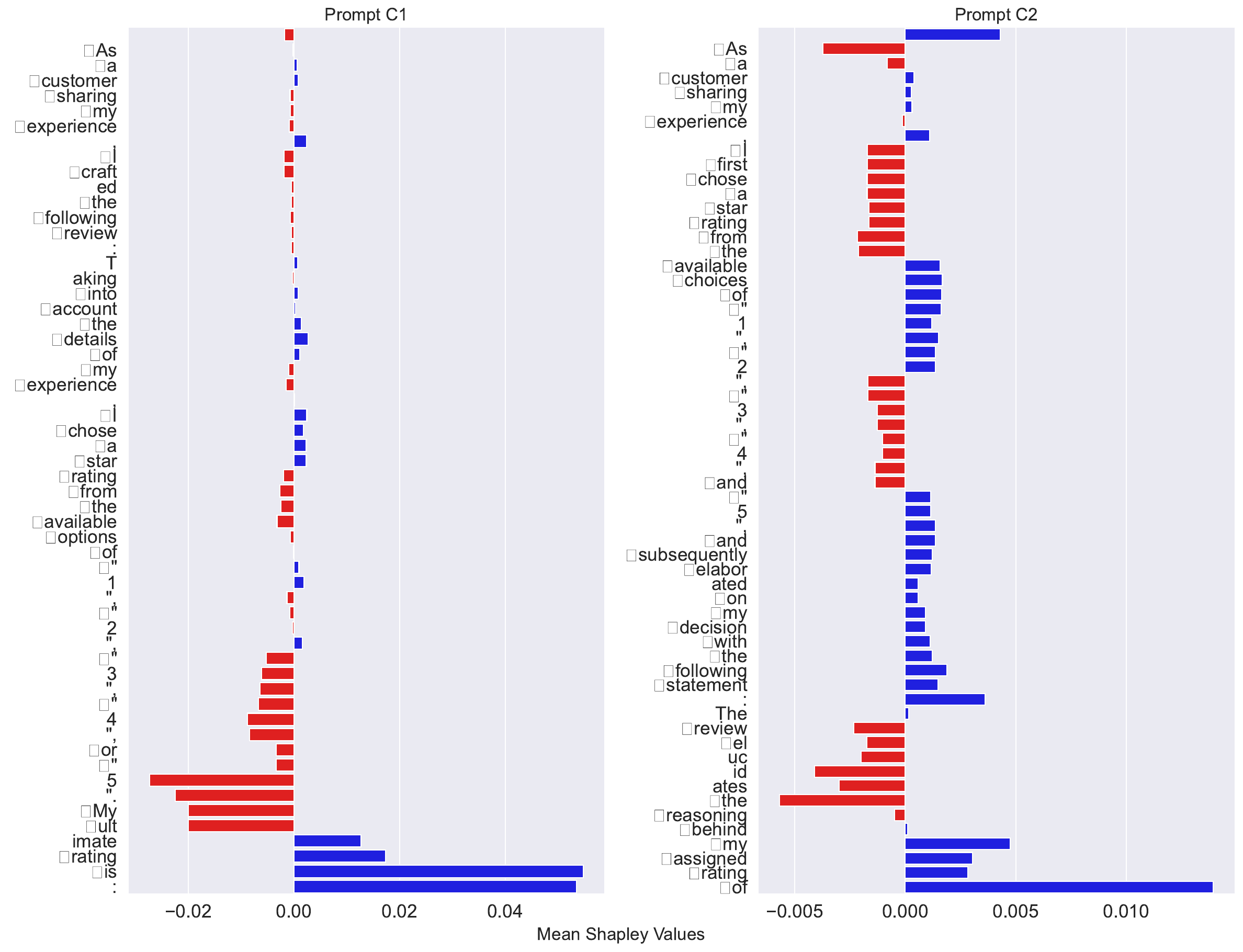}
    \caption{Adjusted shapley values for the two types of prompts.}
    \label{psychcausal:fig:shapley_values_norm}
\end{figure}

We further analyze the effect of each part of the prompts on LLaMa's predictions. Using 50 reviews, we compute the shapley values of each token. In \cref{psychcausal:fig:shapley_values} we observe that the tokens with the largest shapley values are the ones in the end, which is expected since they are the ones helping to form a grammatically correct sentence. To account for that, we subtracted the average shapley values computed for the other possible start rating answers. In \cref{psychcausal:fig:shapley_values_norm} we show the adjusted shapley values. We observe that the tokens in prompt C1 have a larger effect than the tokens in prompt C2. The words introducing the review have a positive effect on C2 but a negative one on C1. Whereas, the phrase ``I chose a star rating'' has a negative effect on C2 but a positive one on C1.

\FloatBarrier

\newpage
\section{Additional Materials for \cref{ch:covidtwitter}}

\subsection{Statistics of our Data}
\subsubsection{COVID Twitter Keywords} \label{covidtwitter:sec:twitter_keyword_chen}
We list the COVID-related Twitter keywords and accounts tracked by \citet{chen2020tracking} in \cref{covidtwitter:tab:chen_twit_keywords,covidtwitter:tab:chen_twit_accounts}. They are used to retrieve the 1.01TB raw Twitter data.
\begin{table}[ht]
    \small
    \centering
    \begin{tabular}{l|lll}
\toprule    
\multicolumn{2}{c}{\textbf{Keywords used by \citet{chen2020tracking}}} \\ \midrule
14DayQuarantine & covidiot \\
CDC & epitwitter \\
COVD & flatten the curve \\
COVID\_\_19 & flattenthecurve \\
COVID-19 & kung flu \\
China & lock down \\
Corona & lockdown \\
Coronavirus & outbreak \\
Coronials & pandemic \\
DontBeASpreader & pandemie \\
DuringMy14DayQuarantine & panic buy \\
Epidemic & panic buying \\
GetMePPE & panic shop \\
InMyQuarantineSurvivalKit & panic shopping \\
Koronavirus & panic-buy \\
Kungflu & panic-shop \\
N95 & panicbuy \\
Ncov & panicbuying \\
PPEshortage & panicshop \\
Sinophobia & quarantinelife \\
Social Distancing & quarentinelife \\
SocialDistancing & saferathome \\
SocialDistancingNow & sars-cov-2 \\
Wuhan & sflockdown \\
Wuhancoronavirus & sheltering in place \\
Wuhanlockdown & shelteringinplace \\
canceleverything & stay at home \\
china virus & stay home \\
chinavirus & stay home challenge \\
chinese virus & stay safe stay home \\
chinesevirus & stayathome \\
corona virus & stayhome \\
coronakindness & stayhomechallenge \\
coronapocalypse & staysafestayhome \\
covid & trump pandemic \\
covid-19 & trumppandemic \\
covid19 & wear a mask \\
covididiot & wearamask \\
\bottomrule
    \end{tabular}
    \caption{Keywords used by \citet{chen2020tracking} to track COVID-related tweets.}
    \label{covidtwitter:tab:chen_twit_keywords}
\end{table}
\FloatBarrier

\begin{table}[ht]
    \small
    \centering
    \begin{tabular}{l|l}
    \toprule
    \multicolumn{2}{c}{\textbf{Accounts tracked by \citet{chen2020tracking}}} \\ \midrule
PneumoniaWuhan & WHO \\
CoronaVirusInfo & HHSGov \\
V2019N & NIAIDNews \\
CDCemergency & DrTedros \\
CDCgov \\
\bottomrule
    \end{tabular}
    \caption{Accounts tracked by \citet{chen2020tracking} to retrieve COVID-related tweets.}
    \label{covidtwitter:tab:chen_twit_accounts}
\end{table}

\subsubsection{Annotation Guidance for Policy Strictness}\label{covidtwitter:appd:annot_policy}

For each state, the annotators are asked to go to the official website that lists all COVID policies of the state. In most cases, the website lists all executive orders (EOs), proclamations, or other forms of policies issued during 2020 -- 2021. Then the annotator is asked to read through the EOs that are related to COVID social distancing policies. For each relevant policy, the annotator is asked to record the start date on which the policy will take effect,\footnote{For consistency, we record 0:01am of the first effective date, but not the 11:59pm of the previous day.} a brief intro of what kind of social distancing policy it is, and a real-valued score in the range of 0 (loosest) to 5 (strictest).

For the scoring criteria, we provide the following guides:

\begin{itemize}[nolistsep]
    \item Score 0: masks are optional, open the schools,, bars, gaming facilities, concert, and almost everything
    \item Score 1: State of emergency, limit gathering, close K-12
    \item Score 2: Open 50\% capacity for retail business, open religious activities like churches to 50\%
    \item Score 3: Open 25\% capacity for retail businesses
    \item Score 4: Open only business for necessities such as supermarkets, only allow delivery and curbside services, gatherings have to be no more than 10 people
    \item Score 5: Strict stay at home policy, close every business
\end{itemize}

\subsubsection{Accuracy of Twitter Sentiment Classifier}\label{covidtwitter:appd:bert_accuracy}
We list the detailed performance report of TextBlob and our COVID BERT in \cref{covidtwitter:tab:accuracy}, including the overall accuracy, weighted and macro F1 scores, precision and recall for each class, and MSE of the average sentiment of random groups of 20 tweets. Note that since TextBlob predicts a real-valued number in the range of -1 to 1 for the sentiment, we regard [-1, -0.33) as negative, [-0.33, 0.33] as neutral, and (0.33, 1] as positive.
\begin{table*}[ht]
    \centering
    \small
    \setlength{\tabcolsep}{3pt}
    \begin{tabular}{lccccccccc|c}
    \toprule
    Model & Acc & \multicolumn{2}{c}{F1 Score} & \multicolumn{2}{c}{Positive} & \multicolumn{2}{c}{Neutral} & \multicolumn{2}{c|}{Negative} & MSE on Groups \\
    & & Weighted & Macro & P & R & P & R & P & R & \\ \hline
    TextBlob & 23.35 & 16.67 & 19.70 & 20.34 & 10.62 & 20.67 & 85.19 & 74.07 & 6.45 & 0.43 %
    \\
    COVID BERT & 60.23 & 62.31 & 55.17 & 51.19 & 76.11 & 26.76 & 35.51 & 83.68 & 62.99 & 0.15 %
    \\
    \bottomrule
    \end{tabular}
    \caption{The detailed performance report of the TextBlob baseline, and our COVID BERT model. We report the overall accuracy (Acc), weighted and macro F1 scores, precision (P) and recall (R) for each class, and MSE of the average sentiment of random groups of 20 tweets.}
    \label{covidtwitter:tab:accuracy}
\end{table*}

\subsection{Additional Analyses}\label{covidtwitter:appd:analyses}
\subsubsection{Correlation across All Variables}
We can see that, averaging over all 50 states, unemployment correlates the most with policy changes, which is consistent with our analysis in \cref{covidtwitter:sec:linear_reg}. Since different states may have different styles to take sentiment into consideration when making policies, the effect of sentiment on policy changes over all 50 states is relatively mild.

For Twitter sentiment, it correlates largely with case numbers, and urbanization rate of the state. 

Interestingly, the case numbers correlate with whether the state governor is a political ally of Trump.
\begin{figure}[ht]
    \centering
    \includegraphics[width=\columnwidth]{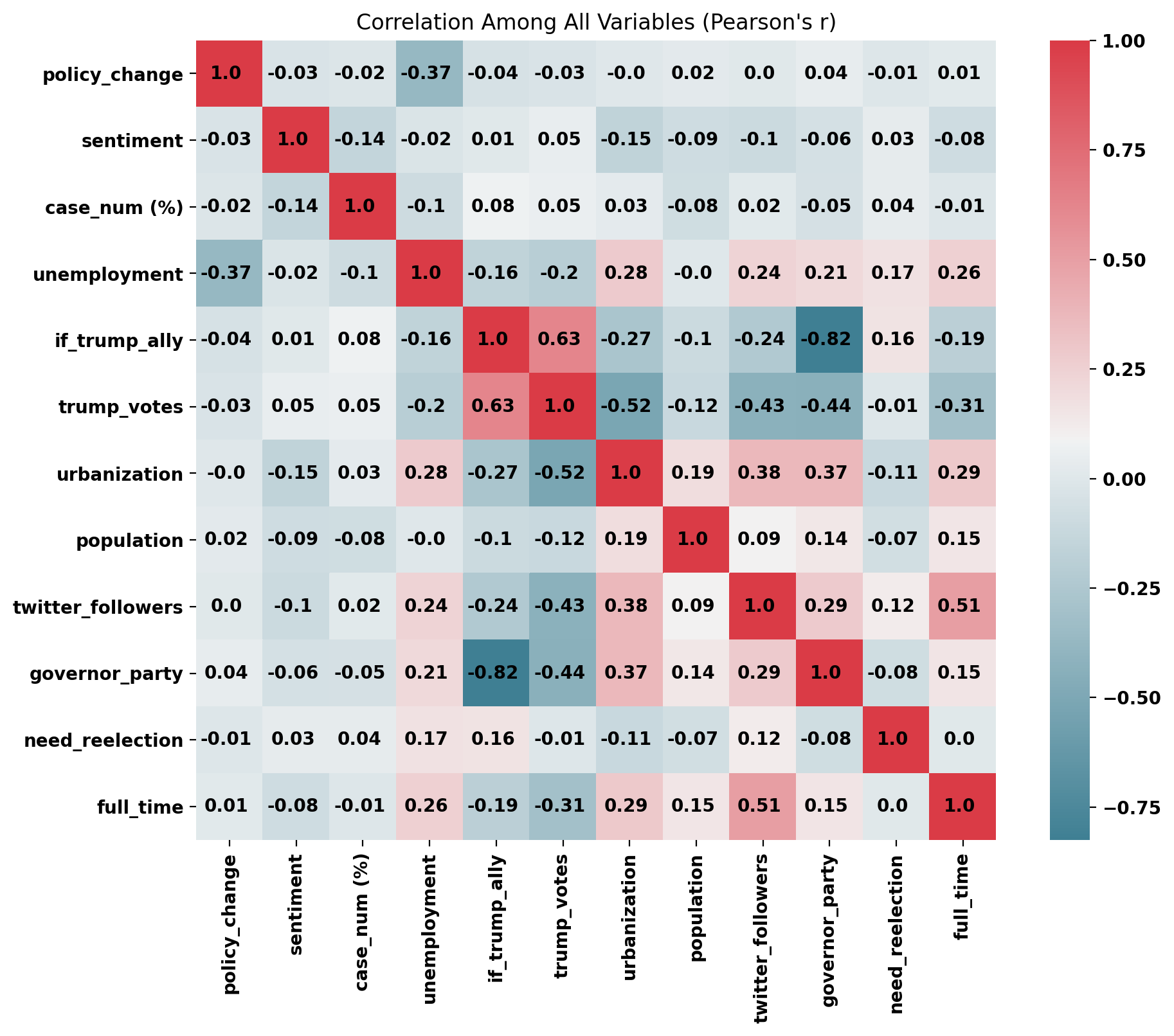}
    \caption{Correlation across all variables.}
    \label{covidtwitter:fig:var_all_all}
\end{figure}

\subsubsection{Alternative Causal Analysis Methods by Potential Outcomes Framework}

There are two commonly used frameworks for causal inference, one is the do-calculus we introduced in \cref{covidtwitter:sec:do_calc}, and the other is the potential outcomes framework \citep{rubin1974estimating,rubin2005causal,imbens2015causal}. We will introduce two alternative causal inference methods on our problem, using the potential outcomes framework.
\paragraph{Difference-in-Differences}
One possible limitation of this study is that we treat the data in an i.i.d. way, following most existing studies. An improvement is to treat it as time series. 
For time series analyses, one commonly used method is the first-difference (FD) estimator, difference in differences (DID) \citep{abadie2005semiparametric}.
Specifically, DID takes in the time series data of the cause $X$, effect $Y$, and confounders $Z$, and solves the following regression:
\begin{align}
    \Delta Y &= \beta \cdot \Delta X + \Delta Z
    \\
    Y_t - Y_{t-1} &= \beta (X_t - X_{t-1}) + Z_t - Z_{t-1}
    ~,
\end{align}
where $t$ is the time step, and $\beta$ is the causal effect of $X$ on $Y$. 

After applying DID on all the policies, we obtain $\beta$ scores for all states, and the top 5 states with largest $\beta$ are Colorado ($\beta=0.67$), Kentucky ($\beta=0.23$), Wyoming ($\beta=0.22$), Oregon ($\beta=0.19$), North Carolina ($\beta=0.17$), Michigan ($\beta=0.14$), and New York ($\beta=0.13$). 

\begin{figure}[ht]
\centering
\begin{subfigure}{.48\columnwidth}
  \centering
  \includegraphics[width=\linewidth]{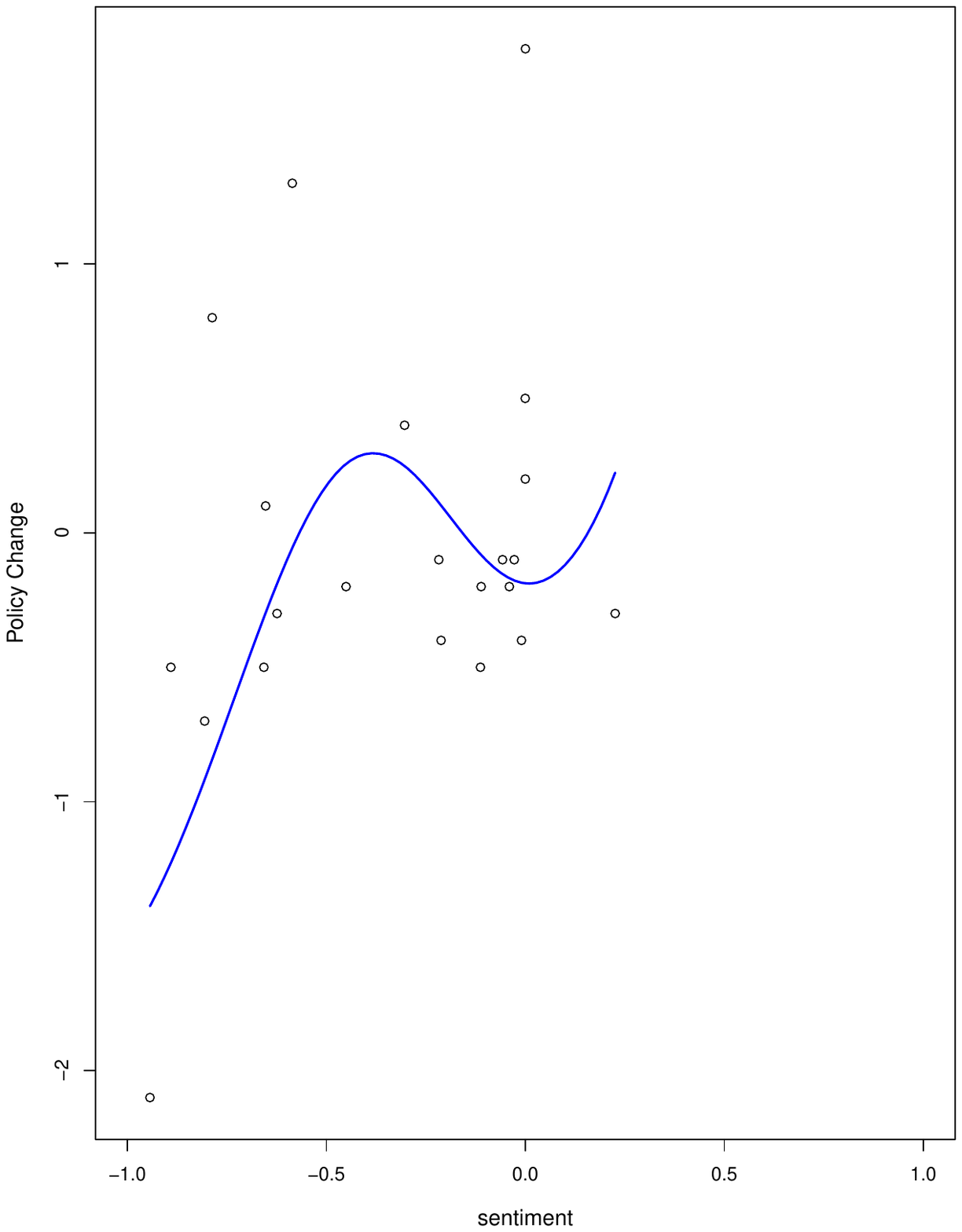}
  \caption{Model of TX.}

\end{subfigure}
\hfill
\begin{subfigure}{.48\columnwidth}
  \centering
  \includegraphics[width=\linewidth]{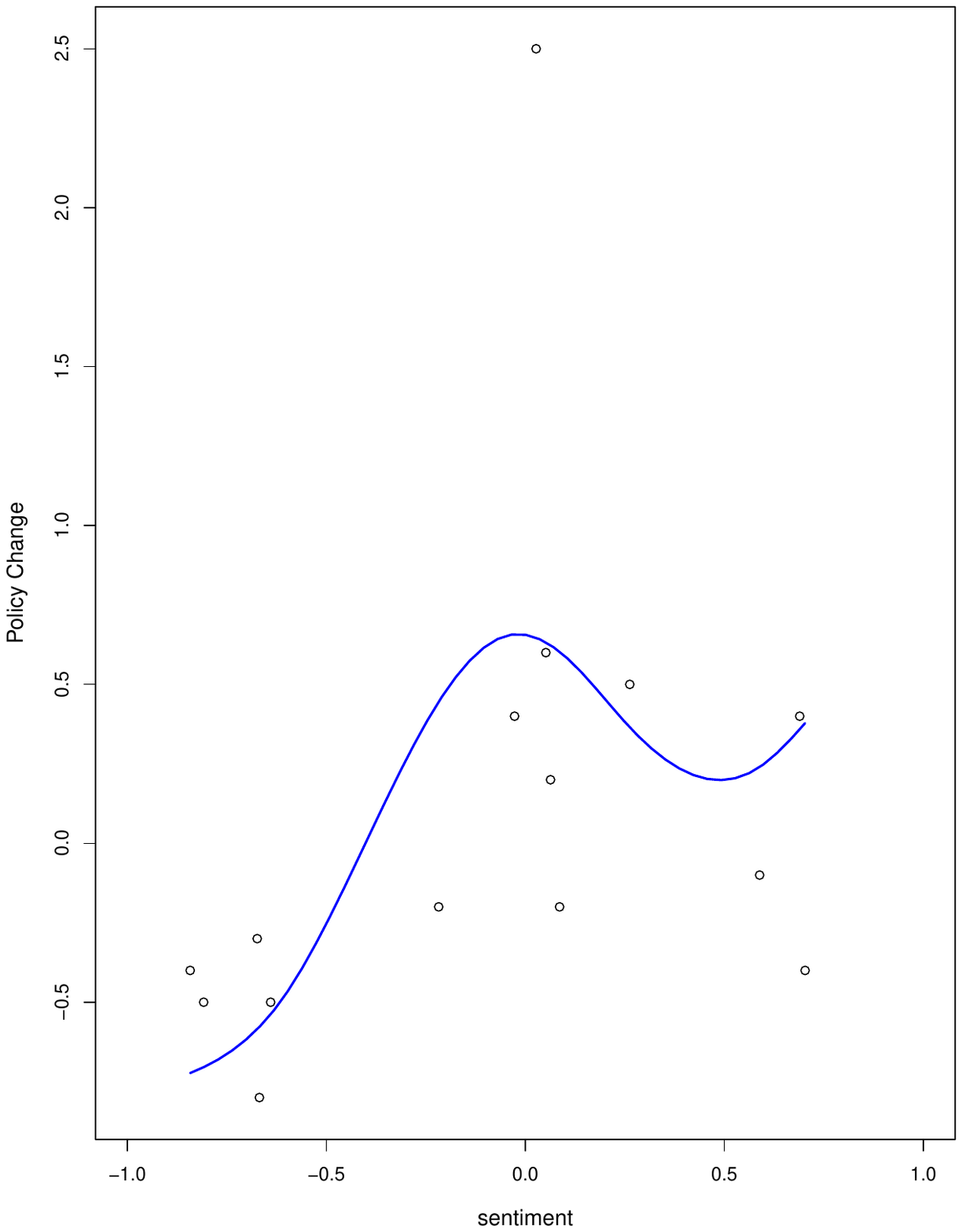}
  \caption{Model of MI.}
\end{subfigure}
\caption{Causal models by continuous-valued propensity score matching of TX and MI.}
\label{covidtwitter:fig:propensity}
\end{figure}

\paragraph{Continuous-Valued Propensity Score Matching}
Another commonly used alternative for causal inference is propensity score matching. However, the challenge in our study is that the cause is not categorical, but takes continuous values. To this end, we follow the extension of propensity score matching to continuous treatment \citep{hirano2004propensity,bia2008stata}. We adopt the stata package of \citet{bia2008stata} for continuous-valued propensity score matching. The resulting prediction of policies based on Twitter sentiment is a polynomial function with an order of three. As examples, We show the predictions of Texas (TX) and Michigan (MI) in \cref{covidtwitter:fig:propensity}.

\newpage
\section{Additional Materials for \cref{ch:causalcite}}

\ifperfect

\subsection{More Method}
\subsubsection{assumptions}
Below are some assumptions  \citep{rubin1980, roberts2020adjusting}.

\begin{assumption}
[Stable Unit Treatment Value Assumption (SUTVA) \citep{rubin1980}]\
For all individuals $i$, $S_i(T) = S_i(T_i)$
\end{assumption}

\begin{assumption}[Conditional Ignorability]
For all individuals $i$,
$T_i \independent S_{i}(T=1), S_{i}(T=0) \vert \bf{Z}$, where $\bf{Z}$ is the set of confounders.
In the language of \it{SCM}, this assumption implies no unobserved confounders that affect both the treatment assignment and the outcome.
\end{assumption}

\begin{assumption}[Positivity]
For all individuals i, $P(T_i = t) > 0$ for all treatments $t$. 
\end{assumption}

\subsubsection{More Algorithm}\label{causalcite:sec:alg}

\begin{algorithm}[ht]
    \caption{Algorithms for RSCI and ITCI}
    \label{causalcite:alg:overview}
    \begin{algorithmic}[1]
        \State \textbf{Input}: Paper $X$
        \State $P \gets $ All its children, namely the receiver paper set
        \State $Q \gets$ Get all its non-descent paper set Q
        \Statex
        \hrulefill
        \Procedure{get\_RSCI}{$X, Y$}
            \State $year_Y \gets$ Publication year of paper $Y$
            \State $s_Y \gets$ Citations of $Y$
            \State $t_Y \gets$ Text embedding of $Y$'s title and abstract
            \State $Q' \gets$ All the papers in $Q$ which share the same $year_Y$ \label{causalcite:alg:year_filter}
            \State $total\_sim \gets 0$
            \For {each $q_i$ in $Q'$}
                \State $s_i \gets$ Citations of $q_i$
                \State $t_i \gets$ Embedding of $q_i$'s title and abstract
                \State $sim_i \gets$ Cosine similarity of $t_i$ and $t_Y$
                \State $total\_sim \gets total\_sim + sim_i$
            \EndFor
            \State $control \gets 0$
            \For{$q_i$ in $Q'$}
                \State $w_i \gets \frac{sim_i}{total\_sim}$
                \State $control \gets control + w_i \cdot s_i$
            \EndFor
            \State \textbf{return} $s_Y - control$
        \EndProcedure
        \Statex    
        \hrulefill
        \Procedure{get\_ACE}{$X$}
            \State $ACE \gets 0$
            \For {each $p_i$ in $P$}
                \State $I_i \gets \text{get\_RSCI}(X, p_i)$
                \State $ACE \gets ACE + \frac{1}{|P|} \cdot I_i$
            \EndFor
            \State \textbf{return} $ACE$
        \EndProcedure
    \end{algorithmic}
\end{algorithm}

\begin{algorithm}[ht]
    \caption{Algorithm for \ourname}\label{causalcite:alg:two}
    \begin{algorithmic}[1]
        \For{$paperA$ in allPapers}
            \If{numChildrenOfPaperA != 0}
                \State paperB = getRandomChild (children\_of\_paperA)
                \State candidatePool = df [ df [ "year" ] == paperB\_year]
                \State get\_closest\_papers = BM25( candidatePool, paperB)
                \State encoded\_paperB = model (paperB\_abstract)
                \For{\_candidate\_paper in get\_closest\_papers}
                    \State descendants\_of\_paperA = BFS(CitationGraph, paperA)
                    \If{\_candidate\_paper not in descendants\_of\_paperA}
                        \State encoded\_candidate\_paper = model(\_candidate\_paper)
                        \State similarity\_score = cosineSimilarity(encoded\_paperB, encoded\_candidate\_paper)
                        \If{similarity\_score > threshold}
                            \State finalSimilarPaperPool.append([similarity\_score, citation\_count]) 
                        \EndIf   
                        \State save that
                    \EndIf
                \EndFor
            \EndIf
        \EndFor
    \end{algorithmic}
\end{algorithm}

\subsubsection{Simpsons curve}
\fi
\subsection{Additional Implementation Details}
\subsubsection{Time and Space Complexity Details}\label{causalcite:appd:impl}

For the time cost of running the causal impact indices, each $\mathrm{PCI}(a,b)$ takes around 1,500 seconds, or 25 minutes. Multiplying this by 40 samples per paper $a$, we spend 16.67 hours to calculate each ACI or TCI for the paper's overall impact.
For a fine-grained division into the time cost, the majority of the time is spend on the BM25 indexing (800s) and the sentence embedding cosine similarities calculation (400s). The rest of the time-consuming steps are the BFS search (150-200s every time) to identify descendants and non-descendants of a paper.

For the space complexity, we loaded the 2.4B edges of the citation graph into a  parquet gzip format for faster loading, and use Dask's lazy load operation to load it part by part to RAM for better parallelization. 
The program can fit into different sizes of RAMs by modifying the number of partitions and reducing the number of workers in Dask, at the cost of an increased computation time. 
On the hard disk, citation graph takes up 19G space, and paper data takes 11G.

\subsubsection{Numerical Estimation Method: Finding the Sample Size}\label{causalcite:appd:sample_size}

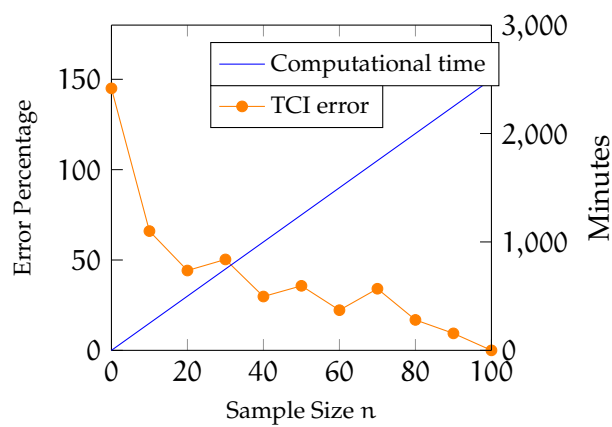
\begin{figure}[ht]
    \centering
    \begin{tikzpicture}
    \pgfplotsset{
        scale only axis,
        xmin=0, xmax=100,
        y axis style/.style={
            yticklabel style=#1,
            ylabel style=#1,
            y axis line style=#1,
            ytick style=#1
       }
    }
    
    \begin{axis}[
      axis y line*=left,
      y axis style=black!75!black,
      ymin=0, ymax=180,
      xlabel={Sample Size $n$
      },
      ylabel={Error Percentage},
    legend style={at={(0.26,0.75)}, anchor=west, font=\footnotesize},
    label style={font=\footnotesize}
    ]
    \addplot[mark=*,orange] 
   coordinates {
    (0, 144.99014469563542)
 (10, 66.00385767461768)
 (20, 44.17677805641458)
 (30, 50.314017672941745)
 (40, 29.788802957937282)
 (50, 35.71244075970475)
 (60, 22.235910683478743)
 (70, 34.097782499124584)
 (80, 16.845642879961183)
 (90, 9.392781270621153)
 (100, 0.0)
};
\addlegendentry{TCI error}
    \end{axis}
    
    \begin{axis}[
      axis y line*=right,
      axis x line=none,
      ymin=0, ymax=3000,
      ylabel={Minutes},
      y axis style=black!75!black,
      legend style={at={(0.26,0.88)}, anchor=west, font=\footnotesize}
    ]
    \addplot[color=blue, domain=0:120]{25*x};
    \addlegendentry{Computational time}
    \end{axis}
    
    \end{tikzpicture}
    \caption{
    We show the trade-off of two curves: the error curve (orange), and the time cost curve (blue). For the error curve, we see an elbow point at around $n=40$, when the error starts to be small. The curve for the computational time is linear, taking 25 minutes for each paper. Balancing the trade-offs, we decided to choose the sample size $n=40$.
        }
    \label{causalcite:fig:simpson}
\end{figure}

For our numerical estimation method, we first calculate the ACI on a subset of carefully sampled papers and then aggregate it to TCI. One design choice question is how to
decide the size of this random subset. In our case, we need to balance both the computation time (25 minutes per pairwise paper impact) and the estimation accuracy.
To identify the best sample size,
we conduct a small-scale study, first obtaining the TCI using our upper-bound budget of $n=100$ samples and then gradually decreasing the number of samples to see if there is a stable point in the middle which also leads to a result close to that obtained with 100 samples.
In \cref{causalcite:fig:simpson}, we show the trade-off of the two curves, the error curve and time cost,
where we can see $n=40$ seems to be a good point balancing the two. It is at the elbow of the arrow curve, making it relatively close to the estimation result of $n=100$, and also in the meantime vastly saving our computational budget, enabling us to run efficient experiments for more analyses.

\subsubsection{Experiment to Select the Best Embedding Method}\label{causalcite:appd:emb}
When selecting the text encoder for our \ourmodel method,
we compare among the three LLMs pre-trained on scientific papers, SciBERT, MPNet, and SPECTER. Specifically, we conduct a small-scale experiment to see how much the similarities scores based on the embedding of each model align with human annotations.
As for the annotation process, we first collect a set of random papers, and for each such paper (which we call a pivot paper), we identify ten papers, from the most similar to the least, with monotonically decreasing similarity. We collect a total of {100} papers consisting of {ten} such collections, for which we show an example in \cref{causalcite:tab:emb}.
Then we see how the resulting similarity scores conform to this order by deducting the percentage of papers that are out of place in the ranking. 

We find that MPNet correlates the best with human judgments, achieving an accuracy of 82\%, which is 10 points better the second best one, SPECTER, which gets 72\%, and 18 points better than SciBERT with a score of 64\%. 
It also gives more distinct scores to papers with different levels of similarity.
This capability advantage
may be attributed to its Siamese network objectives in the training process \citep{song2020mpnet}.
We open-sourced our annotated data in the codebase.

\ifperfect
In order to compare embeddings for our task we take a pivot paper and a sample of papers with decreasing levels of similarity between them. One example of such a comparison table is given in \cref{causalcite:tab:emb}. We find that SciBERT doesn’t perform very well for this sentence-level embedding task as we see all the similarity scores lie within a range and we don’t see a decreasing trend in the similarity scores. SPECTER and MPNet were very close in performance, but we found that the similarity score between the pivot paper and a completely different paper (Paper 10) was a magnitude lower in the case of MPNet whereas it was only 0.1 lower in SPECTER. Due to this reason, we decided to go with MPNet as our embedding model since having a greater distance between unrelated papers is critical to our use case.
\fi
\begin{table*}[t]
\small \centering
\begin{tabular}{llccccc}
\toprule
Paper Index & Title                       & SciBERT & SPECTER & MPNet \\  \midrule
\multicolumn{2}{c}{\textit{Pivot Paper: GPT-3 \citeyearpar{brown2020gpt3}}} \\
1 (Most similar)                         & PaLM \citeyearpar{chowdhery2022palm}                        & 0.9787      & 0.8689  & 0.7679  \\
2                          & GPT-2 \citeyearpar{radford2019language}                        & 0.9346      & 0.9064  & 0.8196  \\
3                          & GPT \citeyearpar{radford2018improving}                        & 0.9488       & 0.8778  & 0.7790  \\
4                          & BERT   \citeyearpar{devlin-etal-2019-bert}                     & 0.9430      & 0.8321   & 0.6784  \\
5                          & Transformers  \citeyearpar{vaswani2017attention}              & 0.9202      & 0.8644  & 0.6385  \\
6                          & SciBERT  \citeyearpar{beltagy-etal-2019-scibert}                   & 0.8396      & 0.8112  & 0.5667  \\
7                          & Latent Diffusion Models \citeyearpar{Rombach2021HighResolutionIS}    & 0.9586      & 0.7755  & 0.4567  \\
8                          & Sentiment Analysis Using DL \citeyearpar{Fang2015SentimentAU}    & 0.7775      & 0.7298  & 0.2911  \\
9                          & Sentiment Analysis Using ML \citeyearpar{Zainuddin2014SentimentAU} & 0.6462      & 0.6403  & 0.2563  \\
10 (Least similar)                        & New High Energy Accelerator \citeyearpar{Courant1952THESS} & 0.8033     & 0.5617  & 0.0359  \\
\bottomrule

\end{tabular}
\caption{
An example collection of papers with monotonically decreasing similarity to the pivot paper. As can be seen from the similarities scores produced by the three text embedding methods, MPNet corresponds to the ground truth the most, and also shows clear score distinctions between less similar and more similar papers.
}
\label{causalcite:tab:emb}
\end{table*}

\subsection{Dataset Overview}\label{causalcite:appd:data}

    \begin{figure}[ht]
        \centering
        \includegraphics[width=.6\linewidth]{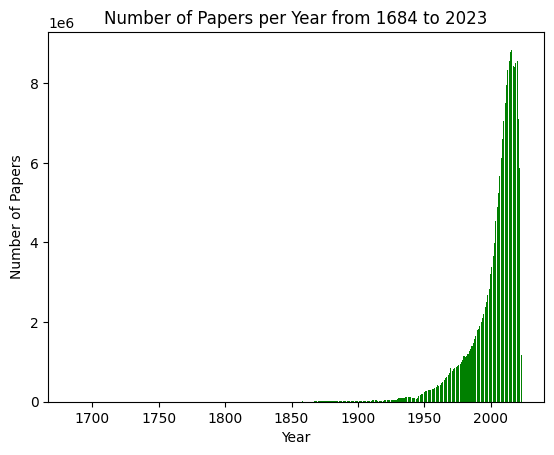}
        \caption{The 
        number of papers published per year
        from 1684 to 2023. We can see that in recent years since 2010, there are more than 7 million papers each year.}
        \label{causalcite:fig:year_distr}
    \end{figure}
    \begin{figure}[ht]
        \centering
        \includegraphics[width=.6\linewidth]{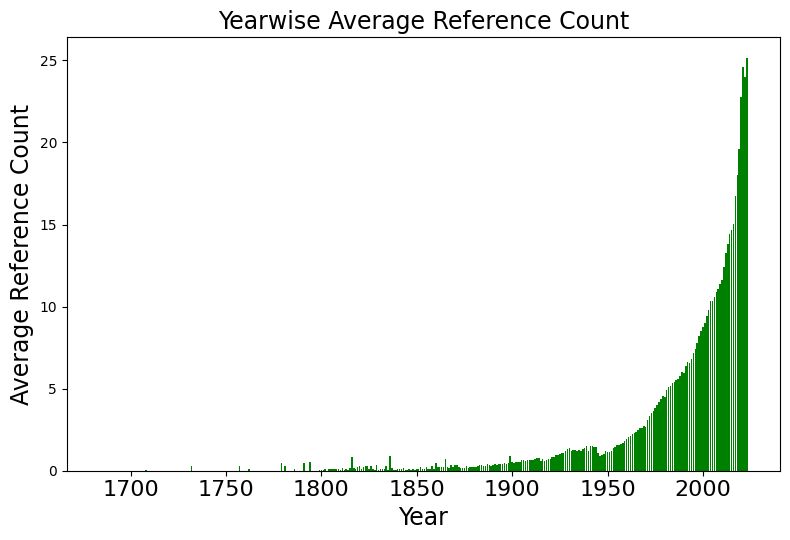}
\caption{
    The 
    year-wise average of the number of references per paper, also with a sharply increasing trend. }
        \label{causalcite:fig:ref_count} 
    \end{figure}
For the Semantic Scholar dataset \citep{Kinney2023TheSS,lo-etal-2020-s2orc}, we obtain the set of 206M papers
using the ``Papers'' endpoint to get the Paper Id, Title, Abstract, Year, Citation Count, Influential Citation Count \citep{valenzuela2015identifying}, and the Reference Count for each paper. 
The papers come from a variety of fields such as law, computer science, linguistics, chemistry, material science, physics, geology, etc.
For the citation network with 2.4B edges, we use the
Semantic Scholar Citations API to get each edge of the citation graph in a triplet format of (fromPaper, toPaper, isInfluentialCitations).

In general, the number of publications shows an explosive increase in recent years. \cref{causalcite:fig:year_distr} shows the number of papers publish the per year, which reaches on average 7.5M per year since 2010. \cref{causalcite:fig:ref_count} shows the number of references each paper cites, which also increases from less than five before 1970s, to around 25 in recent years. 
Both statistics support the need of our paper, which helps distinguish the quality of scientific studies given such massive growths of papers.

\subsection{Additional Analyses}
\subsubsection{Citation Outlier Analysis}\label{causalcite:appd:outlier}

\begin{figure}[t]
    \centering
    \includegraphics[width=.4\linewidth]{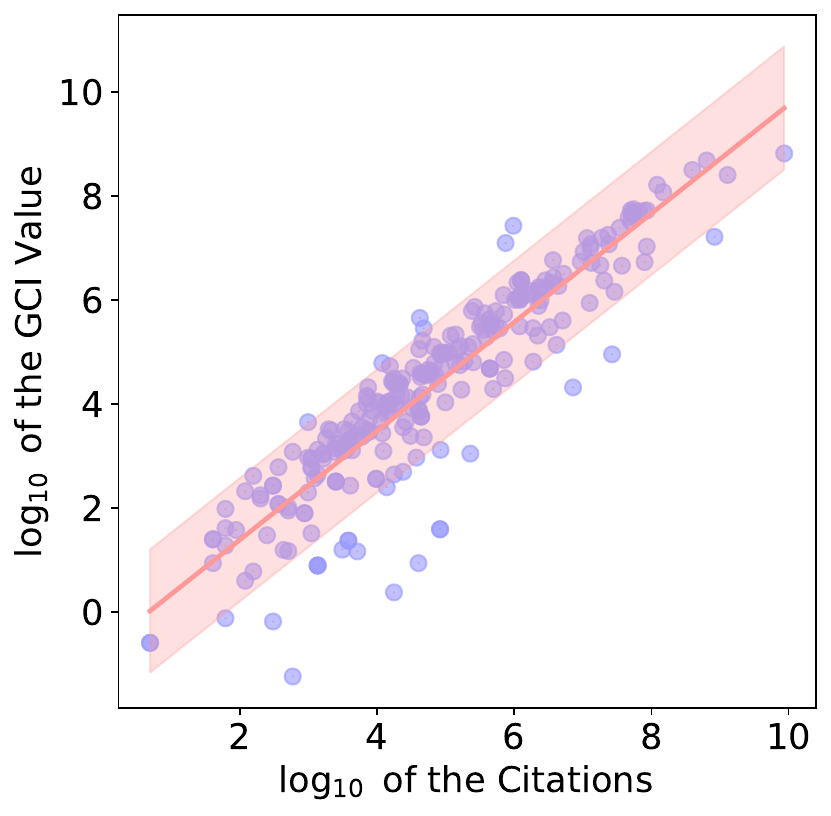}
    \caption{The scatter plot between our \ourname and citations, with the fitted function as $\log (\mathrm{TCI}) = 1.026 *  \log (\mathrm{Cit}) -0.541$, and a non-outlier band width of 0.8809.
    }
    \label{causalcite:fig:correlationPlot}
\end{figure}

For the outlier detection, we first visualize the scatter plot between our \ourname and citations. Then, we fit a log-linear regression to learn the line $\log (\mathrm{TCI}) = 1.026  \log (\mathrm{Cit}) -0.541$, as shown in \cref{causalcite:fig:correlationPlot}, with a root mean squared error (RMSE) of 0.6807.
After fitting the function, we use the interquartile range (IQR) method \citep{outliers},
which identify as outliers any samples that are either lower than the first quartile by over 1.5 IQR, or higher than the third quartile by more than 1.5 IQR, where IQR is the difference between the first and third quartile.

We denote as overcited papers the ones that are identified as outliers by the IQR method due to too many citations than what it should have deserved given the \ourname value. Symmetrically, we denote as undercited papers the ones that are identified as outliers by the IQR method due to too few citations than what it should have deserved given the \ourname value. And we denote the non-outlier papers as the aligned ones.

\subsubsection{Additional Information for the Author-Identified Paper Impact Experiment}\label{causalcite:appd:zhu}
As mentioned in the main paper, the dataset is annotated by pivoting on each paper $b$, and going through each of its references $a$ to label whether $a$ has a significant influence on $b$ or not. We show an example of paper $b$ and all its 31 references in \cref{causalcite:tab:zhu_example}.
We calculate the accuracy of each metric with the spirit that each non-significant paper's impact value should be lower than a significant paper's. Specifically, we go through the score of each non-significant paper, and count its accuracy as 100\% if it is lower than all the significant papers', or the more general form $n_{\mathrm{lower}} / |\mathrm{Sig}|$ of conformity, where $n_{\mathrm{lower}}$ is the number of significant papers which it is lower than, and $|\mathrm{Sig}|$ is the total number of significant papers. Then we report the overall accuracy for each score by averaging the accuracy numbers on each non-significant paper.
To illustrate the idea better, we show the calculated accuracy numbers for all three metrics on our example batch in \cref{causalcite:tab:zhu_example}.

\begin{table*}[t]
    \centering \small
    \begin{tabular}{p{8.5cm}lcccc}
\toprule

\textit{References of the Paper ``Sorting improves word-aligned bitmap indexes''}                                                                                              & Label & \multicolumn{1}{l}{PCI} & \multicolumn{1}{l}{Citations} & SSHI \\ \midrule
- A Quantitative   Analysis and Performance Study for Similarity-Search Methods in   High-Dimensional Spaces & 0                           & \green{3.519}                & 1777                          & 156                        \\
- Optimizing bitmap indices with efficient   compression                                                     & 0                           & \green{3.519}                & 375                           & 40                         \\
- Data Warehouses And Olap: Concepts,   Architectures And Solutions                                          & 0                           & \green{3.526}                & \blackgreen{187}                           & 11                         \\
- Histogram-aware sorting for enhanced   word-aligned compression in bitmap indexes                          & 0                           & \green{3.543}               & \green{17}                            & \green{1}                          \\
- CubiST++: Evaluating Ad-Hoc CUBE Queries   Using Statistics Trees                                          & 0                           & \green{3.543}                & \green{5}                             & \green{1}                          \\
- Improving Performance of Sparse   Matrix-Vector Multiplication                                             & 0                           & \green{3.543}                & \blackgreen{114}                           & 11                         \\
- Binary Gray Codes with Long Bit Runs                                                                       & 0                           & \green{3.543}                & \green{53}                            & \green{4}                          \\
- Analysis of Basic Data Reordering   Techniques                                                             & 0                           & \green{3.543}                & \green{16}                            & \green{1}                          \\
- Tree Based Indexes Versus Bitmap Indexes:   A Performance Study                                            & 0                           & \green{3.543}                & \green{24}                            & \green{0}                          \\
- Secondary indexing in one dimension:   beyond b-trees and bitmap indexes                                   & 0                           & \green{3.543}                & \green{10}                            & \green{1}                          \\
- A comparison of five probabilistic   view-size estimation techniques in OLAP                               & 0                           & \green{3.543}                & \green{24}                            & \green{1}                          \\
- Compression techniques for fast external   sorting                                                         & 0                           & \green{3.543}               & \green{16}                            & \green{0}                          \\
- A Note on Graph Coloring Extensions and   List-Colorings                                                   & 0                           & \green{3.543}                & \green{33}                            & \green{1}                          \\
- Using Multiset Discrimination to Solve   Language Processing Problems Without Hashing                      & 0                           & \green{3.543}                & \green{52}                            & \green{2}                          \\
- Monotone Gray Codes and the Middle Levels   Problem                                                        & 0                           & \green{3.543}                & \blackgreen{80}                            & \blackgreen{5}                          \\
- The Art in Computer Programming                                                                            & 0                           & \green{3.543}                & 9242                          & 678                        \\
- An Efficient Multi-Component Indexing   Embedded Bitmap Compression for Data Reorganization                & 0                           & \green{3.543}                & \green{8}                             & \green{2}                          \\
- The LitOLAP Project: Data Warehousing   with Literature                                                    & 0                           & \green{3.543}                & \green{8}                             & \green{0}                          \\
- Multi-resolution bitmap indexes for   scientific data                                                      & 0                           & \green{3.583}                & \blackgreen{96}                            & \green{3}                          \\
- Notes on design and implementation of   compressed bit vectors                                             & 0                           & \green{3.583}                & \blackgreen{81}                            & 12                         \\
- Compressing Large Boolean Matrices using   Reordering Techniques                                           & 0                           & \green{3.595}                & \blackgreen{88}                            & \blackgreen{7}                          \\
- \textbf{Compressing bitmap indices by data   reorganization}                                                        & \textbf{1}                           & 3.595                & 53                            & 4                          \\
- Model 204 Architecture and Performance                                                                     & 0                           & \blackgreen{3.635}                & 238                           & \blackgreen{10}                         \\
- \textbf{On the performance of bitmap indices for   high cardinality attributes }                                    & \textbf{1}                           & 3.654                 & 196                           & 10                         \\
- A performance comparison of bitmap   indexes                                                               & 0                           & {3.655}                & \blackgreen{86}                            & \blackgreen{9}                          \\
- Minimizing I/O Costs of Multi-Dimensional   Queries with Bitmap Indices                                    & 0                           & {3.692}                & \green{16}                            & \green{0}                          \\
- Evaluation Strategies for Bitmap Indices   with Binning                                                    & 0                           & {3.692}                & \blackgreen{69}                            & \green{3}                          \\
- C-Store: A Column-oriented DBMS                                                                            & 0                           & {3.710}                & 1241                          & 111                        \\
- Byte-aligned bitmap compression                                                                            & 0                           & {3.793}                & 209                           & 48                         \\
- Bit Transposed Files                                                                                       & 0                           & {3.837}                & \blackgreen{84}                            & 10                         \\
- Space efficient bitmap indexing                                                                            & 0                           & {4.011}                & \blackgreen{96}                            & 16  
\\ \bottomrule
\end{tabular}
\caption{All the reference papers for a given study ``Sorting improves word-aligned bitmap indexes.'' 
Among all its 31 references, we \textbf{boldface} the reference papers that are annotated to be significant influencers.
For the three metrics, PCI, citations, and SSHI, we report their impact scores for each reference paper on the given study, where
we mark a score \green{in green} when it conforms to the rule that a non-significant paper's value should be lower than that of a significant paper, and mark a score \blackgreen{in dark green} if it conforms to the rule to have a lower score than one of the significant paper, but violates the rule, i.e., having a higher score than the other significant paper. In this example, our PCI metric has an accuracy score of {79.3}\%, which is higher than both citations ({68.1}\%), and SSHI ({65.0}\%).}
    \label{causalcite:tab:zhu_example}
\end{table*}

\subsubsection{Step Curve for PCI Values Given a Fixed Paper \textit{b}}
\begin{figure}[ht]
    \centering\includegraphics[width=.9\linewidth]{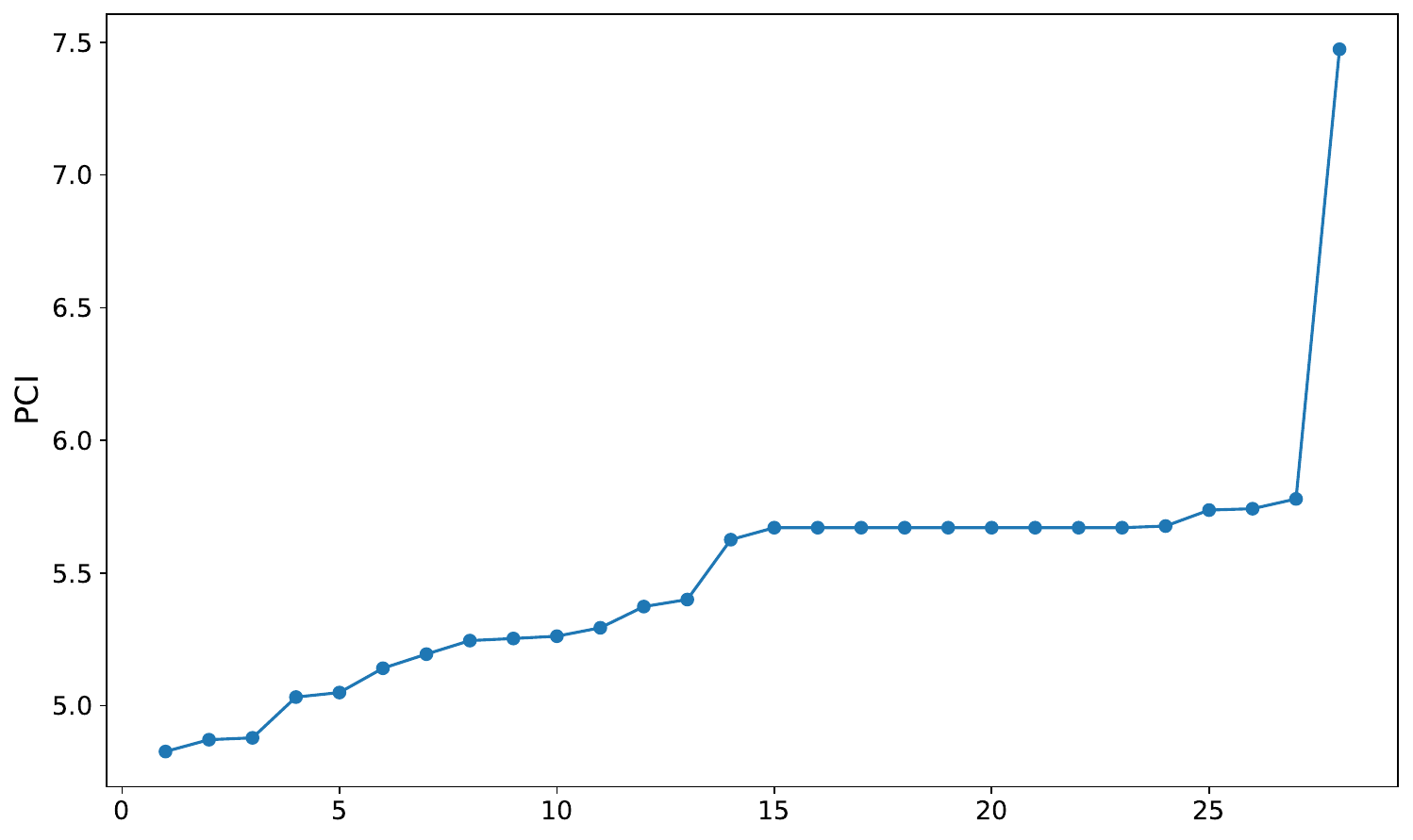}
        \caption{
        We take an example paper $b$, Sentence BERT \citep{Reimers2019SentenceBERTSE}, and plot its PCI values with all its reference paper $a$'s. We can see clearly that there is a plateau in the curve, showing a step function-like nature.
        }
        \label{causalcite:fig:PCIDistributionCurve}
\end{figure}

Apart from the long-tailed curve shape of TCI in \cref{causalcite:sec:curve}, we also look into the pairwise paper impacts by PCI.
If we fix the paper $b$, we can see that $\mathrm{PCI}(\cdot, b)$ often has a step curve shape in \cref{causalcite:fig:PCIDistributionCurve}. The reason behind it lies in the nature of PCI, which is calculated based  on the top K papers that are similar in content with paper $b$, but do not cite paper $a$. When we go through different references, e.g., from $a_1$ to $a_2$ of the same paper $b$, the semantically matched top K papers could still be largely the same pool, and only change when some papers in the pool need to be swapped when releasing the constraint to be that they can cite $a_1$, and adding the constraint that they cannot cite $a_2$.

\newpage
\thispagestyle{empty}
\mbox{}
\newpage

\bibliography{references,refs/refs_causality,refs/refs_cogsci,
refs/refs_causalcite,
refs/refs_corr2cause,
refs/refs_cladder,
refs/refs_compmech,
refs/refs_covidtwitter,
refs/refs_icm,
refs/refs_implicitpers,
refs/refs_mathrobust,
refs/refs_psychcausal,
refs/refs_nlp4sg,refs/refs_semantic_scholar,
refs/refs_zhijing}

\bibliographystyle{Configs/acl_natbib}

\end{document}